\documentclass{article}

\PassOptionsToPackage{numbers, compress}{natbib}
\PassOptionsToPackage{table,dvipsnames}{xcolor}

\usepackage[eandd, final]{neurips_2026}
\usepackage[utf8]{inputenc}
\usepackage[T1]{fontenc}
\usepackage{hyperref}
\usepackage{url}
\usepackage{booktabs}
\usepackage{amsfonts}
\usepackage{amsmath}
\usepackage{amssymb}
\usepackage{nicefrac}
\usepackage{microtype}
\usepackage{graphicx}
\usepackage{tabularx}
\usepackage{xltabular}
\usepackage{longtable}
\usepackage{multirow}
\usepackage{makecell}
\usepackage{array}
\usepackage{colortbl}
\usepackage{float}
\usepackage{wrapfig}
\usepackage{subcaption}
\usepackage{adjustbox}
\usepackage{pdflscape}
\usepackage{siunitx}
\usepackage{bm}
\usepackage{bbm}
\usepackage{mathtools}
\usepackage{textcomp}
\usepackage{verbatim}
\usepackage{pdfpages}
\usepackage[inline]{enumitem}
\usepackage[justification=justified]{caption}
\usepackage{fontawesome5}

\usepackage[normalem]{ulem}
\usepackage{xcolor}

\IfFileExists{algorithm2e.sty}{%
  \usepackage[ruled,vlined,linesnumbered]{algorithm2e}
}{%
  \newenvironment{algorithm}[1][]{%
    \begin{figure}[##1]\begin{minipage}{\linewidth}\small
  }{%
    \end{minipage}\end{figure}
  }
  \newcommand{\KwIn}[1]{\par\noindent\textbf{Input: }##1\par}
  \newcommand{\KwOut}[1]{\par\noindent\textbf{Output: }##1\par}
  \newcommand{\BlankLine}{\par\smallskip}
  \newcommand{\tcp}[1]{\hfill\textit{// ##1}\par}
  \newcommand{\uIf}[2]{\par\noindent\textbf{if }##1\textbf{ then }##2\par}
  \newcommand{\uElseIf}[2]{\par\noindent\textbf{else if }##1\textbf{ then }##2\par}
  \newcommand{\Else}[1]{\par\noindent\textbf{else }##1\par}
  \newcommand{\If}[2]{\par\noindent\textbf{if }##1\textbf{ then }##2\par}
  \newcommand{\lIf}[2]{\par\noindent\textbf{if }##1\textbf{ then }##2\par}
  \newcommand{\Return}[1]{\par\noindent\textbf{return }##1\par}
}

\usepackage{preamble_shared}
\draftcommentsfalse  

\let\origincludegraphics\includegraphics
\RenewDocumentCommand{\includegraphics}{O{} m}{%
  \IfFileExists{plots_camera_ready/#2}%
    {\origincludegraphics[#1]{plots_camera_ready/#2}}%
    {\origincludegraphics[#1]{#2}}}

\providecommand{\justificationYes}[1]{#1}

\title{OpenMHC: Accelerating the Science of Wearable Foundation Models}

\author{\textbf{Narayan Schuetz}$^{1,*}$,
  \textbf{Yuze Bai}$^{2}$,
  \textbf{Lianggang Pan}$^{2}$,
  \textbf{Edgar Eggert}$^{1,3}$, \\
  \textbf{Favour Nerrise}$^{1}$, 
  \textbf{Juan Delgado-SanMartin}$^{2}$,
  \textbf{Max Rosenblattl}$^{1}$,
  \textbf{Milana Gurbanova}$^{1}$, \\
  \textbf{Mohammad Asadi}$^{1}$, 
  \textbf{Anders Johnson}$^{1}$,
  \textbf{Steven G. Hershman}$^{1}$,
  \textbf{Paul Schmiedmayer}$^{1}$, \\
  \textbf{Dennis Wang}$^{2}$, 
  \textbf{Allan Lawrie}$^{2}$, 
  \textbf{Daniel Seung Kim}$^{4}$, 
  \textbf{Xin Liu}$^{4,5}$, \\
  \textbf{Akshay Paruchuri}$^{1,5}$, 
  \textbf{Ehsan Adeli}$^{1}$, 
  \textbf{Euan A. Ashley}$^{1,\dagger}$, 
  \textbf{Kelly W. Zhang}$^{2,*,\dagger}$ \\
  $^1$Stanford University, $^2$Imperial College London, \\
  $^3$Charité -- Universitätsmedizin Berlin, $^4$University of Washington, $^5$Google \\
  $^*$Corresponding authors: \texttt{schuetzn@stanford.edu, kelly.zhang@imperial.ac.uk} \\
  $^\dagger$Co-senior authors 
}

\begin{document}

\maketitle

\begin{abstract}
    Mobile and wearable devices offer an unprecedented opportunity for continuous, passive health monitoring and active health coaching. However, the largest wearable datasets are proprietary, and leading wearable foundation models trained on them are rarely open weight or include reproducible training code. To accelerate open science in wearable health, we introduce \bo{OpenMyHeartCounts (OpenMHC)}, the largest and most comprehensive wearable health dataset broadly accessible to qualified researchers, alongside open-source implementations and weights for recent wearable foundation models. Derived from over a decade of data collected through the My Heart Counts study app, \openmhc{} comprises over 67M participant-hours of wearable recordings across up to 19 sensor channels (e.g., step count, heart rate, sleep, workouts) and up to 170 health, lifestyle, mood, and behavioral variables from 11,894 participants. Furthermore, we establish a benchmark for standardized evaluation across three tasks: (1) health and behavioral outcome prediction, (2) multivariate data imputation, and (3) time-series forecasting. Benchmarking classical methods alongside recent wearable and time-series foundation models, we find that while foundation models perform strongly, carefully tuned classical baselines like XGBoost remain highly competitive. By uniting broad-access wearable data with open code and model weights, \openmhc{} provides a framework needed to drive reproducible advances in wearable health AI.

\end{abstract}

\newenvironment{Itemize}{
    \begin{itemize}[leftmargin=*]
    \setlength{\itemsep}{0pt}
    \setlength{\topsep}{0pt}
    \setlength{\partopsep}{0pt}
    \setlength{\parskip}{1pt}}
{\end{itemize}}

\addtocontents{toc}{\protect\setcounter{tocdepth}{-1}}

\section{Introduction}

\begin{figure}[!t]
  \centering
  \includegraphics[width=\linewidth]{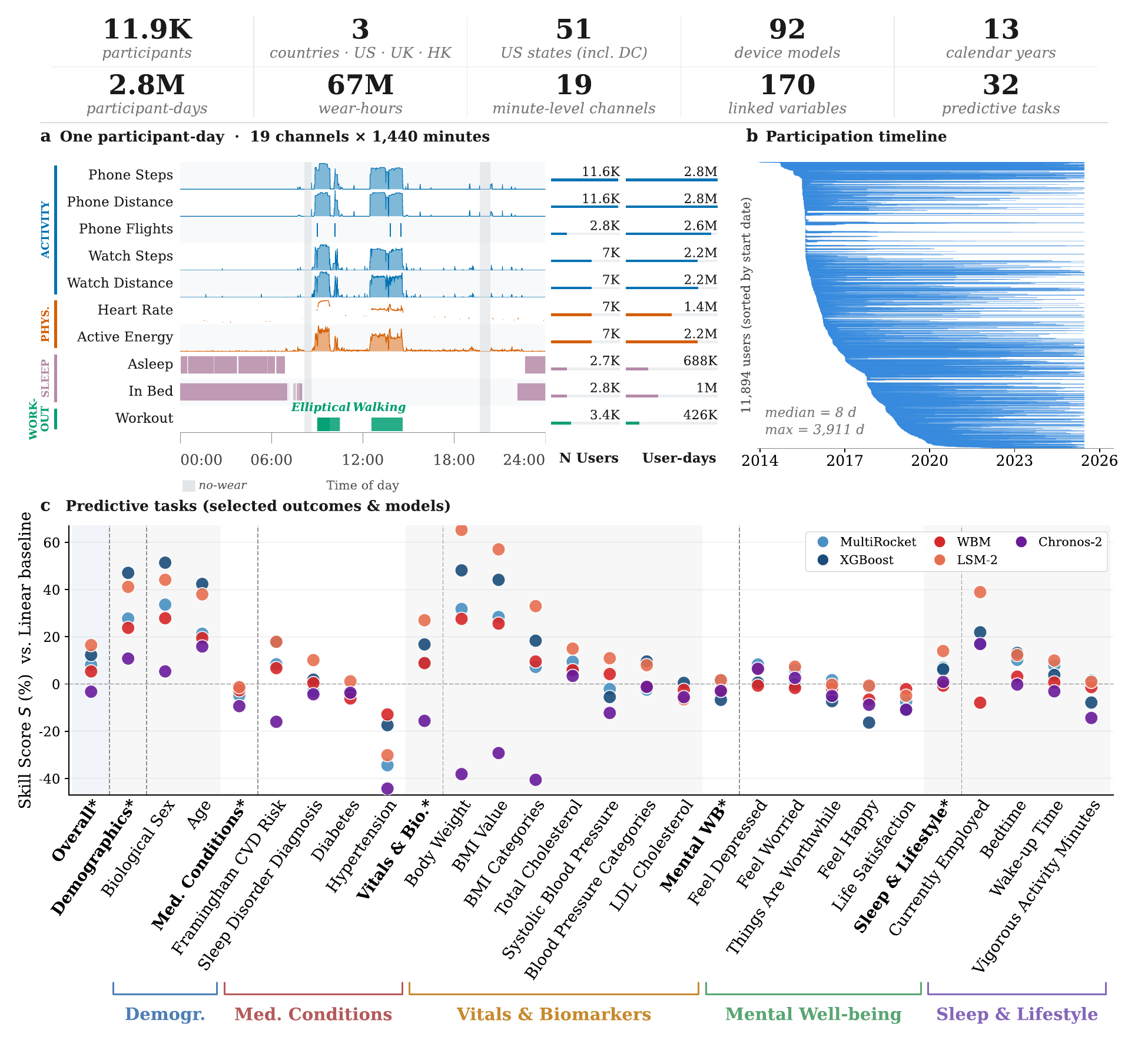}
  \caption{
    \textbf{The \openmhc{} dataset at a glance.} 
    11{,}894 participants across 3 countries (US, UK, HK), 50 US states plus DC, and 92 device models, contributing 2.82M participant-days (over 67M participant-hours) of minute-level sensor data over a 13-year calendar span (earliest to latest recorded day), alongside 170 linked self-report and sparse HealthKit variables and 32 predictive tasks.
    \textbf{(a) Example of minute-level wearable data participant-day} for one user as a 19$\times$1{,}440 matrix spanning Activity, Watch-derived Physiology, Sleep, and 10 Workout types (combined into a single channel for this visualization). Right columns report cohort-wide coverage: number of users contributing to the channel and total user-days. 
    \textbf{(b) Participation timeline:} each line is one of the 11{,}894 participants, sorted by enrollment date, spanning their first to
    last contributed day. 
    \textbf{(c) Skill Scores ($S$, \% vs.\ Linear baseline)} for selected predictive tasks, grouped into five health domains (Demographics, Medical Conditions, Vitals \& Biomarkers, Mental Well-being, Sleep \& Lifestyle), for five methods: \textsc{MultiRocket}, \textsc{XGBoost}, \textsc{WBM}, \textsc{LSM-2}, and \textsc{Chronos-2}. Overall and per-domain summary scores are marked with $\ast$. \vspace{-3mm}
  }
  \label{fig:mhc-overview}
\end{figure}
Mobile and wearable devices have enabled the continuous, passive collection of longitudinal health and behavior data at unprecedented scale \citep{piwek2016rise}. The richness of these streams, spanning physical activity, cardio-respiratory fitness, sleep, and more, has fueled a growing range of applications and research around health and wellness, ranging from chronic disease management and early detection \citep{perez2019large,lubitz2022detection,ajufo2025accelerometer} to digital phenotyping  \citep{matias2026digital,shim2024circadian}, just-in-time adaptive interventions \citep{javed2023personalized,schmiedmayer2026design,klasnja2019efficacy,liao2020personalized}, and personalized health coaching~\citep{schmiedmayer2026design, joerke2026bloom}. Machine learning is often critical for understanding, processing, and effectively utilizing mobile and wearable data for these health applications.
Progress in machine learning application areas has historically been driven by large-scale, high-quality, open datasets and benchmarks (e.g., computer vision advancements were catalyzed by ImageNet \citep{deng2009imagenet}). Yet for wearable and mobile health data, one of the most promising emerging health data modalities, no such broadly accessible foundational datasets exist. To date, larger-scale longitudinal wearable and mobile health datasets are private or strongly access-gated, limiting reproducibility and broader research participation \citep{truslow2024understanding,xu2025lsm,narayanswamy2025scaling}. Moreover, many existing smaller datasets use specialized research-grade devices and sensors, making it harder to establish broadly applicable benchmarks.

This gap is particularly costly given recent evidence of the potential of wearable data: large-scale efforts from industry have demonstrated that modeling wearable and mobile data at scale yields impressive results, with scaling laws reminiscent of those observed in language and vision \citep{narayanswamy2025scaling,narayanswamy2026towards}. These results suggest that limited data access and sharing are major bottlenecks for progress in wearable health modeling. Without open, large-scale datasets, progress risks foreclosing the kind of broad community participation that has propelled other fields.
Our primary contributions are as follows:
\begin{enumerate}[leftmargin=*]
    \item \bo{Curation and release of the largest and most comprehensive broadly accessible wearable health dataset.} 
    We release \openmhc{}, an open science dataset which provides longitudinal records collected over more than a decade from $11,894$ consenting participants. The wearable dataset consists of over 67M participant-hours of wearable data across up to 19 sensor channels, including step count, heart rate, sleep, and workouts, collected via Apple HealthKit from iPhones and Apple Watches. The dataset also includes 170 self-reported and sparse HealthKit variables (diet, lifestyle, medical conditions, mood, etc.). This resource substantially surpasses  existing wearable health datasets broadly accessible to the research community (i.e. without major gating or excessive restrictions) in the number of participants, duration of data collection, and comprehensiveness of linked health and lifestyle variables.
    The scale and longitudinal depth of the dataset enable, for the first time, large-scale pretraining and evaluation of foundation models for wearable health on real-world data, a regime previously inaccessible to the open science community.
    \item \bo{Establishing the first large-scale public benchmark for evaluating wearable health models.}
    To allow the community to better measure progress on this type of data, we develop a comprehensive set of evaluation tasks for wearable and mobile health models, covering three major tasks: prediction, imputation, and forecasting. Our benchmark reflects the real-world complexity of wearable health data, including missingness, irregular sampling, and heterogeneity across individuals.
    \begin{enumerate*}
        \item \bo{Predictive:} We define supervised, predictive tasks for a wide range of health and behavior variables, including self-reported medical conditions such as cardiovascular diseases and diabetes, as well as mental well-being. 
        \item \bo{Generative (Imputation and Forecasting):} We conduct imputation on minute-level sensor data and 24-hour forecasting tasks for hourly data to address realistic missingness and future time series trajectory prediction over rolling windows.
    \end{enumerate*}
    \item \bo{Open-source implementations of wearable foundation models and baselines.} 
    While many 
    wearable foundation models have been recently proposed, their implementations, model weights, and training data have often remained 
    inaccessible. We provide the \emph{first, open-source implementations} of Apple's Wearable Health Behavior model (\textsc{WBM}) \citep{erturk2025beyond} and Google's \textsc{LSM-2} \citep{xu2025lsm}, alongside a suite of classical ML and deep learning baselines. Paired with the \openmhc{} dataset, this creates a 
    reproducible ecosystem for developing and benchmarking wearable health models.

\end{enumerate}
We release the \openmhc{} benchmark \url{https://myheartcounts.stanford.edu/openmhc} (code, weights, leaderboards) and dataset \url{https://doi.org/10.7910/DVN/XNBITM}.

\section{Related Work}

\bo{Wearable and Mobile Health Datasets.}
In Table \ref{tab:wearable-datasets}, we provide an overview of existing wearable and mobile health datasets. We focus on health-related datasets and not those for other tasks like activity recognition. All of Us \citep{singh2024analysis,bailey2025fitbit} is a large-scale US biobank that has high-quality health records and FitBit data from over $30$k individuals; while this dataset is available via application, one must use the All of Us workbench to work with the data and be approved for access, which makes large-scale model training and development challenging for the broader community. Many large-scale biobank studies like the UK Biobank \citep{doherty2017large} and NAKO Germany \citep{weber2024large} only have accelerometer data from individuals for <10 days and are heavily access gated.
Other key studies like the Apple Heart and Movement study \citep{truslow2024understanding} and Google's Fitbit research dataset \citep{xu2025lsm,narayanswamy2025scaling} 
are not publicly available. There are a variety of smaller studies ($\leq 500$ individuals) \citep{jalin2026digital,klasnja2019efficacy,rossi2020multilevel,xu2023globem,xu2023globemdataset}.
An exception is HomeKit, which has $5$k participants, but is limited to $\leq 4$ months of monitoring per individual and has few sensor channel types and outcomes.

\begin{table}[h]
  \centering
  \footnotesize
  \setlength{\tabcolsep}{3pt} 
  \renewcommand{\arraystretch}{1.05} 
  \caption{\textbf{Wearable Health Datasets.} Modalities: ACC (Accelerometer), ST (Steps), HR (Heart rate), SL (Sleep), DT (Distance), FC (Floors Climbed), CB (Calories), WO (Workouts). N: participants in the reported cohort. Access: \checkmark Credentialed and Downloadable Access (registration or light-touch request), $\square$ Credentialed and Platform-Only Access (approved researchers; accessible only within a research platform), $\triangle$ Controlled Access (substantive project/application review), $\times$ Private (no external access). Unreported hours are upper bounds from cohort size and study duration; coverage definitions differ (Appendix~\ref{app:proprietary-comparison}).}
  \label{tab:wearable-datasets}
  \begin{tabular*}{\textwidth}{@{\extracolsep{\fill}} l r r l l l c @{}}
    \toprule
    \textbf{Dataset} & \textbf{N} & \textbf{Hours} & \textbf{Source} & \textbf{Duration} & \textbf{Modalities} & \textbf{Access} \\
    \midrule
    China Kadoorie \citep{chen2023device} & 22k & 3M & Axivity & 7 days & ACC & $\triangle$ \\
    NAKO Germany \citep{weber2024large} & 74k & 10M & ActiGraph & 7 days & ACC & $\triangle$ \\
    UK Biobank \citep{doherty2017large} & 100k & 17M & Axivity & 7 days & ACC & $\square$ \\
    NHANES \citep{aguiar2024daily} & 4k & <868k & ActiGraph & 9 days & ACC & $\checkmark$ \\
    All of Us \citep{singh2024analysis,fulda202611} & 30k & 264M & Fitbit & Multi-year & ST,HR,SL & $\square$ \\
    \midrule
    Fitbit (LSM-1) \citep{narayanswamy2025scaling} & 165k & 40M & Fitbit & 18 mo. & 26 features* & $\times$ \\
    AHMS (WBM) \citep{erturk2025beyond} & 162k & 2.5B & HealthKit & 5 years & 27 variables* & $\times$ \\
    HomeKit \citep{homekit} & 5k & 14M & Fitbit & 4 mo. & ST,HR,SL & $\checkmark$ \\
    MHC Legacy \citep{hershman2019physical} & 3.5k & <19M & HealthKit & 8 mo. & ST,HR,SL,DT,FC,CB,WO & $\checkmark$ \\
    MMASH \citep{rossi2020multilevel} & 22 & 528 & BioBeats & 1 day & ACC,ST,SL,HR & $\checkmark$ \\
    HeartSteps V1 \citep{klasnja2019efficacy} & 37 & 44k & Jawbone & 6 wks & ST & $\checkmark$ \\
    Roadmap HCT \citep{jalin2026digital} & 332 & <478k & Fitbit & 4 mo. & ST,SL,HR & $\checkmark$ \\
    GLOBEM \citep{xu2023globem,xu2023globemdataset} & 500 & <8M & Fitbit & 2 years & ST,SL,DT & $\checkmark$ \\
    \midrule
    \textbf{OpenMHC (ours)} & \textbf{12k} & $\mathbf{>67}$\textbf{M} & \textbf{HealthKit} & \textbf{10 years} & \textbf{ST,HR,SL,DT,FC,CB,WO} & \checkmark \\
    \bottomrule
    \multicolumn{7}{l}{\scriptsize *Model inputs: sensor-derived features (LSM-1); derived HealthKit variables, excluding sleep (WBM).}
  \end{tabular*}
\end{table}

\bo{Foundation Models for Wearable Health Data.}
The predominant focus in foundation model (FM) research for wearable health data has been on learning representations from raw physiological signals, e.g., PPG \citep{saha2025pulse,pillai2025papagei}, ECG \citep{abbaspourazad2024largescale}. Exceptions include wearable accelerometer data \citep{xu2025relcon,yuan2024self,abbaspourazad2024wearable}. These low-level sensor data are used for tasks such as activity recognition \citep{xu2025relcon,yuan2024self} and detecting health conditions like atrial fibrillation \citep{perez2019large,guo2019mobile,lubitz2021rationale,lubitz2022detection}.
In contrast, far less work has explored FMs operating at the level of behavioral and lifestyle measures, e.g., step counts, activity bouts, 
and sleep patterns. 
The few wearable FMs in this space include Apple's wearable health behavior model (\textsc{WBM}) \citep{erturk2025beyond}, which was pretrained using contrastive learning and evaluated via linear probes on self-reported health outcomes and time-varying health detection tasks, and Google's Large Sensor Model (\textsc{LSM}) series \citep{narayanswamy2025scaling,xu2025lsm}, which utilized a pretrained masked autoencoder for both predictive and generative tasks. 
However, these existing efforts \citep{erturk2025beyond,narayanswamy2025scaling,xu2025lsm} were developed and evaluated exclusively on large-scale proprietary datasets, and neither model weights nor training code have been publicly released, making comparison, replication, or extension by the broader research community challenging. Our work aims to fill this gap by providing a first large-scale, broadly accessible dataset for (pre-)training and evaluation, as well as public re-implementations of current wearable foundation models \citep{erturk2025beyond,xu2025lsm}, to enable reproducible research.
By reported data hours, \openmhc{} is of the same order of magnitude as \textsc{LSM-2}'s pretraining corpus and approximately 2--3\% of \textsc{WBM}'s; see Appendix~\ref{app:proprietary-comparison} for a more detailed comparison. 

\section{OpenMHC-Dataset}
\bo{Dataset Collection.}
OpenMHC is derived from the My Heart Counts (MHC) study, a large-scale, smartphone-based cardiovascular health study developed at Stanford University~\citep{mcconnell2017feasibility, hershman2019physical}. The study's iOS application was built using Apple's ResearchKit and launched on the US App Store in March 2015, open to English-speaking individuals aged $\geq 18$ with a US-registered iPhone (iOS~8+). 
The study has since expanded to include a UK arm and has also been available in Hong Kong~\citep{javed2023personalized}, with an Android version planned for 2027~\citep{schmiedmayer2026design}.
The MHC dataset includes de-identified data from its original release up until December 2025  ($\approx 85\%$ US, $\approx 14\%$ UK, $\approx 1\%$ Hong Kong participants). 
Participants enrolled digitally through an eConsent process and could designate their de-identified data for sharing either with Stanford only (``narrow'') or with qualified researchers worldwide (``broad''), the latter of which makes up this dataset, while the former will serve as a private holdout set for future competitions. 
The app collects data both passively and actively. Passive streams include Apple HealthKit data sourced from built-in iPhone sensors as well as compatible wearable devices (e.g., Apple Watch). 
Active data collection consists of health and lifestyle questionnaires (Appendix \ref{app:health_outcome_types}).
The original collection protocol and study design are described by \citet{hershman2019physical}; further questionnaires, such as fo COVID-19 and mindset were added subsequently.
Ethical oversight 
was obtained from Stanford University's Research Compliance Office (Protocol \#IRB-31409).

\bo{Dataset Characteristics.}
The total dataset contains data from 16,993 users (>80M hours of passive Apple HealthKit data), of which we are permitted to share data from 11,894 users (over 67M participant-hours), including users with >10 years of data history. Prevalent passive data are shared as minute-level daily matrices $d \in \real^{19 \times 1440}$, covering activity, physiology, sleep, and workouts (Figure \ref{fig:mhc-overview}a). Seven sparse HealthKit variables (e.g., VO2max) are shared separately, together with 163 self-reported variables related to demographics, disease status, lifestyle, diet, well-being, mindset, Covid, geographical location, and more (Figure \ref{fig:context-overview}).  Participants come predominantly from the US, with around 1,145 from the UK and 80 from Hong Kong. US participants come from all over the US, with a focus in California (Figure \ref{fig:geography-distribution}).
The median participant is 39 years old, male (77.1\%), and white (82.4\%). 
Due to high attrition, participants contributed a median of 8 days of data: approximately 50\% contributed less than one week, about 30\% more than one month, and 13.1\% at least one year (maximum 10.7 years; Figure~\ref{fig:mhc-distributions}b), enrollment/attrition behavior typical for first-generation mHealth apps \citep{amagai2022challenges}. We discuss these demographic and participation limitations in the context of other digital health studies in Appendix~\ref{sec:limitations}. The cohort spans a wide range of consumer devices: 14 iPhone generations and 16 Apple Watch generations released between 2012 and 2025, spanning 92 distinct phone and watch device models in total (see Appendix~\ref{app:dataset_characteristics} for additional dataset statistics).

\bo{Data Preprocessing and Release.}
All benchmark tasks use a shared data preprocessing pipeline, which includes removing anomalous or physiologically impossible values. Additionally, since HealthKit does not track missingness explicitly, it is not known whether an individual is inactive or simply not wearing/carrying the device. Thus, we developed heuristics for non-wear and missing periods (Appendix~\ref{app:shared_wearable_preprocessing}; Figure~\ref{fig:weartime-coverage-samples}) and report resulting coverage summaries in Figure~\ref{fig:mhc-distributions}. 
For reproducible research, we release official dataset splits at participant level ($60\%$ train, $10\%$ val, $30\%$ test), detailed participant and per-split statistics are found in Table \ref{tab:demographics}, and effective per-track sample sizes after filtering in Table~\ref{tab:effective_sample_sizes}. For development purposes, we additionally define a tiny variant, \mbox{\openmhc{}-XS}, comprising 5\% of the total users subsampled from the main split. Additional data, like self-reported variables or sparse HealthKit metrics, are provided as json files.
Data are distributed to all qualified researchers free of charge via Harvard Dataverse under a click-through Data Use Agreement (Appendix~\ref{app:data_access}).

\section{OpenMHC-Benchmark}
\label{sec:benchmark}

\bo{Track 1: Predictive Tasks.}
Increasingly, wearable data is used to predict a variety of health outcomes, including cardiometabolic conditions \citep{guo2019mobile,lubitz2021rationale,lubitz2022detection,metwally2026insulin,perez2019large,delgado2026assessing}, mental health outcomes \citep{abd-alrazaq_systematic_2023, ahmed_wearable_2023}, and sleep-related outcomes \citep{Walch2019SleepSP, retamales_towards_2024}. We develop prediction tasks that aim to probe a model's ability to predict a variety of health and behavior-related characteristics \citep{belinkov2022probing} based on \textbf{32} health and behavior outcomes across five domains:  Demographics (2), Medical conditions \& risk (12), Vitals \& blood biomarkers (8), Mental Well-being (5), Sleep \& Lifestyle (5).

\begin{wrapfigure}{r}{0.65\textwidth}
  \centering
  \vspace{-2em}
  \includegraphics[width=.65\textwidth]{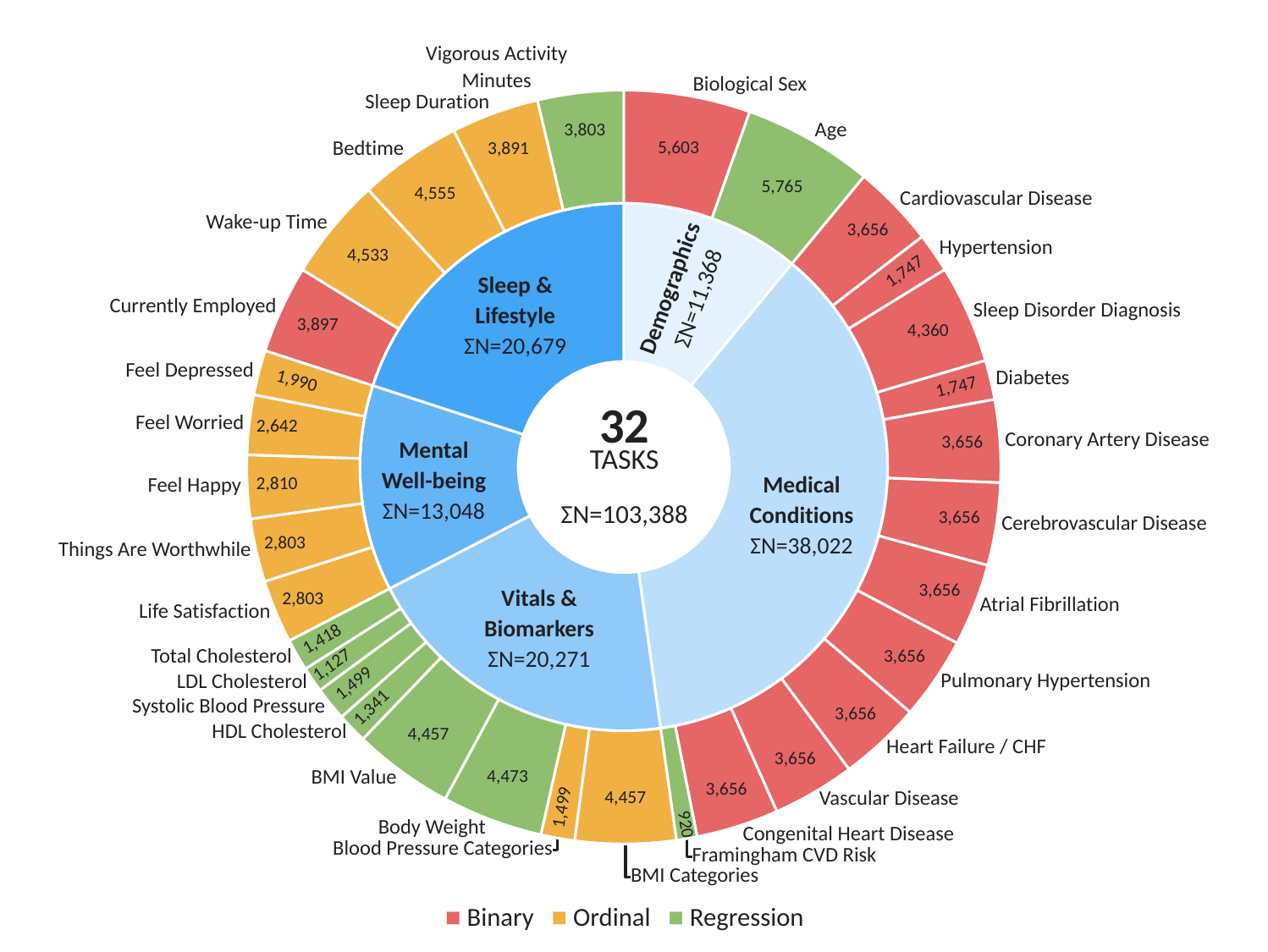}
  \caption{Health and behavior measures used for prediction (\textbf{32 tasks} total), organized by domain. Numbers denote included participants per task (Appendix~\ref{app:health_outcome_inclusion_and_context}), and sums represent the totals across all tasks in each domain.}
  \vspace{-0.75em}
  \label{fig:target_labels}
\end{wrapfigure}

The 32 outcomes include survey responses, enrollment and app profile information, HealthKit fields, and quantities derived from these sources, such as BMI and Framingham CVD risk.
Medical condition labels are self-reported and the labels do not include diagnosis times, only the time at which the survey was conducted. They are therefore a proxy for the underlying health state rather than clinical ground truth. 
Accordingly, these tasks are intended as \emph{retrospective digital-phenotyping  benchmarks}, and should not be interpreted as diagnosis, screening, prospective risk prediction, or clinical decision support; they evaluate whether models can recover health and behavior characteristics from all available wearable data collected. 
We view this evaluation as a step toward 
future, clinical studies of wearable-based risk detection, as well as a useful way to assess the representations learned by wearable foundation models. See Appendix~\ref{app:prediction_models} for additional details on how we set up the prediction tasks.
Note, the benchmark allows models to use different choices of time resolutions and filtering approaches for data quality. 

\bo{Track 2A: Generative (Imputation).}
Missingness in free-living wearable data is the norm rather than the exception, e.g., due to removal for showering or charging, sensor malfunctions, and users forgetting to wear their device. 
Because any model operating on such data must either tolerate or recover from these gaps, imputation quality directly affects the utility of wearable data for almost all applications.
We define a \textit{Daily Imputation} task to evaluate imputation methods across six masking approaches that reflect real-world failure modes in minute-level, single-day sensor data (Figure~\ref{fig:imputation_task_structure}). In this setting, models observe 24 hours of multichannel sensor data with subsets of channels and time intervals masked, and are tasked with reconstructing the masked-out values. Masking approaches are organized into two tiers: \textit{(1) Structural masks} simulate data-collection failures that randomly mask out blocks of times and/or channels in a way that does not depend on the underlying signal values, similar to \cite{xu2025lsm}. \textit{(2) Semantic masks} simulate missingness driven by user activity (e.g., sleep and workouts)
and require a model to learn relationships across channels.
We further consider a \textit{Long-Context Imputation} task, in which models may leverage historical user data beyond a single day, as well as additional contextual features (e.g., demographics), to improve performance on the same 24-hour multichannel imputation task.
 See Appendix~\ref{sec:masking_scenarios} for additional details.

\bo{Track 2B: Generative (Forecasting).}
Wearable data forecasting---such as predicting step count, heart rate trends, sleep duration, or activity levels---has the potential to improve the timeliness and relevance of health coaching and interventions \citep{park2023development,nahum2016just,lee2024examining,schmiedmayer2026design,khasentino2025personal}. 
In this task, we evaluate the ability of models to forecast future physical activity and health behaviors derived from iPhone and Apple Watch data, including step count, flights climbed, heart rate, sleep, and workout. 
Models are trained to forecast the next 24 hours of wearable data at an hourly resolution, given all available historical data (subject to model capacity constraints). We use a rolling evaluation setup: for each trajectory, the model predicts the next 24 hours, after which the history is advanced by 24 hours, and the process is repeated. 
For more details 
see Appendix~\ref{sec:forecasting_data_preprocessing}.

\subsection{Evaluation Approach}
\label{sec:evaluation}
Each of our three tasks comprise of multiple sub-tasks.
We first describe evaluation metrics for each sub-task, then present unified metrics that aggregate performance across sub-tasks.
\begin{itemize}[leftmargin=*]
    \item \bo{Track 1: Predictive Tasks.} Outcomes are either binary, ordinal, or real-valued. For evaluation, we use Area Under the Precision-Recall Curve (AUPRC) for binary outcomes (due to class imbalance; see Appendix~\ref{app:health_outcome_types}), Spearman $\rho$ for ordinal outcomes, and Pearson's $r$ for real-valued outcomes.
    \item \bo{Track 2: Generative (Imputation and Forecasting).} 
    We use Mean Absolute Error (MAE) for continuous channels and ROC AUC for binary channels.
\end{itemize}

\bo{Unified Evaluation Metrics: Rank and Skill Score.}
For each of the tracks, to provide an aggregated model performance metric across sub-tasks, we report (i) the \textit{average rank} of each model across sub-tasks (rank 1 is the best model), and (ii) the \textit{Skill Score} \citep{hyndman2018forecasting,shchur2025fevbench} of each model, which measures relative error reduction with respect to a fixed reference model, aggregated across sub-tasks. Specifically, the Skill Score for model $j$ is computed as the following geometric mean across tasks:
\begin{equation}
    S_j = 1 - \text{GeometricMean}\left( \text{clip}\left( \frac{E_{r,j}}{E_{r,b}}, \ell, u \right) \right),
    \label{eqn:skillscore}
\end{equation}
where for task $r$, $E_{r,j}$ is the error of model $j$ and $E_{r,b}$ is the error of the reference model $b$. Error ratios are clipped for numerical stability; use $\ell = 0.01$ and $u = 100$ (see Appendix~\ref{sec:scoring_methodology} for more details).
A Skill Score of $0$ indicates performance on par with the reference, while a score of $0.2$ indicates a $20\%$ geometric-mean reduction in error. Negative scores indicate performance worse than the reference. The geometric mean balances proportional improvements and degradations. Halving the error on one task and doubling it on another equally weighted task gives a mean error ratio of $1$, compared with $1.25$ under arithmetic averaging \citep{hyndman2018forecasting,shchur2025fevbench}.
Appendix~\ref{app:aggregation_sensitivity} reports sensitivity analyses, including to the reference model choice, clipping bounds, and classification metrics used.


\bo{Subgroup Reporting.}
Following prior wearable-model evaluations~\citep{erturk2025beyond,xu2025lsm}, we report rank and skill score for \textit{female}, \textit{male}, \textit{age $<$ 40}, and \textit{age $\geq$ 40} (40 is the median age).\footnote{We do not stratify by ethnicity, which only 3,769 of the 11,894 participants reported (Table~\ref{tab:demographics}); similar limitations apply to geography and device subgroups (Appendix~\ref{app:dataset_characteristics}).} 
Participants with an unreported attribute are excluded only from that attribute's groups. Each subgroup uses the reference evaluated within that subgroup. Tables~\ref{tab:health_outcome_main_results}--\ref{tab:forecasting_grouped_model_summary} report $S_{\TN{worst}} = \min_{g \in \MC{G}} S^{(g)}$, the lowest skill score over these four groups. Full subgroup results 
are in Appendix~\ref{app:subgroup_reporting}; interactive views are available on the \openmhc{} leaderboard: \url{https://myheartcounts-openmhc.hf.space}.

\section{Experiments and Results}
We train and evaluate a range of models across our benchmark tracks on the full \openmhc dataset. 
While we reasonably optimize each model given a limited compute budget, our goal is not to provide perfectly optimized models  but rather to establish reproducible baselines for training and evaluation on \openmhc as well as providing points of comparison 
across different model types. 
\subsection{Track 1: Predictive Tasks}

\bo{Models.}
Following \cite{erturk2025beyond}, our first baseline is a (generalized) \textsc{Linear} model that takes age, sex, BMI, and summary statistics (i.e., mean and standard deviation) computed from hourly wearable data as input. (Note, we only include age, sex, and BMI as inputs to the \textsc{Linear} model.)
Specifically, \textsc{Linear} uses ordinary least squares for continuous outcomes and logistic regression for binary outcomes. A commonly missing part in evaluations of wearable (foundation) models is comparing to 
domain-expertise guided feature engineering %
methods. To this end, we implement 
\textsc{XGBoost} \citep{chen_xgboost_2016,shwartz2022tabular,grinsztajn2022tree} with 
495 hand-crafted features extracted from minute-level wearable data,
spanning physical activity, sleep, circadian rhythm. 
We also compare to 
\textsc{MultiRocket}~\citep{tan_multirocket_2022}, a 
method that automatically extracts convolutional features for time-series classification,
as well as \textsc{GRU-D} \citep{che_recurrent_2018}, a  supervised neural approach applied to hourly-level wearable data. 

Additionally, we compare to probes trained on two pre-trained self-supervised foundation models for wearable data:
our reimplementation of Apple's wearable behavior foundation model  (\textsc{WBM}) \citep{erturk2025beyond}, a Mamba2-based contrastive encoder that learns representations of user weeks using hourly-level data, 
and our reimplementation of Google's \textsc{LSM-2}, a ViT-style masked autoencoder operating on minute-level daily segments \citep{xu2025lsm}. The probes use the last hidden layer and freeze the pre-trained models. 
We compare also to probes trained on representations from multi-variate time series foundation models 
fine-tuned on forecasting (\textsc{Chronos-2}~\citep{ansari2025chronos} and \textsc{Toto 1.0} \citep{cohen2025time}). See Appendix~\ref{app:prediction_models} for details. 
Pretraining our \textsc{WBM} and \textsc{LSM-2} reimplementations required approximately 3 and 45 GPU-hours, respectively (Appendix~\ref{sec:computing_resources}). 
We additionally evaluate Gemini-family LLMs on zero and few-shot prediction on these tasks, but they generally underperform baselines (Appendix~\ref{app:gemini_baselines}).

\begin{table}[h!]
    \vspace{-2mm}
    \renewcommand{\arraystretch}{1.05}
    \centering
    \captionsetup{width=\textwidth}
    \providecommand{\est}[3]{\ensuremath{#1^{\scriptscriptstyle +#2}_{\scriptscriptstyle -#3}}}
    \caption{\textbf{Prediction Tasks.} We report Average Rank $R$, Aggregate Skill Score $S$ (in \%; $0=$\textsc{Linear} is the reference), and 
    domain-level Skill Scores across the five 
    domains: \textit{Demographics, Medical Conditions \& Risk, Vitals \& Blood Biomarkers, Mental Well-Being, Sleep \& Lifestyle}.
    Below, FT denotes fine-tuned; $^{*}$ denotes reimplementations of the original paper. Methods are grouped by class and ordered by Average Rank within each group. Values are point estimates on the held-out test split; subscripts and superscripts indicate the $95\%$ bootstrap confidence interval ($1$k resamples). 
    For worst-subgroup skill $S_{\TN{worst}}$, see Appendix~\ref{app:subgroup_reporting}. Additional experimental details are in Appendix~\ref{app:prediction_models}.}
    \label{tab:health_outcome_main_results}
    \small
    \setlength{\tabcolsep}{3pt}
    \resizebox{\linewidth}{!}{%
    \begin{tabular}{l ccc ccccc}
    \toprule[1.5pt]
    \textbf{Method} & $R \downarrow$ & $S \uparrow$ & $S_{\TN{worst}}\uparrow$ & \makecell{Demo-\\graphics\,$\uparrow$} & \makecell{Medical\\Conditions \& Risk\,$\uparrow$} & \makecell{Vitals \& Blood\\ Biomarkers\,$\uparrow$} & \makecell{Mental\\Well-Being\,$\uparrow$} &\makecell{Sleep \&\\Lifestyle\,$\uparrow$} \\
    \hline
\multicolumn{9}{l}{\cellcolor[HTML]{EFEFEF}\textit{Statistical Models}} \\
\textsc{MultiRocket}~\citep{tan_multirocket_2022} & \cellcolor{customblue!71}\est{3.53}{0.70}{0.22} & \cellcolor{customblue!62}\est{+8.1}{2.3}{2.5} & \cellcolor{customblue!59}\est{+5.6}{3.0}{3.7} & \cellcolor{customblue!59}\est{+27.7}{5.7}{5.4} & \cellcolor{customblue!48}\est{-4.8}{6.1}{5.0} & \cellcolor{customblue!58}\est{+9.2}{4.4}{5.8} & \cellcolor{customblue!100}\est{+1.6}{4.9}{5.7} & \cellcolor{customblue!55}\est{+6.9}{4.7}{5.0} \\
\textsc{XGBoost}~\citep{chen_xgboost_2016} & \cellcolor{customblue!68}\est{3.65}{0.49}{0.34} & \cellcolor{customblue!81}\est{+12.3}{2.3}{2.5} & \cellcolor{customblue!77}\est{+9.6}{4.8}{4.4} & \cellcolor{customblue!100}\est{\mathbf{+47.1}}{5.5}{5.2} & \cellcolor{customblue!80}\est{-1.8}{5.3}{4.9} & \cellcolor{customblue!76}\est{+16.8}{4.3}{5.2} & \cellcolor{customblue!5}\est{-6.7}{5.7}{6.0} & \cellcolor{customblue!50}\est{+6.2}{5.0}{5.2} \\
\textsc{Linear} \textit{(reference)} & \cellcolor{customblue!40}\est{4.71}{0.55}{0.39} & $0.0$ & $0.0$ & $0.0$ & $0.0$ & $0.0$ & $0.0$ & $0.0$ \\
\hline
\multicolumn{9}{l}{\cellcolor[HTML]{EFEFEF}\textit{Supervised Neural Models}} \\
\textsc{GRU-D}~\citep{che_recurrent_2018} & \cellcolor{customblue!27}\est{5.19}{0.44}{0.57} & \cellcolor{customblue!42}\est{+3.6}{2.6}{2.9} & \cellcolor{customblue!36}\est{+0.6}{3.4}{3.9} & \cellcolor{customblue!42}\est{+19.6}{7.1}{7.2} & \cellcolor{customblue!56}\est{-4.1}{4.5}{5.1} & \cellcolor{customblue!61}\est{+10.4}{4.2}{5.7} & \est{-7.1}{6.3}{6.3} & \cellcolor{customblue!6}\est{-0.6}{5.1}{5.6} \\
\hline
\multicolumn{9}{l}{\cellcolor[HTML]{EFEFEF}\textit{Time-Series Foundation Models}} \\
\textsc{Chronos-2} (FT)~\citep{ansari2025chronos} & \cellcolor{customblue!9}\est{5.88}{0.37}{0.67} & \cellcolor{customblue!10}\est{-3.3}{2.6}{3.3} & \cellcolor{customblue!15}\est{-4.1}{4.5}{7.3} & \cellcolor{customblue!23}\est{+10.8}{7.5}{7.9} & \est{-9.4}{3.5}{5.3} & \est{-15.6}{5.6}{7.6} & \cellcolor{customblue!48}\est{-2.9}{6.1}{6.4} & \cellcolor{customblue!15}\est{+0.9}{5.4}{5.7} \\
\textsc{Toto} (FT)~\citep{cohen2025time} & \est{6.23}{0.30}{0.64} & \est{-5.5}{2.7}{3.1} & \est{-7.5}{4.4}{6.0} & \cellcolor{customblue!2}\est{+1.0}{8.4}{8.7} & \cellcolor{customblue!6}\est{-8.8}{3.5}{5.2} & \cellcolor{customblue!10}\est{-11.5}{5.8}{7.4} & \cellcolor{customblue!3}\est{-6.8}{5.9}{6.8} & \est{-1.5}{5.4}{5.7} \\
\hline
\multicolumn{9}{l}{\cellcolor[HTML]{EFEFEF}\textit{Wearable Foundation Models}} \\
\textsc{LSM-2}$^{*}$~\citep{xu2025lsm} & \cellcolor{customblue!100}\est{\mathbf{2.43}}{0.41}{0.40} & \cellcolor{customblue!100}\est{\mathbf{+16.5}}{2.3}{2.4} & \cellcolor{customblue!100}\est{\mathbf{+14.8}}{2.8}{4.0} & \cellcolor{customblue!87}\est{+41.1}{5.4}{5.5} & \cellcolor{customblue!86}\est{-1.3}{4.4}{4.8} & \cellcolor{customblue!100}\est{\mathbf{+27.0}}{4.3}{4.9} & \cellcolor{customblue!100}\est{\mathbf{+1.6}}{5.6}{5.8} & \cellcolor{customblue!100}\est{\mathbf{+14.0}}{4.3}{4.2} \\
\textsc{WBM}$^{*}$~\citep{erturk2025beyond} & \cellcolor{customblue!48}\est{4.40}{0.49}{0.55} & \cellcolor{customblue!49}\est{+5.3}{1.7}{1.9} & \cellcolor{customblue!44}\est{+2.3}{4.7}{6.3} & \cellcolor{customblue!51}\est{+23.8}{5.1}{5.3} & \cellcolor{customblue!72}\est{-2.6}{2.3}{3.3} & \cellcolor{customblue!57}\est{+8.8}{2.9}{3.2} & \cellcolor{customblue!52}\est{-2.6}{3.7}{3.6} & \cellcolor{customblue!5}\est{-0.7}{3.0}{3.4} \\
    \bottomrule[1.5pt]
    \end{tabular}%
    }
    \vspace{-2mm}
\end{table}

\bo{Results.}
Table~\ref{tab:health_outcome_main_results} shows that no single model consistently performs the best across predicting all 32 outcomes or across all five domains. \textsc{LSM-2} performs best overall, achieving the best Average Rank ($R=2.43^{+0.41}_{-0.40}$), and aggregate Skill Score ($S=+16.5^{+2.3}_{-2.4}$). 
\textsc{XGBoost} is the closest competitor by Skill Score ($S=+12.3^{+2.3}_{-2.5}$), performing best on \textit{Demographics}, while \textsc{MultiRocket} achieves the second-best Average Rank ($R=3.53^{+0.70}_{-0.22}$), narrowly ahead of \textsc{XGBoost}.\textsc{LSM-2} also leads on \textit{Vitals \& Blood Biomarkers}, \textit{Mental Well-Being}, and \textit{Sleep \& Lifestyle}. On \textit{Medical Conditions \& Risk}, no model surpasses the \textsc{Linear} reference, though \textsc{XGBoost}, \textsc{LSM-2}, and \textsc{WBM} come closest. Note that only the linear baseline includes demographics. Overall, \textsc{LSM-2} is the most consistent method, but \textsc{XGBoost} remains highly competitive.
Across age and sex subgroups, \textsc{LSM-2} retains the best Average Rank and skill, while the ordering of other methods varies (Appendix~\ref{app:subgroup_reporting}).

Note that the \textsc{WBM} learns participant week representations and is trained using participant weeks that meet a sufficient wear-time criterion. For a participant in the evaluation set, if the weekly wear-time criterion is met for zero weeks, the model falls back to the \textsc{Linear} baseline to produce a prediction; this occurred for $62.7\%$ of evaluation set outcome instances (no other methods used a fallback). In Appendix~\ref{app:wbm_lsm2_ablation} we provide additional evaluations that further compare \textsc{WBM} to other methods evaluated on the same subset of outcome instances without fallback and find that \textsc{LSM-2} still substantially outperforms \textsc{WBM}. We additionally trained a version of \textsc{LSM-2} that used hourly level information like \textsc{WBM} does (instead of minute-level) on the 
same pretraining participant cohort as WBM and find that it lowered the aggregate Skill Score of \textsc{LSM-2} from $+16.6$ to $+4.5$.

\subsection{Track 2A: Imputation (Generative)}

\bo{Methods.}
For the \textit{single-day imputation} task, we evaluate eleven imputation methods spanning statistical baselines (\textsc{Mean, Mode, Linear Interpolation, Last Observed Carry Forward (LOCF)}, and baseline methods based on diurnal temporal statistics), existing neural imputation models (\textsc{BRITS}~\citep{cao2018brits}, \textsc{DLinear}~\citep{zeng2023dlinear}, \textsc{FEDformer}~\citep{zhou2022fedformer}, \textsc{TimesNet}~\citep{wu2023timesnet}), and \textsc{LSM-2}, our reimplementation of Google's LSM-2 masked autoencoder~\citep{xu2025lsm}. For \textsc{BRITS}, \textsc{DLinear}, \textsc{FEDformer}, and \textsc{TimesNet}, we use implementations from the PyPOTS library~\citep{du2023pypots}. 
For \textit{long-context imputation}, we evaluate methods that can use more than one day of a user's historical data. We use baseline methods based on personalized statistics, a \textsc{7-day DLinear}, and \textsc{LSM-2-Sparse}, which pairs the frozen daily encoder with a sparse cross-day decoder (Appendix~\ref{sec:extended_history_methods}). For computational reasons we limited neural methods to 7 days of context. Methods are described in detail in Appendix~\ref{sec:baseline_methods}; hyperparameter search spaces and selected configurations are given in Appendix~\ref{app:imputation_models}.

\begin{table}[h!]
    \vspace{-2mm}
    \renewcommand{\arraystretch}{1.05}
    \centering
    \captionsetup{width=\textwidth}
    \caption{\textbf{Imputation Results.} We report Average Rank $R$, Aggregate Skill Score $S$ (in \%; $0=\TN{LOCF}$ reference), and Channel-Specific Skill Scores for the following channels: \textit{Activity, Physiology, Sleep, Workout}. Finally, we also report performance on all \textit{Semantic} masking approaches (see Appendix \ref{sec:imputation}). Single-day imputation method results are in the upper section of the table; long-context imputation method results ($\geq 7\times 1440$ time steps) are below. %
    Values are point estimates on the held-out test split; sub/superscripts give the $95\%$ bootstrap confidence interval ($1$k resamples).
    For worst-subgroup skill $S_{\TN{worst}}$, see Appendix~\ref{app:subgroup_reporting}.
    }
    \label{tab:imputation_main_results}
    \small
    \setlength{\tabcolsep}{1.5pt}
    \resizebox{\linewidth}{!}{%
    \begin{tabular}{l cccccccc}
    \toprule[1.5pt]
    \textbf{Method} & $R\downarrow$ & $S\uparrow$ & $S_{\TN{worst}}\uparrow$ & Activity\,$\uparrow$ & Physio.\,$\uparrow$ & Sleep\,$\uparrow$ & Workout\,$\uparrow$ & Semantic\,$\uparrow$ \\
    \midrule
    \multicolumn{9}{l}{\textbf{\emph{Single-day imputation}}} \\
    \hline
    \multicolumn{9}{l}{\cellcolor[HTML]{EFEFEF}\textit{Statistical Models}} \\
    Linear & \cellcolor{customblue!69}$6.4^{+0.1}_{-0.1}$ & \cellcolor{customblue!78}$+21.5^{+0.7}_{-1.2}$ & \cellcolor{customblue!80}\est{+20.0}{2.7}{2.9} & \cellcolor{customblue!49}$+4.5^{+0.5}_{-0.5}$ & \cellcolor{customblue!64}$+9.8^{+0.3}_{-0.4}$ & \cellcolor{customblue!94}$+62.6^{+0.4}_{-1.9}$ & \cellcolor{customblue!77}$+56.5^{+2.2}_{-3.0}$ & \cellcolor{customblue!83}$-0.8^{+1.9}_{-1.9}$ \\
    LOCF \textit{(reference)} & \cellcolor{customblue!54}$7.7^{+0.1}_{-0.1}$ & $0.0$ & $0.0$ & $0.0$ & $0.0$ & $0.0$ & $0.0$ & $0.0$ \\
    Temporal mode & \cellcolor{customblue!36}$9.3^{+0.1}_{-0.1}$ & \cellcolor{customblue!63}$-6.2^{+2.4}_{-2.3}$ & \cellcolor{customblue!63}\est{-12.2}{7.2}{7.0} & \cellcolor{customblue!100}$\mathbf{+46.5}^{+0.6}_{-0.6}$ & \cellcolor{customblue!46}$-0.7^{+1.1}_{-1.1}$ & \cellcolor{customblue!78}$-13.9^{+5.8}_{-5.0}$ & $-69.4^{+7.3}_{-7.6}$ & \cellcolor{customblue!77}$-11.8^{+4.2}_{-4.3}$ \\
    Mode & \cellcolor{customblue!29}$9.9^{+0.0}_{-0.1}$ & \cellcolor{customblue!51}$-27.3^{+2.7}_{-2.5}$ & \cellcolor{customblue!53}\est{-31.5}{7.5}{7.6} & \cellcolor{customblue!100}$\mathbf{+46.5}^{+0.6}_{-0.6}$ & \cellcolor{customblue!46}$-0.8^{+1.1}_{-1.0}$ & $-380.7^{+19.7}_{-18.1}$ & $-69.4^{+7.3}_{-7.6}$ & \cellcolor{customblue!77}$-12.0^{+4.2}_{-4.2}$ \\
    Temporal mean & \cellcolor{customblue!34}$9.5^{+0.1}_{-0.1}$ & \cellcolor{customblue!49}$-31.2^{+3.1}_{-3.2}$ & \cellcolor{customblue!47}\est{-41.7}{9.2}{10.6} & \cellcolor{customblue!7}$-30.7^{+2.1}_{-2.4}$ & \cellcolor{customblue!17}$-18.4^{+1.3}_{-1.4}$ & \cellcolor{customblue!94}$+59.9^{+2.9}_{-3.0}$ & \cellcolor{customblue!41}$-2.1^{+6.1}_{-6.0}$ & \cellcolor{customblue!32}$-93.0^{+6.7}_{-7.2}$ \\
    Mean & $12.4^{+0.0}_{-0.0}$ & $-119.7^{+4.7}_{-4.4}$ & \est{-133.1}{12.2}{13.3} & $-36.3^{+2.2}_{-2.4}$ & \cellcolor{customblue!5}$-25.5^{+1.3}_{-1.4}$ & $-380.7^{+19.7}_{-18.1}$ & $-69.4^{+7.3}_{-7.6}$ & $-149.8^{+8.5}_{-9.4}$ \\
    \hline
    \multicolumn{9}{l}{\cellcolor[HTML]{EFEFEF}\textit{Neural Models}} \\
    \textsc{LSM-2}~\citep{xu2025lsm} & \cellcolor{customblue!100}$\mathbf{3.6}^{+0.1}_{-0.0}$ & \cellcolor{customblue!100}$\mathbf{+61.4}^{+0.5}_{-1.2}$ & \cellcolor{customblue!100}\est{\mathbf{+59.4}}{2.3}{2.3} & \cellcolor{customblue!92}$+40.0^{+0.7}_{-0.8}$ & \cellcolor{customblue!100}$\mathbf{+31.4}^{+0.5}_{-0.5}$ & \cellcolor{customblue!100}$\mathbf{+90.0}^{+0.2}_{-0.8}$ & \cellcolor{customblue!100}$\mathbf{+94.9}^{+0.2}_{-0.6}$ & \cellcolor{customblue!100}$\mathbf{+30.2}^{+2.2}_{-2.4}$ \\
    BRITS~\citep{cao2018brits} & \cellcolor{customblue!60}$7.1^{+0.1}_{-0.1}$ & \cellcolor{customblue!70}$+6.8^{+1.8}_{-1.9}$ & \cellcolor{customblue!69}\est{+0.4}{5.7}{5.8} & \cellcolor{customblue!67}$+18.8^{+1.1}_{-1.1}$ & $-28.5^{+1.7}_{-1.7}$ & \cellcolor{customblue!89}$+39.0^{+2.1}_{-2.7}$ & \cellcolor{customblue!59}$+28.0^{+4.8}_{-5.1}$ & \cellcolor{customblue!80}$-5.7^{+3.3}_{-3.6}$ \\
    DLinear~\citep{zeng2023dlinear} & \cellcolor{customblue!57}$7.4^{+0.1}_{-0.1}$ & \cellcolor{customblue!63}$-5.7^{+2.1}_{-2.1}$ & \cellcolor{customblue!63}\est{-11.5}{5.9}{6.0} & \cellcolor{customblue!79}$+29.3^{+0.7}_{-0.8}$ & \cellcolor{customblue!39}$-5.1^{+1.0}_{-1.0}$ & \cellcolor{customblue!79}$-11.1^{+4.0}_{-3.7}$ & \cellcolor{customblue!78}$+58.2^{+3.0}_{-3.4}$ & \cellcolor{customblue!58}$-45.9^{+4.9}_{-5.0}$ \\
    FEDformer~\citep{zhou2022fedformer} & \cellcolor{customblue!23}$10.4^{+0.1}_{-0.1}$ & \cellcolor{customblue!36}$-53.7^{+3.3}_{-3.0}$ & \cellcolor{customblue!36}\est{-64.1}{8.9}{9.6} & \cellcolor{customblue!79}$+28.9^{+0.7}_{-0.8}$ & \cellcolor{customblue!23}$-14.6^{+1.1}_{-1.1}$ & \cellcolor{customblue!35}$-214.6^{+13.2}_{-12.2}$ & \cellcolor{customblue!10}$-53.7^{+7.6}_{-7.7}$ & \cellcolor{customblue!46}$-67.7^{+5.8}_{-6.3}$ \\
    TimesNet~\citep{wu2023timesnet} & \cellcolor{customblue!28}$10.0^{+0.1}_{-0.1}$ & \cellcolor{customblue!30}$-66.0^{+3.5}_{-3.5}$ & \cellcolor{customblue!28}\est{-79.1}{10.5}{10.3} & \cellcolor{customblue!55}$+9.6^{+1.2}_{-1.4}$ & \cellcolor{customblue!16}$-18.6^{+1.3}_{-1.3}$ & \cellcolor{customblue!35}$-216.2^{+14.0}_{-13.0}$ & \cellcolor{customblue!42}$+0.4^{+7.2}_{-7.3}$ & \cellcolor{customblue!26}$-103.2^{+6.7}_{-7.9}$ \\
    \midrule
    \multicolumn{9}{l}{\textbf{\emph{Long-context imputation ($\geq 7 \times 1440$ time steps)}}} \\
    \hline
    \multicolumn{9}{l}{\cellcolor[HTML]{EFEFEF}\textit{Statistical Models}} \\
    Personalized\ temp.\ mean & \cellcolor{customblue!45}$8.2^{+0.1}_{-0.1}$ & \cellcolor{customblue!59}$-7.7^{+2.8}_{-2.8}$ & \cellcolor{customblue!58}\est{-13.3}{4.8}{5.6} & \cellcolor{customblue!10}$+1.1^{+1.5}_{-1.6}$ & \cellcolor{customblue!14}$-5.9^{+1.1}_{-1.2}$ & \cellcolor{customblue!94}$+58.9^{+3.3}_{-3.6}$ & \cellcolor{customblue!66}$+15.7^{+6.3}_{-6.5}$ & \cellcolor{customblue!50}$-49.5^{+5.3}_{-5.4}$ \\
    Personalized\ mode & \cellcolor{customblue!28}$9.8^{+0.1}_{-0.1}$ & \cellcolor{customblue!49}$-26.1^{+2.6}_{-2.4}$ & \cellcolor{customblue!50}\est{-29.4}{6.4}{7.3} & \cellcolor{customblue!100}$\mathbf{+46.6}^{+0.6}_{-0.6}$ & \cellcolor{customblue!30}$+1.8^{+1.0}_{-1.0}$ & \cellcolor{customblue!10}$-383.1^{+19.7}_{-18.7}$ & \cellcolor{customblue!30}$-69.4^{+7.3}_{-7.6}$ & \cellcolor{customblue!73}$-10.5^{+4.1}_{-4.1}$ \\
    Personalized\ mean & $12.4^{+0.1}_{-0.1}$ & $-114.1^{+4.4}_{-4.3}$ & \est{-120.8}{11.0}{12.0} & $-4.1^{+1.6}_{-1.7}$ & $-12.6^{+1.2}_{-1.2}$ & $-437.7^{+20.9}_{-19.3}$ & $-140.0^{+10.4}_{-11.2}$ & $-132.5^{+8.2}_{-9.2}$ \\
    \hline
    \multicolumn{9}{l}{\cellcolor[HTML]{EFEFEF}\textit{Neural Models}} \\
    \textsc{LSM-2-Sparse} (7-day) & \cellcolor{customblue!100}$\mathbf{3.2}^{+0.1}_{-0.1}$ & \cellcolor{customblue!100}$\mathbf{+64.7}^{+0.4}_{-1.2}$ & \cellcolor{customblue!100}\est{\mathbf{+63.1}}{2.1}{2.1} & \cellcolor{customblue!89}$+41.0^{+0.7}_{-0.7}$ & \cellcolor{customblue!100}$\mathbf{+34.6}^{+0.5}_{-0.5}$ & \cellcolor{customblue!100}$\mathbf{+92.2}^{+0.0}_{-0.7}$ & \cellcolor{customblue!100}$\mathbf{+95.7}^{+0.1}_{-0.5}$ & \cellcolor{customblue!100}$\mathbf{+34.6}^{+2.1}_{-2.5}$ \\
    DLinear (7-day)~\citep{zeng2023dlinear} & \cellcolor{customblue!41}$8.6^{+0.1}_{-0.1}$ & \cellcolor{customblue!48}$-28.3^{+2.5}_{-2.6}$ & \cellcolor{customblue!46}\est{-36.7}{7.2}{7.1} & \cellcolor{customblue!47}$+19.9^{+0.9}_{-1.0}$ & \cellcolor{customblue!21}$-2.7^{+0.9}_{-0.9}$ & \cellcolor{customblue!75}$-40.0^{+5.3}_{-4.9}$ & \cellcolor{customblue!69}$+22.9^{+4.5}_{-5.6}$ & \cellcolor{customblue!38}$-69.5^{+5.8}_{-6.0}$ \\
    \bottomrule[1.5pt]
    \end{tabular}%
    }
    \vspace{-2mm}
\end{table}

\bo{Results.}
Table~\ref{tab:imputation_main_results} summarises aggregate performance across all masking scenarios. \textsc{LSM-2-Sparse} (7-day) attains the best Average Rank and aggregate Skill Score, followed by the daily \textsc{LSM-2}; the two also lead the Physiology, Sleep, and Workout channels and the Semantic scenarios. The main exception is the Activity channels, where the constant mode-based baselines score highest because some activity channels are mostly zero. Among the statistical baselines, only \textsc{Linear} interpolation exceeds the LOCF reference. The rest fall below it. Among the PyPOTS-trained neural baselines, only \textsc{BRITS} ($S{=}+6.8$) exceeds \textsc{LOCF}, while \textsc{DLinear}, \textsc{FEDformer}, and \textsc{TimesNet} fall below it; all remain far below \textsc{LSM-2}. Additional experiments with \textsc{DLinear} and \textsc{BRITS} show that training with random 10-minute patches substantially improves performance, suggesting that the training masking regime and objective contribute to neural baselines' poor performance  (Appendix~\ref{app:mask_alignment}).
Both \textsc{LSM-2} variants lead every age and sex subgroup, although females have the lowest skill for 14 of the 15 non-reference methods (Appendix~\ref{app:subgroup_reporting}).
Full results across masking scenarios are in Appendix~\ref{sec:imputation_results}.

\subsection{Track 2B: Forecasting (Generative)}

\bo{Models.}
We evaluate a set of statistical baselines (\textsc{Seasonal Naive}, \textsc{AutoARIMA}, \textsc{AutoETS}~\citep{hyndman2018forecasting}), which are applied independently to each channel and fitted separately for each participant using only their observed trajectory at test time (i.e., without access to training data). We also evaluate several deep learning sequence models trained from scratch (\textsc{DLinear}~\citep{zeng2023dlinear}, \textsc{MixLinear}~\citep{ma2024mixlinear}, and \textsc{SegRNN}~\citep{lin2025segrnn}), which are trained as global models across participants in the training set using multivariate inputs from all channels. Finally, we evaluate two time-series foundation models \textsc{Chronos-2}~\citep{ansari2025chronos} and \textsc{Toto 1.0}~\citep{cohen2025time}, which we chose due to their ability to incorporate multi-channel inputs. For the foundation models, we evaluate both their zero-shot performance and their performance after fine-tuning on the participants in the training set (Appendix \ref{sec:forecastingModels}). 

\begin{table}[h!]
\centering
\captionsetup{width=0.98\textwidth}
\caption{
\textbf{Forecasting Results.}
We report Average Rank $R$, Aggregate Skill Score $S$
(in \%; $0=\textsc{Seasonal Naive}$ reference),
and category-specific
Skill Scores for \textit{Activity}, \textit{Physiology}, \textit{Sleep},
and \textit{Workout}. FT denotes fine-tuned. Values are point estimates on the
held-out test split; subscripts and superscripts indicate the $95\%$ bootstrap
confidence interval ($1$k resamples).
For worst-subgroup skill $S_{\TN{worst}}$, see Appendix~\ref{app:subgroup_reporting}.
}
\label{tab:forecasting_grouped_model_summary}

\providecommand{\est}[3]{%
  \ensuremath{#1^{\scriptscriptstyle +#2}_{\scriptscriptstyle -#3}}%
}

\small
\renewcommand{\arraystretch}{1.05}
\setlength{\tabcolsep}{1.5pt}
\adjustbox{max width=\linewidth}{%
\begin{tabular}{l ccccccc}
\toprule[1.4pt]

\textbf{Method}
& \mbox{$R\,\downarrow$}
& \mbox{$S\,\uparrow$}
& \mbox{$S_{\TN{worst}}\uparrow$}
& \mbox{Activity~$\uparrow$}
& \mbox{Physio.~$\uparrow$}
& \mbox{Sleep~$\uparrow$}
& \mbox{Workout~$\uparrow$} \\

\midrule
\rowcolor[HTML]{EFEFEF}
\multicolumn{8}{l}{\textit{Statistical Methods}} \\

\textsc{Seasonal Naive}
& \est{7.72}{0.08}{0.08}
& $0.0$
& $0.0$
& $0.0$
& $0.0$
& $0.0$
& $0.0$ \\

\textsc{AutoARIMA}
& \cellcolor{customblue!2}\est{7.64}{0.07}{0.07}
& \cellcolor{customblue!16}\est{+5.9}{2.5}{2.6}
& \cellcolor{customblue!6}\est{+2.0}{4.5}{4.2}
& \est{-1.8}{1.0}{1.2}
& \cellcolor{customblue!33}\est{-9.0}{1.7}{1.6}
& \cellcolor{customblue!10}\est{+7.0}{5.1}{5.3}
& \cellcolor{customblue!81}\est{+24.0}{6.7}{6.7} \\

\textsc{AutoETS}
& \cellcolor{customblue!16}\est{7.07}{0.08}{0.08}
& \cellcolor{customblue!38}\est{+14.3}{2.5}{2.1}
& \cellcolor{customblue!30}\est{+10.2}{4.3}{4.3}
& \cellcolor{customblue!8}\est{+0.6}{1.0}{1.0}
& \est{-26.8}{2.3}{2.8}
& \cellcolor{customblue!53}\est{+37.6}{3.3}{3.3}
& \cellcolor{customblue!100}\est{\mathbf{+31.4}}{5.7}{5.6} \\

\specialrule{\lightrulewidth}{0pt}{0pt}
\rowcolor[HTML]{EFEFEF}
\multicolumn{8}{l}{\textit{Neural Models}} \\

\textsc{MixLinear}~\citep{ma2024mixlinear}
& \cellcolor{customblue!49}\est{5.68}{0.09}{0.10}
& \cellcolor{customblue!78}\est{+29.2}{1.8}{1.7}
& \cellcolor{customblue!78}\est{+27.1}{3.2}{2.9}
& \cellcolor{customblue!78}\est{+23.4}{1.1}{1.1}
& \cellcolor{customblue!75}\est{+13.4}{1.8}{1.9}
& \cellcolor{customblue!90}\est{+64.6}{1.8}{1.7}
& \est{-7.2}{9.2}{9.3} \\

\textsc{DLinear}~\citep{zeng2023dlinear}
& \cellcolor{customblue!74}\est{4.63}{0.10}{0.10}
& \cellcolor{customblue!96}\est{+35.9}{1.8}{1.9}
& \cellcolor{customblue!96}\est{+33.1}{3.0}{3.3}
& \cellcolor{customblue!83}\est{+25.0}{0.9}{1.0}
& \cellcolor{customblue!81}\est{+16.5}{1.7}{1.7}
& \cellcolor{customblue!100}\est{\mathbf{+71.5}}{1.6}{1.6}
& \cellcolor{customblue!33}\est{+5.5}{9.4}{10.7} \\

\textsc{SegRNN}~\citep{lin2025segrnn}
& \cellcolor{customblue!81}\est{4.35}{0.09}{0.08}
& \cellcolor{customblue!92}\est{+34.6}{1.4}{1.6}
& \cellcolor{customblue!97}\est{+33.4}{2.4}{2.6}
& \cellcolor{customblue!84}\est{+25.4}{1.0}{1.1}
& \cellcolor{customblue!89}\est{+20.8}{1.4}{1.4}
& \cellcolor{customblue!95}\est{+68.2}{1.8}{1.9}
& \cellcolor{customblue!25}\est{+2.5}{6.5}{8.7} \\

\specialrule{\lightrulewidth}{0pt}{0pt}
\rowcolor[HTML]{EFEFEF}
\multicolumn{8}{l}{\textit{Time-Series Foundation Models}} \\

\textsc{Toto}~\citep{cohen2025time}
& \cellcolor{customblue!54}\est{5.49}{0.09}{0.10}
& \cellcolor{customblue!71}\est{+26.8}{1.7}{1.7}
& \cellcolor{customblue!70}\est{+24.1}{3.5}{3.5}
& \cellcolor{customblue!95}\est{+29.2}{1.1}{1.2}
& \cellcolor{customblue!63}\est{+6.8}{1.9}{1.7}
& \cellcolor{customblue!71}\est{+50.5}{2.5}{2.4}
& \cellcolor{customblue!49}\est{+11.9}{6.1}{5.9} \\

\textsc{Toto} (FT)
& \cellcolor{customblue!73}\est{4.68}{0.09}{0.09}
& \cellcolor{customblue!82}\est{+30.9}{2.8}{2.2}
& \cellcolor{customblue!83}\est{+28.7}{3.4}{3.4}
& \cellcolor{customblue!96}\est{+29.5}{1.0}{1.1}
& \cellcolor{customblue!98}\est{+26.1}{0.8}{0.8}
& \cellcolor{customblue!64}\est{+46.1}{2.4}{2.5}
& \cellcolor{customblue!67}\est{+18.8}{11.5}{9.8} \\

\textsc{Chronos-2}~\citep{ansari2025chronos}
& \cellcolor{customblue!85}\est{4.17}{0.09}{0.09}
& \cellcolor{customblue!97}\est{+36.4}{2.0}{1.8}
& \cellcolor{customblue!98}\est{+33.8}{3.5}{3.3}
& \cellcolor{customblue!99}\est{+30.5}{1.0}{1.1}
& \cellcolor{customblue!99}\est{+26.5}{0.8}{0.8}
& \cellcolor{customblue!87}\est{+62.3}{2.2}{2.1}
& \cellcolor{customblue!57}\est{+14.8}{8.0}{8.1} \\

\textsc{Chronos-2} (FT)
& \cellcolor{customblue!100}\est{\mathbf{3.56}}{0.08}{0.09}
& \cellcolor{customblue!100}\est{\mathbf{+37.6}}{2.1}{1.9}
& \cellcolor{customblue!100}\est{\mathbf{+34.6}}{3.4}{3.4}
& \cellcolor{customblue!100}\est{\mathbf{+30.7}}{1.0}{1.1}
& \cellcolor{customblue!100}\est{\mathbf{+26.9}}{0.8}{0.8}
& \cellcolor{customblue!89}\est{+63.9}{2.1}{2.0}
& \cellcolor{customblue!63}\est{+17.0}{8.1}{8.7} \\

\bottomrule[1.4pt]
\end{tabular}%
}
\end{table}

\bo{Results.} Table~\ref{tab:forecasting_grouped_model_summary} shows that the fine-tuned \textsc{Chronos-2} achieves the strongest overall performance, obtaining the best average rank ($R=3.56$) and aggregate Skill Score ($S=+37.6$). %
The non-fine-tuned \textsc{Chronos-2} is the closest competitor, ranking second in both average rank ($R=4.17$) and aggregate Skill Score ($S=+36.4$), with the from-scratch \textsc{DLinear} close behind ($S=+35.9$, the strongest model trained from scratch). We also report skill scores across 4 sensor categories: \textit{Activity, Physiology, Sleep, Workout}. At the category level, \textsc{Chronos-2} (FT) performs best on Activity and Physiology and \textsc{DLinear} on Sleep, while the highest Workout skill comes from the statistical \textsc{AutoETS} baseline ($+31.4$), albeit with a poor overall score. See Appendix~\ref{sec:forecasting} for further details.
Fine-tuned \textsc{Chronos-2} retains the best average rank in every subgroup; \textsc{SegRNN} outranks zero-shot \textsc{Chronos-2} for age $<$40 and females (Appendix~\ref{app:subgroup_reporting}).

\section{Discussion}
\openmhc{} introduces, to our knowledge, the first broadly accessible, AI-ready wearable health dataset at a scale sufficient to support and democratize the development of foundation models on real-world consumer wearable device data. Through this work 
we have several key findings:

\bo{1. Self-supervised pretraining can improve performance, but its effectiveness depends on input resolution and the pretraining cohort.} 
We compare three classes of foundation models trained on \openmhc{}: \textsc{WBM}~\citep{erturk2025beyond}, trained with a contrastive objective over dense weekly segments; \textsc{LSM-2}~\citep{xu2025lsm}, trained with a reconstruction (generative) objective; and time-series FMs fine-tuned on OpenMHC for forecasting. Note that our \textsc{LSM-2} and \textsc{WBM} models are best-effort reimplementations and may not reproduce the performance of the original proprietary models. We observe substantial variation in downstream performance across these approaches (Table~\ref{tab:health_outcome_main_results}).
In particular, 
\textsc{LSM-2} achieves the best overall performance on prediction and imputation tasks. 
One possible explanation is that masked reconstruction is well suited to the sparse and structurally missing nature of these data. 
Our ablations (Appendix~\ref{app:wbm_lsm2_ablation}) show that restricting the pretraining cohort or reducing input resolution lowers \textsc{LSM-2}'s aggregate performance. However, a minute-level \textsc{LSM-2} trained and evaluated on the same cohorts used by \textsc{WBM} led to a higher aggregate Skill Score than \textsc{WBM}, which suggests the strengths of \textsc{LSM-2} is not due to using a larger pretraining cohort alone. The greatest drop in \textsc{LSM-2} came from switching from minute-level input resolution to hourly.

\bo{2. Well-crafted simple models remain highly competitive with, and often outperform, more complex architectures.}
Well-tuned \textsc{XGBoost} and \textsc{MultiRocket} models achieve the second-best aggregate Skill Score and Average Rank, respectively, across prediction tasks. Both outperform many neural models, trailing only \textsc{LSM-2}. This highlights the importance of including strong simple baselines when evaluating models on wearable health tasks but also highlights that some of the downstream medical and mental health conditions are hard to improve upon beyond simple baselines. 

\bo{3. Leveraging longitudinal data is a promising direction to improve performance.}
We find that incorporating longitudinal context provides a clear benefit: extending \textsc{LSM-2} with a 7-day sparse cross-day decoder yields a consistent improvement in imputation performance. This highlights that leveraging an extended personal history is a promising frontier for future wearable ML research.



\bo{Limitations.}
The cohort is predominantly male, white, and US-based among participants with available geographic information, with high attrition and data collected through the Apple HealthKit ecosystem. These characteristics limit how well the findings apply to other populations. Moreover, models may perform worse when applied to new populations without adaptation, and their performance may differ across demographic groups. Aggregate model rankings may therefore not transfer uniformly across demographic groups, geographic settings, or device platforms (Appendix~\ref{sec:limitations}). Age/sex subgroup scores provide partial diagnostics of uneven performance and do not mitigate demographic imbalance. Even within the HealthKit ecosystem, variation across device generations and software versions may affect sensor measurements and data availability, complicating comparisons across devices but also adding realism. 
Our analyses do not isolate the effects of changes in HealthKit data definitions across software versions or test whether models exploit device generation or collection period as shortcuts.

\bo{Future Work.} Looking ahead, \openmhc{} enables several concrete research directions. First, its scale makes it possible to study scaling laws for wearable foundation models. 
The dataset also enables controlled comparisons of model scaling and pretraining efficiency, as well as studies that vary architecture and pretraining objective independently to examine their effects (Appendix~\ref{app:wbm_lsm2_ablation}). Second, our results highlight the need to understand cross-device and cross-cohort transfer, particularly given differences in data quality and missingness patterns (e.g., extending to Fitbit datasets). Third, while the Gemini-family LLM and agentic probes evaluated here perform poorly (Appendix~\ref{app:gemini_baselines}), the benchmark provides a controlled setting for developing more effective interfaces between large language models and longitudinal health data. Finally, future releases of My Heart Counts will further expand the dataset and broaden its applicability. Broadening geographic representation is also an important direction for future data collection.


\begin{ack}
We would like to thank everyone who was involved in My Heart Counts who helped made this decade-long study possible, especially Steve Hershman, Anna Sherbina, Michael V McConnell, Ali Javed, Alexander Tolas, and the whole team involved at Sage Bionetworks and Apple, and in particular all the My Heart Counts participants who contributed their data. We would also like to thank Alia Crum for providing mindset questionnaires used in the My Heart Counts app.

\paragraph{Funding.} This work is supported by the Imperial BHF Research Excellence Award (RE/18/4/34215, RE/24/130023, RG/F/25/110167) FS/18/52/33808 (AL)  and NIHR Imperial Biomedical Research Centre. N.S. was supported by the Wu-Tsai Human Performance Alliance as a Postdoctoral Fellow and by Swiss National Science Foundation under Postdoc Mobility fellowship 210803. D.S.K. was supported by the Wu-Tsai Human Performance Alliance as a Clinician-Scientist Fellow, the Stanford Center for Digital Health as a Digital Health Scholar, the Pilot Grant from the Stanford Center for Digital Health, and NIH 1L30HL170306. D.S.K. is presently supported by NIH 9L30DK144879-02, the Robert A. Winn Excellence in Clinical Trials Career Development Award, the American Heart Association (AHA) Career Development Award (AHA 25CDA1436622), and the American Diabetes Association (ADA) Pathway to Stop Diabetes Initiator Award (7-25-INI-11). K.W.Z is supported by the UK Research and Innovation Engineering and Physical Sciences Research Council grant number UKRI4510. E.E. was supported by a doctoral scholarship from the German Academic Scholarship Foundation (Studienstiftung des deutschen Volkes). M.A. was supported by the Amazon AI PhD Fellowship and the Stanford HAI Graduate Fellowship. This project was also supported by Stanford HAI's Google Cloud Credits and directly by Google's GCP research credits program. Apple Watch Series 4 devices were provided via Apple Investigator Awards (AL and EAA).

\paragraph{Competing interests.} Outside this work, E.E. was employed by McKinsey \& Company (on educational leave since March 2024) and holds equity in Zeitgeist Labs. Outside this work, Euan Ashley reports founder roles at Personalis, Deepcell, Svexa, Saturnus Bio, and Swift Bio; founder-advisor roles at Candela and Parameter Health; an advisory role at Pacific Biosciences; non-executive directorships at AstraZeneca and Dexcom; and stock ownership in Personalis, Pacific Biosciences, and AstraZeneca. He also reports collaborative support in kind from Illumina, Pacific Biosciences, Oxford Nanopore, Cache, and Cellsonics; none of this support contributed to the present work.

\end{ack}









\bibliographystyle{plainnat}
\bibliography{references}

\clearpage
\appendix

\addtocontents{toc}{\protect\setcounter{tocdepth}{2}}
\tableofcontents

\newpage

\section{Research Implementation and Oversight} 
\label{sec:implementation}

\subsection{Limitations}
\label{sec:limitations}
A major limitation of our work is that despite the size and diversity, the demographics of our dataset and benchmark reflect the study populations that choose to use such digital health apps, and are thus skewed towards white, male, based around US metropolitan areas, and in their late thirties, which may limit generalizability to the broader public. Similar demographic imbalances have been reported in other large wearable cohorts. For example, the Apple Heart \& Movement Study cohort included 82,809 participants: 74\% male at birth and 72\% White, with a mean age at enrollment of 39.3 years~\citep{truslow2024understanding}. While we make subgroup performance transparent and visible on the live leaderboard---reporting worst-subgroup skill and allowing results to be viewed per demographic cohort (\url{https://myheartcounts-openmhc.hf.space})---this surfaces the imbalance rather than resolving it. Moreover, the dataset includes self-reported information on health, lifestyle, and well-being, which may be subject to recall and reporting biases.

Keeping participants engaged over time is another challenge in fully digital health research. A scoping review of 37 entirely remote studies reported a median study completion rate of 48\% among the 27 studies reporting this outcome, although definitions of completion varied~\citep{daniore2022enrollment}. Across eight remote studies involving 109,914 participants, including an earlier My Heart Counts cohort, more than half stopped completing active app tasks within the first week. The median duration of active participation across studies was 5.5 days, measured using active tasks during participants’ first 12 weeks rather than passive sensor-data collection. This analysis also found that racial and ethnic minority participants were disproportionately represented in the groups with the least engagement~\citep{pratap2020indicators}. Participation challenges have also been observed in passive wearable-data collection. In the Apple Heart \& Movement Study, 34\% of participants had stopped sharing Watch-derived Stand Hour data by one year after enrollment, with no further contributions of that type through the analysis cutoff. The authors also reported that dropout during the first year amplified the demographic imbalances present at recruitment~\citep{truslow2024understanding}. These studies show that other digital health datasets also face demographic imbalance and difficulties keeping participants engaged. This highlights the need to evaluate models across different populations and device settings.

The dataset reflects real-world conditions and evolved over time, which is a strength and limitation; the data are not perfect, and despite our efforts to clean them and include quality indicators such as coverage metrics, substantial data are missing for some users, and not all variables and channels are available uniformly. It should be noted that our sensitivity analyses (Appendix \ref{app:inclusion_ablations}), indicate that, e.g., data quality filtering introduces no meaningful bias. Finally, the models presented in our benchmark tasks are best-effort implementations that may not always be optimal due to resource constraints and should be viewed as a reference point, not a reflection of what will ultimately be possible.


\subsection{Compute Resources}
\label{sec:computing_resources}
Experiments were run on a mix of academic HPC clusters (Imperial College London, Stanford) and commercial cloud infrastructure, using a heterogeneous mix of NVIDIA GPUs (including A100, H100, L40S, and L4). Single-run evaluation wall-clock ranges from minutes for statistical baselines and linear probes on precomputed features, to up to roughly two days for slower neural method in imputation and forecasting evaluation. 
Reported experiments on academic clusters used approximately 400 GPU-hours, excluding fine-tuning of the foundation models in the forecasting track, which required up to 1000 additional GPU-hours. Tables~\ref{tab:compute_pretraining} and~\ref{tab:compute_inference} report selected component costs and do not provide a complete breakdown of these cumulative budgets. The large language model baseline evaluation was performed via Google's Gemini API and consumed no local GPU compute; the at-list-price equivalent on Vertex AI is approximately \$3,400 (actual cost zero through a courtesy research allocation). Hyperparameter sweeps and exploratory or preliminary experiments that did not contribute to reported results required approximately 2,200 additional GPU-hours on academic clusters.

\begin{table}[ht]
\centering
\footnotesize
\caption{\textbf{Pretraining compute.} Approximate costs for our
reimplementations, excluding hyperparameter search. Training data
counts describe available data. Hardware and input representations
differ, so these timings do not isolate model efficiency.}
\label{tab:compute_pretraining}
\begin{tabular}{llll}
\toprule
\textbf{Model} & \textbf{Training data} &
\textbf{Hardware} & \textbf{Time} \\
\midrule
\textsc{WBM} & 83{,}889 eligible weeks &
1$\times$80-GB GPU & ${\sim}$3 GPU-h / 20 epochs \\
\textsc{LSM-2} & 666{,}175 daily samples &
1$\times$A100 40-GB & ${\sim}$45 GPU-h / 25 epochs \\
\bottomrule
\end{tabular}
\end{table}

\begin{table}[ht]
\centering
\footnotesize
\caption{\textbf{Inference and benchmark evaluation compute.}
Approximate timings for selected components on the listed hardware.
These component costs are distinct from the cumulative experimental
budgets reported above. Hardware differs across rows, so timings
are not directly comparable.}
\label{tab:compute_inference}
\begin{tabular}{lll}
\toprule
\textbf{Component} & \textbf{Hardware} & \textbf{Time} \\
\midrule
\textsc{WBM} inference &
1$\times$L40S & ${\sim}$10 GPU-min \\
\textsc{LSM-2} inference &
1$\times$A100 40-GB & ${\sim}$6 GPU-h \\
Forecasting (\textsc{Chronos-2}) &
1$\times$A100 80-GB & ${\sim}$1 GPU-h \\
Forecasting (\textsc{Toto}) &
1$\times$A100 80-GB & ${\sim}$33 GPU-h \\
Forecasting (\textsc{MixLinear}/\textsc{DLinear}/\textsc{SegRNN}) &
1$\times$A100 80-GB & ${\sim}$6 GPU-min \\
Imputation (\textsc{LSM-2}) &
1$\times$RTX 3090 & ${\sim}$5 GPU-h \\
Imputation (\textsc{BRITS}) &
1$\times$RTX 3090 & ${\sim}$10 GPU-h \\
\bottomrule
\end{tabular}
\end{table}



\subsection{My Heart Counts App and Screenshots}
\label{subsec:app_details}
\begin{figure}[H]
    \centering
    \includegraphics[width=\textwidth]{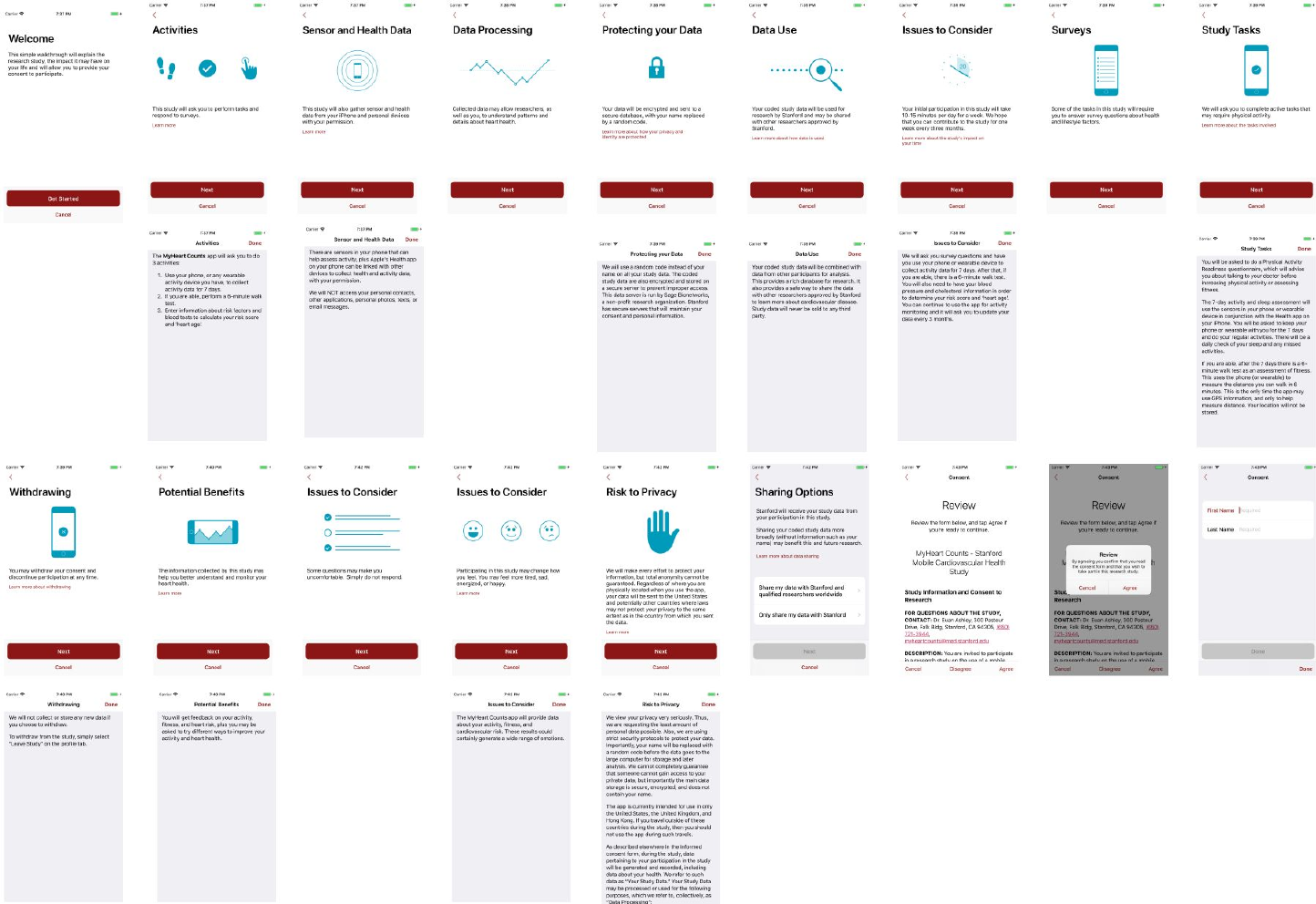} 
    \caption{
    Exemplar screenshots of consent screens as they appear in the US / HK version of the app.
    }
    \label{fig:app_screenshots_US_HK}
\end{figure}

The data for this study were collected via a custom-built mobile application called My Heart Counts (\url{https://myheartcounts.stanford.edu/}) built on Apple ResearchKit and released as one of the first flagship apps in collaboration with Apple. The application was designed to ensure user consent, data integrity, and user privacy. 
Informed consent was obtained digitally from all participants prior to data collection. Participants may withdraw at any point during the study without justification. Participants also have granular control on what individual metrics they are comfortable sharing (e.g., share step count but not heart rate) \citep{mcconnell2017feasibility}.
Below, in Figure \ref{fig:app_screenshots_US_HK} we show the actual consent screens as they appeared in the app. 

\subsection{Release Notes}
\label{app:release_notes}


\subsubsection{Data Access and Sharing}
\label{app:data_access}
\openmhc{} is available to all qualified researchers free of charge via Harvard Dataverse as two deposits: the full dataset (\url{https://doi.org/10.7910/DVN/XNBITM}) and the 5\% development subset \mbox{\openmhc{}-XS} (\url{https://doi.org/10.7910/DVN/RIKSK4}). Both deposits are distributed behind a click-through Data Use Agreement (DUA; the Stanford University School of Medicine Research Use Agreement for \openmhc{} and its subsets), approved by Stanford's Privacy Office, that every downloader accepts when registering for download.
The DUA grants use free of charge for personal, non-commercial research only and prohibits any commercial use, sale, or other monetization. Every user must register individually: users may not share the download link, and may not distribute, publish, or reproduce any portion of the data without prior written permission from the Stanford University School of Medicine. Derivative works that would give third parties access to the data are prohibited. Training generative AI models on \openmhc{} is permitted only if no \openmhc{} data can be obtained or recreated from them. The data are for non-clinical research use only and may not be used in diagnosis or patient care. Users must not attempt to re-identify any participant, must promptly report to Stanford's Privacy Office any information they believe might permit identification, and must immediately report any re-identification. Violations are grounds for immediate termination of use, and the School of Medicine may require that all copies be returned and none retained.

To ensure privacy, Stanford's Privacy Office and multiple researchers independently and manually reviewed all variables: the ZIP code/postcode field was removed, with state or territory and a rounded mean primary Rural-Urban Commuting Area (RUCA) code derived from three-digit US ZIP prefixes using the 2020 USDA ZIP–RUCA reference~\citep{usda2025ruca}. Fitzpatrick skin type was bucketed to three levels, race and ethnicity categories were coarsened within each field, and four rare outcomes were removed. 
As an additional voluntary safety measure, the dates in the dataset are shifted by a random offset drawn independently per participant and applied uniformly to every record of that participant, across both passive sensor streams and survey timestamps. The offset is constrained to a whole number of weeks. The shift is designed to make it harder to anchor a participant's timeline to absolute calendar time, and thus to link records against external date-stamped auxiliary information (a hospital admission, a news-reported event, a social-media post).

\subsubsection{Data, Model and Code Availability}
The official Github repository can be found under \url{https://github.com/AshleyLab/OpenMHC}, research code can be found under \url{https://github.com/NarayanSchuetz/OpenMHC}.
We make available an already fully imputed version of the dataset based on the best-performing \textsc{LSM-2-Sparse} imputation model.
Neural model checkpoints of methods used in this article are made available on Hugging Face (\url{https://huggingface.co/MyHeartCounts}).

\subsubsection{Third-Party Assets and Licenses}
\label{app:asset_licenses}
External implementations and pretrained checkpoints are listed below, with links to their upstream licenses or model cards. The model subsections cite the original methods and describe our adaptations.

\begin{center}
\small
\renewcommand{\arraystretch}{1.15}
\begin{tabularx}{\linewidth}{@{}>{\raggedright\arraybackslash}p{0.23\linewidth} >{\raggedright\arraybackslash}X >{\raggedright\arraybackslash}p{0.19\linewidth}@{}}
\toprule
\textbf{Asset} & \textbf{Use in the benchmark} & \textbf{License} \\
\midrule
\href{https://github.com/amazon-science/chronos-forecasting/blob/v2.2.2/LICENSE}{Chronos-2 code} and \href{https://huggingface.co/amazon/chronos-2}{checkpoint} &
Package 2.2.2; \texttt{amazon/chronos-2}; zero-shot forecasting, fine-tuning and feature extraction & Apache 2.0 \\
\href{https://github.com/DataDog/toto/blob/v0.1.4/LICENSE}{Toto code} and \href{https://huggingface.co/Datadog/Toto-Open-Base-1.0}{checkpoint} &
Package 0.1.4; \texttt{Datadog/Toto-Open-Base-1.0}; zero-shot forecasting, fine-tuning and feature extraction & Apache 2.0 \\
\href{https://github.com/WenjieDu/PyPOTS/blob/main/LICENSE}{PyPOTS} &
BRITS, GRU-D, DLinear, FEDformer, TimesNet, MixLinear and SegRNN implementations & BSD 3-Clause \\
\href{https://github.com/WenjieDu/PyGrinder/blob/main/LICENSE}{PyGrinder} &
Missing-value masking utilities & BSD 3-Clause \\
\href{https://github.com/sktime/sktime/blob/main/LICENSE}{sktime} &
MultiRocket and the AutoARIMA/AutoETS forecasting interfaces & BSD 3-Clause \\
\href{https://github.com/dmlc/xgboost/blob/master/LICENSE}{XGBoost} &
Prediction models using handcrafted features & Apache 2.0 \\
\href{https://github.com/GAA-UAM/scikit-fda/blob/develop/LICENSE.txt}{scikit-fda} &
Functional principal components and curve features & BSD 3-Clause \\
\href{https://github.com/Google-Health/consumer-health-research/blob/main/personal_health_agent/LICENSE.txt}{PHA framework} &
Data Science Agent and prompts, with benchmark-specific adaptations & Apache 2.0 \\
\href{https://github.com/googleapis/python-genai/blob/v1.73.1/LICENSE}{google-genai} &
SDK version 1.73.1 for Gemini API access & Apache 2.0 \\
\bottomrule
\end{tabularx}
\end{center}

The Chronos-2 and Toto integrations adapt upstream code. Our \textsc{LSM-2} and \textsc{WBM} reimplementations are trained on \openmhc{} and do not use the original proprietary checkpoints.

Shared scientific libraries include \href{https://github.com/pytorch/pytorch/blob/main/LICENSE}{PyTorch}, \href{https://github.com/scikit-learn/scikit-learn/blob/main/COPYING}{scikit-learn}, \href{https://github.com/statsmodels/statsmodels/blob/main/LICENSE.txt}{statsmodels}, \href{https://github.com/numpy/numpy/blob/main/LICENSE.txt}{NumPy}, \href{https://github.com/scipy/scipy/blob/main/LICENSE.txt}{SciPy}, and \href{https://github.com/pandas-dev/pandas/blob/main/LICENSE}{pandas} (each BSD 3-Clause); \href{https://github.com/Lightning-AI/pytorch-lightning/blob/master/LICENSE}{Lightning}, \href{https://github.com/huggingface/peft/blob/main/LICENSE}{PEFT}, and \href{https://github.com/huggingface/datasets/blob/main/LICENSE}{Hugging Face Datasets} (each Apache 2.0); and \href{https://github.com/alkaline-ml/pmdarima/blob/master/LICENSE}{pmdarima} and \href{https://github.com/pola-rs/polars/blob/main/LICENSE}{Polars} (each MIT). AutoARIMA uses pmdarima, while AutoETS uses statsmodels.

Gemini API access is governed by the \href{https://developers.google.com/terms}{Google APIs Terms of Service} and \href{https://ai.google.dev/gemini-api/terms}{Gemini API Additional Terms of Service}; the SDK's Apache 2.0 license covers the SDK code.

The original \href{https://github.com/ResearchKit/MyHeartCounts/blob/master/LICENSE}{My Heart Counts app}, \href{https://github.com/ResearchKit/AppCore/blob/master/LICENSE}{AppCore}, and \href{https://github.com/ResearchKit/ResearchKit/blob/main/LICENSE}{ResearchKit} use BSD 3-Clause-style licenses with a trademark reservation. The \href{https://catalog.data.gov/dataset/rural-urban-commuting-area-codes}{federal RUCA catalogue} lists CC0 1.0; the reference is cited in Appendix~\ref{app:data_access}. Participant data remain under the \openmhc{} DUA.

\section{Evaluation Metrics}
\label{sec:scoring_methodology}
Evaluating models across diverse tasks in wearable and mobile health research presents a significant challenge due to the varied nature of the data and prediction targets. For instance, forecasting and imputation tasks span multiple sensor channels (e.g., heart rate, step count) with vastly different units, scales, and variances. Similarly, downstream prediction tasks encompass binary classification, ordinal classification, and regression, each evaluated using different metrics (e.g., AUROC, PRAUC, Pearson $R$). Averaging raw metrics across these tasks is mathematically unsound.

To address this, we adopt a unified evaluation methodology inspired by large-scale time series benchmarks \citep{hyndman2018forecasting,shchur2025fevbench}, utilizing a \textbf{Skill Score} and an \textbf{Average Rank} metric. These approaches aggregate performance reliably across heterogeneous tasks.

\subsection{Skill Score}
\label{app:skill_score}

Recall from \eqref{eqn:skillscore} that the Skill Score quantifies the average relative improvement of a model over a fixed reference or baseline model. We now formally define the clipping function:
\begin{align*}
    \text{clip}\left( \frac{E_{r,j}}{E_{r,b}}, \ell, u \right)
    = \min \left( \max \left( \frac{E_{r,j}}{E_{r,b}}, \ell \right), u \right)
\end{align*}
Specifically, we use lower clipping value $\ell = 0.01$ and an upper clipping value of $u = 100$.
We adapt the definition of the ``error'' term $E$ depending on the domain:
\begin{itemize}[leftmargin=*]
    \item \bo{Forecasting and Imputation:} For tasks predicting continuous sensor values, $E$ is a standard error metric such as Mean Absolute Error (MAE) or Mean Squared Error (MSE). The baseline $\beta$ is a simple heuristic, such as a Seasonal Naive forecaster (for forecasting) or LOCF (for imputation).
    \item \bo{Prediction Tasks:} For our prediction tasks we are using metrics that better models maximize: AUPRC for binary outcomes, Spearman's $\rho$ for ordinal outcomes, and Pearson's $r$ for continuous outcomes. Each has a maximum attainable value of $1$. To apply the Skill Score, we convert each metric into an error-transformed score measuring distance from the optimum:
    \begin{equation}
        E = 1 - \mathrm{Metric}.
    \end{equation}
    For example, an AUPRC of $0.85$ corresponds to an error-transformed score of $0.15$. For health outcome prediction, the baseline $\beta$ is the \textsc{Linear} model.
\end{itemize}

\bo{Uncertainty quantification.}
We report uncertainty using $95\%$ bootstrap confidence intervals (CIs). The reported value for every metric is its point estimate, computed on the original held-out test split. CIs are estimated by participant-level bootstrap resampling of the test split with $1{,}000$ replicates: participants are sampled with replacement using a single shared, seeded draw matrix per split, and each metric (Skill Score, Average Rank) is recomputed on every replicate. We report the percentile interval---the $2.5$th and $97.5$th percentiles of the bootstrap distribution.

\subsection{Average Rank}

While the Skill Score measures the magnitude of improvement, the \textbf{Average Rank} measures consistency across tasks. For each task $r$, all evaluated models are ranked from $1$ (best) to $N$ (worst) based on their respective metrics. The Average Rank for model $j$ is the arithmetic mean of its ranks across all tasks.
The Average Rank uses neither a reference model nor clipping. Therefore, comparing the model orderings produced by Average Rank and Skill Score shows whether the choice of aggregation method changes which models perform best. The Average Rank is entirely invariant to the scale of the metric and robust to outliers, providing a reliable secondary measure to confirm that a model's high Skill Score is due to consistent performance across the board rather than massive gains on a small subset of tasks.
We report a 95\% confidence interval constructed using same bootstrap approach used for the Skill Score (Appendix \ref{app:skill_score}).

\subsection{Aggregation Sensitivity}
\label{app:aggregation_sensitivity}
We examine whether model orderings change under four alternative evaluation choices. These include varying the reference model and clipping bounds, comparing Average Rank with aggregate Skill Score, replacing AUPRC with AUROC for binary prediction tasks, and excluding binary prediction tasks with low test split prevalence. No models were retrained for these analyses. 

\bo{Reference model and clipping bounds.}
We recompute aggregate Skill Scores using three reference models per track: \textsc{Linear}, \textsc{XGBoost} and \textsc{MultiRocket} for prediction; \textsc{LOCF}, \textsc{Linear} and \textsc{Mean} for imputation; and \textsc{Seasonal Naive}, \textsc{AutoARIMA} and \textsc{AutoETS} for forecasting.
Each reference is combined with clipping intervals $[0.001,1000]$, $[0.01,100]$ and $[0.1,10]$, which gives nine combinations per track and 27 in total.
Every combination preserves the best-performing model and the top-three models. At the original bounds $[0.01,100]$, the proportions of clipped error ratios are $0.0\%$, $1.8\%$ and $0.17\%$ for prediction, imputation and forecasting, respectively.

\bo{Classification metric.}
We replace AUPRC with AUROC on all 13 binary Track~1 tasks, retaining the original metrics for the other tasks and the same aggregation procedure. The Average Rank ordering is unchanged ($\rho=1.000$), while the aggregate Skill Score ordering remains similar ($\rho=0.929$). Both metrics preserve the best-performing model and the set of top-three models, although individual aggregate Skill Scores differ from their original AUPRC-based values by up to $8.45$ points.

\bo{Rare-Outcome Exclusion.}
We exclude the four binary Track~1 tasks with test split prevalence below $2\%$ (Congenital Heart Disease, Heart Failure/CHF, Pulmonary Hypertension and Vascular Disease). Both the aggregate Skill Score and Average Rank orderings remain unchanged, with aggregate Skill Scores changing by at most $0.889$ points. We also test prevalence thresholds of $0\%$, $1\%$, $2\%$, $3\%$, $5\%$ and $10\%$, excluding up to eight binary tasks.
The aggregate Skill Score ordering is unchanged at every tested threshold. The Average Rank ordering changes at $5\%$ and $10\%$, but the best-performing model and the top-three models are preserved.


\subsection{Subgroup Reporting}
\label{app:subgroup_reporting}

Tables~\ref{tab:subgroup_track1}--\ref{tab:subgroup_track2b} report Average Rank $R$ and aggregate Skill Score $S$ within four key sugroups, together with $S_{\TN{worst}}$, the minimum Skill Score over the four subgroups. They use the same method sets as Tables~\ref{tab:health_outcome_main_results}--\ref{tab:forecasting_grouped_model_summary}. 
Each participant has one age group across tracks ($<40$ and $\geq 40$), based on enrollment year minus reported birth year; sex (male or female) uses the latest reported BiologicalSex label. 
Missing age or sex excludes a participant only from that attribute's groups, while retaining them in All and in any known group for the other attribute. 

\bo{Subgroup findings.}
For prediction (Table~\ref{tab:subgroup_track1}) and imputation (Table~\ref{tab:subgroup_track2a}), \textsc{LSM-2} has the best Average Rank and Skill Score on the entire population, as well as for each of the  four subgroups. For forecasting (Table~\ref{tab:subgroup_track2b}), fine-tuned \textsc{Chronos-2} has the best Average Rank and Skill score on the entire population, as well as on all four sub-groups---except on the female subgroup, for which \textsc{SegRNN} performs the best in terms of Skill Score.
Overall, for all three tasks, the rank ordering and skill score of models on the entire population is similar to their on each of the four subgroups.
\begin{table}[H]
\centering
\captionsetup{width=\textwidth}
\caption{\textbf{Prediction tasks (Track 1) stratified by age and sex.} Average Rank $R$ ($\downarrow$) and Skill Score $S$ ($\uparrow$, in \%) are recomputed within each cohort; \textsc{Linear} is the within-cohort reference. Sub/superscripts are offsets from the point estimate to the $95\%$ percentile interval ($1{,}000$ participant resamples); $S_{\TN{worst}}$ uses the projection interval described in the text. All gives point estimates only. Darker blue indicates better performance within each column and metric; bold marks the best point estimate. FT denotes fine-tuned; $^{*}$ denotes a reimplementation. $n$ counts evaluated participants, with smaller task-specific samples and undefined tasks omitted as described above.}
\label{tab:subgroup_track1}
\providecommand{\est}[3]{\ensuremath{#1^{\scriptscriptstyle +#2}_{\scriptscriptstyle -#3}}}
\small
\renewcommand{\arraystretch}{1.05}
\setlength{\tabcolsep}{3pt}
\resizebox{\linewidth}{!}{%
\begin{tabular}{lc c cccc c}
\toprule[1.5pt]
\textbf{Method} & & \makecell{All\\$n=1{,}762$} & \makecell{Age $<$40\\$n=895$} & \makecell{Age $\geq$40\\$n=867$} & \makecell{Female\\$n=343$} & \makecell{Male\\$n=1{,}331$} & $S_{\TN{worst}}\uparrow$ \\
\midrule
\multicolumn{8}{l}{\cellcolor[HTML]{EFEFEF}\textit{Statistical Models}} \\
\multirow{2}{*}{\textsc{MultiRocket}} & $R\downarrow$ & \cellcolor{customblue!71}$3.53$ & \cellcolor{customblue!49}\est{4.01}{0.70}{0.53} & \cellcolor{customblue!65}\est{3.80}{0.65}{0.51} & \cellcolor{customblue!50}\est{4.00}{0.59}{0.49} & \cellcolor{customblue!52}\est{3.98}{0.72}{0.48} & \cellcolor{customblue!59} \\
 & $S\uparrow$ & \cellcolor{customblue!62}$+8.1$ & \cellcolor{customblue!59}\est{+6.9}{3.5}{3.1} & \cellcolor{customblue!66}\est{+7.8}{3.4}{3.8} & \cellcolor{customblue!65}\est{+12.8}{8.1}{6.4} & \cellcolor{customblue!45}\est{+5.6}{2.3}{3.0} & \cellcolor{customblue!59}\multirow{-2}{*}{\est{+5.6}{3.0}{3.7}} \\
\addlinespace[2pt]
\multirow{2}{*}{\textsc{XGBoost}} & $R\downarrow$ & \cellcolor{customblue!68}$3.65$ & \cellcolor{customblue!61}\est{3.56}{0.83}{0.46} & \cellcolor{customblue!61}\est{3.94}{0.36}{0.67} & \cellcolor{customblue!72}\est{3.29}{0.88}{0.32} & \cellcolor{customblue!63}\est{3.60}{0.46}{0.42} & \cellcolor{customblue!77} \\
 & $S\uparrow$ & \cellcolor{customblue!81}$+12.3$ & \cellcolor{customblue!73}\est{+9.6}{4.1}{3.6} & \cellcolor{customblue!83}\est{+11.7}{3.3}{3.8} & \cellcolor{customblue!74}\est{+14.9}{8.4}{5.7} & \cellcolor{customblue!77}\est{+10.9}{2.2}{3.0} & \cellcolor{customblue!77}\multirow{-2}{*}{\est{+9.6}{4.8}{4.4}} \\
\addlinespace[2pt]
\multirow{2}{*}{\textsc{Linear}} & $R\downarrow$ & \cellcolor{customblue!40}$4.71$ & \cellcolor{customblue!10}\est{5.39}{0.58}{0.72} & \cellcolor{customblue!48}\est{4.45}{0.77}{0.33} & \cellcolor{customblue!14}\est{5.20}{0.55}{0.53} & \cellcolor{customblue!22}\est{4.94}{0.60}{0.28} &  \\
 & $S\uparrow$ & $0.0$ & $0.0$ & $0.0$ & $0.0$ & $0.0$ & \multirow{-2}{*}{$0.0$} \\
\hline
\multicolumn{8}{l}{\cellcolor[HTML]{EFEFEF}\textit{Supervised Neural Models}} \\
\multirow{2}{*}{\textsc{GRU-D}} & $R\downarrow$ & \cellcolor{customblue!27}$5.19$ & \cellcolor{customblue!9}\est{5.44}{0.26}{1.06} & \cellcolor{customblue!33}\est{5.07}{0.45}{0.75} & \cellcolor{customblue!28}\est{4.73}{0.69}{0.56} & \est{5.68}{0.34}{0.65} & \cellcolor{customblue!36} \\
 & $S\uparrow$ & \cellcolor{customblue!42}$+3.6$ & \cellcolor{customblue!44}\est{+3.8}{4.2}{3.5} & \cellcolor{customblue!47}\est{+3.4}{3.3}{4.4} & \cellcolor{customblue!34}\est{+5.3}{9.1}{7.1} & \cellcolor{customblue!15}\est{+0.6}{2.7}{2.9} & \cellcolor{customblue!36}\multirow{-2}{*}{\est{+0.6}{3.4}{3.9}} \\
\hline
\multicolumn{8}{l}{\cellcolor[HTML]{EFEFEF}\textit{Time-Series Foundation Models}} \\
\multirow{2}{*}{\textsc{Chronos-2} (FT)} & $R\downarrow$ & \cellcolor{customblue!9}$5.88$ & \est{5.77}{0.46}{0.77} & \cellcolor{customblue!22}\est{5.49}{0.59}{0.60} & \cellcolor{customblue!7}\est{5.40}{0.58}{0.73} & \est{5.68}{0.36}{0.87} & \cellcolor{customblue!15} \\
 & $S\uparrow$ & \cellcolor{customblue!10}$-3.3$ & \cellcolor{customblue!5}\est{-4.1}{4.1}{4.5} & \cellcolor{customblue!16}\est{-3.7}{3.3}{4.4} & \est{-2.7}{5.2}{6.7} & \est{-1.9}{2.7}{3.4} & \cellcolor{customblue!15}\multirow{-2}{*}{\est{-4.1}{4.5}{7.3}} \\
\addlinespace[2pt]
\multirow{2}{*}{\textsc{Toto} (FT)} & $R\downarrow$ & $6.23$ & \cellcolor{customblue!8}\est{5.49}{0.61}{0.63} & \est{6.37}{0.25}{0.77} & \cellcolor{customblue!8}\est{5.38}{0.47}{0.79} & \cellcolor{customblue!1}\est{5.66}{0.48}{0.70} &  \\
 & $S\uparrow$ & $-5.5$ & \est{-5.1}{4.6}{4.7} & \est{-7.5}{3.4}{4.8} & \cellcolor{customblue!1}\est{-2.4}{4.7}{6.3} & \cellcolor{customblue!4}\est{-1.2}{2.6}{3.3} & \multirow{-2}{*}{\est{-7.5}{4.4}{6.0}} \\
\hline
\multicolumn{8}{l}{\cellcolor[HTML]{EFEFEF}\textit{Wearable Foundation Models}} \\
\multirow{2}{*}{\textsc{LSM-2}$^{*}$} & $R\downarrow$ & \cellcolor{customblue!100}$\mathbf{2.43}$ & \cellcolor{customblue!100}\est{\mathbf{2.15}}{0.91}{0.22} & \cellcolor{customblue!100}\est{\mathbf{2.41}}{0.75}{0.21} & \cellcolor{customblue!100}\est{\mathbf{2.37}}{0.86}{0.26} & \cellcolor{customblue!100}\est{\mathbf{2.39}}{0.58}{0.23} & \cellcolor{customblue!100} \\
 & $S\uparrow$ & \cellcolor{customblue!100}$\mathbf{+16.5}$ & \cellcolor{customblue!100}\est{\mathbf{+15.2}}{4.0}{3.2} & \cellcolor{customblue!100}\est{\mathbf{+15.8}}{3.0}{3.6} & \cellcolor{customblue!100}\est{\mathbf{+21.2}}{9.2}{5.4} & \cellcolor{customblue!100}\est{\mathbf{+14.8}}{2.2}{2.9} & \cellcolor{customblue!100}\multirow{-2}{*}{\est{\mathbf{+14.8}}{2.8}{4.0}} \\
\addlinespace[2pt]
\multirow{2}{*}{\textsc{WBM}$^{*}$} & $R\downarrow$ & \cellcolor{customblue!48}$4.40$ & \cellcolor{customblue!43}\est{4.19}{0.64}{0.57} & \cellcolor{customblue!48}\est{4.47}{0.59}{0.58} & \est{5.64}{0.34}{1.09} & \cellcolor{customblue!49}\est{4.07}{0.53}{0.50} & \cellcolor{customblue!44} \\
 & $S\uparrow$ & \cellcolor{customblue!49}$+5.3$ & \cellcolor{customblue!54}\est{+5.8}{2.6}{2.0} & \cellcolor{customblue!47}\est{+3.5}{2.8}{3.0} & \cellcolor{customblue!21}\est{+2.3}{4.5}{4.8} & \cellcolor{customblue!39}\est{+4.7}{1.9}{2.1} & \cellcolor{customblue!44}\multirow{-2}{*}{\est{+2.3}{4.7}{6.3}} \\
\bottomrule[1.5pt]
\end{tabular}}
\end{table}

\begin{table}[H]
\centering
\captionsetup{width=\textwidth}
\caption{\textbf{Imputation (Track 2A) stratified by age and sex.} Average Rank $R$ ($\downarrow$) and Skill Score $S$ ($\uparrow$, in \%) are recomputed within each cohort; \textsc{LOCF} is the within-cohort reference. Sub/superscripts are offsets from the point estimate to the $95\%$ percentile interval ($1{,}000$ participant resamples); $S_{\TN{worst}}$ uses the projection interval described in the text. All gives point estimates only. Darker blue indicates better performance within each column and metric; bold marks the best point estimate within each context section. $n$ counts evaluated participants, with smaller task-specific samples and undefined tasks omitted as described above.}
\label{tab:subgroup_track2a}
\providecommand{\est}[3]{\ensuremath{#1^{\scriptscriptstyle +#2}_{\scriptscriptstyle -#3}}}
\small
\renewcommand{\arraystretch}{1.05}
\setlength{\tabcolsep}{3pt}
\resizebox{\linewidth}{!}{%
\begin{tabular}{lc c cccc c}
\toprule[1.5pt]
\textbf{Method} & & \makecell{All\\$n=1{,}777$} & \makecell{Age $<$40\\$n=904$} & \makecell{Age $\geq$40\\$n=873$} & \makecell{Female\\$n=343$} & \makecell{Male\\$n=1{,}331$} & $S_{\TN{worst}}\uparrow$ \\
\midrule
\multicolumn{8}{l}{\textbf{\emph{Single-day imputation}}} \\
\hline
\multicolumn{8}{l}{\cellcolor[HTML]{EFEFEF}\textit{Statistical Models}} \\
\multirow{2}{*}{Linear} & $R\downarrow$ & \cellcolor{customblue!69}$6.4$ & \cellcolor{customblue!69}\est{6.3}{0.1}{0.1} & \cellcolor{customblue!68}\est{6.4}{0.1}{0.1} & \cellcolor{customblue!69}\est{6.4}{0.1}{0.1} & \cellcolor{customblue!68}\est{6.4}{0.1}{0.1} & \cellcolor{customblue!80} \\
 & $S\uparrow$ & \cellcolor{customblue!78}$+21.5$ & \cellcolor{customblue!78}\est{+21.9}{1.6}{1.7} & \cellcolor{customblue!78}\est{+21.2}{1.3}{1.4} & \cellcolor{customblue!80}\est{+20.0}{2.1}{2.3} & \cellcolor{customblue!77}\est{+21.8}{1.2}{1.2} & \cellcolor{customblue!80}\multirow{-2}{*}{\est{+20.0}{2.7}{2.9}} \\
\addlinespace[2pt]
\multirow{2}{*}{LOCF} & $R\downarrow$ & \cellcolor{customblue!54}$7.7$ & \cellcolor{customblue!55}\est{7.6}{0.1}{0.1} & \cellcolor{customblue!53}\est{7.8}{0.1}{0.1} & \cellcolor{customblue!55}\est{7.6}{0.2}{0.1} & \cellcolor{customblue!53}\est{7.8}{0.1}{0.1} &  \\
 & $S\uparrow$ & $0.0$ & $0.0$ & $0.0$ & $0.0$ & $0.0$ & \multirow{-2}{*}{$0.0$} \\
\addlinespace[2pt]
\multirow{2}{*}{Temporal mode} & $R\downarrow$ & \cellcolor{customblue!36}$9.3$ & \cellcolor{customblue!35}\est{9.4}{0.1}{0.1} & \cellcolor{customblue!37}\est{9.2}{0.1}{0.1} & \cellcolor{customblue!33}\est{9.5}{0.1}{0.1} & \cellcolor{customblue!37}\est{9.2}{0.1}{0.1} & \cellcolor{customblue!63} \\
 & $S\uparrow$ & \cellcolor{customblue!63}$-6.2$ & \cellcolor{customblue!63}\est{-6.0}{3.2}{3.4} & \cellcolor{customblue!62}\est{-6.8}{3.0}{3.4} & \cellcolor{customblue!63}\est{-12.2}{5.7}{5.8} & \cellcolor{customblue!63}\est{-3.4}{2.5}{2.6} & \cellcolor{customblue!63}\multirow{-2}{*}{\est{-12.2}{7.2}{7.0}} \\
\addlinespace[2pt]
\multirow{2}{*}{Mode} & $R\downarrow$ & \cellcolor{customblue!29}$9.9$ & \cellcolor{customblue!28}\est{9.9}{0.1}{0.1} & \cellcolor{customblue!30}\est{9.8}{0.1}{0.1} & \cellcolor{customblue!29}\est{9.9}{0.1}{0.1} & \cellcolor{customblue!30}\est{9.8}{0.1}{0.1} & \cellcolor{customblue!53} \\
 & $S\uparrow$ & \cellcolor{customblue!51}$-27.3$ & \cellcolor{customblue!52}\est{-26.6}{3.3}{3.6} & \cellcolor{customblue!50}\est{-28.2}{3.6}{3.8} & \cellcolor{customblue!53}\est{-31.5}{6.4}{6.4} & \cellcolor{customblue!51}\est{-25.2}{2.7}{3.0} & \cellcolor{customblue!53}\multirow{-2}{*}{\est{-31.5}{7.5}{7.6}} \\
\addlinespace[2pt]
\multirow{2}{*}{Temporal mean} & $R\downarrow$ & \cellcolor{customblue!34}$9.5$ & \cellcolor{customblue!33}\est{9.5}{0.1}{0.1} & \cellcolor{customblue!34}\est{9.4}{0.1}{0.1} & \cellcolor{customblue!32}\est{9.6}{0.2}{0.2} & \cellcolor{customblue!35}\est{9.4}{0.1}{0.1} & \cellcolor{customblue!47} \\
 & $S\uparrow$ & \cellcolor{customblue!49}$-31.2$ & \cellcolor{customblue!48}\est{-34.0}{4.6}{4.6} & \cellcolor{customblue!49}\est{-29.8}{3.7}{4.6} & \cellcolor{customblue!47}\est{-41.7}{7.4}{8.5} & \cellcolor{customblue!50}\est{-27.5}{3.3}{3.5} & \cellcolor{customblue!47}\multirow{-2}{*}{\est{-41.7}{9.2}{10.6}} \\
\addlinespace[2pt]
\multirow{2}{*}{Mean} & $R\downarrow$ & $12.4$ & \est{12.4}{0.1}{0.1} & \est{12.5}{0.1}{0.1} & \est{12.4}{0.1}{0.1} & \est{12.5}{0.0}{0.0} &  \\
 & $S\uparrow$ & $-119.7$ & \est{-121.5}{6.8}{6.5} & \est{-119.1}{5.8}{6.6} & \est{-133.1}{10.5}{10.4} & \est{-116.5}{5.0}{5.2} & \multirow{-2}{*}{\est{-133.1}{12.2}{13.3}} \\
\hline
\multicolumn{8}{l}{\cellcolor[HTML]{EFEFEF}\textit{Neural Models}} \\
\multirow{2}{*}{\textsc{LSM-2}} & $R\downarrow$ & \cellcolor{customblue!100}$\mathbf{3.6}$ & \cellcolor{customblue!100}\est{\mathbf{3.7}}{0.1}{0.1} & \cellcolor{customblue!100}\est{\mathbf{3.6}}{0.1}{0.1} & \cellcolor{customblue!100}\est{\mathbf{3.6}}{0.1}{0.1} & \cellcolor{customblue!100}\est{\mathbf{3.6}}{0.1}{0.1} & \cellcolor{customblue!100} \\
 & $S\uparrow$ & \cellcolor{customblue!100}$\mathbf{+61.4}$ & \cellcolor{customblue!100}\est{\mathbf{+61.3}}{1.2}{1.2} & \cellcolor{customblue!100}\est{\mathbf{+61.4}}{1.1}{1.2} & \cellcolor{customblue!100}\est{\mathbf{+59.4}}{1.9}{1.9} & \cellcolor{customblue!100}\est{\mathbf{+62.1}}{1.0}{1.0} & \cellcolor{customblue!100}\multirow{-2}{*}{\est{\mathbf{+59.4}}{2.3}{2.3}} \\
\addlinespace[2pt]
\multirow{2}{*}{BRITS} & $R\downarrow$ & \cellcolor{customblue!60}$7.1$ & \cellcolor{customblue!62}\est{7.0}{0.1}{0.1} & \cellcolor{customblue!59}\est{7.2}{0.1}{0.1} & \cellcolor{customblue!60}\est{7.2}{0.1}{0.1} & \cellcolor{customblue!61}\est{7.1}{0.1}{0.1} & \cellcolor{customblue!69} \\
 & $S\uparrow$ & \cellcolor{customblue!70}$+6.8$ & \cellcolor{customblue!71}\est{+8.0}{2.8}{2.9} & \cellcolor{customblue!69}\est{+5.4}{2.7}{2.7} & \cellcolor{customblue!69}\est{+0.4}{4.8}{4.9} & \cellcolor{customblue!70}\est{+9.3}{2.0}{2.1} & \cellcolor{customblue!69}\multirow{-2}{*}{\est{+0.4}{5.7}{5.8}} \\
\addlinespace[2pt]
\multirow{2}{*}{DLinear} & $R\downarrow$ & \cellcolor{customblue!57}$7.4$ & \cellcolor{customblue!58}\est{7.3}{0.1}{0.1} & \cellcolor{customblue!56}\est{7.5}{0.1}{0.1} & \cellcolor{customblue!58}\est{7.4}{0.1}{0.1} & \cellcolor{customblue!57}\est{7.4}{0.1}{0.1} & \cellcolor{customblue!63} \\
 & $S\uparrow$ & \cellcolor{customblue!63}$-5.7$ & \cellcolor{customblue!64}\est{-4.8}{3.0}{3.2} & \cellcolor{customblue!62}\est{-6.7}{2.5}{3.1} & \cellcolor{customblue!63}\est{-11.5}{4.7}{5.0} & \cellcolor{customblue!63}\est{-3.8}{2.4}{2.5} & \cellcolor{customblue!63}\multirow{-2}{*}{\est{-11.5}{5.9}{6.0}} \\
\addlinespace[2pt]
\multirow{2}{*}{FEDformer} & $R\downarrow$ & \cellcolor{customblue!23}$10.4$ & \cellcolor{customblue!24}\est{10.3}{0.1}{0.1} & \cellcolor{customblue!23}\est{10.4}{0.1}{0.1} & \cellcolor{customblue!21}\est{10.5}{0.1}{0.1} & \cellcolor{customblue!24}\est{10.3}{0.1}{0.1} & \cellcolor{customblue!36} \\
 & $S\uparrow$ & \cellcolor{customblue!36}$-53.7$ & \cellcolor{customblue!37}\est{-54.6}{4.5}{4.5} & \cellcolor{customblue!36}\est{-53.3}{3.9}{4.6} & \cellcolor{customblue!36}\est{-64.1}{7.5}{7.4} & \cellcolor{customblue!37}\est{-50.3}{3.4}{3.6} & \cellcolor{customblue!36}\multirow{-2}{*}{\est{-64.1}{8.9}{9.6}} \\
\addlinespace[2pt]
\multirow{2}{*}{TimesNet} & $R\downarrow$ & \cellcolor{customblue!28}$10.0$ & \cellcolor{customblue!28}\est{10.0}{0.1}{0.1} & \cellcolor{customblue!28}\est{10.0}{0.1}{0.1} & \cellcolor{customblue!27}\est{10.1}{0.1}{0.1} & \cellcolor{customblue!28}\est{10.0}{0.1}{0.1} & \cellcolor{customblue!28} \\
 & $S\uparrow$ & \cellcolor{customblue!30}$-66.0$ & \cellcolor{customblue!31}\est{-65.5}{5.4}{5.1} & \cellcolor{customblue!29}\est{-67.1}{4.8}{5.2} & \cellcolor{customblue!28}\est{-79.1}{8.4}{8.7} & \cellcolor{customblue!30}\est{-62.5}{3.9}{4.1} & \cellcolor{customblue!28}\multirow{-2}{*}{\est{-79.1}{10.5}{10.3}} \\
\midrule
\multicolumn{8}{l}{\textbf{\emph{Long-context imputation ($\geq 7 \times 1440$ time steps)}}} \\
\hline
\multicolumn{8}{l}{\cellcolor[HTML]{EFEFEF}\textit{Statistical Models}} \\
\multirow{2}{*}{Personalized temp. mean} & $R\downarrow$ & \cellcolor{customblue!45}$8.2$ & \cellcolor{customblue!42}\est{8.5}{0.1}{0.1} & \cellcolor{customblue!48}\est{8.0}{0.1}{0.1} & \cellcolor{customblue!46}\est{8.0}{0.2}{0.2} & \cellcolor{customblue!45}\est{8.3}{0.1}{0.1} & \cellcolor{customblue!58} \\
 & $S\uparrow$ & \cellcolor{customblue!59}$-7.7$ & \cellcolor{customblue!57}\est{-13.3}{4.1}{4.4} & \cellcolor{customblue!61}\est{-3.9}{3.6}{3.9} & \cellcolor{customblue!60}\est{-10.8}{6.0}{6.4} & \cellcolor{customblue!60}\est{-6.1}{3.2}{3.4} & \cellcolor{customblue!58}\multirow{-2}{*}{\est{-13.3}{4.8}{5.6}} \\
\addlinespace[2pt]
\multirow{2}{*}{Personalized mode} & $R\downarrow$ & \cellcolor{customblue!28}$9.8$ & \cellcolor{customblue!28}\est{9.8}{0.1}{0.1} & \cellcolor{customblue!28}\est{9.9}{0.1}{0.1} & \cellcolor{customblue!28}\est{9.7}{0.1}{0.1} & \cellcolor{customblue!28}\est{9.8}{0.1}{0.1} & \cellcolor{customblue!50} \\
 & $S\uparrow$ & \cellcolor{customblue!49}$-26.1$ & \cellcolor{customblue!51}\est{-25.2}{3.2}{3.4} & \cellcolor{customblue!48}\est{-27.3}{3.5}{3.7} & \cellcolor{customblue!50}\est{-29.4}{6.0}{5.9} & \cellcolor{customblue!49}\est{-24.4}{2.6}{2.9} & \cellcolor{customblue!50}\multirow{-2}{*}{\est{-29.4}{6.4}{7.3}} \\
\addlinespace[2pt]
\multirow{2}{*}{Personalized mean} & $R\downarrow$ & $12.4$ & \est{12.4}{0.1}{0.1} & \est{12.4}{0.1}{0.1} & \est{12.2}{0.1}{0.1} & \est{12.4}{0.1}{0.1} &  \\
 & $S\uparrow$ & $-114.1$ & \est{-117.6}{6.3}{6.3} & \est{-112.1}{5.6}{5.9} & \est{-120.8}{9.1}{9.8} & \est{-112.2}{4.7}{5.4} & \multirow{-2}{*}{\est{-120.8}{11.0}{12.0}} \\
\hline
\multicolumn{8}{l}{\cellcolor[HTML]{EFEFEF}\textit{Neural Models}} \\
\multirow{2}{*}{\textsc{LSM-2-Sparse} (7-day)} & $R\downarrow$ & \cellcolor{customblue!100}$\mathbf{3.2}$ & \cellcolor{customblue!100}\est{\mathbf{3.3}}{0.1}{0.1} & \cellcolor{customblue!100}\est{\mathbf{3.1}}{0.1}{0.1} & \cellcolor{customblue!100}\est{\mathbf{3.1}}{0.1}{0.1} & \cellcolor{customblue!100}\est{\mathbf{3.2}}{0.1}{0.1} & \cellcolor{customblue!100} \\
 & $S\uparrow$ & \cellcolor{customblue!100}$\mathbf{+64.7}$ & \cellcolor{customblue!100}\est{\mathbf{+64.4}}{1.1}{1.1} & \cellcolor{customblue!100}\est{\mathbf{+64.9}}{1.0}{1.1} & \cellcolor{customblue!100}\est{\mathbf{+63.1}}{1.8}{1.8} & \cellcolor{customblue!100}\est{\mathbf{+65.3}}{0.9}{0.9} & \cellcolor{customblue!100}\multirow{-2}{*}{\est{\mathbf{+63.1}}{2.1}{2.1}} \\
\addlinespace[2pt]
\multirow{2}{*}{DLinear (7-day)} & $R\downarrow$ & \cellcolor{customblue!41}$8.6$ & \cellcolor{customblue!41}\est{8.6}{0.1}{0.1} & \cellcolor{customblue!41}\est{8.7}{0.1}{0.1} & \cellcolor{customblue!38}\est{8.7}{0.1}{0.1} & \cellcolor{customblue!42}\est{8.6}{0.1}{0.1} & \cellcolor{customblue!46} \\
 & $S\uparrow$ & \cellcolor{customblue!48}$-28.3$ & \cellcolor{customblue!49}\est{-28.0}{3.7}{3.6} & \cellcolor{customblue!47}\est{-28.8}{2.9}{3.5} & \cellcolor{customblue!46}\est{-36.7}{5.9}{5.6} & \cellcolor{customblue!49}\est{-25.8}{2.9}{2.9} & \cellcolor{customblue!46}\multirow{-2}{*}{\est{-36.7}{7.2}{7.1}} \\
\bottomrule[1.5pt]
\end{tabular}}
\end{table}

\begin{table}[H]
\centering
\captionsetup{width=\textwidth}
\caption{\textbf{Forecasting (Track 2B) stratified by age and sex.} Average Rank $R$ ($\downarrow$) and Skill Score $S$ ($\uparrow$, in \%) are recomputed within each cohort; \textsc{Seasonal Naive} is the within-cohort reference. Sub/superscripts are offsets from the point estimate to the $95\%$ percentile interval ($1{,}000$ participant resamples); $S_{\TN{worst}}$ uses the projection interval described in the text. All gives point estimates only. Darker blue indicates better performance within each column and metric; bold marks the best point estimate. FT denotes fine-tuned. $n$ counts evaluated participants, with smaller task-specific samples and undefined tasks omitted as described above.}
\label{tab:subgroup_track2b}
\providecommand{\est}[3]{\ensuremath{#1^{\scriptscriptstyle +#2}_{\scriptscriptstyle -#3}}}
\small
\renewcommand{\arraystretch}{1.05}
\setlength{\tabcolsep}{3pt}
\resizebox{\linewidth}{!}{%
\begin{tabular}{lc c cccc c}
\toprule[1.5pt]
\textbf{Method} & & \makecell{All\\$n=827$} & \makecell{Age $<$40\\$n=356$} & \makecell{Age $\geq$40\\$n=471$} & \makecell{Female\\$n=195$} & \makecell{Male\\$n=569$} & $S_{\TN{worst}}\uparrow$ \\
\midrule
\multicolumn{8}{l}{\cellcolor[HTML]{EFEFEF}\textit{Statistical Methods}} \\
\multirow{2}{*}{\textsc{Seasonal Naive}} & $R\downarrow$ & $7.72$ & \cellcolor{customblue!3}\est{7.51}{0.11}{0.11} & \est{7.88}{0.11}{0.11} & \est{7.79}{0.15}{0.13} & \est{7.68}{0.09}{0.08} &  \\
 & $S\uparrow$ & $0.0$ & $0.0$ & $0.0$ & $0.0$ & $0.0$ & \multirow{-2}{*}{$0.0$} \\
\addlinespace[2pt]
\multirow{2}{*}{\textsc{AutoARIMA}} & $R\downarrow$ & \cellcolor{customblue!2}$7.64$ & \est{7.62}{0.10}{0.10} & \cellcolor{customblue!5}\est{7.65}{0.09}{0.10} & \cellcolor{customblue!2}\est{7.69}{0.13}{0.15} & \cellcolor{customblue!2}\est{7.62}{0.08}{0.08} & \cellcolor{customblue!6} \\
 & $S\uparrow$ & \cellcolor{customblue!16}$+5.9$ & \cellcolor{customblue!6}\est{+2.0}{3.3}{3.4} & \cellcolor{customblue!22}\est{+9.0}{3.9}{3.7} & \cellcolor{customblue!12}\est{+4.5}{4.7}{4.8} & \cellcolor{customblue!16}\est{+5.9}{3.0}{3.0} & \cellcolor{customblue!6}\multirow{-2}{*}{\est{+2.0}{4.5}{4.2}} \\
\addlinespace[2pt]
\multirow{2}{*}{\textsc{AutoETS}} & $R\downarrow$ & \cellcolor{customblue!16}$7.07$ & \cellcolor{customblue!12}\est{7.13}{0.13}{0.13} & \cellcolor{customblue!19}\est{7.04}{0.13}{0.12} & \cellcolor{customblue!16}\est{7.13}{0.15}{0.22} & \cellcolor{customblue!15}\est{7.05}{0.10}{0.09} & \cellcolor{customblue!30} \\
 & $S\uparrow$ & \cellcolor{customblue!38}$+14.3$ & \cellcolor{customblue!30}\est{+10.2}{3.2}{3.2} & \cellcolor{customblue!44}\est{+17.8}{3.6}{3.4} & \cellcolor{customblue!35}\est{+13.3}{4.3}{3.6} & \cellcolor{customblue!38}\est{+14.0}{2.9}{2.8} & \cellcolor{customblue!30}\multirow{-2}{*}{\est{+10.2}{4.3}{4.3}} \\
\hline
\multicolumn{8}{l}{\cellcolor[HTML]{EFEFEF}\textit{Neural Models}} \\
\multirow{2}{*}{\textsc{MixLinear}} & $R\downarrow$ & \cellcolor{customblue!49}$5.68$ & \cellcolor{customblue!50}\est{5.65}{0.15}{0.14} & \cellcolor{customblue!49}\est{5.69}{0.13}{0.11} & \cellcolor{customblue!55}\est{5.54}{0.20}{0.15} & \cellcolor{customblue!47}\est{5.73}{0.10}{0.10} & \cellcolor{customblue!78} \\
 & $S\uparrow$ & \cellcolor{customblue!78}$+29.2$ & \cellcolor{customblue!78}\est{+27.1}{2.4}{2.2} & \cellcolor{customblue!78}\est{+31.4}{2.1}{3.0} & \cellcolor{customblue!84}\est{+31.7}{3.0}{4.2} & \cellcolor{customblue!75}\est{+28.1}{2.3}{2.0} & \cellcolor{customblue!78}\multirow{-2}{*}{\est{+27.1}{3.2}{2.9}} \\
\addlinespace[2pt]
\multirow{2}{*}{\textsc{DLinear}} & $R\downarrow$ & \cellcolor{customblue!74}$4.63$ & \cellcolor{customblue!75}\est{4.70}{0.15}{0.13} & \cellcolor{customblue!75}\est{4.55}{0.15}{0.12} & \cellcolor{customblue!77}\est{4.62}{0.19}{0.20} & \cellcolor{customblue!74}\est{4.61}{0.12}{0.11} & \cellcolor{customblue!96} \\
 & $S\uparrow$ & \cellcolor{customblue!96}$+35.9$ & \cellcolor{customblue!96}\est{+33.1}{2.7}{2.6} & \cellcolor{customblue!95}\est{+38.4}{3.0}{2.9} & \cellcolor{customblue!97}\est{+36.8}{3.1}{3.4} & \cellcolor{customblue!96}\est{+35.7}{2.3}{2.3} & \cellcolor{customblue!96}\multirow{-2}{*}{\est{+33.1}{3.0}{3.3}} \\
\addlinespace[2pt]
\multirow{2}{*}{\textsc{SegRNN}} & $R\downarrow$ & \cellcolor{customblue!81}$4.35$ & \cellcolor{customblue!86}\est{4.24}{0.13}{0.13} & \cellcolor{customblue!77}\est{4.46}{0.11}{0.13} & \cellcolor{customblue!89}\est{4.14}{0.17}{0.16} & \cellcolor{customblue!79}\est{4.42}{0.10}{0.11} & \cellcolor{customblue!97} \\
 & $S\uparrow$ & \cellcolor{customblue!92}$+34.6$ & \cellcolor{customblue!98}\est{+33.8}{2.2}{2.4} & \cellcolor{customblue!86}\est{+34.8}{2.3}{2.3} & \cellcolor{customblue!100}\est{\mathbf{+37.8}}{3.1}{3.0} & \cellcolor{customblue!89}\est{+33.4}{1.9}{2.0} & \cellcolor{customblue!97}\multirow{-2}{*}{\est{+33.4}{2.4}{2.6}} \\
\hline
\multicolumn{8}{l}{\cellcolor[HTML]{EFEFEF}\textit{Time-Series Foundation Models}} \\
\multirow{2}{*}{\textsc{Toto}} & $R\downarrow$ & \cellcolor{customblue!54}$5.49$ & \cellcolor{customblue!52}\est{5.57}{0.16}{0.16} & \cellcolor{customblue!55}\est{5.43}{0.13}{0.12} & \cellcolor{customblue!58}\est{5.42}{0.18}{0.15} & \cellcolor{customblue!52}\est{5.52}{0.12}{0.12} & \cellcolor{customblue!70} \\
 & $S\uparrow$ & \cellcolor{customblue!71}$+26.8$ & \cellcolor{customblue!70}\est{+24.1}{2.8}{2.9} & \cellcolor{customblue!71}\est{+28.5}{2.1}{2.2} & \cellcolor{customblue!74}\est{+27.9}{3.0}{3.3} & \cellcolor{customblue!70}\est{+26.2}{2.0}{2.1} & \cellcolor{customblue!70}\multirow{-2}{*}{\est{+24.1}{3.5}{3.5}} \\
\addlinespace[2pt]
\multirow{2}{*}{\textsc{Toto} (FT)} & $R\downarrow$ & \cellcolor{customblue!73}$4.68$ & \cellcolor{customblue!78}\est{4.56}{0.14}{0.12} & \cellcolor{customblue!70}\est{4.77}{0.10}{0.12} & \cellcolor{customblue!75}\est{4.70}{0.13}{0.15} & \cellcolor{customblue!73}\est{4.67}{0.10}{0.09} & \cellcolor{customblue!83} \\
 & $S\uparrow$ & \cellcolor{customblue!82}$+30.9$ & \cellcolor{customblue!90}\est{+31.0}{4.0}{3.3} & \cellcolor{customblue!77}\est{+30.9}{3.4}{3.1} & \cellcolor{customblue!76}\est{+28.7}{2.6}{2.8} & \cellcolor{customblue!85}\est{+31.6}{3.0}{2.7} & \cellcolor{customblue!83}\multirow{-2}{*}{\est{+28.7}{3.4}{3.4}} \\
\addlinespace[2pt]
\multirow{2}{*}{\textsc{Chronos-2}} & $R\downarrow$ & \cellcolor{customblue!85}$4.17$ & \cellcolor{customblue!84}\est{4.31}{0.13}{0.13} & \cellcolor{customblue!86}\est{4.07}{0.14}{0.13} & \cellcolor{customblue!86}\est{4.27}{0.16}{0.17} & \cellcolor{customblue!85}\est{4.16}{0.10}{0.10} & \cellcolor{customblue!98} \\
 & $S\uparrow$ & \cellcolor{customblue!97}$+36.4$ & \cellcolor{customblue!98}\est{+33.8}{2.7}{2.5} & \cellcolor{customblue!96}\est{+38.7}{2.5}{3.1} & \cellcolor{customblue!96}\est{+36.2}{2.9}{2.9} & \cellcolor{customblue!97}\est{+36.2}{2.3}{2.1} & \cellcolor{customblue!98}\multirow{-2}{*}{\est{+33.8}{3.5}{3.3}} \\
\addlinespace[2pt]
\multirow{2}{*}{\textsc{Chronos-2} (FT)} & $R\downarrow$ & \cellcolor{customblue!100}$\mathbf{3.56}$ & \cellcolor{customblue!100}\est{\mathbf{3.70}}{0.13}{0.12} & \cellcolor{customblue!100}\est{\mathbf{3.45}}{0.13}{0.13} & \cellcolor{customblue!100}\est{\mathbf{3.69}}{0.14}{0.18} & \cellcolor{customblue!100}\est{\mathbf{3.54}}{0.09}{0.09} & \cellcolor{customblue!100} \\
 & $S\uparrow$ & \cellcolor{customblue!100}$\mathbf{+37.6}$ & \cellcolor{customblue!100}\est{\mathbf{+34.6}}{2.7}{2.6} & \cellcolor{customblue!100}\est{\mathbf{+40.3}}{2.3}{3.1} & \cellcolor{customblue!98}\est{+37.1}{3.5}{2.9} & \cellcolor{customblue!100}\est{\mathbf{+37.3}}{2.2}{2.0} & \cellcolor{customblue!100}\multirow{-2}{*}{\est{\mathbf{+34.6}}{3.4}{3.4}} \\
\bottomrule[1.5pt]
\end{tabular}}
\end{table}

\clearpage
\section{Detailed Dataset Characteristics}
\label{app:dataset_characteristics}
\label{app:dataset_channel_overview}

This appendix provides additional descriptive statistics for the \openmhc{} participant cohort, linked variables, device coverage, geography, wearable-channel coverage, and official dataset splits. Data preprocessing is described separately in Appendix~\ref{app:shared_wearable_preprocessing}.

\begin{figure}[h]
    \centering
    \includegraphics[width=0.7\textwidth]{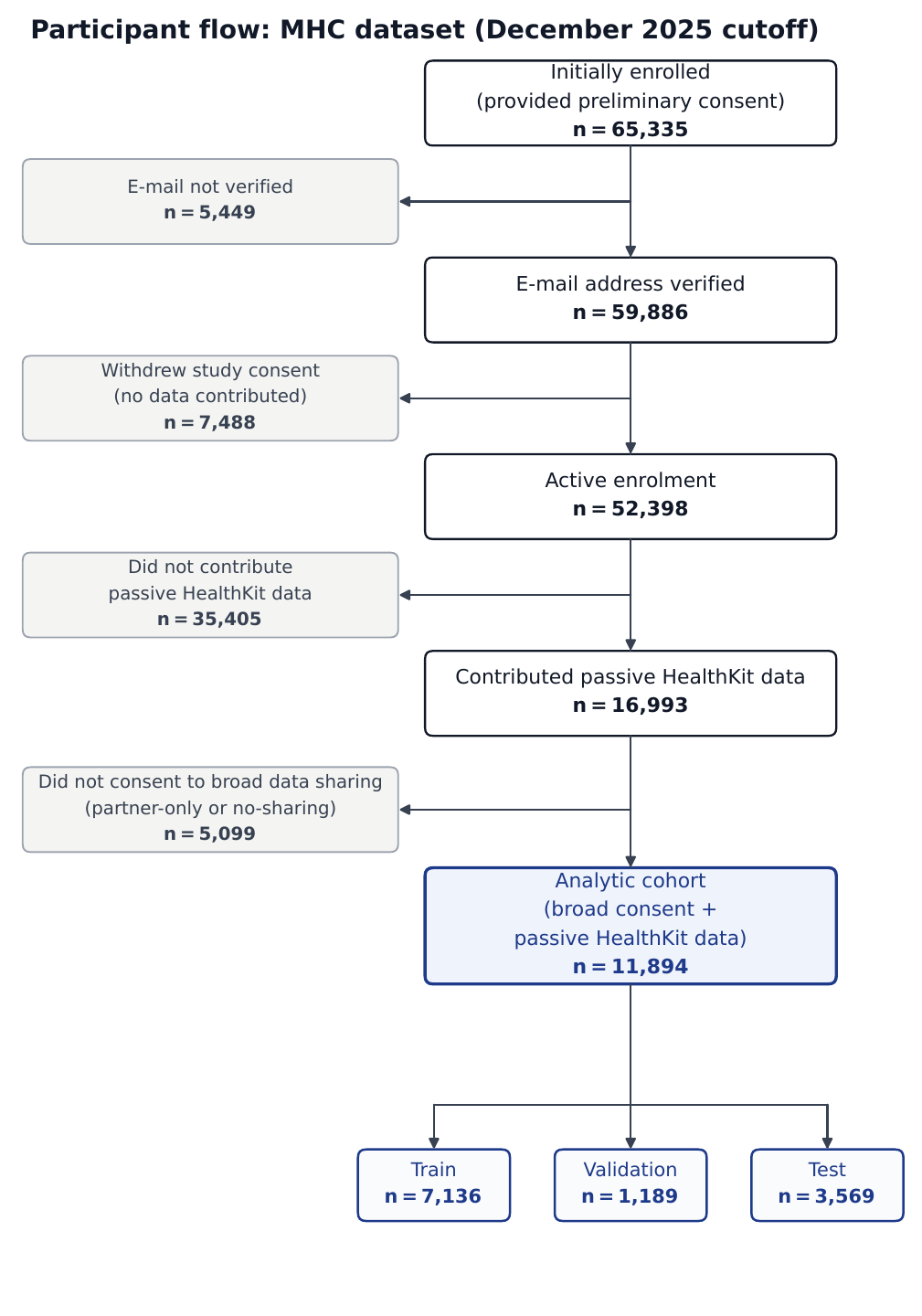} 
    \caption{
    Participant flow diagram for the MHC wearable benchmark cohort. Counts $n$ denote the number of unique participants at each cohort-construction step. The figure reports cohort construction and wearable-data availability. Task-specific downstream prediction participant counts are reported separately in Table~\ref{tab:health_outcome_full}.
    }
    \label{fig:consort}
\end{figure}

\begin{figure}[H]
  \centering
  \includegraphics[width=\linewidth]{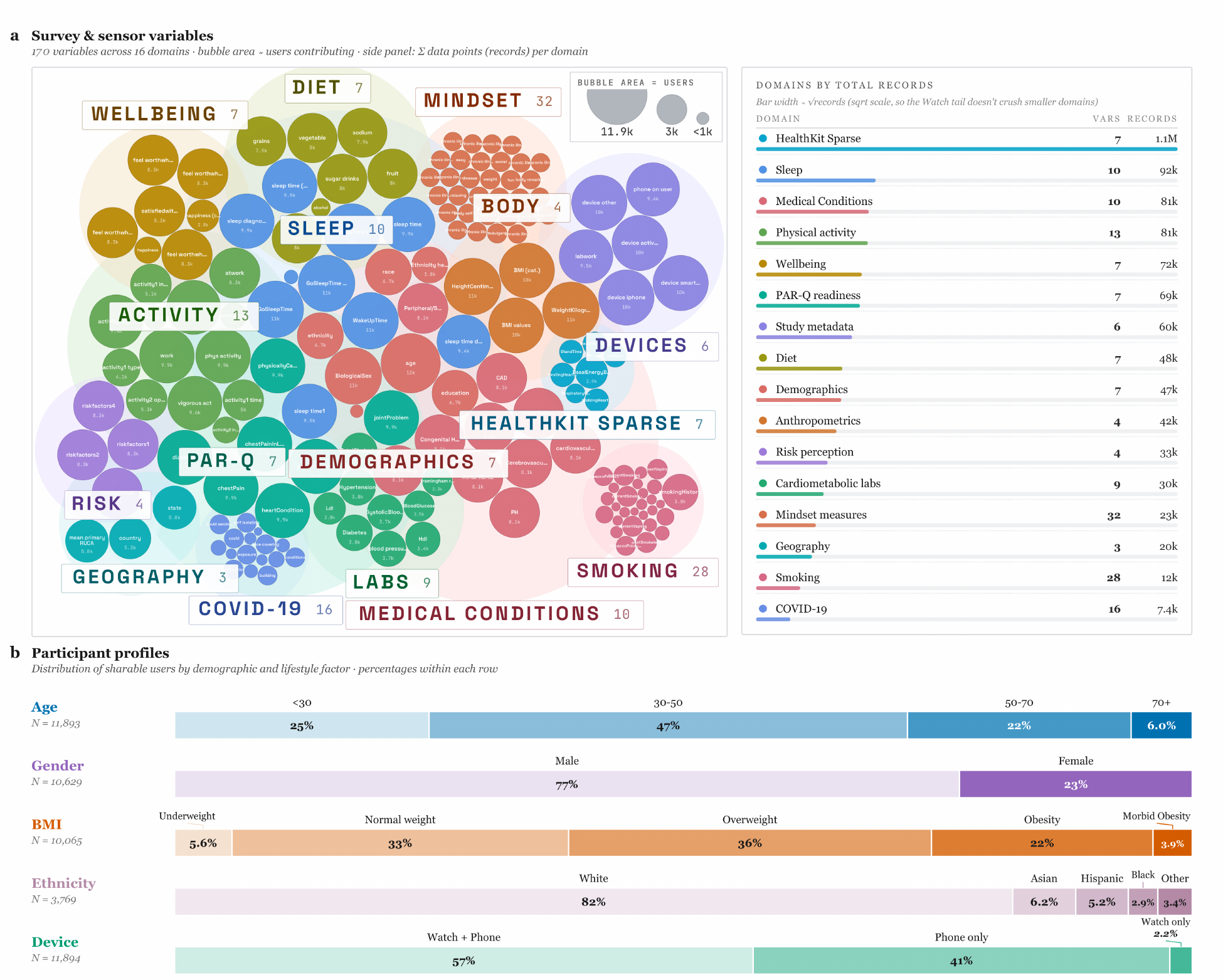}
  \caption{%
    \textbf{Linked variables and participant profiles.}
    \textbf{(a) Linked variables:} coverage of 170 linked variables across 16 source domains. Each circle denotes one variable, and circle area is proportional to the number of participants with at least one observation for that variable.
    \textbf{(b) Participant profiles:} distributions of age ($N=11{,}893$), biological sex ($N=10{,}629$), BMI ($N=10{,}047$), ethnicity ($N=3{,}769$), and device availability ($N=11{,}894$).
  }
  \label{fig:context-overview}
\end{figure}

The \openmhc{} dataset is built from 11{,}894 sharable participants in the My Heart Counts study. The wearable data include four channel groups: phone-derived activity, watch-derived activity and physiology, sleep, and workouts. 
Figure~\ref{fig:consort} describes participant flow for the wearable cohort and Figure~\ref{fig:context-overview} summarizes linked variable coverage and participant demographics. 


\begin{figure}[H]
  \centering
  \includegraphics[width=\linewidth]{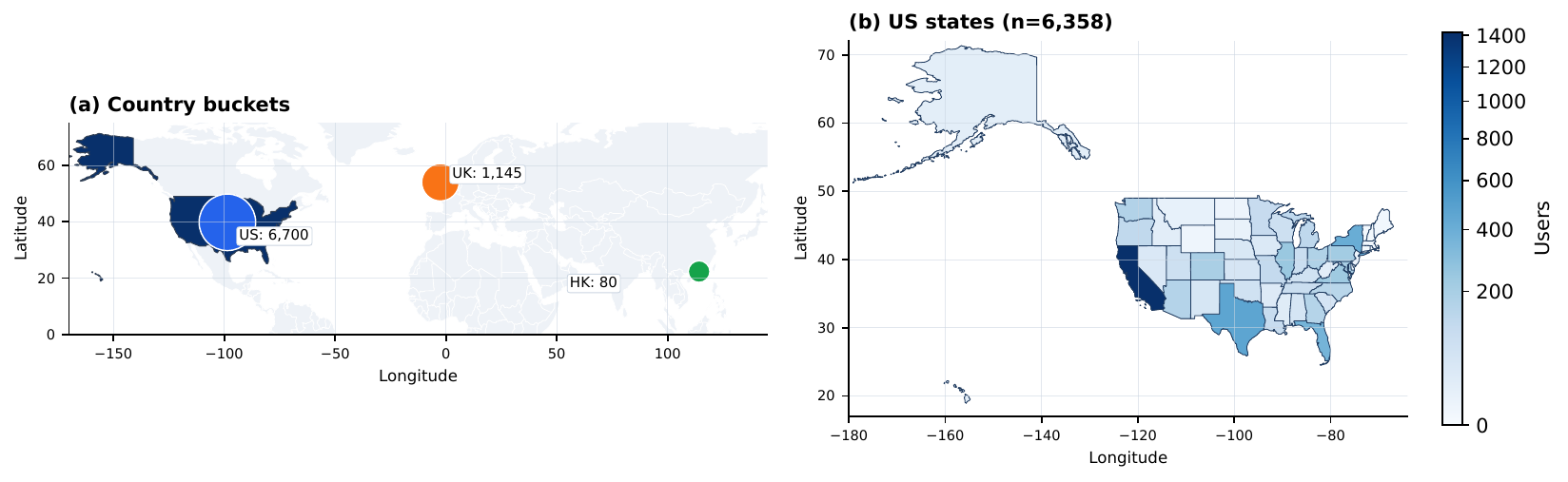}
  \caption{%
    \textbf{
    Participant geographic distribution.}
    \textbf{(a) Country buckets:} participant counts resolved to US, UK, and HK from the geography context labels.
    \textbf{(b) US states:}  state-level distribution among the $6{,}372$ participants. 
  }
  \label{fig:geography-distribution}
\end{figure}

\bo{Geographic Coverage.}
We summarize participant geography using \texttt{field\_country}. 
The released data contain derived state/territory and rounded mean RUCA fields instead of ZIP codes (Appendix~\ref{app:data_access}). 
The resulting country and US state distributions are shown in Figure~\ref{fig:geography-distribution}. 

\begin{figure}[H]
    \centering
    \includegraphics[width=0.8\linewidth]{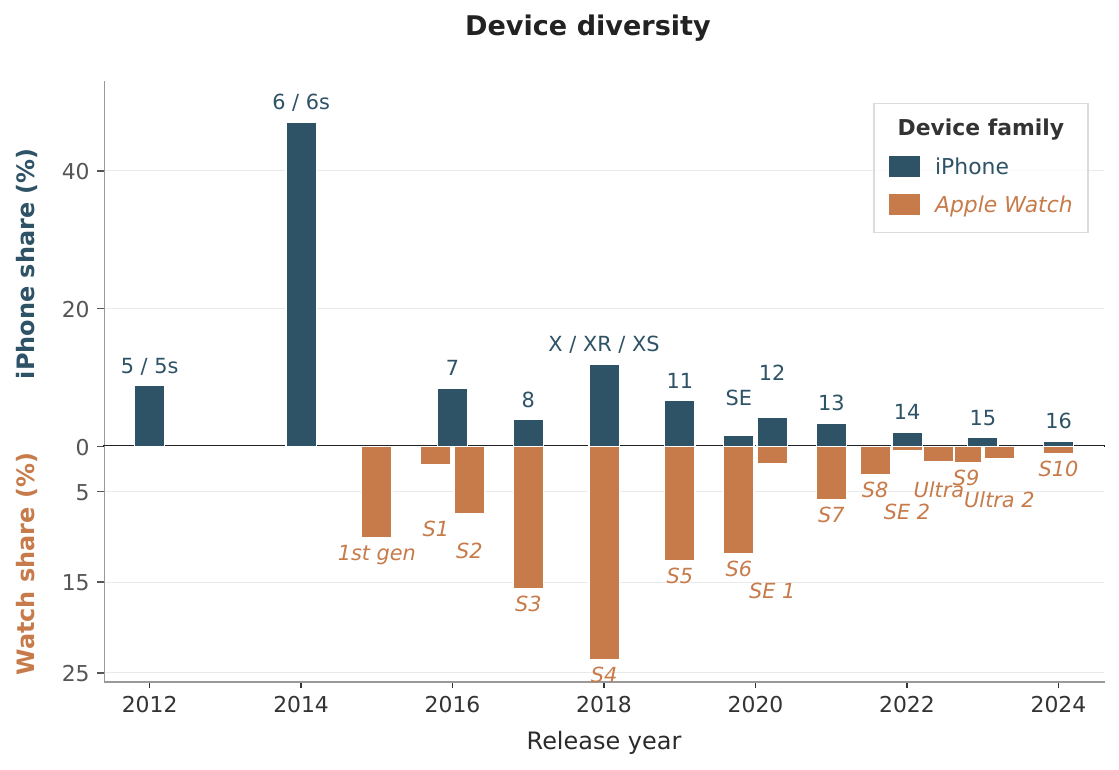}
    \caption{\textbf{Device generation distribution in the sharable
    cohort}. Bars show each generation's share of known device entries within its category (iPhone top, Watch
    bottom; bars sum to 100\% per panel). Users are counted once per
    generation owned. Unknown entries and iPhone~4 ($n = 1$) are
    excluded; SE generations are aggregated.}
    \label{fig:device-diversity}
\end{figure}

\bo{Device coverage.}
Table~\ref{tab:devices} summarises the iPhone and Apple Watch generations present in the sharable cohort. Each user is counted once per generation
family they ever owned 
(Plus, Pro, Pro Max, mini, case size, GPS versus GPS+Cellular). 
The column sums therefore exceed the number of unique users, because a user who upgrades across generations contributes to multiple rows. 
\emph{Unknown} entries arise when a user renames their device in iOS Settings, which overwrites the model identifier exposed via HealthKit and the MHC app did not store the device version specifically otherwise. 
Across the cohort, 18\,\% of phone records and 40\,\% of watch records carry no clear model identifier. 
The distribution of device generations, ordered by release year, is visualized in Figure~\ref{fig:device-diversity}. 
The range of device generations offers an opportunity to study robustness across device settings. However, comparisons between device groups may also reflect differences in study entry year, cohort composition, and data availability. 

\begin{table}[H]
\centering
\small
\begin{minipage}[t]{0.48\linewidth}
\centering
\textbf{(a) iPhone families}\\[2pt]
\begin{tabular}{lr}
\toprule
Family & Users \\
\midrule
iPhone 4              &     1 \\
iPhone 5 / 5s         &   906 \\
iPhone 6 / 6s (+ Plus)& 4{,}855 \\
iPhone 7 (+ Plus)     &   872 \\
iPhone 8 (+ Plus)     &   400 \\
iPhone X / XR / XS / XS Max & 1{,}234 \\
iPhone SE (1st--3rd gen) &   172 \\
iPhone 11 (+ Pro / Pro Max) &   682 \\
iPhone 12 (+ mini / Pro / Pro Max) & 428 \\
iPhone 13 (+ mini / Pro / Pro Max) & 346 \\
iPhone 14 (+ Plus / Pro / Pro Max) & 207 \\
iPhone 15 (+ Plus / Pro / Pro Max) & 130 \\
iPhone 16 (+ Pro / Pro Max) &    79 \\
iPhone Unknown        & 6{,}499 \\
\midrule
Any phone (unique users)   & 11{,}642 \\
\bottomrule
\end{tabular}
\end{minipage}\hfill
\begin{minipage}[t]{0.48\linewidth}
\centering
\textbf{(b) Apple Watch families}\\[2pt]
\begin{tabular}{lr}
\toprule
Family & Users \\
\midrule
Apple Watch (1st gen)  &   304 \\
Series 1               &    61 \\
Series 2               &   224 \\
Series 3               &   474 \\
Series 4               &   710 \\
Series 5               &   382 \\
Series 6               &   357 \\
Series 7               &   176 \\
Series 8               &    93 \\
Series 9               &    53 \\
Series 10              &    23 \\
SE (1st gen)           &    56 \\
SE (2nd gen)           &    15 \\
Ultra                  &    52 \\
Ultra 2                &    42 \\
Apple Watch Unknown    & 6{,}485 \\
\midrule
Any watch (unique users)   & 7{,}338 \\
\bottomrule
\end{tabular}
\end{minipage}
\vspace{2mm}
\caption{\bo{Device generations represented in the sharable cohort
($n = 11{,}894$).} 
}
\label{tab:devices}
\end{table}

\begin{figure}[H]
  \centering
  \includegraphics[width=0.7\linewidth]{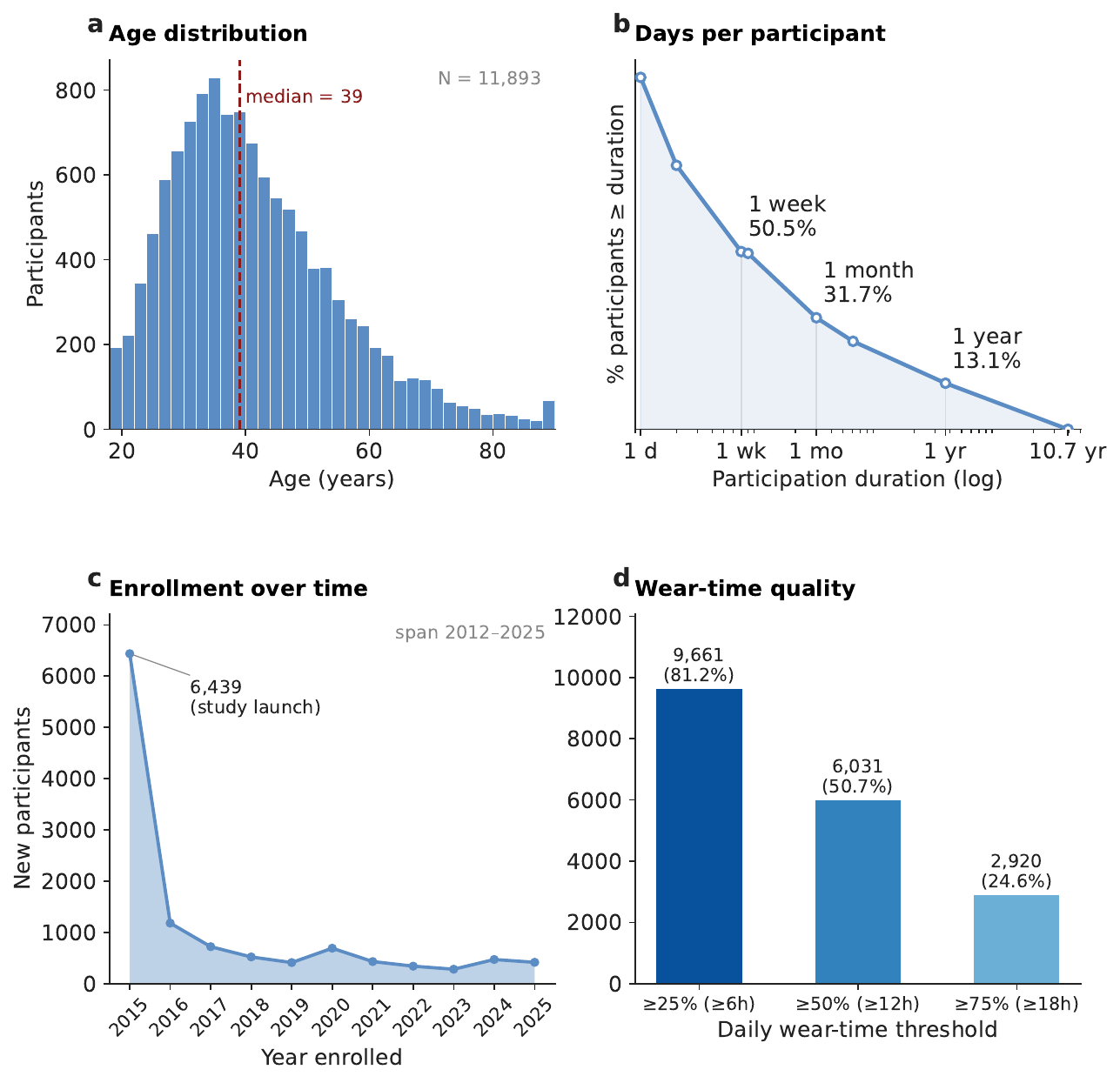}
  \caption{%
    \textbf{Overview of the My Heart Counts (MHC) dataset distributions.}
    \textbf{(a)} Age distribution of all reporting participants
    ($N=11{,}893$, median 39, IQR $[30, 52]$).
    \textbf{(b)} Survival curve of participation duration (number of
    days with passive data) on a log x-axis. 
    \textbf{(c)} New participants by year of enrollment. 
    \textbf{(d)} Wear-time quality: number of participants with at
    least one day meeting each daily wear-time threshold. 
  }
  \label{fig:mhc-distributions}
\end{figure}
\begin{table}[H]
\centering
\footnotesize
\renewcommand{\arraystretch}{1.15}
\caption{Summary of the 19 channels in the daily matrix $d \in \real^{19 \times 1440}$, computed over all 11{,}894 sharable participants prior to any quality filtering. \textbf{n}: number of participants with $\geq$1 day of data for that channel. \textbf{Avg.\ Days}: mean number of days with data per participant. \textbf{Span}: calendar-day range from first to last observation across all participants.}
\label{tab:healthkit_summary}
\begin{tabular}{llrrrr}
\toprule
\textbf{Category} & \textbf{Channel} & \textbf{Unit} & \textbf{n} & \textbf{Avg.\ Days} & \textbf{Span} \\
\midrule
\multirow{3}{*}{Phone} & StepCount & steps/min & 11,631 & 234.4 & 4,730 \\
 & DistanceWalkingRunning & meter/min & 11,630 & 234.6 & 4,730 \\
 & FlightsClimbed & count/min & 2,752 & 430.6 & 4,730 \\
\midrule
\multirow{4}{*}{Watch} 
 & StepCount & steps/min & 7,010 & 202.2 & 4,307 \\
 & DistanceWalkingRunning & meter/min & 7,006 & 200.6 & 4,307 \\
 & HeartRate & count/s & 7,020 & 202.0 & 4,307 \\
 & ActiveEnergyBurned & cal/min & 6,993 & 201.4 & 4,307 \\
\midrule
\multirow{2}{*}{Sleep} & Asleep & binary & 2,704 & 254.3 & 4,730 \\
 & InBed & binary & 2,784 & 361.7 & 4,730 \\
\midrule
\multirow{10}{*}{Workout} & Walking & binary & 2,607 & 90.3 & 3,909 \\
 & Cycling & binary & 1,325 & 52.9 & 3,912 \\
 & Running & binary & 1,544 & 39.5 & 3,911 \\
 & Other & binary & 1,164 & 43.4 & 3,871 \\
 & Mixed Metabolic Cardio & binary & 176 & 20.5 & 3,240 \\
 & Strength Training & binary & 557 & 52.9 & 3,827 \\
 & Elliptical & binary & 677 & 35.1 & 3,868 \\
 & HIIT & binary & 379 & 43.1 & 3,579 \\
 & Functional Strength & binary & 493 & 36.7 & 3,858 \\
 & Yoga & binary & 498 & 32.1 & 3,833 \\
\bottomrule
\end{tabular}
\end{table}

\begin{table}[h]
\centering
\footnotesize %
\renewcommand{\arraystretch}{1.15}
\caption{Participant demographics by dataset split. Continuous variables: mean $\pm$ SD, median [P25--P75]. Categorical variables: $n$ (\% of covered). Coverage: fraction of split with a non-null value. Height excludes 418 values of exactly 0\,cm, which remain in the released data.}
\label{tab:demographics}
\begin{tabular}{lcccc}
\toprule
\textbf{Characteristic} & \textbf{Train} ($n{=}7{,}136$) & \textbf{Validation} ($n{=}1{,}189$) & \textbf{Test} ($n{=}3{,}569$) & \textbf{Overall} ($N{=}11{,}894$) \\
\midrule
\multicolumn{5}{l}{\textit{Age (years)}} \\
\quad Coverage & 7,136 (100\%) & 1,189 (100\%) & 3,568 (100\%) & 11,893 (100\%) \\
\quad Mean $\pm$ SD & 42.2 $\pm$ 15.5 & 41.1 $\pm$ 14.7 & 41.6 $\pm$ 15.4 & 41.9 $\pm$ 15.4 \\
\quad Median [IQR] & 39 [30--53] & 39 [30--50] & 39 [30--52] & 39 [30--52] \\
\midrule
\multicolumn{5}{l}{\textit{Biological Sex}} \\
\quad Coverage & 6,395 (89.6\%) & 1,064 (89.5\%) & 3,170 (88.8\%) & 10,629 (89.4\%) \\
\quad Male & 4,945 (77.3\%) & 813 (76.4\%) & 2,439 (76.9\%) & 8,197 (77.1\%) \\
\quad Female & 1,450 (22.7\%) & 251 (23.6\%) & 731 (23.1\%) & 2,432 (22.9\%) \\
\midrule
\multicolumn{5}{l}{\textit{Height (cm)}} \\
\quad Coverage & 6,203 (86.9\%) & 1,037 (87.2\%) & 3,053 (85.5\%) & 10,293 (86.5\%) \\
\quad Mean $\pm$ SD & 175.5 $\pm$ 10.2 & 175.3 $\pm$ 11.1 & 175.3 $\pm$ 10.5 & 175.4 $\pm$ 10.4 \\
\quad Median [IQR] & 175.3 [170.2--182.9] & 177.8 [170.2--182.9] & 175.3 [170.2--182.9] & 175.3 [170.2--182.9] \\
\midrule
\multicolumn{5}{l}{\textit{Weight (kg)}} \\
\quad Coverage & 6,097 (85.4\%) & 1,024 (86.1\%) & 3,005 (84.2\%) & 10,126 (85.1\%) \\
\quad Mean $\pm$ SD & 83.9 $\pm$ 20.8 & 84.4 $\pm$ 20.8 & 84.3 $\pm$ 21.1 & 84.1 $\pm$ 20.9 \\
\quad Median [IQR] & 81.6 [70.3--95.3] & 81.6 [70.3--94.3] & 81.6 [70.3--95.3] & 81.6 [70.3--95.3] \\
\midrule
\multicolumn{5}{l}{\textit{BMI (kg/m$^2$)}} \\
\quad Coverage & 6,051 (84.8\%) & 1,015 (85.4\%) & 2,981 (83.5\%) & 10,047 (84.5\%) \\
\quad Mean $\pm$ SD & 27.2 $\pm$ 5.8 & 27.3 $\pm$ 5.9 & 27.3 $\pm$ 6.0 & 27.2 $\pm$ 5.8 \\
\quad Median [IQR] & 26.1 [23.3--30.1] & 26.2 [23.6--30.1] & 26.2 [23.3--30.0] & 26.1 [23.3--30.0] \\
\midrule
\multicolumn{5}{l}{\textit{Ethnicity (self-reported)}} \\
\quad Coverage & 2,277 (31.9\%) & 375 (31.5\%) & 1,117 (31.3\%) & 3,769 (31.7\%) \\
\quad White & 1,872 (82.2\%) & 294 (78.4\%) & 938 (84.0\%) & 3,104 (82.4\%) \\
\quad Asian & 138 (6.1\%) & 31 (8.3\%) & 64 (5.7\%) & 233 (6.2\%) \\
\quad Hispanic & 119 (5.2\%) & 23 (6.1\%) & 53 (4.7\%) & 195 (5.2\%) \\
\quad Black & 69 (3.0\%) & 11 (2.9\%) & 29 (2.6\%) & 109 (2.9\%) \\
\quad Other & 68 (3.0\%) & 14 (3.7\%) & 27 (2.4\%) & 109 (2.9\%) \\
\quad Prefer not to say & 11 (0.5\%) & 2 (0.5\%) & 6 (0.5\%) & 19 (0.5\%) \\
\bottomrule
\end{tabular}
\end{table}

\begin{figure}[p]
  \centering
  \includegraphics[width=\linewidth]{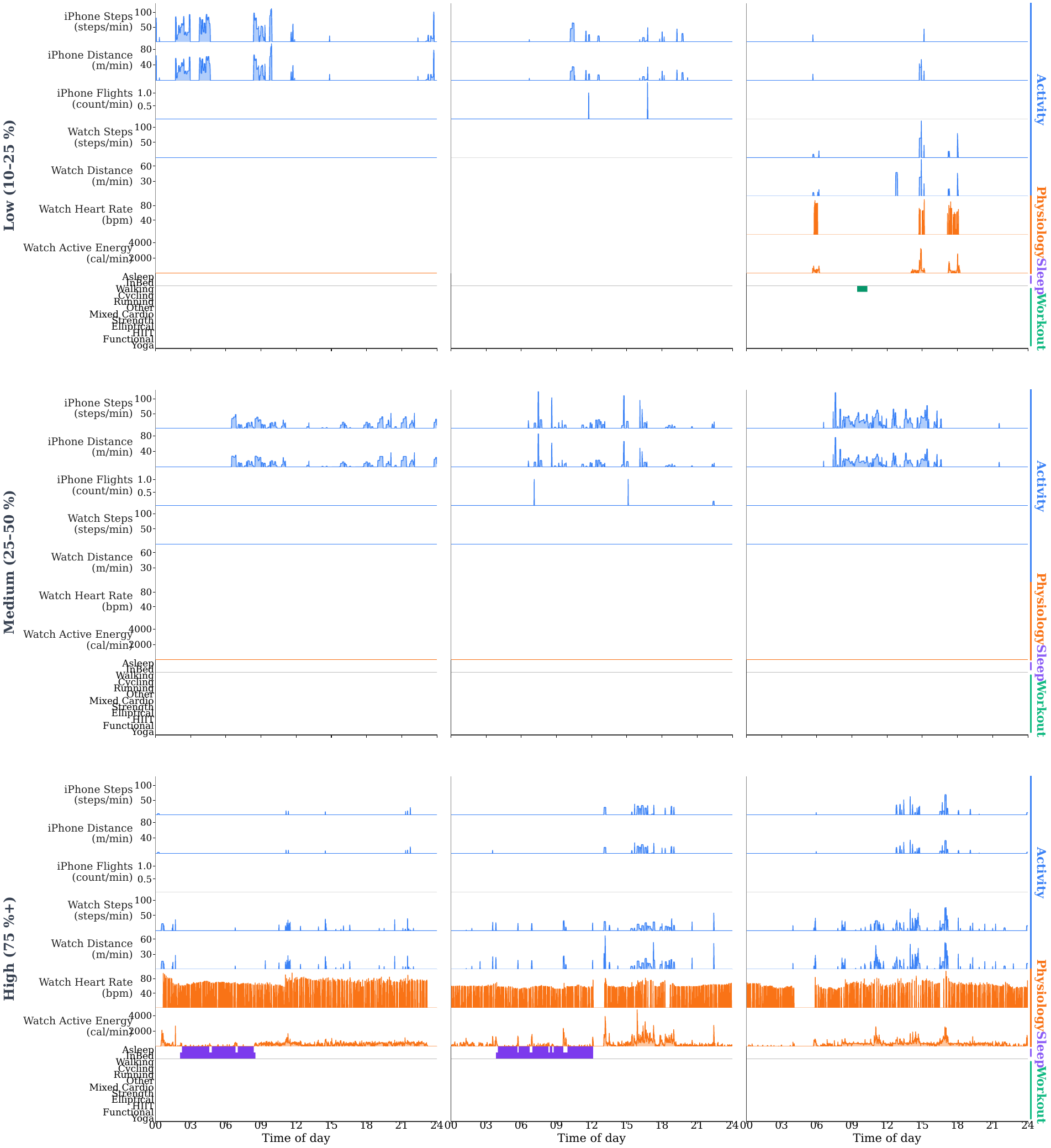}
  \caption{%
    \textbf{Daily wearable samples at different wear-time coverage levels.}
    Nine randomly selected daily samples from the released dataset, arranged in a $3 \times 3$ grid by wear-time coverage tier: low (10--25\,\%), medium (25--50\,\%), and high ($\geq$75\,\%). Each cell shows all 19 channels at minute resolution (1{,}440 time steps): seven continuous channels (phone and watch activity, heart rate, active energy) as sparkline traces and twelve binary channels (sleep, workouts) as heatmap lanes. Gaps in the traces correspond to periods with no recorded data. The figure illustrates the heterogeneity of data completeness across real-world participants.
  }
  \label{fig:weartime-coverage-samples}
\end{figure}

\subsection{Temporal Shifts in Demographics and Sensor Coverage}
\label{app:temporal_shift}
In Table~\ref{tab:entry_year_demographics}, we show how age and sex composition vary across survey year groups. 
Table~\ref{tab:channel_coverage_by_year} shows that Watch and sleep-channel availability generally increases through 2024, while iPhone step count availability remains consistently high.




\vspace{-3mm}
\begingroup
\small
\renewcommand{\arraystretch}{1.15}
\setlength{\LTcapwidth}{\linewidth}
\begin{longtable}{rrrrr}
\caption{Demographic composition by year of the earliest available survey record. Participants are grouped by the year of their earliest available survey record, which may differ from their enrollment year shown in panel~(c) of Figure~\ref{fig:mhc-distributions}.} 
\label{tab:entry_year_demographics} \\
\toprule
Year & Male (\%) & Sex available, $N$ & Age $<40$ (\%) & Age available, $N$ \\
\midrule
\endfirsthead
\caption[]{Demographic composition by earliest survey year (continued).} \\
\toprule
Year & Male (\%) & Sex available, $N$ & Age $<40$ (\%) & Age available, $N$ \\
\midrule
\endhead
\midrule
\multicolumn{5}{r}{\small Continued on next page} \\
\endfoot
\bottomrule
\endlastfoot
2015 & 84.9 & 4{,}873 & 55.2 & 5{,}944 \\
2016 & 73.9 & 2{,}015 & 47.5 & 2{,}043 \\
2017 & 78.8 & 1{,}189 & 44.2 & 1{,}194 \\
2018 & 65.0 & 1{,}229 & 45.3 & 1{,}233 \\
2019 & 66.9 & 534 & 51.5 & 540 \\
2020 & 62.9 & 464 & 42.3 & 468 \\
2021 & 59.5 & 168 & 49.4 & 170 \\
2022 & 58.5 & 135 & 49.6 & 135 \\
2023 & 36.4 & 22 & 46.2 & 78 \\
2024 & --- & 0 & 44.4 & 72 \\
2025 & --- & 0 & 25.0 & 16 \\
\end{longtable}
\endgroup

{\footnotesize
\renewcommand{\arraystretch}{1.1}
\setlength{\tabcolsep}{1pt}
\setlength{\LTcapwidth}{\linewidth}
\begin{longtable}{@{}>{\raggedright\arraybackslash}p{0.27\linewidth}*{11}{>{\raggedleft\arraybackslash}p{\dimexpr0.73\linewidth/11-2\tabcolsep\relax}}@{}}
\caption{For each calendar year, we report the percentage of participants with at least one non-NaN value in each channel, among participants with at least one non-NaN value in any channel during that year. Calendar years refer to the original study dates.}
\label{tab:channel_coverage_by_year} \\
\toprule
Channel (\%) & 2015 & 2016 & 2017 & 2018 & 2019 & 2020 & 2021 & 2022 & 2023 & 2024 & 2025 \\
\midrule
\endfirsthead
\caption[]{Sensor-channel coverage by year (continued).} \\
\toprule
Channel (\%) & 2015 & 2016 & 2017 & 2018 & 2019 & 2020 & 2021 & 2022 & 2023 & 2024 & 2025 \\
\midrule
\endhead
\midrule
\multicolumn{12}{r}{\footnotesize Continued on next page} \\
\endfoot
\endlastfoot
iPhone step count & 98.6 & 98.0 & 97.8 & 98.3 & 99.2 & 99.7 & 99.8 & 99.9 & 99.9 & 99.8 & 99.7 \\
iPhone walking/running distance & 98.6 & 98.0 & 97.8 & 98.3 & 99.2 & 99.7 & 99.8 & 99.9 & 99.9 & 99.8 & 99.7 \\
iPhone flights climbed & 75.4 & 80.3 & 85.5 & 86.8 & 89.8 & 91.3 & 92.0 & 92.8 & 92.8 & 92.9 & 92.1 \\
\midrule
Watch step count & 46.5 & 60.0 & 76.8 & 78.5 & 77.7 & 78.4 & 81.9 & 81.2 & 80.6 & 80.4 & 81.9 \\
Watch walking/running distance & 46.6 & 60.2 & 77.0 & 78.7 & 77.8 & 78.6 & 82.0 & 81.2 & 80.6 & 80.4 & 81.9 \\
Watch heart rate & 46.5 & 60.1 & 76.9 & 78.7 & 77.8 & 78.6 & 82.0 & 81.2 & 80.6 & 80.4 & 81.9 \\
Watch active energy burned & 46.5 & 59.9 & 76.6 & 78.3 & 77.1 & 77.6 & 81.0 & 80.6 & 80.2 & 80.4 & 81.9 \\
\midrule
Asleep & 12.9 & 19.5 & 30.5 & 38.3 & 45.8 & 59.1 & 68.1 & 72.2 & 75.2 & 80.0 & 66.3 \\
In bed & 12.9 & 19.5 & 30.5 & 38.3 & 45.8 & 59.1 & 68.1 & 72.2 & 75.2 & 80.0 & 66.3 \\
\midrule
Walking workout & 11.2 & 16.3 & 24.2 & 32.6 & 39.8 & 56.5 & 62.6 & 64.9 & 66.5 & 70.2 & 61.6 \\
Cycling workout & 6.8 & 9.7 & 15.3 & 20.5 & 28.1 & 40.9 & 47.9 & 50.2 & 51.6 & 54.1 & 47.3 \\
Running workout & 8.6 & 10.9 & 15.4 & 21.3 & 27.8 & 40.8 & 46.4 & 49.5 & 51.5 & 53.1 & 49.5 \\
Other workout & 6.9 & 9.5 & 13.5 & 18.5 & 24.0 & 34.4 & 40.3 & 43.1 & 44.9 & 48.4 & 41.6 \\
Mixed cardio workout & 2.0 & 2.0 & 2.6 & 2.8 & 3.9 & 4.5 & 5.0 & 4.7 & 4.5 & 4.1 & 3.8 \\
Traditional strength workout & 1.8 & 3.9 & 7.8 & 10.4 & 15.7 & 21.0 & 24.4 & 25.4 & 26.3 & 26.3 & 22.5 \\
Elliptical workout & 3.1 & 5.0 & 9.1 & 12.7 & 17.9 & 25.1 & 28.8 & 30.3 & 31.6 & 33.9 & 29.8 \\
HIIT workout & 1.2 & 2.6 & 4.9 & 7.9 & 12.8 & 19.6 & 23.4 & 24.9 & 24.2 & 23.1 & 21.9 \\
Functional strength workout & 1.7 & 3.8 & 6.9 & 9.8 & 15.3 & 23.0 & 29.9 & 32.3 & 33.6 & 34.5 & 31.4 \\
Yoga workout & 1.6 & 3.4 & 7.2 & 9.6 & 15.9 & 23.1 & 27.5 & 29.3 & 29.3 & 29.6 & 23.5 \\*
Participants with sensor data, $N$ & 5{,}601 & 3{,}851 & 3{,}021 & 2{,}820 & 2{,}242 & 1{,}783 & 1{,}231 & 932 & 711 & 510 & 315 \\*
\noalign{\nobreak\hrule height\heavyrulewidth\nobreak\vskip\belowrulesep\nobreak}
\multicolumn{12}{@{}p{\linewidth}@{}}{\footnotesize Data for 2025 extend through 13 June. Records from 2012--2014 are omitted because some are assigned to Watch channels before the Apple Watch was available, and their source could not be verified.} 
\end{longtable}
}

\vspace{-3mm}
\subsection{Comparison with Proprietary Pretraining Datasets}
\label{app:proprietary-comparison}

\bo{Scale.}
\openmhc{} releases 2{,}822{,}500 daily records from 11{,}894 participants (Table~\ref{tab:effective_sample_sizes}), representing 67.74M participant-hours at 24 hours per daily matrix, summarized throughout as over 67M participant-hours. This counts the full daily matrices, including missing periods, rather than measured wear time or the sum of observed channel-hours. \textsc{LSM-2} reports approximately 40M pretraining hours from 60{,}440 participants~\citep{xu2025lsm}; \textsc{WBM} reports 15.14M participant-weeks (approximately 2.5B hours) from 161{,}855 participants~\citep{erturk2025beyond}. By these reported volumes, \openmhc{} and \textsc{LSM-2} are of the same order of magnitude, while \openmhc{} is approximately 2--3\% of \textsc{WBM}. 

\bo{Cohort composition.}
The \textsc{WBM} pretraining cohort is 62.0\% male and 79.1\% White, with mean age 40.7 years~\citep[Table~6]{erturk2025beyond}. These statistics describe the model's cohort (distinct from the AHMS cohort characterized in \citet{truslow2024understanding}). \textsc{LSM-2} reports approximately 62\% male and 38\% female participants, with mean age 42.5 years~\citep{xu2025lsm}. \openmhc{} is also predominantly male (Figure \ref{fig:context-overview}).


\bo{Inputs and temporal resolution.}
\textsc{WBM} uses 27 irregularly sampled HealthKit variables aggregated hourly, covering activity, cardiovascular measures, vitals, mobility, body measurements, and fitness~\citep{erturk2025beyond}. 
\textsc{LSM-2} uses 26 minute-level features derived from PPG, accelerometry, skin conductance, skin temperature, and altimetry~\citep{xu2025lsm}. \openmhc{} provides 19 minute-level channels covering activity, heart rate, sleep, and workout types (Table~\ref{tab:healthkit_summary}); seven sparse HealthKit variables, including heart-rate variability, resting heart rate, and VO\textsubscript{2}max, are released separately as contextual variables (Appendix~\ref{app:label_registry}). Thus, \openmhc{} shares HealthKit-derived behavioral measures with \textsc{WBM} and minute-level resolution with \textsc{LSM-2}, while measuring different sets of variables.


\clearpage
\section{Data Preprocessing}
\label{app:shared_wearable_preprocessing}
\label{app:preprocessing}

This appendix describes how raw My Heart Counts app collected HealthKit records are turned into activity, physiology, sleep, and workout records, which are released and used throughout the benchmark. We first define the daily minute-level matrix construction pipeline, then describe the additional filtering and aggregation steps used to construct daily, hourly, and weekly model inputs.

\subsection{Basic Data Cleaning: Construction of Daily Matrices}
\label{app:daily_matrix_construction}

This section describes how raw wearable data from the MHC study are converted into daily minute-level matrices. The raw data are heterogeneous: participants use different devices and iOS versions, records arrive with free-text source identifiers rather than structured device metadata, and measurement coverage varies substantially across participants and time periods as the app evolved together with HealthKit and included more and more HealthKit types as time went on. Initial database files are stored as CSV records on a daily level, leading to TBs of raw data. In a first step, we extract and transform these raw files with evolving schemas, remove corrupted entries, and de-duplicate rows to create an initial data lake version of the raw HealthKit data in parquet format (for full transparency, this process is documented here: \url{https://github.com/NarayanSchuetz/myheart-counts-client}). 

We transform per-participant Parquet exports of HealthKit, sleep, and workout records into per-participant HDF5 files, each containing daily matrices $d \in \real^{19 \times 1440}$ (19 channels at minute resolution). The pipeline proceeds through device type inference, dominant source selection, record-level cleaning, temporal alignment, data stream availability detection, daily matrix generation, and hourly aggregation.

\bo{Device Type Inference.} 
From the raw HealthKit data, it is often unclear what type of device each individual has (e.g., iPhone, Apple Watch, Garmin, etc.). 
Therefore, we first assign each raw source to a device class: iPhone, Apple Watch, or ambiguous. In the raw HealthKit records, each source is represented by a free-text source name (e.g., ``iPhone~14 Pro'', ``Apple Watch Series~8'') rather than structured device type identifiers. We apply the staged heuristic in Algorithm~\ref{alg:device_inference} to classify each record's source. Let $\mathcal{T}_{\text{watch}} = \{\text{StepCount}, \text{DistanceWalkingRunning}, \text{ActiveEnergyBurned}, \text{HeartRate}\}$ and $\mathcal{T}_{\text{phone}} = \{\text{StepCount}, \text{DistanceWalkingRunning}, \text{FlightsClimbed}\}$ denote the expected type signatures for each device class. Ambiguous devices are filtered out as we found their data to be too inconsistent, while likely iPhone and likely Apple Watch sources are propagated to their respective Apple device class. This decision was made because manual inspection showed that the vast majority are most likely Apple devices, and retaining them maximizes data availability.

\begin{algorithm}[h]
\caption{Device Type Inference}\label{alg:device_inference}
\KwIn{Source name $n$, bundle identifier $b$, set of reported HealthKit types $\mathcal{T}$}
\KwOut{Device type label $\ell \in \{\text{iPhone}, \text{AppleWatch}, \text{ambiguous}\}$}
\BlankLine
\tcp{Stage 1: Name-based hints}
\uIf{$n$ contains ``phone'' (case-insensitive)}{$\ell \leftarrow \text{likely\_iPhone}$}
\uElseIf{$n$ contains ``watch'' (case-insensitive)}{$\ell \leftarrow \text{likely\_AppleWatch}$}
\Else{$\ell \leftarrow \text{unassigned}$}
\BlankLine
\tcp{Stage 2: Apple identifier confirmation}
\If{$b$ starts with \texttt{com.apple}}{
    \lIf{$\ell = \text{likely\_iPhone}$}{$\ell \leftarrow \text{iPhone}$}
    \lIf{$\ell = \text{likely\_AppleWatch}$}{$\ell \leftarrow \text{AppleWatch}$}
}
\BlankLine
\tcp{Stage 3: Type signature classification (unresolved sources only)}
\If{$\ell \notin \{\text{iPhone}, \text{AppleWatch}\}$}{
    \uIf{$\mathcal{T}_{\text{watch}} \subseteq \mathcal{T}$}{$\ell \leftarrow \text{likely\_AppleWatch}$}
    \uElseIf{$\mathcal{T}_{\text{phone}} \subseteq \mathcal{T}$ \textbf{and} $\{\text{ActiveEnergyBurned}, \text{HeartRate}\} \cap \mathcal{T} = \emptyset$}{$\ell \leftarrow \text{likely\_iPhone}$}
    \tcp{Repeat Apple identifier confirmation for newly labeled sources}
    \If{$b$ starts with \texttt{com.apple}}{
        \lIf{$\ell = \text{likely\_iPhone}$}{$\ell \leftarrow \text{iPhone}$}
        \lIf{$\ell = \text{likely\_AppleWatch}$}{$\ell \leftarrow \text{AppleWatch}$}
    }
}
\BlankLine
\tcp{Stage 4: Fallback}
\lIf{$\ell \notin \{\text{iPhone}, \text{AppleWatch}, \text{likely\_iPhone}, \text{likely\_AppleWatch}\}$}{$\ell \leftarrow \text{ambiguous}$}
\Return{$\ell$}
\end{algorithm}

\bo{Dominant Source Selection.}
Another issue we encountered is that some participants have multiple devices of the same type contributing data simultaneously, for instance, when upgrading to a new Apple Watch while the previous device continues to sync. To avoid double-counting, we select a single \emph{dominant source} per device type (iPhone, Apple Watch) per calendar day. A source is considered valid for a given day only if it reported all required HealthKit types for its device class: StepCount and DistanceWalkingRunning for iPhones; those two plus HeartRate and ActiveEnergyBurned for Apple Watches. Among valid sources, the one with the highest record count is selected as dominant; ties result in no dominant source for that device type on that day. Only records from the dominant iPhone and dominant Apple Watch sources are retained for later matrix construction.

\bo{Record-level cleaning.}
Before the temporal aggregation, we remove records with inconsistent timestamps, invalid durations, or implausible measurement values. Three universal checks are applied to all record types: records with mismatched timezone offsets between start and end times (indicating timezone transitions during recording, creating potential artifacts), records with negative duration, and records exceeding 24 hours are removed (which we found to mostly be anomalies). We additionally apply type-specific rate and range thresholds to remove physiologically implausible values. Active energy values are converted from kilocalories to calories where applicable. Specific threshold settings are defined in Table~\ref{tab:quality_check_thresholds}.

\begin{table}[ht]
\centering
\small
\renewcommand{\arraystretch}{1.15}
\caption{Per-type quality filter thresholds. Records exceeding the maximum rate or falling outside the valid range are removed.}
\label{tab:quality_check_thresholds}
\begin{tabular}{lll}
\toprule
\bo{HealthKit Type} & \bo{Filter} & \bo{Threshold} \\
\midrule
StepCount & Max rate & 5 steps/s \\
DistanceWalkingRunning & Max rate & 10 m/s \\
ActiveEnergyBurned & Max rate & 250 cal/s \\
DistanceCycling & Max rate & 25 m/s \\
AppleStandTime & Max rate & 1 min/min \\
HeartRate & Valid range & 40--200 bpm \\
\bottomrule
\end{tabular}
\end{table}

\bo{Temporal Alignment (Midnight Splitting).}
To provide daily matrices, records spanning midnight are split at the day boundary so that each day receives its respective part of the record. For count-based quantities (steps, distance, energy, flights climbed, stand time), the value is allocated proportionally to the fraction of the record's duration falling on each side of midnight. For rate quantities (heart rate) the value is preserved unchanged on both sides. Sleep and workout intervals are similarly split at midnight with adjusted durations. Note, Apple HealthKit data is recorded in the user's local time, and while it logs timezones, those are unreliable and only valid for live records since the recorded timezone is based on the timezone at the time the data is extracted from HealthKit; to avoid confusion, we do not adjust timezones and simply provide data in the user's local time.

\begin{table}[ht]
\centering
\caption{Value semantics in the daily matrix.}
\label{tab:nan_semantics}
\resizebox{0.9\textwidth}{!}{%
\begin{tabular}{llll}
\toprule
\textbf{Channel Group} & \textbf{Presence Criterion} & \textbf{Zero} & \textbf{NaN} \\
\midrule
Continuous HK (0--6) & $\geq$10\% of days & No activity & Stream unavailable \\
Sleep (7--8) & Any occurrence & Not asleep & No sleep data \\
Workouts (9--18) & Any occurrence & Not exercising & Never logged \\
\bottomrule
\end{tabular}
}
\end{table}

\begin{figure}
    \centering
    \includegraphics[width=0.8\linewidth]{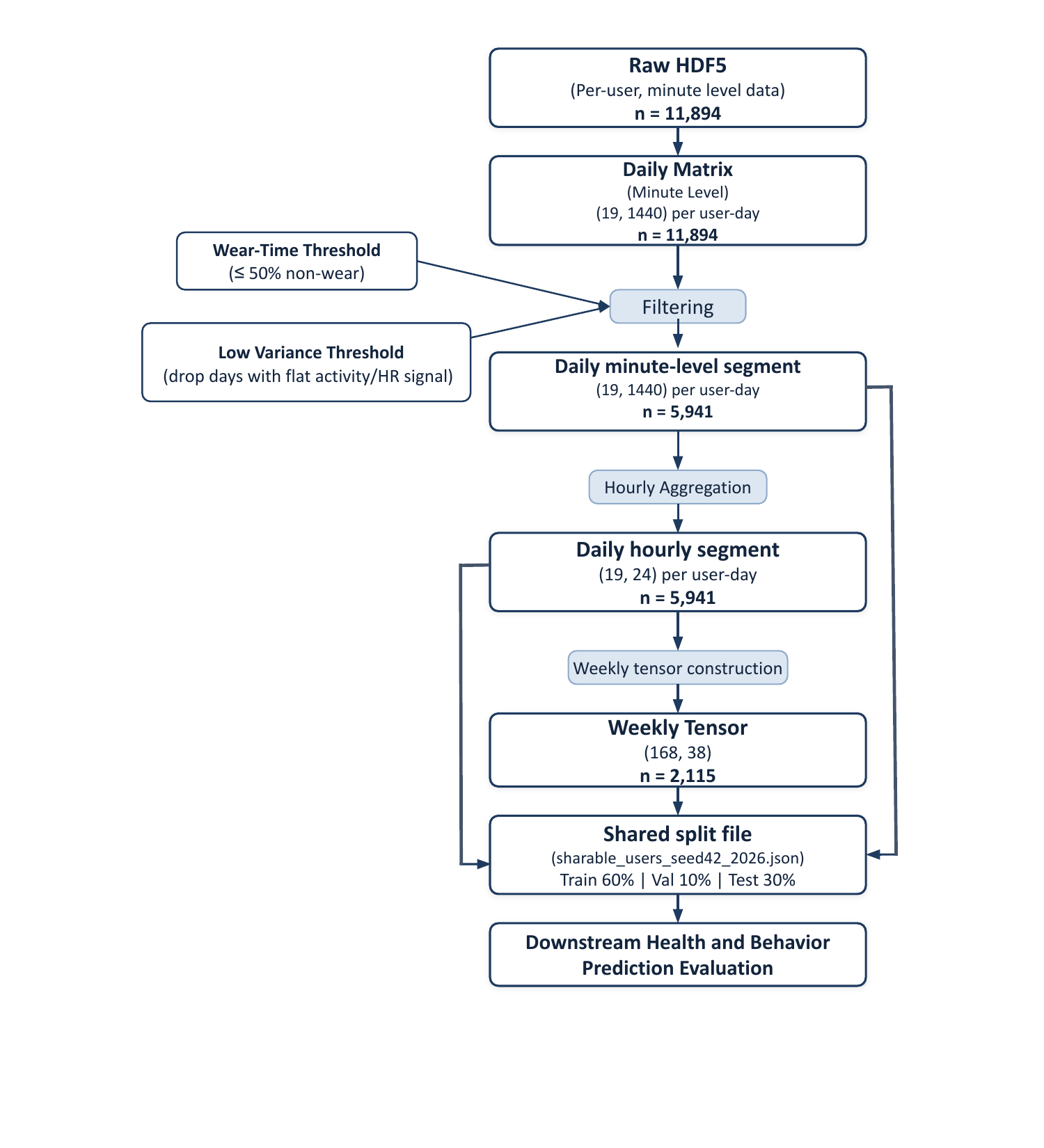}
    \caption{
    Data preprocessing flow for the downstream health and behavior prediction evaluation. The initial participant filtering process is shown in Figure~\ref{fig:consort}, and daily matrix construction is described in Section~\ref{app:daily_matrix_construction}.
    }
    \label{fig:data_preprocessing_downstream_eval}
\end{figure}

\bo{Daily Matrix Generation.}
Since HealthKit data is recorded as intervals and values for many types, and we require minute-level data, records were first spread across an 86{,}400-element second-level array (one value per second), then resampled to 1{,}440 minute-level bins: count-based channels (steps, distance, energy, flights, stand time) are summed per minute, while heart rate is averaged. Overlapping records at the same second bin are resolved via \texttt{nanmean}. Sleep and workout intervals produce binary minute-level indicator arrays (1 during the interval, 0 otherwise; overlapping intervals use OR semantics). The 7 continuous, 2 sleep, and 10 workout channels are stacked to form the daily matrix $d \in \real^{19 \times 1440}$.

\bo{Data Stream Detection and Non-Wear Time Heuristic.}
\label{par:nan_semantics}
Not all participants have the same wearable channels: many lack an Apple Watch entirely, and workout logging varies widely. To distinguish \emph{true inactivity} (zero) from \emph{structural absence} (NaN), we determine per-participant data stream availability before generating daily matrices. For each participant and continuous HealthKit channel, we compute the fraction of the participant's observed calendar days on which that channel reports any data. If this fraction is below 10\%, the channel is treated as unavailable for that participant and filled with NaN on all days. Otherwise, the channel is treated as available, and missing values on individual days are represented through the standard NaN/mask convention. For sleep and workout channels, any occurrence across the participant's history is sufficient to mark the stream as available. Table~\ref{tab:nan_semantics} summarizes the resulting value semantics.

We try to estimate non-wear periods by identifying consecutive time intervals where all 7 continuous HealthKit channels (indices 0--6) simultaneously record zero or NaN. Consecutive runs exceeding 30 minutes are flagged as non-wear. Sleep and workout channels (indices 7--18) are excluded from this detection, as they represent behavioral signals rather than passive sensor readings (for instance sleep can be non-wear). The resulting binary non-wear vector is used to compute per-day wear-time statistics.

\bo{Per-Channel Missing Data Detection.}
Beyond general non-wear periods that affect all channels, we independently for each of the 19 channels, identify consecutive runs of $\geq$120 minutes of zero or NaN values as missing intervals. These per-channel missing intervals are stored as metadata alongside each daily matrix in a HDF5 file, providing a compact summary of long gaps in each data stream.

\bo{Output Format.}
The pipeline produces one gzip-compressed HDF5 file per participant, with one dataset per calendar day. Each dataset stores the $19 \times 1440$ daily matrix together with metadata attributes including the non-wear vector, per-channel missing intervals, and total wear-time minutes.

\subsection{Self-Reported Variable Sanitation}
\label{app:label_cleaning}
Self-reported outcomes were cleaned to ensure physiological validity. For instance, to address implausible outliers in height and weight, specific thresholds were implemented based on clinical guidelines and physiological limits. Table~\ref{tab:label_assumptions} summarizes the inclusion/filtering criteria and processing steps applied to the self-reported variables exposed through the labels API in the code base.

\begin{table}[H]
\centering
\caption{Assumptions and filtering of final self-reported outcome labels}
\label{tab:label_assumptions}
\begin{tabular}{@{}lll@{}}
\toprule
\textbf{Metrics} & \textbf{Variable Names} & \textbf{Considerations} \\ \midrule
Blood Pressure & \begin{tabular}[c]{@{}l@{}}SystolicBloodPressure, \\ blood\_pressure\_categories\end{tabular} & \begin{tabular}[c]{@{}l@{}}Swapped if Systolic BP was lower \\ than Diastolic BP; both values \\ required. Diastolic BP is not \\ released; it is used only for this \\ swap and the categories\end{tabular} \\ \addlinespace
BMI & \begin{tabular}[c]{@{}l@{}}BMI\_values, \\ BMI\_categories\end{tabular} & \begin{tabular}[c]{@{}l@{}}Height 1.4--2.1m, weight $\ge$ 40kg, \\ and BMI $\le$ 60\end{tabular} \\ \addlinespace
Body weight & WeightKilograms & \begin{tabular}[c]{@{}l@{}}Entries of exactly 30kg or 300kg \\ removed as sentinel values\end{tabular} \\ \addlinespace
Wake-up time \& bedtime & \begin{tabular}[c]{@{}l@{}}WakeUpTime\_categories, \\ GoSleepTime\_categories\end{tabular} & \begin{tabular}[c]{@{}l@{}}Profile times converted to \\ decimal hours, then binned\end{tabular} \\ \addlinespace
Psychological & \begin{tabular}[c]{@{}l@{}}feel\_worthwhile1--4, \\ satisfiedwith\_life, \\ happiness\_categories\end{tabular} & Thresholds based on distributions \\ \addlinespace
Happiness (long) & happiness & Filtered to $\ge$ 3 responses per participant \\ \addlinespace
Activity & vigorous\_act & \begin{tabular}[c]{@{}l@{}}Minutes per week as entered; \\ no range filter or categories applied\end{tabular} \\ \bottomrule
\end{tabular}
\end{table}

\subsection{Wearable Data Preprocessing for the Benchmark}
\label{app:model_input_representations}

After constructing the daily wearable matrices, we apply additional preprocessing to define the wearable data used by the models we train for the benchmark. Note that other dataset preprocessing approaches could be used at this step. To standardize the evaluation, we apply these filtering steps and require users who want to submit their models to our leaderboard to evaluate on these samples (it is up to the submitter whether they use the same filtering for their training dataset).

\subsubsection{Data Filtering}
\label{app:data_filtering}
\bo{Wear-Time Filtering.}
For all evaluation tasks, we retain only calendar days whose non-wear vector contains at most 720 non-wear minutes, equivalently to at least 12 hours of estimated wear time ($\leq 50\%$ non-wear). The 12-hour cutoff was chosen to balance day-level signal quality, cohort size, and cohort representativeness; Appendix~\ref{app:inclusion_ablations} reports a sensitivity analysis supporting this choice.

\bo{Low-Variance Filtering.}
\label{par:low_variance_filter}
As a second day-level removal criterion applied to the retained days, we remove days with near-constant sensor traces while being above zero (likely sensor artifacts/glitches) by requiring minimum within-day variance on the monitored continuous channels. The thresholds are $1.0$ for iPhone steps, iPhone distance, Watch steps, Watch distance, and Watch active energy, and $10^{-4}$ for Watch heart rate; the flights-climbed channel is excluded. Channels with undefined variance because of insufficient observed values are not removed by this rule. We refer to days that pass both the wear-time and low-variance filters as \emph{retained days}.

\bo{Representation-level NaN handling.}
On the days that survive the wear-time and low-variance filters, we apply a value-level zero-to-NaN transform that re-labels missingness without removing any further days. It reinterprets values that are semantically implausible or unreliable as missing: heart-rate values equal to zero are set to NaN; for the step, distance, and active-energy channels, the entire day of that channel is set to NaN if it is all zero throughout the day; and for the two sleep channels, zero-valued minutes are set to NaN when the total detected sleep on that day is less than 3 hours. Flights climbed and workout channels are left unchanged. These filtering and re-labeling steps are applied when constructing the hourly and weekly benchmark representations and related feature-extraction pipelines.

\bo{Weekly Data Filtering.}
We define a \emph{retained week} as a candidate 7-day window containing at least five retained days. Candidate windows with fewer than five retained days are discarded and are not used for weekly-tensor model inputs.

\subsubsection{Data Aggregation}
\label{app:data_aggregation}
\bo{Daily minute-level segment.}
A daily minute-level segment corresponds to one filtered participant-day at minute resolution, represented as
\begin{equation}
    X^{\mathrm{min}}_{i,d} \in \mathbb{R}^{19 \times 1440},
\end{equation}
where $i$ denotes the participant and $d$ the calendar day. This segment preserves within-day minute resolution and is used by methods that operate directly on minute-level daily records.

\bo{Hourly Aggregation.}
To transform a daily matrix from minute-level to hourly level, it is aggregated from minute resolution to one value per hour. Count-like continuous channels are summed (NaNs removed), whereas heart rate is averaged (NaNs removed). Sleep and workout channels are binary and use OR semantics: an hourly value is 1 if any minute in that hour is 1, and 0 if all observed minutes are 0. If all minute-level values for a channel within an hour are NaN, the hourly entry is treated as missing and is stored as a zero-filled value together with a binary indicator of missingness.

\bo{Daily hourly segment.}
The hourly segment for a retained day is stored as a pair of arrays
\begin{equation}
    V^{\mathrm{hr}}_{i,d} \in \mathbb{R}^{19 \times 24},
    \qquad
    M^{\mathrm{hr}}_{i,d} \in \{0,1\}^{19 \times 24},
\end{equation}
where $V^{\mathrm{hr}}_{i,d}$ contains hourly aggregated values and $M^{\mathrm{hr}}_{i,d}$ is the corresponding binary missingness indicator. For methods that require a time-major segment, the hourly values and missingness indicators are transposed to shape $(24,19)$ and concatenated to form a $(24,38)$ tensor.

\bo{Weekly Tensors.}
\label{app:weekly_tensors}
A weekly tensor corresponds to a 7-day sequence of hourly aggregated wearable data. We use rolling 7-day windows to find each available week window. For each participant, we generate candidate 7-day windows starting from the participant's first available wearable date and advance the window seven calendar days at a time. Each candidate window is represented as
\begin{equation}
    X^{\mathrm{week}}_{i,s} \in \mathbb{R}^{168 \times 38},
\end{equation}
where $s$ indexes the candidate 7-day window and the 168 time steps correspond to 7 consecutive days at hourly resolution. Each hourly step contains 19 aggregated wearable values and 19 binary missingness indicators.

\subsection{Effective Sample Sizes per Track}
\label{app:effective_sample_sizes}
Table~\ref{tab:effective_sample_sizes} reports the participants and samples that remain after data filtering for each track. 
Note that these filters are not required for models utilizing the benchmark and were chosen to balance dataset size and signal quality. 
The wear-time filter (Appendix~\ref{app:data_filtering}) removes 59.0\% of the 2,822,500 released participant-days, and the low-variance filter removes a further 1.8\% of the remaining days. This leaves 1,136,862 retained days (40.3\% of released days) from 5,941 of the 11,894 participants (Figure~\ref{fig:data_preprocessing_downstream_eval}).
Note that Track~1 (Prediction) cohorts differ by outcome because of the task-specific inclusion windows (Appendix~\ref{app:health_outcome_inclusion_and_context}). Table~\ref{tab:track1_split_sizes} gives per-outcome participants and participant-days for each split, and Table~\ref{tab:health_outcome_full} outcome prevalences for binary/categorical/ordinal outcomes. 

\bo{Channel coverage.}
Table~\ref{tab:channel_hours} reports, for each sensor channel and split, the participants with at least one non-zero value on retained days (counted as in Table~\ref{tab:healthkit_summary}) and the observed channel-hours. A channel-hour is observed if at least one minute of that channel is non-missing after the zero-to-NaN transform (Appendix~\ref{app:data_filtering}). Observed hours may have value zero, for example no steps during sleep. For the binary sleep and workout channels, an observed zero means not asleep, not in bed, or not exercising, so their observed hours far exceed the time spent in these states.

\begin{table}[h]
\centering
\footnotesize
\setlength{\tabcolsep}{5pt}
\renewcommand{\arraystretch}{1.1}
\caption{\textbf{Effective sample sizes by stage, track, and split (participants / samples).} Samples are participant-days. Track~1 sizes vary by outcome (Table~\ref{tab:track1_split_sizes}). Track~2A validation and test samples are capped at 91 days per participant, and Track~2B validation and test windows at 100 per participant; training samples are not capped. 
}
\label{tab:effective_sample_sizes}
\begin{tabular}{@{}lrrr@{}}
\toprule
\textbf{Stage} & \textbf{Train} & \textbf{Validation} & \textbf{Test} \\
\midrule
\multicolumn{4}{@{}l}{\textit{Day-level filtering (Appendix~\ref{app:data_filtering})}} \\
\quad Released cohort (all days) & 7,136 / 1,682,478 & 1,189 / 285,540 & 3,569 / 854,482 \\
\quad Wear-time filter ($\geq$12\,h wear) & 3,602 / 681,251 & 630 / 124,327 & 1,799 / 352,570 \\
\quad Retained days (+ low variance) & 3,542 / 666,175 & 622 / 122,107 & 1,777 / 348,580 \\
\multicolumn{4}{@{}l}{\textit{Track~1: Health and Behavior Prediction}} \\
\quad Largest outcome (Age) & 3,429 / 656,086 & 606 / 118,941 & 1,730 / 342,145 \\
\quad Smallest outcome (Framingham) & 544 / 179,169 & 100 / 29,308 & 276 / 100,046 \\
\quad Participant--outcome pairs (32 outcomes) & 61,797 & 10,592 & 30,999 \\
\multicolumn{4}{@{}l}{\textit{Track~2A: Imputation}} \\
\quad Samples (participant-days) & 3,542 / 666,175 & 622 / 17,923 & 1,777 / 49,942 \\
\multicolumn{4}{@{}l}{\textit{Track~2B: Forecasting}} \\
\quad 24-hour forecast windows & 1,621 / 596,047 & 289 / 16,061 & 827 / 43,563 \\
\bottomrule
\end{tabular}
\end{table}

\begin{table}[h]
\centering
\footnotesize
\setlength{\tabcolsep}{5pt}
\renewcommand{\arraystretch}{1.1}
\caption{\textbf{Track~1 sample sizes per outcome and split} (included participants / participant-days). Participant-days are all retained days of the included participants. Outcomes listed together have identical cohorts; the cardiovascular conditions are Atrial Fibrillation, Cardiovascular Disease, Cerebrovascular Disease, Congenital Heart Disease, Coronary Artery Disease, Heart Failure / CHF, Pulmonary Hypertension, and Vascular Disease. Totals across splits equal the Included Participants column of Table~\ref{tab:health_outcome_full}.}
\label{tab:track1_split_sizes}
\begin{tabular}{@{}lrrr@{}}
\toprule
\textbf{Outcome} & \textbf{Train} & \textbf{Validation} & \textbf{Test} \\
\midrule
\multicolumn{4}{@{}l}{\textit{Demographics}} \\
\quad Age & 3,429 / 656,086 & 606 / 118,941 & 1,730 / 342,145 \\
\quad Biological Sex & 3,341 / 602,448 & 588 / 111,903 & 1,674 / 317,426 \\
\multicolumn{4}{@{}l}{\textit{Medical Conditions \& Risk}} \\
\quad Cardiovascular conditions (8) & 2,197 / 446,973 & 368 / 69,925 & 1,091 / 242,231 \\
\quad Diabetes, Hypertension & 1,048 / 291,237 & 174 / 47,287 & 525 / 173,133 \\
\quad Framingham CVD Risk & 544 / 179,169 & 100 / 29,308 & 276 / 100,046 \\
\quad Sleep Disorder Diagnosis & 2,596 / 484,452 & 444 / 84,079 & 1,320 / 264,867 \\
\multicolumn{4}{@{}l}{\textit{Vitals \& Blood Biomarkers}} \\
\quad BMI Categories, BMI Value & 2,660 / 506,320 & 469 / 91,708 & 1,328 / 262,561 \\
\quad Blood Pressure Categories, Systolic BP & 890 / 278,202 & 151 / 41,419 & 458 / 162,767 \\
\quad Body Weight & 2,669 / 509,399 & 471 / 93,875 & 1,333 / 262,890 \\
\quad HDL Cholesterol & 797 / 259,741 & 138 / 43,052 & 406 / 152,983 \\
\quad LDL Cholesterol & 668 / 251,316 & 107 / 37,005 & 352 / 138,905 \\
\quad Total Cholesterol & 841 / 268,836 & 146 / 43,552 & 431 / 161,238 \\
\multicolumn{4}{@{}l}{\textit{Mental Well-being}} \\
\quad Feel Depressed & 1,183 / 268,925 & 211 / 42,672 & 596 / 123,580 \\
\quad Feel Happy & 1,680 / 402,937 & 287 / 62,333 & 843 / 215,245 \\
\quad Feel Worried & 1,570 / 355,099 & 273 / 57,701 & 799 / 188,671 \\
\quad Life Satisfaction & 1,676 / 400,122 & 285 / 62,271 & 842 / 215,213 \\
\quad Things Are Worthwhile & 1,677 / 398,280 & 284 / 61,687 & 842 / 214,061 \\
\multicolumn{4}{@{}l}{\textit{Sleep \& Lifestyle}} \\
\quad Bedtime & 2,725 / 519,019 & 476 / 97,662 & 1,354 / 276,578 \\
\quad Currently Employed & 2,325 / 465,616 & 394 / 78,047 & 1,178 / 252,362 \\
\quad Sleep Duration & 2,324 / 465,707 & 393 / 78,046 & 1,174 / 247,776 \\
\quad Vigorous Activity Minutes & 2,278 / 462,212 & 384 / 76,668 & 1,141 / 249,024 \\
\quad Wake-up Time & 2,702 / 520,544 & 473 / 96,306 & 1,358 / 274,874 \\
\bottomrule
\end{tabular}
\end{table}

\clearpage
\begin{table}[H]
\centering
\footnotesize
\setlength{\tabcolsep}{4pt}
\renewcommand{\arraystretch}{1.1}
\caption{\textbf{Channel coverage on retained days by split.} \emph{Participants}: participants with at least one non-zero value on retained days, counted as in Table~\ref{tab:healthkit_summary}. \emph{Hours}: observed channel-hours, i.e., hours in which at least one minute of the channel is non-missing; k = thousand hours. For the binary sleep and workout channels, observed hours include hours with value 0 (not asleep, not in bed, not exercising).}
\label{tab:channel_hours}
\begin{tabular}{@{}lrrrrrr@{}}
\toprule
 & \multicolumn{2}{c}{\textbf{Train}} & \multicolumn{2}{c}{\textbf{Validation}} & \multicolumn{2}{c}{\textbf{Test}} \\
\cmidrule(lr){2-3} \cmidrule(lr){4-5} \cmidrule(l){6-7}
\textbf{Channel} & \textbf{Participants} & \textbf{Hours} & \textbf{Participants} & \textbf{Hours} & \textbf{Participants} & \textbf{Hours} \\
\midrule
\multicolumn{7}{@{}l}{\textit{Phone}} \\
\quad StepCount & 3,437 & 15,345k & 603 & 2,726k & 1,719 & 8,057k \\
\quad DistanceWalkingRunning & 3,437 & 15,349k & 602 & 2,727k & 1,719 & 8,059k \\
\quad FlightsClimbed & 1,217 & 15,204k & 220 & 2,696k & 647 & 7,708k \\
\multicolumn{7}{@{}l}{\textit{Watch}} \\
\quad StepCount & 3,004 & 15,265k & 532 & 2,812k & 1,531 & 8,038k \\
\quad DistanceWalkingRunning & 2,997 & 15,048k & 532 & 2,810k & 1,528 & 7,981k \\
\quad HeartRate & 3,004 & 11,721k & 532 & 2,195k & 1,531 & 6,102k \\
\quad ActiveEnergyBurned & 2,991 & 15,049k & 532 & 2,811k & 1,524 & 7,981k \\
\multicolumn{7}{@{}l}{\textit{Sleep}} \\
\quad Asleep & 1,052 & 8,692k & 205 & 1,559k & 583 & 4,584k \\
\quad InBed & 1,176 & 8,668k & 221 & 1,561k & 630 & 4,526k \\
\multicolumn{7}{@{}l}{\textit{Workout}} \\
\quad Walking & 1,033 & 4,711k & 182 & 757k & 534 & 2,739k \\
\quad Cycling & 574 & 4,168k & 107 & 625k & 268 & 2,314k \\
\quad Running & 575 & 3,859k & 102 & 626k & 295 & 2,383k \\
\quad Other & 485 & 3,881k & 100 & 581k & 237 & 2,236k \\
\quad Mixed Metabolic Cardio & 49 & 480k & 10 & 80k & 24 & 242k \\
\quad Strength Training & 236 & 2,189k & 50 & 382k & 133 & 1,338k \\
\quad Elliptical & 299 & 2,713k & 58 & 479k & 158 & 1,769k \\
\quad HIIT & 184 & 2,062k & 40 & 336k & 97 & 1,306k \\
\quad Functional Strength & 225 & 2,633k & 48 & 466k & 120 & 1,642k \\
\quad Yoga & 212 & 2,454k & 41 & 393k & 135 & 1,656k \\
\midrule
All 19 channels &  & 149,490k &  & 26,623k &  & 80,660k \\
\bottomrule
\end{tabular}
\end{table}

\subsection{Participant Demographics for Different Wear-Time Filtering Choices}
\label{app:inclusion_ablations}

We assessed whether the choice of wear-time filtering requirements affected participant demographics. We evaluated four wear-time filtering thresholds: 
estimated wear times of 
at least 18, 12, 6, or 0 hours per day (equivalently at most 25\%, 50\%, 75\%, or 100\% non-wear). The 12-hour threshold ($\leq 50\%$ non-wear) is the benchmark default (Appendix~\ref{app:data_filtering}). 
Overall the choice of threshold did not greatly change participant demographics; Table~\ref{tab:nonwear_cohort} summarizes our results. 

\begin{table}[H]
  \centering
  \small
  \renewcommand{\arraystretch}{1.15}
  \caption{Cohort size and demographic composition across non-wear thresholds. Values marked $^{\ast}$ are statistically significantly different ($p<0.05$) from the unfiltered baseline (KS test for continuous variables, $\chi^2$ test for categorical variables).} 
  \label{tab:nonwear_cohort}
  \begin{tabular}{lrrrr}
  \toprule
   & \bo{$\leq 25\%$} & \bo{$\leq 50\%$} & \bo{$\leq 75\%$} & No filter \\
   & \small$\geq 18\,$h wear & \small$\geq 12\,$h wear & \small$\geq 6\,$h wear & \small(all days) \\
  \midrule
  Participants Retained                & 2{,}920 & 6{,}031 & 9{,}661 & 11{,}894 \\
  Age (mean $\pm$ SD)   & $46.1\pm15.4^{\ast}$ & $44.8\pm14.9^{\ast}$ & $44.6\pm15.2$ & $44.4\pm15.1$ \\
  BMI (mean $\pm$ SD)   & $27.7\pm6.1$         & $27.5\pm5.9^{\ast}$ & $27.3\pm5.8$ & $27.2\pm5.8$ \\
  Hypertension \%       & $29.4^{\ast}$        & 26.8                 & 26.2         & 25.9 \\
  Diabetes \%           & 7.2                  & 6.0                  & 5.8          & 5.8 \\
  CVD \%                & $14.1^{\ast}$        & 11.4                 & 10.9         & 10.9 \\
  Male \%               & 75.5                 & $78.9^{\ast}$        & 78.2         & 77.1 \\
  \bottomrule
  \end{tabular}
  \end{table}

\clearpage
\section{Prediction Tasks}
\label{app:health_outcome_prediction}

This appendix section specifies the prediction tasks and their evaluation protocol. The benchmark contains 32 static prediction tasks constructed from MHC source variables, including survey fields, HealthKit measurements, and derived quantities.

\subsection{Prediction Task Definitions}
\label{app:health_outcome_tasks}

\subsubsection{Outcome Types and Construction}
\label{app:health_outcome_types}

{\small
\renewcommand{\arraystretch}{1.3}
\begin{longtable}{>{\raggedright\arraybackslash}p{0.20\linewidth}
                  >{\raggedright\arraybackslash}p{0.32\linewidth}
                  >{\raggedright\arraybackslash}p{0.10\linewidth}
                  >{\raggedright\arraybackslash}p{0.30\linewidth}}
\caption{32 prediction tasks from self-reported health outcomes and HealthKit measurements.}
\label{tab:health_outcome_construction}\\
\toprule
\textbf{Outcome} & \textbf{Self-report question/source} & \textbf{Type} & \textbf{Construction details} \\
\midrule
\endfirsthead
\toprule
\textbf{Outcome} & \textbf{Self-report question/source} & \textbf{Type} & \textbf{Construction details} \\
\midrule
\endhead

\rowcolor{white}\multicolumn{4}{l}{\textit{Demographics (2 outcomes)}} \\
\midrule
Age & \texttt{enrollment\_info.json} birthdate and \texttt{last\_labels.json} last-survey timestamp & Continuous & Whole years from birthdate to last-survey timestamp; participants with computed age outside $[0,100]$ are dropped. \\
Biological Sex & ``Gender'', recorded by Apple's HealthKit as one of \texttt{Female} / \texttt{Male} / \texttt{Other} (HealthKit field \texttt{HKBiologicalSex}) & Binary & Male is encoded as $1$ and Female as $0$; "Other" option absent in observed data. \\

\midrule
\rowcolor{white}\multicolumn{4}{l}{\textit{Medical Conditions \& Risk (12 outcomes)}} \\
\midrule
Atrial Fibrillation & \texttt{heart\_disease}: ``Have you been diagnosed with any of the below diseases?'' & Binary & $1$ if the participant selected ``Atrial fibrillation (Afib)''. \\
Cardiovascular Disease & \texttt{heart\_disease}: ``Have you been diagnosed with any of the below diseases?''; \texttt{vascular}: ``Which vascular disease diagnosis have you received?'' & Binary & $1$ if the participant selected any non-``None of the above'' option in either multi-select question. \\
Cerebrovascular Disease & \texttt{vascular}: ``Which vascular disease diagnosis have you received?'' & Binary & $1$ if the participant selected Stroke, Transient Ischemic Attack, Carotid Artery Blockage/Stenosis, or Carotid Artery Surgery/Stent. \\
Congenital Heart Disease & \texttt{heart\_disease}: ``Have you been diagnosed with any of the below diseases?'' & Binary & $1$ if the participant selected ``Congenital Heart Defect''. \\
Coronary Artery Disease & \texttt{heart\_disease}: ``Have you been diagnosed with any of the below diseases?'' & Binary & $1$ if the participant selected any of Heart Attack/Myocardial Infarction, Heart Bypass Surgery, Coronary Blockage/Stenosis, Coronary Stent/Angioplasty, Angina, or High Coronary Calcium Score. \\
Diabetes & ``Do you have Diabetes?'' & Binary & Yes/no response. \\
Framingham CVD Risk & Age, Biological Sex, Total Cholesterol, HDL Cholesterol, Systolic Blood Pressure, hypertension-treatment status, smoking, and Diabetes & Continuous, $[0,1]$ & \cite{framingham_2008} Framingham 10-year CVD risk; values are clipped to $[0,1]$; cohort restricted to age 30--79, with participants with prior CVD or diabetes excluded. The age range and diabetes exclusion are benchmark-specific; ages 75--79 extend beyond the original 30--74 derivation range. \\
Heart Failure / CHF & \texttt{heart\_disease}: ``Have you been diagnosed with any of the below diseases?'' & Binary & $1$ if the participant selected ``Heart Failure or CHF''. \\
Hypertension & ``Are you being treated for Hypertension (High Blood Pressure)?'' & Binary & Yes/no response; denotes treatment status, not measured blood-pressure elevation. \\
Pulmonary Hypertension & \texttt{heart\_disease}: ``Have you been diagnosed with any of the below diseases?''; \texttt{vascular}: ``Which vascular disease diagnosis have you received?'' & Binary & $1$ if the participant selected ``Pulmonary Hypertension'' in \texttt{heart\_disease} or ``Pulmonary Arterial Hypertension'' in \texttt{vascular}. \\
Sleep Disorder Diagnosis & ``Have you ever been told by a doctor or other health professional that you have a sleep disorder?'' & Binary & Yes/no response. \\
Vascular Disease & \texttt{vascular}: ``Which vascular disease diagnosis have you received?'' & Binary & $1$ if the participant selected Peripheral Vascular Disease or Abdominal Aortic Aneurysm. \\

\midrule
\rowcolor{white}\multicolumn{4}{l}{\textit{Vitals \& Blood Biomarkers (8 outcomes)}} \\
\midrule
BMI Categories & BMI Value & Ordinal (K$=$5) & BMI binned using cut points 19.9, 24.9, 29.9, and 39.9 into Underweight, Normal weight, Overweight, Obesity, and Morbid Obesity. \\
BMI Value & HealthKit \texttt{BodyMass} and HealthKit \texttt{Height} & Continuous & Computed as $\text{weight}_\text{kg}/\text{height}_\text{m}^2$. \\
Blood Pressure Categories & Systolic Blood Pressure and Diastolic Blood Pressure & Ordinal (K$=$4) & Compound rule: Normal if SBP$<120$ and DBP$<80$; Elevated if SBP 120--129 and DBP$<80$; Hypertension Stage~1 if SBP 130--139 or DBP 80--89; Hypertension Stage~2 if SBP$\geq140$ or DBP$\geq90$. Swapped if DBP > SBP \\
Body Weight & HealthKit \texttt{BodyMass} passive measurement & Continuous & Numeric value in kilograms. \\
HDL Cholesterol & ``HDL Cholesterol'' & Continuous & Numeric response (US entry range 10--140\,mg/dL), harmonized to mmol/L$^{\dagger}$. \\
LDL Cholesterol & ``LDL Cholesterol'' & Continuous & Numeric response (US entry range 0--1000\,mg/dL), harmonized to mmol/L$^{\dagger}$. \\
Systolic Blood Pressure & ``Systolic Blood Pressure'' & Continuous & Numeric response, range 90--200 mmHg. \\
Total Cholesterol & ``Total Cholesterol'' & Continuous & Numeric response (US entry range 80--400\,mg/dL), harmonized to mmol/L$^{\dagger}$. \\

\midrule
\rowcolor{white}\multicolumn{4}{l}{\textit{Mental Well-Being (5 outcomes)}} \\
\midrule
Feel Depressed & ``How about depressed?'' & Ordinal (K$=$4) & 0--10 Likert response binned using cut points 1, 3, and 5 into Low, Medium, High, and Very High; higher classes correspond to higher reported distress. \\
Feel Happy & ``How about happy?'' & Ordinal (K$=$4) & 0--10 Likert response binned using cut points 4, 6, and 8 into Low, Medium, High, and Very High. \\
Feel Worried & ``How about worried?'' & Ordinal (K$=$4) & 0--10 Likert response binned using cut points 4, 6, and 8 into Low, Medium, High, and Very High. \\
Life Satisfaction & ``Overall, how satisfied are you with life as a whole these days?'' & Ordinal (K$=$4) & 0--10 Likert response binned using cut points 4, 6, and 8 into Low, Medium, High, and Very High. \\
Things Are Worthwhile & ``Overall, to what extent do you feel the things you do in your life are worthwhile?'' & Ordinal (K$=$4) & 0--10 Likert response binned using cut points 4, 6, and 8 into Low, Medium, High, and Very High. \\

\midrule
\rowcolor{white}\multicolumn{4}{l}{\textit{Sleep \& Lifestyle (5 outcomes)}} \\
\midrule
Bedtime & AppCore profile \texttt{GoSleepTime}, represented as 24-hour decimal time & Ordinal & Binned using cut points 1, 7, 19, 21, and 23; Late Sleeper appears on both sides of midnight. \\
Currently Employed & ``Do you do regular work?'' & Binary & Yes/no response. \\
Sleep Duration & ``How much sleep do you think you need every night to be rested?'' & Ordinal (K$=$4) & Continuous hours binned using cut points 6, 7, and 9 into Insufficient, Short, Normal, and Too Long. \\
Vigorous Activity Minutes & ``Overall, how many minutes of vigorous activity do you get in a week?'' & Continuous & Numeric slider, 0--2000 minutes/week. \\
Wake-up Time & AppCore profile \texttt{WakeUpTime}, represented as 24-hour decimal time & Ordinal (K$=$4) & Binned using cut points 5, 7, and 9 into Early Riser, Normal Riser, Late Riser, and Very Late Riser. \\
\bottomrule
\multicolumn{4}{@{}p{0.96\linewidth}@{}}{\scriptsize $^{\dagger}$Lipid values were entered in mg/dL (US) or mmol/L (UK, HK). We assign each response a unit using the participant's country, then ZIP code or postcode, and otherwise whether the value is a whole-number mg/dL entry; mg/dL responses are converted to mmol/L (divided by 38.67), and responses whose unit cannot be determined are excluded.} \\
\end{longtable}
}

\subsubsection{Inclusion Criteria}
\label{app:health_outcome_inclusion_and_context}

In order to ensure that participants have enough wearable data for us to consider doing prediction, we define a formal inclusion criteria that determines this. For each prediction task $\tau$ and label date $t$ associated with a participant's non-missing label for that task, we define a task-specific window
\begin{equation}
    W_{\tau}(t) = [t - \Delta^-_{\tau},\; t + \Delta^+_{\tau}],
\end{equation}
where $\Delta^-_{\tau}, \Delta^+_{\tau} \geq 0$ are the task-specific offsets specified in days. Table~\ref{tab:health_outcome_full} reports the window used for each task. A retained day is a calendar day that satisfies both the wear-time and the low-variance filters defined in the Appendix~\ref{app:data_filtering}. A participant is included if (i) the participant's label is not missing and (ii) the task-specific window contains at least one retained day.

{\footnotesize
\setlength{\LTcapwidth}{\textwidth}
\renewcommand{\arraystretch}{1.15}
\setlength{\LTleft}{\fill}
\setlength{\LTright}{\fill}
\begin{longtable}{@{} @{\hspace{12pt}}l c l c l @{}}
\caption{\bo{Overview of the 32 prediction tasks.} Each row corresponds to one prediction task defined from a health or behavior outcome, with outcomes grouped by domain. For each task, we report the task-specific offsets $\Delta^-_{\tau}$ and $\Delta^+_{\tau}$, in days, that define the inclusion window $W_\tau$. Included Participants gives the number of participants meeting the inclusion criteria (Appendix~\ref{app:health_outcome_inclusion_and_context}), pooled over splits, and their percentage of the 5,941 participants with at least one retained day after data filtering (Figure~\ref{fig:data_preprocessing_downstream_eval}); Table~\ref{tab:track1_split_sizes} gives the counts per split. Label summaries: mean\,$\pm$\,SD for continuous (regression) outcomes,
positive-class prevalence for binary outcomes, and the category distribution for
ordinal outcomes.}
\label{tab:health_outcome_full}\\
\toprule
\multicolumn{1}{@{}l}{\textbf{Outcome}} & \textbf{Source} & \textbf{Outcome Type} & \textbf{Included Participants} & \textbf{Label Summary} \\
\midrule
\endfirsthead

\multicolumn{5}{@{}l}{\footnotesize\itshape Table~\ref{tab:health_outcome_full} continued}\\[2pt]
\toprule
\multicolumn{1}{@{}l}{\textbf{Outcome}} & \textbf{Source} & \textbf{Outcome Type} & \textbf{Included} & \textbf{Label Summary} \\
\midrule
\endhead

\midrule
\multicolumn{5}{r@{}}{\footnotesize\itshape Continued on next page}\\
\endfoot

\bottomrule
\\[-6pt]
\multicolumn{5}{@{}p{0.96\linewidth}@{}}{\scriptsize
Enrollment = participant enrollment metadata.\enspace
Profile = participant-entered app profile field.\enspace
Derived = computed from one or more source variables.}
\endlastfoot

\multicolumn{5}{@{}l}{\quad\textit{Demographics}}\\
\cmidrule{1-5}
Age\enspace($\Delta^-_{\tau}=\Delta^+_{\tau}=$1095\,days)                    & Enrollment & Continuous & 5,765 (97.0\%) & 43.0 $\pm$ 15.1 years \\
Biological Sex         & HealthKit & Binary     & 5,603 (94.3\%) & Male: 79.0\% \\
\addlinespace[3pt]

\multicolumn{5}{@{}l}{\quad\textit{Medical Conditions \& Risk\enspace($\Delta^-_{\tau}=\Delta^+_{\tau}=$365\,days)}}\\
\cmidrule{1-5}
Atrial Fibrillation        & Survey & Binary & 3,656 (61.5\%) & Prevalence\ 2.0\% \\
Cardiovascular Disease     & Survey & Binary & 3,656 (61.5\%) & Prevalence\ 11.2\% \\
Cerebrovascular Disease    & Survey & Binary & 3,656 (61.5\%) & Prevalence\ 1.8\% \\
Congenital Heart Disease   & Survey & Binary & 3,656 (61.5\%) & Prevalence\ 1.0\% \\
Coronary Artery Disease    & Survey & Binary & 3,656 (61.5\%) & Prevalence\ 4.2\% \\
Diabetes                   & Survey & Binary & 1,747 (29.4\%) & Prevalence\ 6.4\% \\
Framingham CVD Risk & Derived & Continuous & 920 (15.5\%) & 0.1 $\pm$ 0.1 (10-yr prob.) \\
Heart Failure / CHF        & Survey & Binary & 3,656 (61.5\%) & Prevalence\ 0.7\% \\
Hypertension               & Survey & Binary & 1,747 (29.4\%) & Prevalence\ 27.0\% \\
Pulmonary Hypertension     & Survey & Binary & 3,656 (61.5\%) & Prevalence\ 0.9\% \\
Sleep Disorder Diagnosis   & Survey & Binary & 4,360 (73.4\%) & Prevalence\ 15.7\% \\
Vascular Disease           & Survey & Binary & 3,656 (61.5\%) & Prevalence\ 0.7\% \\
\addlinespace[3pt]

\multicolumn{5}{@{}l}{\quad\textit{Vitals \& Blood Biomarkers\enspace($\Delta^-_{\tau}=\Delta^+_{\tau}=$91\,days)}}\\
\cmidrule{1-5}
BMI Categories             & Derived & Ordinal    & 4,457 (75.0\%) & $K{=}5$ \\
BMI Value                  & Derived & Continuous & 4,457 (75.0\%) & 27.6 $\pm$ 5.9 kg/m$^2$ \\
Blood Pressure Categories  & Survey & Ordinal    & 1,499 (25.2\%) & $K{=}4$ \\
Body Weight                & HealthKit & Continuous & 4,473 (75.3\%) & 85.6 $\pm$ 20.6 kg \\
HDL Cholesterol            & Survey & Continuous & 1,341 (22.6\%) & 1.4 $\pm$ 0.6 mmol/L \\
LDL Cholesterol            & Survey & Continuous & 1,127 (19.0\%) & 2.7 $\pm$ 1.1 mmol/L \\
Systolic Blood Pressure    & Survey & Continuous & 1,499 (25.2\%) & 120.7 $\pm$ 13.3 mmHg \\
Total Cholesterol          & Survey & Continuous & 1,418 (23.9\%) & 4.7 $\pm$ 1.7 mmol/L \\
\addlinespace[3pt]

\multicolumn{5}{@{}l}{\quad\textit{Mental Well-being\enspace($\Delta^-_{\tau}=\Delta^+_{\tau}=$14\,days)}}\\
\cmidrule{1-5}
Feel Depressed          & Survey & Ordinal & 1,990 (33.5\%) & $K{=}4$ \\
Feel Happy              & Survey & Ordinal & 2,810 (47.3\%) & $K{=}4$ \\
Feel Worried            & Survey & Ordinal & 2,642 (44.5\%) & $K{=}4$ \\
Life Satisfaction       & Survey & Ordinal & 2,803 (47.2\%) & $K{=}4$ \\
Things Are Worthwhile   & Survey & Ordinal & 2,803 (47.2\%) & $K{=}4$ \\
\addlinespace[3pt]

\multicolumn{5}{@{}l}{\quad\textit{Sleep \& Lifestyle\enspace($\Delta^-_{\tau}=\Delta^+_{\tau}=$91\,days)}}\\
\cmidrule{1-5}
Bedtime               & Profile & Ordinal    & 4,555 (76.7\%) & $K{=}5$ \\
Currently Employed    & Survey & Binary     & 3,897 (65.6\%) & Prevalence\ 80.9\% \\
Sleep Duration        & Survey & Ordinal    & 3,891 (65.5\%) & $K{=}4$ \\
Vigorous Activity Minutes & Survey & Continuous & 3,803 (64.0\%) & 71.8 $\pm$ 124.1 min/week \\
Wake-up Time          & Profile & Ordinal    & 4,533 (76.3\%) & $K{=}4$ 
\end{longtable}
}

\subsection{Consistency of Self-Reported Labels}
\label{app:label_consistency}
We assessed the consistency of self-reported labels using repeated survey responses (Table~\ref{tab:label_consistency}). Agreement was 92.2--98.5\% for the three medical condition labels. Some disagreement may reflect changes over time. In addition, we compared self-reported medical condition labels with medication reports from separate surveys among 3,653 respondents with the relevant items. Agreement was 96.2\% for hypertension ($\kappa=0.90$; PABAK $=0.92$) and 97.9\% for diabetes ($\kappa=0.81$; PABAK $=0.96$). Together, these findings suggest internal consistency in self-reported medical conditions across repeated surveys and agreement with related medication reports. These checks do not establish clinical validity or a performance ceiling and cannot rule out systematic misreporting.

\begin{table}[ht]
\centering
\small
\caption{\bo{Agreement between repeated self-reported survey responses.}
$n$: reported sample size; mean gap: average time between compared
responses. A dash indicates an unreported $\kappa$.}
\label{tab:label_consistency}
\begin{tabular}{lrrrr}
\toprule
Item & $n$ & Mean gap & Agreement & $\kappa$ \\
\midrule
Biological sex & 4,014 & 19.4 wk & 99.7\% & 0.99 \\
Diabetes & 963 & 16.6 wk & 98.5\% & 0.85 \\
Hypertension & 963 & 16.6 wk & 93.1\% & 0.83 \\
Sleep disorder & 2,040 & 35.1 wk & 92.2\% & 0.70 \\
Sense of purpose (time-varying) & 3,115 & 15.4 wk & 31.0\% & --- \\
\bottomrule
\end{tabular}
\end{table}

\subsection{Prediction Task Modeling}
\label{app:prediction_models}


For each prediction task, models are evaluated on participants that meet the Inclusion Criteria (defined in Appendix~\ref{app:health_outcome_inclusion_and_context}). For each included participant, the models are allowed to use the complete history of the wearable data as input. 


\bo{Auxiliary Covariates.}
We include age, biological sex, and BMI value as direct inputs only for \textsc{Linear}. These covariates are excluded for tasks where they correspond to, or directly determine, the prediction target: Age, Biological Sex, BMI Value, BMI Category, Body Weight, and Framingham CVD Risk.

\bo{Outcome-specific prediction heads.}
For binary outcomes, all models except \textsc{XGBoost} and \textsc{GRU-D} use a logistic regression head. Similarly, for ordinal outcomes, all models use the Frank--Hall method~\citep{Frank_eibe_2001}. Specifically, for an outcome with ordered classes $\{0,1,\dots,K-1\}$ and ground-truth label $y \in \{0,1,\dots,K-1\}$, we train $K-1$ binary threshold predictors to estimate whether the label exceeds threshold $k$, for $k=0,\dots,K-2$. For continuous outcomes, all models except \textsc{XGBoost} and \textsc{GRU-D} use ordinary least-squares regression. \textsc{XGBoost} and \textsc{GRU-D} use model-specific prediction heads and losses, described in their respective sections.


\subsubsection{Linear}
\textsc{Linear} uses summary-statistic features computed from the retained daily hourly segments (defined in Appendix~\ref{app:data_aggregation}). For each
retained daily hourly segment, we compute the mean and standard deviation of each of the 19 sensor channels (ignoring NaNs), yielding a 38-dimensional segment-level feature vector~\citep{erturk2025beyond}. The segment-level feature vectors are averaged across retained daily segments in the model input context to obtain a single 38-dimensional representation.

\subsubsection{XGBoost}
\label{app:health_outcome_fexgb}
\textsc{XGBoost} uses hand-crafted features from daily minute-level segments. Features are organized in three types:
\begin{enumerate}[label=(\roman*),leftmargin=*]
  \item \textbf{Daily descriptors} (177 features): for each retained day we extract activity totals, heart-rate summaries (resting, daytime, nighttime), physical-activity intensity zones, sleep duration and timing, workout statistics, and non-parametric circadian rhythm indices
(interdaily stability, intradaily variability, L5/M10 relative amplitude); these are then aggregated across the participant's full available wearable history using robust summary statistics (P5, median, P95, IQR).
  \item \textbf{Day-to-day dynamics} (266 features): 14 daily metrics spanning steps, distance, flights, heart-rate percentiles, energy expenditure, sleep and in-bed minutes, active minutes, and workout time are treated as longitudinal time series over each participant's         
observation period, from which we extract distributional statistics, ARIMA(2,1,0) parameters (fit only on participants with at least 15 retained days), and pairwise cross-correlations at lags $0$--$2$ over an 8-metric subset ($28$ pairs $\times$ $3$ lags).                                 
  \item \textbf{Curve analysis} (52 features): each participant's minute-level daily curves are averaged across days, yielding mean 24-hour profiles on which we compute functional PCA scores ($10$ components $\times$ $4$ channels; fit on the training set split), single-component cosinor parameters for heart   
rate (MESOR, amplitude, acrophase, fit $R^2$, $p$-value, amplitude-to-MESOR ratio, and fitted-minute count; $7$ features), and $5$ heart-rate-over-steps (HROS) profile statistics (mean, standard deviation, daytime mean, nighttime mean, and day-to-night ratio).
\end{enumerate}

\noindent A separate \textsc{XGBoost} model is fitted for each prediction task using the task-specific prediction heads. We use a hyperparameter configuration selected using validation-set performance ($1000$ estimators, tree depth $2$, learning rate
$0.05$, row subsampling $0.8$, column subsampling $0.3$, $\ell_1=0.1$, $\ell_2=1.0$). NaN features are passed to XGBoost natively without imputation.

\subsubsection{MultiRocket}
MultiRocket is a deterministic convolutional transform for multivariate time series \citep{tan_multirocket_2022,LoningSktime2019}. We apply it on hourly data derived according to  Appendix~\ref{app:data_aggregation}, treating each retained daily segment as a multivariate sequence over 19 sensor channels. We z-score normalize the hourly values and zero-fill missing entries before applying the transform.
The resulting 19-channel tensor is then passed to MultiRocket. We apply 6,216 kernels combined with four pooling operators to both the hourly value sequence and its first differences, yielding a 49,728-dimensional segment-level representation. Segment-level \textsc{MultiRocket} features are averaged across retained daily segments in the model input context to obtain a 49{,}728-dimensional participant-level representation. This representation is reduced to 50 dimensions using PCA fitted on the training split.

\subsubsection{GRU-D}
GRU-D is a recurrent neural network designed for multivariate time series with informative missingness \citep{che_recurrent_2018}. GRU-D is trained end-to-end on the prediction tasks and on day-level hourly segments  (Appendix~\ref{app:data_aggregation}). We train a shared \textsc{GRU-D} model end-to-end with task-specific linear prediction heads. Hidden states from a participant's retained daily segments are mean-pooled to obtain one 64-dimensional representation. Binary outcomes use classification heads trained with cross-entropy loss, ordinal outcomes use $K{-}1$ sigmoid threshold heads trained with binary cross-entropy loss, and continuous outcomes use regression heads trained with mean-squared-error loss.

\subsubsection{WBM}
\label{app:health_outcome_wbm}
We follow the Mamba-2-based \textsc{WBM} architecture of Erturk et al.~\citep{erturk2025beyond} with a bidirectional Mamba-2 encoder over 168-step weekly wearable tensors trained with a contrastive objective. After pretraining, the encoder is frozen. For each participant, each retained week (a seven-day window containing at least five retained days; Appendix~\ref{app:data_filtering}) is fed into the model to produce a 256-dimensional embedding; weekly embeddings are averaged. This representation is reduced to 50 dimensions using PCA fitted on the training split and then passed to the task-specific linear prediction heads. If a participant has no retained weeks, we use the \textsc{Linear} fallback predictor.

\bo{Architecture.}
\vspace{-3mm}
\begin{itemize}[leftmargin=*]
    \item \bo{Input representation.} The model receives weekly tensors $\mathbf{X} \in \mathbb{R}^{168 \times 38}$ consisting of 168 hourly time steps with 19 sensor channels and 19 binary missingness-mask channels (Appendix~\ref{app:weekly_tensors}). The hourly values are z-score normalized using training-set hourly statistics, and missing entries are zero-filled before encoder input. The missingness is conveyed explicitly through the corresponding binary mask channels.
    \item \bo{Hour patch embedding.}
    Each hourly input vector $\mathbf{x}_t \in \mathbb{R}^{38}$ is independently projected to a $d$-dimensional token embedding using a two-layer feedforward network:
    \begin{equation*}
      \mathbf{e}_t = W_2 \,\mathrm{GELU}(W_1 \mathbf{x}_t + b_1) + b_2, \quad \mathbf{e}_t \in \mathbb{R}^d,
      \label{eq:wbm_tokenizer}
    \end{equation*}
    where $W_1 \in \mathbb{R}^{h \times 38}$, $W_2 \in \mathbb{R}^{d \times h}$, with hidden dimension $h{=}64$ and output dimension $d{=}256$. This produces a token sequence $(\mathbf{e}_1, \ldots, \mathbf{e}_{168}) \in \mathbb{R}^{168 \times 256}$.
    \item \bo{Bidirectional Mamba2 encoder.}
    The token sequence is processed by a stack of $L{=}4$ bidirectional Mamba2 blocks. Each block applies a forward Mamba2 pass and a backward (time-reversed) Mamba2 pass in parallel, concatenates their outputs, and projects back to the model dimension:
    \begin{equation*}
      \mathbf{z}_t = W_{\mathrm{proj}} \bigl[\mathrm{Mamba2}_\mathrm{fwd}(\mathbf{e})_t \;\|\; \mathrm{Mamba2}_\mathrm{bwd}(\mathbf{e})_t \bigr] + b_{\mathrm{proj}},
      \label{eq:wbm_bimamba}
    \end{equation*}
    where $W_{\mathrm{proj}} \in \mathbb{R}^{d \times 2d}$ and $\|$ denotes concatenation. This is followed by LayerNorm, a residual connection, and a feedforward network (FFN) with expansion factor 4:
    \begin{equation*}
      \mathrm{FFN}(\mathbf{z}) = W_4 \,\mathrm{GELU}(W_3 \mathbf{z} + b_3) + b_4, \quad W_3 \in \mathbb{R}^{4d \times d},\; W_4 \in \mathbb{R}^{d \times 4d},
      \label{eq:wbm_ffn}
    \end{equation*}
    with a second LayerNorm and residual connection. Dropout is applied within the FFN.
    \item \bo{Pooling and projection.}
    The encoder output is aggregated via masked mean pooling over the time dimension. During pretraining, a binary keep-mask $\mathbf{m} \in \{0,1\}^{168}$ (from time-step dropout augmentation) determines which tokens contribute to the pool:
    \begin{equation*}
      \mathbf{r} = \frac{\sum_{t=1}^{168} m_t \cdot \mathbf{z}_t}{\sum_{t=1}^{168} m_t}, \quad \mathbf{r} \in \mathbb{R}^{256}.
      \label{eq:wbm_pool}
    \end{equation*}
    The representation $\mathbf{r}$ is used for downstream tasks. For the contrastive loss, a linear projection head maps $\mathbf{r}$ to a lower-dimensional embedding that is $L_2$-normalized:
    \begin{equation*}
      \mathbf{h} = \frac{W_p \mathbf{r} + b_p}{\|W_p \mathbf{r} + b_p\|_2}, \quad \mathbf{h} \in \mathbb{R}^{128}.
      \label{eq:wbm_proj}
    \end{equation*}
\end{itemize}

\bo{Pretraining Objective.}
\vspace{-3mm}
\begin{itemize}[leftmargin=*]
    \item \bo{Contrastive views.}
    Each training batch samples $B$ participants, drawing one random week per participant. Two augmented views of each weekly tensor are created using time-step dropout: each hourly token is independently dropped with probability $p_{\mathrm{drop}}{=}0.223$, and the surviving tokens are pooled as in Eq.~\eqref{eq:wbm_pool}. This yields two normalized embeddings $\mathbf{h}_i^{(1)}, \mathbf{h}_i^{(2)}$ per participant $i$.
    \item \bo{Symmetric InfoNCE.}
    We use a symmetric InfoNCE loss with temperature $\tau{=}0.2$. Only diagonal pairs (same participant, different views) serve as positives; off-diagonal pairs from the same participant are masked from the denominator to prevent trivial shortcuts:
    \begin{equation*}
      \mathcal{L}_{\mathrm{InfoNCE}} = -\frac{1}{2B}\sum_{i=1}^{B}\left[
        \log\frac{\exp(\mathbf{h}_i^{(1)\top}\mathbf{h}_i^{(2)}/\tau)}{\sum_{j \in \mathcal{N}_i}\exp(\mathbf{h}_i^{(1)\top}\mathbf{h}_j^{(2)}/\tau)}
        + \log\frac{\exp(\mathbf{h}_i^{(2)\top}\mathbf{h}_i^{(1)}/\tau)}{\sum_{j \in \mathcal{N}_i}\exp(\mathbf{h}_i^{(2)\top}\mathbf{h}_j^{(1)}/\tau)}
      \right],
      \label{eq:wbm_infonce}
    \end{equation*}
    where $\mathcal{N}_i$ excludes other samples from the same participant in the batch.
    \item \bo{KoLeo regularization.}
    To encourage uniform coverage of the embedding space, we use a squared-distance variant of KoLeo regularization~\citep{sablayrolles2019spreadingvectorssimilaritysearch}, averaged over the two views with numerical stabilization:
    \begin{equation*}
      \mathcal{L}_{\mathrm{KoLeo}} =
      -\frac{1}{2B}\sum_{v \in \{1,2\}}\sum_{i=1}^{B}
      \log\left(
      \max\left\{
      \min_{j \neq i}
      \|\mathbf{h}_i^{(v)} - \mathbf{h}_j^{(v)}\|_2^2,
      \epsilon
      \right\}
      \right).
      \label{eq:wbm_koleo}
    \end{equation*}
    Here $\epsilon=10^{-12}$.
    \item \bo{Total loss.}
    The final pretraining objective is:
    \begin{equation*}
      \mathcal{L} = \mathcal{L}_{\mathrm{InfoNCE}} + \lambda_{\mathrm{KoLeo}}\,\mathcal{L}_{\mathrm{KoLeo}},
      \label{eq:wbm_total_loss}
    \end{equation*}
    with $\lambda_{\mathrm{KoLeo}}{=}0.689$ selected by hyperparameter optimization.
\end{itemize}

\bo{Training Configuration.} Table~\ref{tab:wbm_training_config} summarizes the final \textsc{WBM} pretraining configuration used for the selected checkpoint. Architecture, batching, normalization, and optimizer settings are fixed across pretraining runs, while the token dropout probability, KoLeo weight, and learning rate are selected through the hyperparameter search described below.

\begin{table}[h!]
  \caption{\textsc{WBM} pretraining configuration. All hyperparameters not in Table~\ref{tab:hpo_search_space} are fixed at the values below.}
  \label{tab:wbm_training_config}
  \centering
  \small
  \renewcommand{\arraystretch}{1.15}
  \begin{tabularx}{0.92\linewidth}{>{\raggedright\arraybackslash}p{0.45\linewidth} >{\raggedright\arraybackslash}X}
    \toprule
    \textbf{Parameter} & \textbf{Value} \\
    \midrule
    \multicolumn{2}{l}{\textit{Architecture}} \\
    Encoder type & Bidirectional Mamba2 \\
    Encoder layers & 4 \\
    Embedding dimension ($d$) & 256 \\
    Tokenizer hidden dimension ($h$) & 64 \\
    FFN expansion factor & 4 \\
    Mamba2 internal heads  & 8 \\
    Projection head & Linear \\
    Projection dimension & 128 \\
    \midrule
    \multicolumn{2}{l}{\textit{Loss}} \\
    Temperature ($\tau$) & 0.2 \\
    Loss variant & Masked (diagonal positives only) \\
    KoLeo weight ($\lambda_\mathrm{KoLeo}$) & 0.689 \\
    \midrule
    \multicolumn{2}{l}{\textit{Optimization}} \\
    Optimizer & AdamW \\
    Learning rate & $1.3 \times 10^{-5}$ \\
    Weight decay & 0.01 \\
    Scheduler & Cosine ($\eta_\mathrm{min}{=}10^{-6}$) \\
    Warmup ratio & 0.09 \\
    Gradient clipping & 1.0 \\
    Precision & bf16-mixed \\
    Epochs & 20 \\
    \midrule
    \multicolumn{2}{l}{\textit{Data and batching}} \\
    Segment type & Weekly (168 hours) \\
    Input dimension & 38 (19 sensors + 19 masks) \\
    Normalization & Hourly z-score (training-set stats) \\
    Minimum retained days per week & 5 \\
    Participants per batch & 128 \\
    Weeks per participants per batch & 1 \\
    Maximum weeks per participants (cap) & 50 \\
    \midrule
    \multicolumn{2}{l}{\textit{Augmentation}} \\
    Time-step dropout ($p_\mathrm{drop}$) & 0.223 \\
    \bottomrule
  \end{tabularx}
\end{table}

\bo{Hyperparameter Selection.}
\textsc{WBM} requires a separate pretraining stage before health outcome evaluation. We select its pretraining configuration using the self-supervised validation loss. After pretraining, the selected encoder is frozen and reused across health outcome tasks.

Because exhaustive pretraining sweeps are computationally expensive, we use a two-stage selection procedure. The first stage identifies which pretraining hyperparameters most affect validation InfoNCE loss; the second stage performs a targeted Bayesian sweep over this reduced set of hyperparameters.
\begin{itemize}[leftmargin=*]
    \item \bo{Phase 1: Sensitivity screening.} We first run a one-factor-at-a-time sensitivity screen on a proxy subset of 1{,}783 participants. In each run, one candidate hyperparameter is varied while the remaining hyperparameters are held fixed at their reference values. Runs are compared using validation InfoNCE loss on held-out weekly tensors. This phase is used only to reduce the search space, not to select the final pretrained encoder. Based on this screen, we retain three hyperparameters for the final sweep: token drop probability (\texttt{drop\_prob}), KoLeo regularization weight (\texttt{lambda\_koleo}), and learning rate (\texttt{lr}).
    \item \bo{Phase 2: Bayesian optimization.}
After Phase 1, we run targeted Bayesian optimization~\citep{snoek2012practical} using Weights \& Biases sweeps. For \textsc{WBM}, the search is restricted to the three hyperparameters retained from sensitivity screening: token drop probability (\texttt{drop\_prob}), KoLeo regularization weight (\texttt{lambda\_koleo}), and learning rate (\texttt{lr}). Each trial consists of pretraining \textsc{WBM} from scratch, extracting frozen representations, and evaluating validation InfoNCE loss. To limit compute cost, the sweep is capped at 15 trials and uses Hyperband early termination \citep{li2018hyperband} with minimum iterations $=5$ and reduction factor $\eta=3$.
\end{itemize}

The final search space and selected \textsc{WBM} hyperparameters are reported in Table~\ref{tab:hpo_search_space}. All hyperparameters not listed are fixed at their baseline values throughout Phase 2.

\begin{table}[h!]
  \caption{Phase 2 search space and selected \textsc{WBM} pretraining configuration. Selected values correspond to the final pretrained checkpoint used for downstream evaluation.}
  \label{tab:hpo_search_space}
  \centering
  \small
  \renewcommand{\arraystretch}{1.15}
  \begin{tabularx}{0.98\linewidth}{>{\raggedright\arraybackslash}p{0.42\linewidth} >{\raggedright\arraybackslash}X >{\centering\arraybackslash}p{0.2\linewidth}}
    \toprule
    \textbf{Hyperparameter} & \textbf{Search space} & \textbf{Selected value} \\
    \midrule
    Token drop probability (\texttt{drop\_prob}) & Uniform $[0.05, 0.25]$ & $0.223$ \\
    KoLeo regularization weight (\texttt{lambda\_koleo}) & Uniform $[0.0, 1.0]$ & $0.689$ \\
    Learning rate (\texttt{lr}) & Log-uniform $[10^{-5}, 5 \times 10^{-4}]$ & $1.3 \times 10^{-5}$ \\
    \bottomrule
  \end{tabularx}
\end{table}

\bo{Fallback Prediction.}
Since the WBM model requires entire weeks of users with sufficient wear time, some users do not have any weeks that meet the wear time criteria. In these cases, we fall back to using the \textsc{Linear} prediction model. See Appendix~\ref{app:wbm_lsm2_ablation} for additional details.


\subsubsection{LSM-2.}
\label{app:health_outcome_LSM-2}
We additionally evaluate \textsc{LSM-2} \citep{xu2025lsm}, a ViT-1D encoder pretrained on minute-level daily wearable segments (see Appendix \ref{sec:LSM-2_architecture} for details on the pretraining approach). \textsc{LSM-2} uses minute-level daily inputs (as defined in Appendix~\ref{app:data_aggregation}). Minute-level values are z-score normalized using training-split statistics, and missing entries are zero-filled before encoder input. Representations from the frozen model at non-masked positions are mean-pooled to a single 384-dimensional day-level vector, and these day-level vectors are averaged across all retained days in the model input context to obtain one participant-level representation. This representation is reduced to 50 dimensions using PCA fitted on the training split and then passed to the task-specific prediction heads. 

\subsubsection{Toto (Time-series foundation model)}
We use the \textsc{Toto}~\citep{cohen2025time} model fine-tuned on forecasting to extract representations for prediction (see Appendix~\ref{sec:toto_architecture} for fine-tuning details). \textsc{Toto} operates on the hourly wearable representation defined in Appendix~\ref{app:data_aggregation}. Specifically, we extract last-layer latent representations, yielding channel-wise embeddings of shape $(19,768)$, with one 768-dimensional vector per wearable channel. These channel-wise embeddings are mean-pooled across the 19 channels to obtain one 768-dimensional participant-level representation. This representation is reduced to 50 dimensions using PCA fitted on the training split and then passed to the task-specific linear prediction heads.

\subsubsection{Chronos-2 (Time-series foundation model)}
We use a \textsc{Chronos-2} model \citep{ansari2025chronos} fine-tuned on forecasting to extract representations for prediction (see Appendix~\ref{sec:chronos2_architecture} for fine-tuning details). \textsc{Chronos-2} operates on the same hourly wearable representation used by \textsc{Toto}, as defined in Appendix~\ref{app:data_aggregation}. We extract last-layer latent representations, yielding channel-wise embedding matrix of dimension $19 \times 768$, with one 768-dimensional vector per wearable channel. As with \textsc{Toto}, these channel-wise embeddings are mean-pooled across the 19 channels to obtain one 768-dimensional participant-level representation. This representation is reduced to 50 dimensions using PCA fitted on the training split and then passed to the task-specific prediction heads.

\subsection{Additional Prediction Task Results}
\label{app:additional_health_prediction_results}
\begin{table}[!htbp]
\centering
\captionsetup{width=\textwidth}
\providecommand{\est}[3]{\ensuremath{#1^{\scriptscriptstyle +#2}_{\scriptscriptstyle -#3}}}
\caption{\textbf{Prediction Per-Task Results.} Per-task primary metric across the 32 outcome tasks in five outcome domains. AUPRC (binary), Spearman~$\rho$ (ordinal), Pearson~$r$ (regression), all $\uparrow$. Values are point estimates on the held-out test split; superscripts and subscripts indicate the $95\%$ percentile bootstrap confidence interval ($B{=}1{,}000$). Average Rank $R$ and aggregate Skill Score $S$ are the arithmetic
means of their respective five domain-level values.
Skill Scores are reported in \%, with $0$ corresponding to the
\textsc{Linear} reference. The overall results also appear in
Table~\ref{tab:health_outcome_main_results}.
See Appendix~\ref{sec:scoring_methodology} for metric definitions.}
\label{tab:main_results}
\small
\setlength{\tabcolsep}{3pt}
\renewcommand{\arraystretch}{1.15}
\resizebox{\linewidth}{!}{%
\begin{tabular}{@{}lll cccccccc@{}}
\toprule
Domain & Task & Metric & \textsc{Linear} & \textsc{MultiRocket} & \textsc{XGBoost} & \textsc{GRU-D} & \textsc{LSM-2} & \textsc{WBM} & \textsc{Toto} & \textsc{Chronos-2} \\
\midrule
 \multirow{4}{*}{\rotatebox[origin=c]{90}{\scriptsize Demographics}}    &    Age    &    Pearson $r$ $\uparrow$    & \est{0.284}{0.042}{0.040} & \cellcolor{customblue!50}\est{0.437}{0.038}{0.037} & \cellcolor{customblue!100}\est{\mathbf{0.587}}{0.034}{0.035} & \cellcolor{customblue!24}\est{0.357}{0.040}{0.041} & \cellcolor{customblue!90}\est{0.556}{0.034}{0.035} & \cellcolor{customblue!46}\est{0.423}{0.042}{0.041} & \cellcolor{customblue!41}\est{0.408}{0.039}{0.040} & \cellcolor{customblue!38}\est{0.398}{0.040}{0.039} \\
     &    Biological Sex    &    AUPRC $\uparrow$    & \cellcolor{customblue!26}\est{0.876}{0.020}{0.021} & \cellcolor{customblue!75}\est{0.918}{0.016}{0.017} & \cellcolor{customblue!100}\est{\mathbf{0.940}}{0.012}{0.014} & \cellcolor{customblue!67}\est{0.911}{0.016}{0.015} & \cellcolor{customblue!90}\est{0.931}{0.013}{0.016} & \cellcolor{customblue!66}\est{0.911}{0.016}{0.017} & \est{0.854}{0.022}{0.022} & \cellcolor{customblue!34}\est{0.883}{0.020}{0.020} \\
    &    \cellcolor{gray!10}\textit{Domain Average Rank}    &    \cellcolor{gray!10}    & \est{7.50}{0.00}{0.50} & \cellcolor{customblue!69}\est{3.00}{1.50}{0.00} & \cellcolor{customblue!100}\est{\mathbf{1.00}}{0.50}{0.00} & \cellcolor{customblue!31}\est{5.50}{0.50}{0.50} & \cellcolor{customblue!85}\est{2.00}{0.50}{0.50} & \cellcolor{customblue!46}\est{4.50}{1.00}{1.00} & \cellcolor{customblue!15}\est{6.50}{0.50}{1.00} & \cellcolor{customblue!23}\est{6.00}{0.50}{1.00} \\
    &    \cellcolor{gray!10}\textit{Domain Skill Score $S$ (\%)}    &    \cellcolor{gray!10}    & $0.0$ & \cellcolor{customblue!59}\est{+27.7}{5.7}{5.4} & \cellcolor{customblue!100}\est{\mathbf{+47.1}}{5.5}{5.2} & \cellcolor{customblue!42}\est{+19.6}{7.1}{7.2} & \cellcolor{customblue!87}\est{+41.1}{5.4}{5.5} & \cellcolor{customblue!51}\est{+23.8}{5.1}{5.3} & \cellcolor{customblue!2}\est{+1.0}{8.4}{8.7} & \cellcolor{customblue!23}\est{+10.8}{7.5}{7.9} \\
\midrule
 \multirow{14}{*}{\rotatebox[origin=c]{90}{\scriptsize Medical conditions \& risk}}    &    Atrial Fibrillation    &    AUPRC $\uparrow$    & \cellcolor{customblue!100}\est{\mathbf{0.099}}{0.119}{0.050} & \cellcolor{customblue!8}\est{0.030}{0.028}{0.012} & \cellcolor{customblue!6}\est{0.028}{0.020}{0.011} & \est{0.024}{0.016}{0.009} & \cellcolor{customblue!17}\est{0.037}{0.026}{0.015} & \cellcolor{customblue!26}\est{0.044}{0.058}{0.022} & \cellcolor{customblue!8}\est{0.030}{0.050}{0.014} & \cellcolor{customblue!1}\est{0.025}{0.018}{0.010} \\
     &    Cardiovascular Disease    &    AUPRC $\uparrow$    & \cellcolor{customblue!100}\est{\mathbf{0.367}}{0.094}{0.072} & \cellcolor{customblue!20}\est{0.231}{0.074}{0.056} & \cellcolor{customblue!64}\est{0.306}{0.082}{0.074} & \cellcolor{customblue!9}\est{0.212}{0.071}{0.049} & \cellcolor{customblue!43}\est{0.270}{0.074}{0.063} & \cellcolor{customblue!54}\est{0.288}{0.084}{0.064} & \cellcolor{customblue!7}\est{0.209}{0.061}{0.044} & \est{0.197}{0.061}{0.041} \\
     &    Cerebrovascular Disease    &    AUPRC $\uparrow$    & \cellcolor{customblue!100}\est{\mathbf{0.065}}{0.061}{0.031} & \cellcolor{customblue!10}\est{0.027}{0.044}{0.012} & \cellcolor{customblue!22}\est{0.032}{0.023}{0.012} & \cellcolor{customblue!19}\est{0.031}{0.029}{0.012} & \cellcolor{customblue!47}\est{0.042}{0.047}{0.020} & \cellcolor{customblue!70}\est{0.052}{0.059}{0.026} & \est{0.023}{0.015}{0.008} & \cellcolor{customblue!14}\est{0.028}{0.024}{0.011} \\
     &    Congenital Heart Disease    &    AUPRC $\uparrow$    & \cellcolor{customblue!32}\est{0.017}{0.037}{0.012} & \cellcolor{customblue!6}\est{0.012}{0.016}{0.007} & \cellcolor{customblue!33}\est{0.017}{0.032}{0.010} & \cellcolor{customblue!100}\est{\mathbf{0.029}}{0.089}{0.020} & \cellcolor{customblue!26}\est{0.016}{0.033}{0.010} & \cellcolor{customblue!47}\est{0.020}{0.055}{0.015} & \est{0.011}{0.022}{0.006} & \cellcolor{customblue!72}\est{0.024}{0.041}{0.016} \\
     &    Coronary Artery Disease    &    AUPRC $\uparrow$    & \cellcolor{customblue!83}\est{0.137}{0.082}{0.063} & \cellcolor{customblue!14}\est{0.072}{0.045}{0.023} & \cellcolor{customblue!14}\est{0.072}{0.056}{0.026} & \cellcolor{customblue!57}\est{0.113}{0.108}{0.052} & \cellcolor{customblue!61}\est{0.116}{0.073}{0.047} & \cellcolor{customblue!100}\est{\mathbf{0.153}}{0.099}{0.065} & \cellcolor{customblue!27}\est{0.085}{0.067}{0.032} & \est{0.059}{0.034}{0.019} \\
     &    Diabetes    &    AUPRC $\uparrow$    & \cellcolor{customblue!73}\est{0.127}{0.097}{0.050} & \cellcolor{customblue!29}\est{0.095}{0.079}{0.037} & \cellcolor{customblue!20}\est{0.088}{0.095}{0.036} & \cellcolor{customblue!100}\est{\mathbf{0.147}}{0.133}{0.081} & \cellcolor{customblue!87}\est{0.137}{0.113}{0.058} & \est{0.074}{0.081}{0.029} & \cellcolor{customblue!25}\est{0.092}{0.087}{0.040} & \cellcolor{customblue!29}\est{0.095}{0.102}{0.041} \\
     &    Framingham CVD Risk    &    Pearson $r$ $\uparrow$    & \cellcolor{customblue!47}\est{0.181}{0.208}{0.159} & \cellcolor{customblue!72}\est{0.249}{0.136}{0.130} & \cellcolor{customblue!100}\est{\mathbf{0.327}}{0.106}{0.109} & \cellcolor{customblue!59}\est{0.213}{0.121}{0.123} & \cellcolor{customblue!100}\est{0.327}{0.082}{0.095} & \cellcolor{customblue!67}\est{0.236}{0.165}{0.138} & \cellcolor{customblue!26}\est{0.122}{0.121}{0.102} & \est{0.049}{0.126}{0.111} \\
     &    Heart Failure / CHF    &    AUPRC $\uparrow$    & \cellcolor{customblue!16}\est{0.035}{0.048}{0.022} & \est{0.012}{0.025}{0.008} & \cellcolor{customblue!4}\est{0.018}{0.041}{0.012} & \cellcolor{customblue!100}\est{\mathbf{0.153}}{0.267}{0.147} & \cellcolor{customblue!53}\est{0.088}{0.272}{0.079} & \cellcolor{customblue!11}\est{0.028}{0.040}{0.017} & \cellcolor{customblue!2}\est{0.015}{0.045}{0.010} & \cellcolor{customblue!3}\est{0.016}{0.029}{0.011} \\
     &    Hypertension    &    AUPRC $\uparrow$    & \cellcolor{customblue!100}\est{\mathbf{0.578}}{0.084}{0.082} & \cellcolor{customblue!31}\est{0.433}{0.086}{0.073} & \cellcolor{customblue!65}\est{0.504}{0.085}{0.076} & \cellcolor{customblue!21}\est{0.410}{0.086}{0.076} & \cellcolor{customblue!40}\est{0.451}{0.086}{0.072} & \cellcolor{customblue!74}\est{0.523}{0.085}{0.081} & \est{0.366}{0.087}{0.065} & \cellcolor{customblue!12}\est{0.391}{0.078}{0.070} \\
     &    Pulmonary Hypertension    &    AUPRC $\uparrow$    & \cellcolor{customblue!1}\est{0.019}{0.045}{0.013} & \cellcolor{customblue!100}\est{\mathbf{0.151}}{0.461}{0.144} & \cellcolor{customblue!65}\est{0.105}{0.408}{0.096} & \cellcolor{customblue!17}\est{0.041}{0.081}{0.029} & \cellcolor{customblue!13}\est{0.035}{0.073}{0.023} & \est{0.017}{0.065}{0.015} & \cellcolor{customblue!3}\est{0.021}{0.063}{0.015} & \cellcolor{customblue!2}\est{0.020}{0.023}{0.012} \\
     &    Sleep Disorder Diagnosis    &    AUPRC $\uparrow$    & \cellcolor{customblue!33}\est{0.269}{0.056}{0.039} & \cellcolor{customblue!12}\est{0.246}{0.054}{0.039} & \cellcolor{customblue!45}\est{0.282}{0.062}{0.047} & \cellcolor{customblue!34}\est{0.270}{0.062}{0.048} & \cellcolor{customblue!100}\est{\mathbf{0.343}}{0.074}{0.054} & \cellcolor{customblue!35}\est{0.272}{0.057}{0.044} & \est{0.233}{0.053}{0.041} & \cellcolor{customblue!4}\est{0.237}{0.055}{0.042} \\
     &    Vascular Disease    &    AUPRC $\uparrow$    & \cellcolor{customblue!100}\est{\mathbf{0.035}}{0.174}{0.030} & \cellcolor{customblue!7}\est{0.007}{0.012}{0.005} & \est{0.005}{0.008}{0.004} & \cellcolor{customblue!34}\est{0.015}{0.036}{0.011} & \cellcolor{customblue!14}\est{0.010}{0.023}{0.007} & \cellcolor{customblue!5}\est{0.007}{0.015}{0.005} & \cellcolor{customblue!1}\est{0.006}{0.007}{0.003} & \cellcolor{customblue!6}\est{0.007}{0.016}{0.005} \\
    &    \cellcolor{gray!10}\textit{Domain Average Rank}    &    \cellcolor{gray!10}    & \cellcolor{customblue!100}\est{3.00}{0.67}{0.75} & \cellcolor{customblue!47}\est{5.00}{1.08}{0.59} & \cellcolor{customblue!67}\est{4.25}{1.00}{0.83} & \cellcolor{customblue!78}\est{3.83}{1.25}{0.58} & \cellcolor{customblue!98}\est{3.08}{0.92}{0.58} & \cellcolor{customblue!78}\est{3.83}{0.92}{0.83} & \est{6.75}{0.33}{1.50} & \cellcolor{customblue!13}\est{6.25}{0.59}{1.17} \\
    &    \cellcolor{gray!10}\textit{Domain Skill Score $S$ (\%)}    &    \cellcolor{gray!10}    & $0.0$ & \cellcolor{customblue!48}\est{-4.8}{6.1}{5.0} & \cellcolor{customblue!80}\est{-1.8}{5.3}{4.9} & \cellcolor{customblue!56}\est{-4.1}{4.5}{5.1} & \cellcolor{customblue!86}\est{-1.3}{4.4}{4.8} & \cellcolor{customblue!72}\est{-2.6}{2.3}{3.3} & \cellcolor{customblue!6}\est{-8.8}{3.5}{5.2} & \est{-9.4}{3.5}{5.3} \\
\midrule
 \multirow{10}{*}{\rotatebox[origin=c]{90}{\scriptsize Vitals \& blood biomarkers}}    &    BMI Categories    &    Spearman $\rho$ $\uparrow$    & \cellcolor{customblue!55}\est{0.338}{0.050}{0.047} & \cellcolor{customblue!65}\est{0.385}{0.045}{0.052} & \cellcolor{customblue!80}\est{0.459}{0.044}{0.045} & \cellcolor{customblue!56}\est{0.344}{0.046}{0.048} & \cellcolor{customblue!100}\est{\mathbf{0.556}}{0.040}{0.041} & \cellcolor{customblue!68}\est{0.401}{0.047}{0.049} & \cellcolor{customblue!7}\est{0.103}{0.052}{0.052} & \est{0.069}{0.053}{0.052} \\
     &    BMI Value    &    Pearson $r$ $\uparrow$    & \cellcolor{customblue!34}\est{0.317}{0.099}{0.074} & \cellcolor{customblue!67}\est{0.511}{0.047}{0.053} & \cellcolor{customblue!85}\est{0.618}{0.042}{0.045} & \cellcolor{customblue!68}\est{0.517}{0.046}{0.052} & \cellcolor{customblue!100}\est{\mathbf{0.706}}{0.037}{0.045} & \cellcolor{customblue!64}\est{0.492}{0.081}{0.077} & \cellcolor{customblue!11}\est{0.180}{0.051}{0.055} & \est{0.117}{0.055}{0.057} \\
     &    Blood Pressure Categories    &    Spearman $\rho$ $\uparrow$    & \cellcolor{customblue!34}\est{0.091}{0.089}{0.096} & \cellcolor{customblue!22}\est{0.069}{0.092}{0.094} & \cellcolor{customblue!79}\est{0.178}{0.093}{0.098} & \cellcolor{customblue!100}\est{\mathbf{0.217}}{0.087}{0.085} & \cellcolor{customblue!72}\est{0.163}{0.091}{0.093} & \cellcolor{customblue!27}\est{0.079}{0.095}{0.094} & \est{0.027}{0.097}{0.095} & \cellcolor{customblue!28}\est{0.080}{0.089}{0.095} \\
     &    Body Weight    &    Pearson $r$ $\uparrow$    & \cellcolor{customblue!37}\est{0.339}{0.115}{0.086} & \cellcolor{customblue!68}\est{0.549}{0.049}{0.057} & \cellcolor{customblue!83}\est{0.657}{0.037}{0.042} & \cellcolor{customblue!68}\est{0.555}{0.043}{0.046} & \cellcolor{customblue!100}\est{\mathbf{0.770}}{0.033}{0.042} & \cellcolor{customblue!64}\est{0.522}{0.099}{0.085} & \cellcolor{customblue!5}\est{0.120}{0.052}{0.057} & \est{0.087}{0.051}{0.053} \\
     &    HDL Cholesterol    &    Pearson $r$ $\uparrow$    & \cellcolor{customblue!100}\est{\mathbf{0.138}}{0.101}{0.091} & \cellcolor{customblue!77}\est{0.119}{0.090}{0.095} & \cellcolor{customblue!13}\est{0.066}{0.087}{0.097} & \cellcolor{customblue!34}\est{0.083}{0.081}{0.085} & \cellcolor{customblue!12}\est{0.065}{0.081}{0.087} & \cellcolor{customblue!53}\est{0.099}{0.082}{0.081} & \cellcolor{customblue!72}\est{0.115}{0.075}{0.076} & \est{0.055}{0.082}{0.081} \\
     &    LDL Cholesterol    &    Pearson $r$ $\uparrow$    & \cellcolor{customblue!43}\est{0.023}{0.130}{0.145} & \cellcolor{customblue!10}\est{-0.027}{0.104}{0.084} & \cellcolor{customblue!46}\est{0.027}{0.111}{0.099} & \cellcolor{customblue!10}\est{-0.026}{0.084}{0.083} & \est{-0.041}{0.139}{0.104} & \cellcolor{customblue!27}\est{-0.002}{0.120}{0.138} & \cellcolor{customblue!100}\est{\mathbf{0.106}}{0.088}{0.073} & \cellcolor{customblue!6}\est{-0.032}{0.090}{0.106} \\
     &    Systolic Blood Pressure    &    Pearson $r$ $\uparrow$    & \cellcolor{customblue!65}\est{0.164}{0.089}{0.091} & \cellcolor{customblue!58}\est{0.146}{0.083}{0.096} & \cellcolor{customblue!48}\est{0.118}{0.086}{0.090} & \cellcolor{customblue!93}\est{0.236}{0.081}{0.087} & \cellcolor{customblue!100}\est{\mathbf{0.255}}{0.085}{0.090} & \cellcolor{customblue!79}\est{0.199}{0.084}{0.083} & \est{-0.007}{0.093}{0.093} & \cellcolor{customblue!26}\est{0.061}{0.092}{0.094} \\
     &    Total Cholesterol    &    Pearson $r$ $\uparrow$    & \cellcolor{customblue!2}\est{0.014}{0.067}{0.040} & \cellcolor{customblue!64}\est{0.107}{0.097}{0.095} & \cellcolor{customblue!38}\est{0.068}{0.089}{0.088} & \est{0.012}{0.086}{0.082} & \cellcolor{customblue!100}\est{\mathbf{0.162}}{0.105}{0.101} & \cellcolor{customblue!40}\est{0.072}{0.104}{0.055} & \cellcolor{customblue!62}\est{0.104}{0.089}{0.100} & \cellcolor{customblue!24}\est{0.048}{0.090}{0.092} \\
    &    \cellcolor{gray!10}\textit{Domain Average Rank}    &    \cellcolor{gray!10}    & \cellcolor{customblue!59}\est{4.62}{1.25}{0.38} & \cellcolor{customblue!68}\est{4.25}{1.00}{0.88} & \cellcolor{customblue!88}\est{3.38}{1.12}{0.75} & \cellcolor{customblue!74}\est{4.00}{1.12}{0.63} & \cellcolor{customblue!100}\est{\mathbf{2.88}}{0.50}{1.12} & \cellcolor{customblue!68}\est{4.25}{0.88}{0.88} & \cellcolor{customblue!38}\est{5.50}{0.88}{0.75} & \est{7.12}{0.50}{1.50} \\
    &    \cellcolor{gray!10}\textit{Domain Skill Score $S$ (\%)}    &    \cellcolor{gray!10}    & $0.0$ & \cellcolor{customblue!58}\est{+9.2}{4.4}{5.8} & \cellcolor{customblue!76}\est{+16.8}{4.3}{5.2} & \cellcolor{customblue!61}\est{+10.4}{4.2}{5.7} & \cellcolor{customblue!100}\est{\mathbf{+27.0}}{4.3}{4.9} & \cellcolor{customblue!57}\est{+8.8}{2.9}{3.2} & \cellcolor{customblue!10}\est{-11.5}{5.8}{7.4} & \est{-15.6}{5.6}{7.6} \\
\midrule
 \multirow{7}{*}{\rotatebox[origin=c]{90}{\scriptsize Mental well-being}}    &    Feel Depressed    &    Spearman $\rho$ $\uparrow$    & \cellcolor{customblue!20}\est{0.056}{0.076}{0.084} & \cellcolor{customblue!100}\est{\mathbf{0.134}}{0.078}{0.078} & \cellcolor{customblue!26}\est{0.062}{0.077}{0.081} & \est{0.037}{0.080}{0.079} & \cellcolor{customblue!80}\est{0.115}{0.081}{0.080} & \cellcolor{customblue!13}\est{0.050}{0.076}{0.080} & \cellcolor{customblue!10}\est{0.046}{0.077}{0.079} & \cellcolor{customblue!82}\est{0.117}{0.078}{0.080} \\
     &    Feel Happy    &    Spearman $\rho$ $\uparrow$    & \cellcolor{customblue!100}\est{\mathbf{0.144}}{0.060}{0.066} & \cellcolor{customblue!92}\est{0.134}{0.066}{0.070} & \est{0.004}{0.067}{0.070} & \cellcolor{customblue!58}\est{0.086}{0.069}{0.070} & \cellcolor{customblue!96}\est{0.138}{0.064}{0.067} & \cellcolor{customblue!60}\est{0.088}{0.071}{0.065} & \cellcolor{customblue!53}\est{0.078}{0.060}{0.067} & \cellcolor{customblue!46}\est{0.069}{0.069}{0.062} \\
     &    Feel Worried    &    Spearman $\rho$ $\uparrow$    & \cellcolor{customblue!25}\est{0.088}{0.065}{0.066} & \cellcolor{customblue!87}\est{0.143}{0.067}{0.072} & \cellcolor{customblue!21}\est{0.084}{0.073}{0.069} & \est{0.065}{0.074}{0.074} & \cellcolor{customblue!100}\est{\mathbf{0.155}}{0.074}{0.066} & \cellcolor{customblue!8}\est{0.072}{0.067}{0.066} & \cellcolor{customblue!29}\est{0.092}{0.071}{0.067} & \cellcolor{customblue!52}\est{0.111}{0.075}{0.064} \\
     &    Life Satisfaction    &    Spearman $\rho$ $\uparrow$    & \cellcolor{customblue!100}\est{\mathbf{0.207}}{0.061}{0.069} & \cellcolor{customblue!64}\est{0.148}{0.065}{0.065} & \cellcolor{customblue!46}\est{0.119}{0.065}{0.068} & \cellcolor{customblue!25}\est{0.083}{0.072}{0.075} & \cellcolor{customblue!76}\est{0.168}{0.068}{0.070} & \cellcolor{customblue!90}\est{0.190}{0.066}{0.067} & \est{0.043}{0.066}{0.068} & \cellcolor{customblue!48}\est{0.121}{0.071}{0.064} \\
     &    Things Are Worthwhile    &    Spearman $\rho$ $\uparrow$    & \cellcolor{customblue!85}\est{0.137}{0.062}{0.062} & \cellcolor{customblue!100}\est{\mathbf{0.151}}{0.066}{0.067} & \cellcolor{customblue!17}\est{0.075}{0.067}{0.067} & \est{0.058}{0.070}{0.072} & \cellcolor{customblue!82}\est{0.135}{0.063}{0.070} & \cellcolor{customblue!67}\est{0.120}{0.067}{0.064} & \cellcolor{customblue!26}\est{0.083}{0.065}{0.069} & \cellcolor{customblue!38}\est{0.093}{0.064}{0.065} \\
    &    \cellcolor{gray!10}\textit{Domain Average Rank}    &    \cellcolor{gray!10}    & \cellcolor{customblue!88}\est{2.80}{2.20}{0.80} & \cellcolor{customblue!100}\est{\mathbf{2.20}}{2.00}{0.80} & \cellcolor{customblue!20}\est{6.20}{1.20}{1.60} & \est{7.20}{0.40}{2.60} & \cellcolor{customblue!96}\est{2.40}{1.60}{1.20} & \cellcolor{customblue!52}\est{4.60}{1.41}{1.40} & \cellcolor{customblue!20}\est{6.20}{1.20}{2.00} & \cellcolor{customblue!56}\est{4.40}{1.60}{1.60} \\
    &    \cellcolor{gray!10}\textit{Domain Skill Score $S$ (\%)}    &    \cellcolor{gray!10}    & $0.0$ & \cellcolor{customblue!100}\est{+1.6}{4.9}{5.7} & \cellcolor{customblue!5}\est{-6.7}{5.7}{6.0} & \est{-7.1}{6.3}{6.3} & \cellcolor{customblue!100}\est{\mathbf{+1.6}}{5.6}{5.8} & \cellcolor{customblue!52}\est{-2.6}{3.7}{3.6} & \cellcolor{customblue!3}\est{-6.8}{5.9}{6.8} & \cellcolor{customblue!48}\est{-2.9}{6.1}{6.4} \\
\midrule
 \multirow{7}{*}{\rotatebox[origin=c]{90}{\scriptsize Sleep \& lifestyle}}    &    Bedtime    &    Spearman $\rho$ $\uparrow$    & \cellcolor{customblue!2}\est{0.100}{0.050}{0.050} & \cellcolor{customblue!78}\est{0.192}{0.053}{0.055} & \cellcolor{customblue!100}\est{\mathbf{0.218}}{0.052}{0.053} & \cellcolor{customblue!9}\est{0.109}{0.056}{0.049} & \cellcolor{customblue!93}\est{0.210}{0.055}{0.055} & \cellcolor{customblue!25}\est{0.128}{0.057}{0.053} & \cellcolor{customblue!12}\est{0.112}{0.056}{0.055} & \est{0.098}{0.052}{0.053} \\
     &    Currently Employed    &    AUPRC $\uparrow$    & \cellcolor{customblue!17}\est{0.871}{0.028}{0.027} & \cellcolor{customblue!54}\est{0.893}{0.024}{0.024} & \cellcolor{customblue!64}\est{0.899}{0.024}{0.022} & \cellcolor{customblue!27}\est{0.877}{0.025}{0.025} & \cellcolor{customblue!100}\est{\mathbf{0.921}}{0.018}{0.018} & \est{0.861}{0.028}{0.028} & \cellcolor{customblue!46}\est{0.889}{0.021}{0.023} & \cellcolor{customblue!53}\est{0.893}{0.022}{0.023} \\
     &    Sleep Duration    &    Spearman $\rho$ $\uparrow$    & \cellcolor{customblue!78}\est{0.059}{0.055}{0.062} & \cellcolor{customblue!46}\est{0.025}{0.054}{0.055} & \cellcolor{customblue!50}\est{0.029}{0.057}{0.056} & \cellcolor{customblue!57}\est{0.036}{0.058}{0.058} & \cellcolor{customblue!91}\est{0.072}{0.056}{0.052} & \cellcolor{customblue!92}\est{0.073}{0.051}{0.057} & \est{-0.023}{0.055}{0.056} & \cellcolor{customblue!100}\est{\mathbf{0.081}}{0.060}{0.058} \\
     &    Vigorous Activity Minutes    &    Pearson $r$ $\uparrow$    & \cellcolor{customblue!93}\est{0.214}{0.095}{0.100} & \cellcolor{customblue!100}\est{\mathbf{0.225}}{0.096}{0.098} & \cellcolor{customblue!53}\est{0.152}{0.065}{0.056} & \cellcolor{customblue!38}\est{0.129}{0.045}{0.045} & \cellcolor{customblue!97}\est{0.221}{0.064}{0.062} & \cellcolor{customblue!86}\est{0.203}{0.081}{0.084} & \est{0.070}{0.053}{0.050} & \cellcolor{customblue!20}\est{0.101}{0.058}{0.056} \\
     &    Wake-up Time    &    Spearman $\rho$ $\uparrow$    & \cellcolor{customblue!24}\est{0.126}{0.053}{0.050} & \cellcolor{customblue!84}\est{0.195}{0.049}{0.053} & \cellcolor{customblue!53}\est{0.160}{0.052}{0.060} & \cellcolor{customblue!49}\est{0.156}{0.052}{0.052} & \cellcolor{customblue!100}\est{\mathbf{0.213}}{0.050}{0.055} & \cellcolor{customblue!29}\est{0.132}{0.051}{0.054} & \cellcolor{customblue!35}\est{0.139}{0.056}{0.052} & \est{0.099}{0.051}{0.055} \\
    &    \cellcolor{gray!10}\textit{Domain Average Rank}    &    \cellcolor{gray!10}    & \cellcolor{customblue!14}\est{5.60}{0.80}{1.60} & \cellcolor{customblue!68}\est{3.20}{1.00}{1.00} & \cellcolor{customblue!64}\est{3.40}{1.40}{0.60} & \cellcolor{customblue!18}\est{5.40}{1.20}{1.20} & \cellcolor{customblue!100}\est{\mathbf{1.80}}{0.80}{0.60} & \cellcolor{customblue!32}\est{4.80}{1.20}{1.00} & \est{6.20}{1.00}{1.00} & \cellcolor{customblue!14}\est{5.60}{0.80}{1.20} \\
    &    \cellcolor{gray!10}\textit{Domain Skill Score $S$ (\%)}    &    \cellcolor{gray!10}    & $0.0$ & \cellcolor{customblue!55}\est{+6.9}{4.7}{5.0} & \cellcolor{customblue!50}\est{+6.2}{5.0}{5.2} & \cellcolor{customblue!6}\est{-0.6}{5.1}{5.6} & \cellcolor{customblue!100}\est{\mathbf{+14.0}}{4.3}{4.2} & \cellcolor{customblue!5}\est{-0.7}{3.0}{3.4} & \est{-1.5}{5.4}{5.7} & \cellcolor{customblue!15}\est{+0.9}{5.4}{5.7} \\
\midrule
    &    \cellcolor{gray!20}\textbf{\textit{Average Rank $R$}}    &    \cellcolor{gray!20}    & \cellcolor{customblue!40}\est{4.71}{0.55}{0.39} & \cellcolor{customblue!71}\est{3.53}{0.70}{0.22} & \cellcolor{customblue!68}\est{3.65}{0.49}{0.34} & \cellcolor{customblue!27}\est{5.19}{0.44}{0.57} & \cellcolor{customblue!100}\est{\mathbf{2.43}}{0.41}{0.40} & \cellcolor{customblue!48}\est{4.40}{0.49}{0.55} & \est{6.23}{0.30}{0.64} & \cellcolor{customblue!9}\est{5.88}{0.37}{0.67} \\
    &    \cellcolor{gray!20}\textbf{\textit{Aggregate Skill Score $S$ (\%)}}    &    \cellcolor{gray!20}    & $0.0$ & \cellcolor{customblue!62}\est{+8.1}{2.3}{2.5} & \cellcolor{customblue!81}\est{+12.3}{2.3}{2.5} & \cellcolor{customblue!42}\est{+3.6}{2.6}{2.9} & \cellcolor{customblue!100}\est{\mathbf{+16.5}}{2.3}{2.4} & \cellcolor{customblue!49}\est{+5.3}{1.7}{1.9} & \est{-5.5}{2.7}{3.1} & \cellcolor{customblue!10}\est{-3.3}{2.6}{3.3} \\
\bottomrule
\end{tabular}%
}
\end{table}

\clearpage
\subsection{Ablations Investigating the \textsc{WBM} vs. \textsc{LSM-2} Performance Gap}
\label{app:wbm_lsm2_ablation}

\textsc{LSM-2} and \textsc{WBM} differ along several dimensions simultaneously: temporal resolution (minute-level vs.\ hourly), required input duration (a single day vs.\ five retained days within a seven-day window), pretraining objective (masked reconstruction vs.\ contrastive learning), backbone (1D-ViT vs.\ bidirectional Mamba-2). 

\bo{Use of the fallback mechanism.}
\textsc{WBM} requires at least one retained week with sufficient weartime per user (Appendix~\ref{app:data_filtering}); otherwise, it falls back to predicting using the \textsc{Linear} baseline (Appendix~\ref{app:prediction_models}). This substitution applies to $62.7\%$ of evaluation set outcome instances for \textsc{WBM}; note that every other method in Table~\ref{tab:health_outcome_main_results} has a fallback rate of $0.0\%$. 
Below, we evaluate \textsc{LSM-2}, \textsc{XGBoost}, \textsc{WBM}, and \textsc{Linear} on the same subset of outcome instances for which \textsc{WBM} does not use the fallback (${\approx}37.3\%$ of evaluation set outcome instances). On this matched-subset evaluation, \textsc{LSM-2} and \textsc{XGBoost} have higher aggregate Skill Scores than \textsc{WBM} ($+24.5$ and $+19.4$, respectively, versus $+13.5$; Table~\ref{tab:wbm_no_fallback}).

\begin{table}[h]
\centering
\footnotesize
\setlength{\tabcolsep}{4pt}
\begin{tabular}{lrrrrrrr}
\toprule
\textbf{Method} & $R \downarrow$ & $S \uparrow$ & \makecell{Demo-\\graphics\,$\uparrow$} & \makecell{Medical\\Conditions \& Risk\,$\uparrow$} & \makecell{Vitals \& Blood\\Biomarkers\,$\uparrow$} & \makecell{Mental\\Well-Being\,$\uparrow$} & \makecell{Sleep \&\\Lifestyle\,$\uparrow$} \\
\midrule
\textsc{LSM-2} & $1.78$ & $+24.5$ & $+59.7$ & $-0.7$ & $+35.7$ & $+6.6$ & $+21.1$ \\
\textsc{XGBoost} & $2.47$ & $+19.4$ & $+65.1$ & $-0.6$ & $+25.0$ & $-1.4$ & $+8.8$ \\
\textsc{WBM} & $2.82$ & $+13.5$ & $+53.4$ & $-3.5$ & $+25.4$ & $-6.1$ & $-1.6$ \\
\textsc{Linear} \textit{(reference)} & $2.93$ & $0.0$ & $0.0$ & $0.0$ & $0.0$ & $0.0$ & $0.0$ \\
\bottomrule
\end{tabular}
\vspace{2mm}
\caption{\textbf{Matched-subset evaluation without fallback.} All four methods are evaluated on the same $11{,}561$ outcome instances from $620$ participants for which \textsc{WBM} produces predictions without \textsc{Linear} fallback. (No changes to model training.)
We report point estimates of Average Rank $R$, Aggregate Skill Score $S$, and domain-level Skill Scores across the five domains (in \%; $0=\textsc{Linear}$). Overall scores are domain-balanced; ranks are computed over these four methods only, and Skill Scores use \textsc{Linear} evaluated on the same subset as the reference.}
\label{tab:wbm_no_fallback}
\end{table}

\bo{Pretraining-data and Temporal Resolution.}
We additionally pretrain a \textsc{LSM-2} 
variant on exactly the participant days that \textsc{WBM} uses (that pass the weekly wear-time filtering criterion); here, \textsc{LSM-2} still uses minute-level representations. 
By restricting the LSM-2's pretraining dataset to the same participant days as \textsc{WBM}, the aggregate skill score of LSM-2 falls from $+24.5$ to $+16.6$, but the LSM-2 still outperforms WBM in four out of five domains (Table \ref{tab:wbm_lsm2_ablation}).

Additionally, we train an additional variant of LSM-2 that uses hourly-level wearable data just as \textsc{WBM} uses (rather than minute-level) and continue to restrict to the same participant days as \textsc{WBM}.
This change leads LSM-2's skill Score to drop from $+16.6$ to $+4.5$. The change in temporal resolution of the wearable data is the largest single effect we measure, and it indicates that temporal resolution matters substantially for these prediction tasks.




\begin{table}[h]
\centering
\footnotesize
\setlength{\tabcolsep}{4pt}
\begin{tabular}{lrrrrrrr}
\toprule
\textbf{Method} & $R \downarrow$ & $S \uparrow$ & \makecell{Demo-\\graphics\,$\uparrow$} & \makecell{Medical\\Conditions\,$\uparrow$} & \makecell{Vitals \& Blood\\Biomarkers\,$\uparrow$} & \makecell{Mental\\Well-Being\,$\uparrow$} & \makecell{Sleep \&\\Lifestyle\,$\uparrow$} \\
\midrule
\textsc{LSM-2} Full Cohort (min) & $1.86$ & $+24.5$ & $+59.7$ & $-0.7$ & $+35.7$ & $+6.6$ & $+21.1$ \\
\textsc{LSM-2} \textsc{WBM} Cohort (min) & $2.87$ & $+16.6$ & $+40.4$ & $-2.3$ & $+35.3$ & $+3.5$ & $+5.9$ \\
\textsc{WBM} & $3.11$ & $+13.5$ & $+53.4$ & $-3.5$ & $+25.4$ & $-6.1$ & $-1.6$ \\
\textsc{LSM-2} \textsc{WBM} Cohort (hour) & $3.60$ & $+4.5$ & $+21.5$ & $-8.4$ & $+4.7$ & $+8.7$ & $-4.1$ \\
\textsc{Linear} \textit{(reference)} & $3.56$ & $0.0$ & $0.0$ & $0.0$ & $0.0$ & $0.0$ & $0.0$ \\
\bottomrule
\end{tabular}
\vspace{2mm}
\caption{\textbf{Cohort- and resolution-matched comparison of \textsc{LSM-2} and \textsc{WBM}.} ``Full Cohort'' and ``\textsc{WBM} Cohort'' refer to the participants used for \emph{pretraining}; the evaluation subset is identical 
and consists of the outcome instances for which \textsc{WBM} does not 
use the \textsc{Linear} fallback. Ranks $R$ are computed over the five methods in this table only, and Skill Scores $S$ are computed on this evaluation subset; 
therefore they are not directly comparable to the values in Table~\ref{tab:health_outcome_main_results}, which are computed over the full method set and the full evaluation cohort.} 
\label{tab:wbm_lsm2_ablation}
\end{table}

\bo{Additional properties that could explain the performance gap.}
These experiments narrow the additional performance gap between the two models to two entangled factors: architecture and pretraining objective. Having controlled for the use of the fallback mechanism, pretraining-data availability, and temporal resolution, the differences that remain between \textsc{WBM} and \textsc{LSM-2} are the backbone (bidirectional Mamba-2 vs.\ 1D-ViT), the daily versus weekly context window, the missingness encoding and tokenization, and the pretraining objective itself. 
We further note that both models are our best-effort reimplementations and may not reproduce the performance of the original proprietary systems, which were trained on different proprietary datasets. In particular, the original \textsc{WBM} was pretrained on substantially more data than our reimplementation~\citep{erturk2025beyond}.


\subsection{Gemini-Family LLM Baselines}
\label{app:gemini_baselines}

\textbf{Scope.} This appendix evaluates the \textsc{Gemini} family of frontier large language models (\textsc{Gemini-2.5-Pro}~\citep{comanici2025gemini} and \textsc{Gemini-3.1-Pro-Preview}~\citep{google2026gemini31pro}) as zero-shot/few-shot baselines on the 32 outcome prediction tasks in our benchmark. \emph{It is not an exhaustive characterization of LLM capabilities on wearable-sensor health tasks}: cross-vendor evaluation and broader sweeps over open-weights LLMs are out of scope and left to future work. Findings here should be read as Gemini-family analysis, not a general LLM-vs-supervised verdict.\footnote{We report these as preliminary results: given their limited performance relative to simple baselines and the cost of scaling API-based inference, we did not pursue these models further or rerun them after final label corrections.}

\textbf{Headline finding.} All five Gemini-family baseline configurations we evaluate underperform the \textsc{Linear} baseline on this cohort in aggregate Skill Score $S$ (Table~\ref{tab:llm_baselines_per_task}): the strongest probe (\textsc{Gem-3.1 Stats}) is $-10.3$ skill points relative to \textsc{Linear} and $\sim$29 points below \textsc{LSM-2}. Per-task wins for the Gemini baselines total 10 of 32 outcomes, of which 8 fall in the rare-event Medical conditions \& risk domain where every method's standard error is wide; the Gemini baselines win zero Demographics, zero Sleep \& lifestyle, and only one Vitals \& blood biomarkers task.

\textbf{Setup.} To complement the supervised methods reported in Table~\ref{tab:health_outcome_main_results}, we evaluate the Gemini baselines under three prompting strategies: \emph{statistics} (38-channel weekly summary statistics rendered as text), \emph{vision} (rendered weekly sensor heatmaps; rendering recipe and prompt template in Appendix~\ref{app:gemini_vision_details}), and \emph{agentic} (a modified version of the PHA Data Science Agent~\citep{merrill2026transforming, heydari2025anatomy} with $k{=}3$ retrieved few-shot exemplars and tool-use; built on Gemini-2.5-Pro). We evaluate both Gemini variants per non-agentic strategy, yielding five probe configurations in total ($2 \times 2 + 1$). Because all probe strategies require a coherent week of data, we restrict evaluation to participants with $\geq 5$ valid days within a 7-day window; per-task participant-method intersection cohorts span 119--600 participants (median 388) across the 32 tasks shown here, drawn from a strict subset of the 1{,}637-participant cohort used for this preliminary analysis. \textsc{Linear}, \textsc{XGBoost}, and \textsc{LSM-2} columns are recomputed on each task's intersection cohort for direct paired comparison and may differ slightly from Table~\ref{tab:health_outcome_main_results}.

\textbf{API configuration and Reproducibility.} We queried \texttt{gemini-2.5-pro} and \texttt{gemini-3.1-pro-preview} through the Gemini Developer API using \texttt{google-genai} 1.73.1. Statistics and vision prompts used an output-token limit of 8{,}192 and a thinking budget of 128 tokens for Gemini 2.5 or thinking level \texttt{low} for Gemini 3.1; vision also used high media resolution. Sampling parameters were omitted from these requests, using the recorded API defaults: temperature 1.0, top-$p$ 0.95, top-$k$ 64, and one candidate. The PHA agent requested \texttt{models/gemini-2.5-pro} with temperature 0.7 and an output-token limit of 16{,}384; top-$p$, top-$k$, and candidate count used the same defaults, with no explicit thinking configuration. No configuration supplied an API sampling seed.

\textbf{Per-task pattern.} The per-task breakdown (Table~\ref{tab:llm_baselines_per_task}) sharpens the picture: the Gemini baselines are competitive on roughly a third of tasks --- almost entirely low-prevalence Medical conditions \& risk binary outcomes where AUPRC standard errors are wide (e.g., Atrial Fibrillation, Cerebrovascular Disease, Heart Failure / CHF, Hypertension, Coronary Artery Disease) --- but lose on every Demographics and Sleep \& lifestyle task and on 7 of 8 Vitals \& blood biomarkers tasks, where the supervised methods exploit signal that the Gemini baselines do not extract from text or vision representations of the same week. Two configurations (\textsc{Gem-2.5 Vis} and the agentic \textsc{PHA-DSA}) win zero tasks outright; \textsc{PHA-DSA} also has the worst aggregate Skill Score $S$ and Average Rank in the table, indicating that agentic tooling does not rescue performance on this benchmark within the Gemini family.

\begin{table}[!htbp]
\centering
\captionsetup{width=\textwidth}
\caption{\textbf{Gemini-Family LLM Baselines: Per-Task Primary Metrics (Cohort-Matched).} Per-task primary metric on the 8-method weekly cohort: participants with $\geq 5$ valid days per 7-day window, per-task participant-method intersection 119--600 participants (median 388), strict subset of the 1{,}637-participant cohort used for this preliminary analysis. AUPRC (binary), Spearman~$\rho$ (ordinal), Pearson~$r$ (regression), all $\uparrow$. \textsc{Linear}, \textsc{XGBoost}, \textsc{LSM-2} recomputed on each task's intersection cohort. Standard errors from paired participant-level bootstrap ($B{=}1{,}000$). Aggregate Skill Score $S$ = mean of the 5 per-domain $S$ (\%; $0{=}$\textsc{Linear} reference).}
\label{tab:llm_baselines_per_task}
\small
\setlength{\tabcolsep}{3pt}
\renewcommand{\arraystretch}{1.15}
\resizebox{\linewidth}{!}{%
\begin{tabular}{@{}lll cccccccc@{}}
\toprule
Domain & Task & Metric & \textsc{Linear} & \textsc{XGBoost} & \textsc{LSM-2} & \textsc{Gem-2.5 St} & \textsc{Gem-3.1 St} & \textsc{Gem-2.5 Vis} & \textsc{Gem-3.1 Vis} & \textsc{PHA-DSA} \\
\midrule
\multirow{4}{*}{\rotatebox[origin=c]{90}{\scriptsize Demographics}} & Age & Pearson $r$ $\uparrow$ & \cellcolor{customblue!23}$0.142_{\pm 0.036}$ & \cellcolor{customblue!100}$\mathbf{0.659_{\pm 0.028}}$ & \cellcolor{customblue!97}$0.638_{\pm 0.026}$ & \cellcolor{customblue!15}$0.086_{\pm 0.042}$ & \cellcolor{customblue!25}$0.155_{\pm 0.042}$ & \cellcolor{customblue!5}$0.018_{\pm 0.042}$ & \cellcolor{customblue!11}$0.055_{\pm 0.040}$ & \cellcolor{customblue!0}$-0.016_{\pm 0.043}$ \\
 & Biological Sex & AUPRC $\uparrow$ & \cellcolor{customblue!43}$0.816_{\pm 0.023}$ & \cellcolor{customblue!100}$\mathbf{0.937_{\pm 0.010}}$ & \cellcolor{customblue!96}$0.929_{\pm 0.010}$ & \cellcolor{customblue!0}$0.726_{\pm 0.023}$ & \cellcolor{customblue!20}$0.769_{\pm 0.023}$ & \cellcolor{customblue!17}$0.762_{\pm 0.023}$ & \cellcolor{customblue!5}$0.737_{\pm 0.023}$ & \cellcolor{customblue!3}$0.732_{\pm 0.026}$ \\
\rowcolor{gray!10}
 & \multicolumn{2}{l}{\textit{Domain Average Rank}} & \cellcolor{customblue!62}$3.50$ & \cellcolor{customblue!100}$\mathbf{1.00}$ & \cellcolor{customblue!85}$2.00$ & \cellcolor{customblue!15}$6.50$ & \cellcolor{customblue!62}$3.50$ & \cellcolor{customblue!23}$6.00$ & \cellcolor{customblue!23}$6.00$ & \cellcolor{customblue!0}$7.50$ \\
\rowcolor{gray!10}
 & \multicolumn{2}{l}{\textit{Domain Skill Score $S$ (\%)}} & $0.0_{\pm 0.0}$ & \cellcolor{customblue!100}$\mathbf{+63.1_{\pm 3.1}}$ & \cellcolor{customblue!96}$+59.5_{\pm 3.1}$ & \cellcolor{customblue!6}$-26.7_{\pm 8.1}$ & \cellcolor{customblue!21}$-11.9_{\pm 7.8}$ & \cellcolor{customblue!10}$-22.6_{\pm 8.5}$ & \cellcolor{customblue!6}$-26.5_{\pm 8.3}$ & \cellcolor{customblue!0}$-32.0_{\pm 9.3}$ \\
\midrule
\multirow{14}{*}{\rotatebox[origin=c]{90}{\scriptsize Medical conditions \& risk}} & Atrial Fibrillation & AUPRC $\uparrow$ & \cellcolor{customblue!94}$0.113_{\pm 0.059}$ & \cellcolor{customblue!0}$0.046_{\pm 0.017}$ & \cellcolor{customblue!80}$0.103_{\pm 0.042}$ & \cellcolor{customblue!100}$\mathbf{0.117_{\pm 0.049}}$ & \cellcolor{customblue!72}$0.097_{\pm 0.052}$ & \cellcolor{customblue!13}$0.055_{\pm 0.017}$ & \cellcolor{customblue!65}$0.092_{\pm 0.057}$ & \cellcolor{customblue!4}$0.049_{\pm 0.018}$ \\
 & Cardiovascular Disease & AUPRC $\uparrow$ & \cellcolor{customblue!100}$\mathbf{0.440_{\pm 0.063}}$ & \cellcolor{customblue!80}$0.411_{\pm 0.058}$ & \cellcolor{customblue!65}$0.388_{\pm 0.057}$ & \cellcolor{customblue!0}$0.292_{\pm 0.046}$ & \cellcolor{customblue!83}$0.415_{\pm 0.059}$ & \cellcolor{customblue!11}$0.308_{\pm 0.049}$ & \cellcolor{customblue!54}$0.372_{\pm 0.057}$ & \cellcolor{customblue!0}$0.292_{\pm 0.049}$ \\
 & Cerebrovascular Disease & AUPRC $\uparrow$ & \cellcolor{customblue!33}$0.043_{\pm 0.025}$ & \cellcolor{customblue!43}$0.048_{\pm 0.034}$ & \cellcolor{customblue!48}$0.051_{\pm 0.032}$ & \cellcolor{customblue!0}$0.025_{\pm 0.012}$ & \cellcolor{customblue!100}$\mathbf{0.079_{\pm 0.070}}$ & \cellcolor{customblue!24}$0.038_{\pm 0.021}$ & \cellcolor{customblue!76}$0.066_{\pm 0.062}$ & \cellcolor{customblue!43}$0.048_{\pm 0.033}$ \\
 & Congenital Heart Disease & AUPRC $\uparrow$ & \cellcolor{customblue!27}$0.027_{\pm 0.021}$ & \cellcolor{customblue!0}$0.016_{\pm 0.010}$ & \cellcolor{customblue!71}$0.045_{\pm 0.060}$ & \cellcolor{customblue!100}$\mathbf{0.057_{\pm 0.057}}$ & \cellcolor{customblue!10}$0.020_{\pm 0.017}$ & \cellcolor{customblue!83}$0.050_{\pm 0.071}$ & \cellcolor{customblue!7}$0.019_{\pm 0.010}$ & \cellcolor{customblue!7}$0.019_{\pm 0.008}$ \\
 & Coronary Artery Disease & AUPRC $\uparrow$ & \cellcolor{customblue!76}$0.160_{\pm 0.055}$ & \cellcolor{customblue!0}$0.112_{\pm 0.037}$ & \cellcolor{customblue!8}$0.117_{\pm 0.042}$ & \cellcolor{customblue!2}$0.113_{\pm 0.038}$ & \cellcolor{customblue!100}$\mathbf{0.175_{\pm 0.062}}$ & \cellcolor{customblue!41}$0.138_{\pm 0.055}$ & \cellcolor{customblue!6}$0.116_{\pm 0.039}$ & \cellcolor{customblue!11}$0.119_{\pm 0.039}$ \\
 & Diabetes & AUPRC $\uparrow$ & \cellcolor{customblue!96}$0.238_{\pm 0.074}$ & \cellcolor{customblue!55}$0.203_{\pm 0.062}$ & \cellcolor{customblue!98}$0.239_{\pm 0.069}$ & \cellcolor{customblue!19}$0.173_{\pm 0.047}$ & \cellcolor{customblue!96}$0.238_{\pm 0.072}$ & \cellcolor{customblue!0}$0.157_{\pm 0.048}$ & \cellcolor{customblue!100}$\mathbf{0.241_{\pm 0.074}}$ & \cellcolor{customblue!18}$0.172_{\pm 0.056}$ \\
 & Framingham CVD Risk & Pearson $r$ $\uparrow$ & \cellcolor{customblue!69}$0.172_{\pm 0.092}$ & \cellcolor{customblue!100}$\mathbf{0.310_{\pm 0.116}}$ & \cellcolor{customblue!52}$0.099_{\pm 0.112}$ & \cellcolor{customblue!22}$-0.035_{\pm 0.089}$ & \cellcolor{customblue!12}$-0.077_{\pm 0.078}$ & \cellcolor{customblue!6}$-0.103_{\pm 0.081}$ & \cellcolor{customblue!0}$-0.131_{\pm 0.070}$ & \cellcolor{customblue!43}$0.058_{\pm 0.067}$ \\
 & Heart Failure / CHF & AUPRC $\uparrow$ & \cellcolor{customblue!0}$0.077_{\pm 0.058}$ & \cellcolor{customblue!22}$0.121_{\pm 0.162}$ & \cellcolor{customblue!10}$0.097_{\pm 0.085}$ & \cellcolor{customblue!12}$0.100_{\pm 0.132}$ & \cellcolor{customblue!27}$0.131_{\pm 0.163}$ & \cellcolor{customblue!26}$0.128_{\pm 0.114}$ & \cellcolor{customblue!100}$\mathbf{0.275_{\pm 0.238}}$ & \cellcolor{customblue!52}$0.180_{\pm 0.167}$ \\
 & Hypertension & AUPRC $\uparrow$ & \cellcolor{customblue!78}$0.574_{\pm 0.056}$ & \cellcolor{customblue!71}$0.566_{\pm 0.056}$ & \cellcolor{customblue!86}$0.583_{\pm 0.060}$ & \cellcolor{customblue!54}$0.548_{\pm 0.050}$ & \cellcolor{customblue!100}$\mathbf{0.598_{\pm 0.054}}$ & \cellcolor{customblue!33}$0.525_{\pm 0.056}$ & \cellcolor{customblue!68}$0.563_{\pm 0.054}$ & \cellcolor{customblue!0}$0.489_{\pm 0.055}$ \\
 & Pulmonary Hypertension & AUPRC $\uparrow$ & \cellcolor{customblue!36}$0.035_{\pm 0.030}$ & \cellcolor{customblue!20}$0.026_{\pm 0.019}$ & \cellcolor{customblue!100}$\mathbf{0.073_{\pm 0.063}}$ & \cellcolor{customblue!7}$0.018_{\pm 0.012}$ & \cellcolor{customblue!15}$0.023_{\pm 0.019}$ & \cellcolor{customblue!3}$0.016_{\pm 0.008}$ & \cellcolor{customblue!0}$0.014_{\pm 0.008}$ & \cellcolor{customblue!27}$0.030_{\pm 0.020}$ \\
 & Sleep Disorder Diagnosis & AUPRC $\uparrow$ & \cellcolor{customblue!56}$0.316_{\pm 0.043}$ & \cellcolor{customblue!100}$\mathbf{0.373_{\pm 0.047}}$ & \cellcolor{customblue!96}$0.368_{\pm 0.047}$ & \cellcolor{customblue!0}$0.242_{\pm 0.027}$ & \cellcolor{customblue!40}$0.295_{\pm 0.037}$ & \cellcolor{customblue!27}$0.278_{\pm 0.034}$ & \cellcolor{customblue!50}$0.307_{\pm 0.039}$ & \cellcolor{customblue!27}$0.278_{\pm 0.041}$ \\
 & Vascular Disease & AUPRC $\uparrow$ & \cellcolor{customblue!0}$0.012_{\pm 0.010}$ & \cellcolor{customblue!0}$0.012_{\pm 0.007}$ & \cellcolor{customblue!4}$0.015_{\pm 0.010}$ & \cellcolor{customblue!27}$0.031_{\pm 0.021}$ & \cellcolor{customblue!27}$0.031_{\pm 0.020}$ & \cellcolor{customblue!82}$0.070_{\pm 0.091}$ & \cellcolor{customblue!100}$\mathbf{0.083_{\pm 0.104}}$ & \cellcolor{customblue!28}$0.032_{\pm 0.019}$ \\
\rowcolor{gray!10}
 & \multicolumn{2}{l}{\textit{Domain Average Rank}} & \cellcolor{customblue!85}$3.67$ & \cellcolor{customblue!29}$4.92$ & \cellcolor{customblue!96}$3.42$ & \cellcolor{customblue!0}$5.58$ & \cellcolor{customblue!100}$\mathbf{3.33}$ & \cellcolor{customblue!7}$5.42$ & \cellcolor{customblue!52}$4.42$ & \cellcolor{customblue!15}$5.25$ \\
\rowcolor{gray!10}
 & \multicolumn{2}{l}{\textit{Domain Skill Score $S$ (\%)}} & $0.0_{\pm 0.0}$ & \cellcolor{customblue!100}$\mathbf{+0.9_{\pm 3.7}}$ & \cellcolor{customblue!92}$+0.3_{\pm 2.4}$ & \cellcolor{customblue!0}$-6.4_{\pm 2.9}$ & \cellcolor{customblue!74}$-1.0_{\pm 3.1}$ & \cellcolor{customblue!0}$-6.4_{\pm 2.9}$ & \cellcolor{customblue!71}$-1.2_{\pm 5.9}$ & \cellcolor{customblue!14}$-5.4_{\pm 3.6}$ \\
\midrule
\multirow{10}{*}{\rotatebox[origin=c]{90}{\scriptsize Vitals \& blood biomarkers}} & BMI Categories & Spearman $\rho$ $\uparrow$ & \cellcolor{customblue!48}$0.331_{\pm 0.043}$ & \cellcolor{customblue!80}$0.541_{\pm 0.037}$ & \cellcolor{customblue!100}$\mathbf{0.666_{\pm 0.031}}$ & \cellcolor{customblue!15}$0.121_{\pm 0.047}$ & \cellcolor{customblue!36}$0.254_{\pm 0.043}$ & \cellcolor{customblue!16}$0.132_{\pm 0.046}$ & \cellcolor{customblue!0}$0.027_{\pm 0.046}$ & \cellcolor{customblue!6}$0.066_{\pm 0.047}$ \\
 & BMI Value & Pearson $r$ $\uparrow$ & \cellcolor{customblue!51}$0.430_{\pm 0.053}$ & \cellcolor{customblue!91}$0.725_{\pm 0.034}$ & \cellcolor{customblue!100}$\mathbf{0.788_{\pm 0.024}}$ & \cellcolor{customblue!29}$0.271_{\pm 0.046}$ & \cellcolor{customblue!28}$0.264_{\pm 0.046}$ & \cellcolor{customblue!17}$0.181_{\pm 0.051}$ & \cellcolor{customblue!10}$0.127_{\pm 0.056}$ & \cellcolor{customblue!0}$0.057_{\pm 0.053}$ \\
 & Blood Pressure Categories & Spearman $\rho$ $\uparrow$ & \cellcolor{customblue!0}$0.089_{\pm 0.069}$ & \cellcolor{customblue!100}$\mathbf{0.261_{\pm 0.066}}$ & \cellcolor{customblue!13}$0.112_{\pm 0.071}$ & \cellcolor{customblue!11}$0.108_{\pm 0.074}$ & \cellcolor{customblue!16}$0.116_{\pm 0.073}$ & \cellcolor{customblue!19}$0.121_{\pm 0.074}$ & \cellcolor{customblue!53}$0.181_{\pm 0.070}$ & \cellcolor{customblue!36}$0.151_{\pm 0.069}$ \\
 & Body Weight & Pearson $r$ $\uparrow$ & \cellcolor{customblue!98}$0.945_{\pm 0.005}$ & \cellcolor{customblue!100}$0.960_{\pm 0.004}$ & \cellcolor{customblue!100}$\mathbf{0.962_{\pm 0.004}}$ & \cellcolor{customblue!64}$0.645_{\pm 0.063}$ & \cellcolor{customblue!82}$0.804_{\pm 0.026}$ & \cellcolor{customblue!77}$0.754_{\pm 0.053}$ & \cellcolor{customblue!81}$0.790_{\pm 0.032}$ & \cellcolor{customblue!0}$0.075_{\pm 0.056}$ \\
 & HDL Cholesterol & Pearson $r$ $\uparrow$ & \cellcolor{customblue!100}$\mathbf{0.241_{\pm 0.070}}$ & \cellcolor{customblue!50}$0.195_{\pm 0.073}$ & \cellcolor{customblue!91}$0.233_{\pm 0.068}$ & \cellcolor{customblue!14}$0.162_{\pm 0.072}$ & \cellcolor{customblue!84}$0.226_{\pm 0.066}$ & \cellcolor{customblue!47}$0.192_{\pm 0.070}$ & \cellcolor{customblue!18}$0.166_{\pm 0.068}$ & \cellcolor{customblue!0}$0.149_{\pm 0.067}$ \\
 & LDL Cholesterol & Pearson $r$ $\uparrow$ & \cellcolor{customblue!100}$\mathbf{0.178_{\pm 0.065}}$ & \cellcolor{customblue!97}$0.170_{\pm 0.065}$ & \cellcolor{customblue!52}$0.065_{\pm 0.061}$ & \cellcolor{customblue!51}$0.062_{\pm 0.069}$ & \cellcolor{customblue!84}$0.141_{\pm 0.065}$ & \cellcolor{customblue!28}$0.008_{\pm 0.073}$ & \cellcolor{customblue!34}$0.024_{\pm 0.060}$ & \cellcolor{customblue!0}$-0.057_{\pm 0.062}$ \\
 & Systolic Blood Pressure & Pearson $r$ $\uparrow$ & \cellcolor{customblue!10}$0.143_{\pm 0.069}$ & \cellcolor{customblue!61}$0.209_{\pm 0.064}$ & \cellcolor{customblue!100}$\mathbf{0.259_{\pm 0.065}}$ & \cellcolor{customblue!78}$0.230_{\pm 0.061}$ & \cellcolor{customblue!86}$0.241_{\pm 0.064}$ & \cellcolor{customblue!90}$0.246_{\pm 0.066}$ & \cellcolor{customblue!5}$0.136_{\pm 0.068}$ & \cellcolor{customblue!0}$0.130_{\pm 0.062}$ \\
 & Total Cholesterol & Pearson $r$ $\uparrow$ & \cellcolor{customblue!96}$0.173_{\pm 0.068}$ & \cellcolor{customblue!77}$0.130_{\pm 0.062}$ & \cellcolor{customblue!48}$0.065_{\pm 0.063}$ & \cellcolor{customblue!0}$-0.043_{\pm 0.066}$ & \cellcolor{customblue!66}$0.105_{\pm 0.065}$ & \cellcolor{customblue!25}$0.013_{\pm 0.066}$ & \cellcolor{customblue!100}$\mathbf{0.182_{\pm 0.064}}$ & \cellcolor{customblue!6}$-0.029_{\pm 0.069}$ \\
\rowcolor{gray!10}
 & \multicolumn{2}{l}{\textit{Domain Average Rank}} & \cellcolor{customblue!83}$3.38$ & \cellcolor{customblue!100}$\mathbf{2.62}$ & \cellcolor{customblue!100}$\mathbf{2.62}$ & \cellcolor{customblue!25}$6.00$ & \cellcolor{customblue!72}$3.88$ & \cellcolor{customblue!44}$5.12$ & \cellcolor{customblue!42}$5.25$ & \cellcolor{customblue!0}$7.12$ \\
\rowcolor{gray!10}
 & \multicolumn{2}{l}{\textit{Domain Skill Score $S$ (\%)}} & $0.0_{\pm 0.0}$ & \cellcolor{customblue!96}$+18.0_{\pm 2.8}$ & \cellcolor{customblue!100}$\mathbf{+21.7_{\pm 2.3}}$ & \cellcolor{customblue!31}$-40.2_{\pm 6.0}$ & \cellcolor{customblue!51}$-22.5_{\pm 4.5}$ & \cellcolor{customblue!38}$-34.5_{\pm 5.9}$ & \cellcolor{customblue!38}$-33.7_{\pm 5.1}$ & \cellcolor{customblue!0}$-68.3_{\pm 5.9}$ \\
\midrule
\multirow{7}{*}{\rotatebox[origin=c]{90}{\scriptsize Mental well-being}} & Feel Depressed & Spearman $\rho$ $\uparrow$ & \cellcolor{customblue!99}$0.153_{\pm 0.066}$ & \cellcolor{customblue!63}$0.034_{\pm 0.069}$ & \cellcolor{customblue!100}$\mathbf{0.156_{\pm 0.067}}$ & \cellcolor{customblue!36}$-0.053_{\pm 0.067}$ & \cellcolor{customblue!0}$-0.172_{\pm 0.074}$ & \cellcolor{customblue!26}$-0.087_{\pm 0.072}$ & \cellcolor{customblue!44}$-0.027_{\pm 0.066}$ & \cellcolor{customblue!55}$0.007_{\pm 0.067}$ \\
 & Feel Happy & Spearman $\rho$ $\uparrow$ & \cellcolor{customblue!100}$\mathbf{0.209_{\pm 0.052}}$ & \cellcolor{customblue!50}$0.105_{\pm 0.054}$ & \cellcolor{customblue!73}$0.152_{\pm 0.054}$ & \cellcolor{customblue!80}$0.167_{\pm 0.053}$ & \cellcolor{customblue!65}$0.137_{\pm 0.058}$ & \cellcolor{customblue!53}$0.112_{\pm 0.057}$ & \cellcolor{customblue!0}$0.001_{\pm 0.054}$ & \cellcolor{customblue!9}$0.020_{\pm 0.056}$ \\
 & Feel Worried & Spearman $\rho$ $\uparrow$ & \cellcolor{customblue!76}$0.129_{\pm 0.060}$ & \cellcolor{customblue!89}$0.177_{\pm 0.061}$ & \cellcolor{customblue!100}$\mathbf{0.217_{\pm 0.058}}$ & \cellcolor{customblue!19}$-0.082_{\pm 0.053}$ & \cellcolor{customblue!7}$-0.128_{\pm 0.062}$ & \cellcolor{customblue!0}$-0.152_{\pm 0.057}$ & \cellcolor{customblue!25}$-0.059_{\pm 0.056}$ & \cellcolor{customblue!43}$0.005_{\pm 0.058}$ \\
 & Life Satisfaction & Spearman $\rho$ $\uparrow$ & \cellcolor{customblue!72}$0.203_{\pm 0.054}$ & \cellcolor{customblue!42}$0.134_{\pm 0.058}$ & \cellcolor{customblue!66}$0.189_{\pm 0.053}$ & \cellcolor{customblue!29}$0.105_{\pm 0.051}$ & \cellcolor{customblue!100}$\mathbf{0.266_{\pm 0.055}}$ & \cellcolor{customblue!39}$0.129_{\pm 0.059}$ & \cellcolor{customblue!14}$0.071_{\pm 0.057}$ & \cellcolor{customblue!0}$0.040_{\pm 0.052}$ \\
 & Things Are Worthwhile & Spearman $\rho$ $\uparrow$ & \cellcolor{customblue!81}$0.131_{\pm 0.056}$ & \cellcolor{customblue!100}$\mathbf{0.156_{\pm 0.053}}$ & \cellcolor{customblue!91}$0.144_{\pm 0.057}$ & \cellcolor{customblue!73}$0.120_{\pm 0.053}$ & \cellcolor{customblue!33}$0.065_{\pm 0.052}$ & \cellcolor{customblue!0}$0.021_{\pm 0.058}$ & \cellcolor{customblue!91}$0.144_{\pm 0.054}$ & \cellcolor{customblue!73}$0.119_{\pm 0.059}$ \\
\rowcolor{gray!10}
 & \multicolumn{2}{l}{\textit{Domain Average Rank}} & \cellcolor{customblue!95}$2.40$ & \cellcolor{customblue!77}$3.20$ & \cellcolor{customblue!100}$\mathbf{2.20}$ & \cellcolor{customblue!36}$5.00$ & \cellcolor{customblue!27}$5.40$ & \cellcolor{customblue!0}$6.60$ & \cellcolor{customblue!27}$5.40$ & \cellcolor{customblue!18}$5.80$ \\
\rowcolor{gray!10}
 & \multicolumn{2}{l}{\textit{Domain Skill Score $S$ (\%)}} & $0.0_{\pm 0.0}$ & \cellcolor{customblue!69}$-5.4_{\pm 3.9}$ & \cellcolor{customblue!100}$\mathbf{+0.7_{\pm 3.1}}$ & \cellcolor{customblue!28}$-13.3_{\pm 4.6}$ & \cellcolor{customblue!22}$-14.4_{\pm 4.8}$ & \cellcolor{customblue!0}$-18.7_{\pm 4.9}$ & \cellcolor{customblue!10}$-16.7_{\pm 4.5}$ & \cellcolor{customblue!17}$-15.4_{\pm 4.9}$ \\
\midrule
\multirow{7}{*}{\rotatebox[origin=c]{90}{\scriptsize Sleep \& lifestyle}} & Bedtime & Spearman $\rho$ $\uparrow$ & \cellcolor{customblue!93}$0.175_{\pm 0.058}$ & \cellcolor{customblue!89}$0.166_{\pm 0.059}$ & \cellcolor{customblue!100}$\mathbf{0.189_{\pm 0.059}}$ & \cellcolor{customblue!23}$0.025_{\pm 0.057}$ & \cellcolor{customblue!0}$-0.024_{\pm 0.055}$ & \cellcolor{customblue!59}$0.102_{\pm 0.059}$ & \cellcolor{customblue!16}$0.011_{\pm 0.055}$ & \cellcolor{customblue!35}$0.050_{\pm 0.056}$ \\
 & Currently Employed & AUPRC $\uparrow$ & \cellcolor{customblue!60}$0.869_{\pm 0.023}$ & \cellcolor{customblue!100}$\mathbf{0.934_{\pm 0.013}}$ & \cellcolor{customblue!89}$0.916_{\pm 0.016}$ & \cellcolor{customblue!52}$0.856_{\pm 0.022}$ & \cellcolor{customblue!83}$0.907_{\pm 0.018}$ & \cellcolor{customblue!43}$0.842_{\pm 0.021}$ & \cellcolor{customblue!66}$0.879_{\pm 0.021}$ & \cellcolor{customblue!0}$0.772_{\pm 0.027}$ \\
 & Sleep Duration & Spearman $\rho$ $\uparrow$ & \cellcolor{customblue!82}$0.089_{\pm 0.051}$ & \cellcolor{customblue!96}$0.109_{\pm 0.050}$ & \cellcolor{customblue!100}$\mathbf{0.114_{\pm 0.050}}$ & \cellcolor{customblue!11}$-0.010_{\pm 0.051}$ & \cellcolor{customblue!64}$0.064_{\pm 0.050}$ & \cellcolor{customblue!51}$0.045_{\pm 0.050}$ & \cellcolor{customblue!0}$-0.026_{\pm 0.050}$ & \cellcolor{customblue!16}$-0.003_{\pm 0.049}$ \\
 & Vigorous Activity Minutes & Pearson $r$ $\uparrow$ & \cellcolor{customblue!80}$0.300_{\pm 0.072}$ & \cellcolor{customblue!57}$0.208_{\pm 0.055}$ & \cellcolor{customblue!100}$\mathbf{0.377_{\pm 0.057}}$ & \cellcolor{customblue!53}$0.192_{\pm 0.055}$ & \cellcolor{customblue!58}$0.213_{\pm 0.066}$ & \cellcolor{customblue!57}$0.210_{\pm 0.062}$ & \cellcolor{customblue!59}$0.215_{\pm 0.068}$ & \cellcolor{customblue!0}$-0.015_{\pm 0.063}$ \\
 & Wake-up Time & Spearman $\rho$ $\uparrow$ & \cellcolor{customblue!81}$0.079_{\pm 0.052}$ & \cellcolor{customblue!91}$0.100_{\pm 0.058}$ & \cellcolor{customblue!100}$\mathbf{0.118_{\pm 0.053}}$ & \cellcolor{customblue!20}$-0.048_{\pm 0.054}$ & \cellcolor{customblue!57}$0.029_{\pm 0.038}$ & \cellcolor{customblue!0}$-0.090_{\pm 0.049}$ & \cellcolor{customblue!57}$0.028_{\pm 0.045}$ & \cellcolor{customblue!37}$-0.014_{\pm 0.052}$ \\
\rowcolor{gray!10}
 & \multicolumn{2}{l}{\textit{Domain Average Rank}} & \cellcolor{customblue!67}$3.00$ & \cellcolor{customblue!70}$2.80$ & \cellcolor{customblue!100}$\mathbf{1.20}$ & \cellcolor{customblue!0}$6.60$ & \cellcolor{customblue!37}$4.60$ & \cellcolor{customblue!15}$5.80$ & \cellcolor{customblue!22}$5.40$ & \cellcolor{customblue!0}$6.60$ \\
\rowcolor{gray!10}
 & \multicolumn{2}{l}{\textit{Domain Skill Score $S$ (\%)}} & $0.0_{\pm 0.0}$ & \cellcolor{customblue!98}$+11.3_{\pm 4.4}$ & \cellcolor{customblue!100}$\mathbf{+12.1_{\pm 2.9}}$ & \cellcolor{customblue!35}$-13.9_{\pm 5.4}$ & \cellcolor{customblue!66}$-1.5_{\pm 4.6}$ & \cellcolor{customblue!36}$-13.3_{\pm 5.5}$ & \cellcolor{customblue!49}$-8.2_{\pm 5.2}$ & \cellcolor{customblue!0}$-27.7_{\pm 6.0}$ \\
\midrule
\rowcolor{gray!20}
 & \multicolumn{2}{l}{\textbf{\textit{Average Rank $R$}}} & \cellcolor{customblue!78}$3.19$ & \cellcolor{customblue!85}$2.91$ & \cellcolor{customblue!100}$\mathbf{2.29}$ & \cellcolor{customblue!12}$5.94$ & \cellcolor{customblue!56}$4.14$ & \cellcolor{customblue!16}$5.79$ & \cellcolor{customblue!28}$5.29$ & \cellcolor{customblue!0}$6.46$ \\
\rowcolor{gray!20}
 & \multicolumn{2}{l}{\textbf{\textit{Aggregate Skill Score $S$ (\%)}}} & $0.0_{\pm 0.0}$ & \cellcolor{customblue!97}$+17.6_{\pm 1.6}$ & \cellcolor{customblue!100}$\mathbf{+18.8_{\pm 1.2}}$ & \cellcolor{customblue!20}$-20.1_{\pm 2.6}$ & \cellcolor{customblue!40}$-10.3_{\pm 2.3}$ & \cellcolor{customblue!22}$-19.1_{\pm 2.7}$ & \cellcolor{customblue!28}$-17.2_{\pm 2.7}$ & \cellcolor{customblue!0}$-29.8_{\pm 2.8}$ \\
\bottomrule
\end{tabular}
}
\end{table}

\textbf{Related Work on LLMs for Wearable Health Prediction.} While our evaluation is scoped to the Gemini family, concurrent work documents a similar pattern across vendors and prompting modes, suggesting our findings are not idiosyncratic to a single vendor: PH-LLM~\citep{cosentino2024towards} reports parity (not superiority) of fine-tuned Gemini-Ultra-1.0 with logistic regression on numerical wearable predictions; OpenTSLM~\citep{langer2025opentslm} shows GPT-4o below random on human-activity recognition; \cite{tan2024language} demonstrate that removing the LLM from popular time-series-LLM pipelines often \emph{improves} performance. We position our results as a within-Gemini analysis whose qualitative direction is consistent with these vendor-diverse reports; a controlled cross-vendor benchmark on this exact task suite remains future work.

\subsubsection{Gemini Vision: Rendering and Prompt Template}
\label{app:gemini_vision_details}
This subsection documents the Vision-mode rendering and prompt template used by the \textsc{Gem-2.5 Vis} and \textsc{Gem-3.1 Vis} probes. Per-task results across all three Gemini-family probe configurations (Statistics, Vision, and Agentic) are reported in Appendix~\ref{app:gemini_baselines}.

\bo{Rendering wearable weeks as images.} Each $(168, 19)$ weekly tensor is rendered as a single stacked-subplot figure, following the multivariate layout of \citet{he2025harnessingvisionlanguagemodelstime}: the seven continuous channels (iPhone and Apple Watch step counts and distances, flights climbed, heart rate, active energy) appear as line plots on top, and the twelve binary activity/sleep indicators as filled ``active'' strips below. We omit binary channels with no positive samples in the week to keep the figure compact. We use minimal axis decoration (no tick marks, frames, or legend blocks), but retain a small per-channel title of the form \texttt{"<channel> (<unit>) [vmin - vmax]"} so the VLM can identify each row and recover absolute scale (we set per-channel y-limits rather than a shared global range). Each channel is drawn in a unique, fixed color, following the per-channel coloring scheme of \citet{liu2025mllm4tsleveragingvisionmultimodal}. Gray vertical lines every 24 hours mark day boundaries. The accompanying prompt (Figure~\ref{fig:vlm-prompt}) pairs the image with a fixed channel-legend block describing the layout, a per-task description with the index-to-label map, and a forced \texttt{Answer: <\dots>} suffix for parsing. We rely on the cross-VLM ablation in \citet[Appendix~B]{he2025harnessingvisionlanguagemodelstime}, which finds the rendering transfers across GPT-4o, Claude~3.5, Gemini~2.0, and Qwen-2.5-72B, to motivate applying the same design to Gemini~3.1~Pro.

\begin{figure}[h!]
\centering
\fbox{%
\begin{minipage}{0.95\linewidth}
\textbf{Wearable VLM Prompt}\par\medskip
\hrule\medskip
\small
You are shown one image and asked to make a single prediction. The image will contain:
\begin{itemize}\itemsep0pt
    \item A plot of one week (168 hours) of wearable sensor data from a single participant.
    \item \textbf{Top panels:} seven continuous signals as line plots -- iPhone step count, iPhone distance, flights climbed, Apple Watch step count, Apple Watch distance, heart rate (bpm), and active energy (cal/hr). Each panel title states the channel name, unit, and observed value range \texttt{[vmin--vmax]}.
    \item \textbf{Bottom panels:} binary activity and sleep indicators rendered as colored strips (filled = active). Channels with no activity in the week are omitted.
    \item Faint vertical gray lines mark day boundaries every 24 hours.
\end{itemize}

\textbf{Task.} \textit{$\langle$task description$\rangle$}
\begin{itemize}\itemsep2pt
    \item \emph{Binary (BiologicalSex):}\\
    \texttt{Predict the person's biological sex.\textbackslash n 0=Female, 1=Male}
    \item \emph{Ordinal (BMI\_categories):}\\
    \texttt{Predict the person's BMI category.\textbackslash n 0=Underweight, 1=Normal weight, 2=Overweight, 3=Obesity, 4=Morbid Obesity}
    \item \emph{Regression (age):}\\
    \texttt{Predict the person's age in years.}
\end{itemize}

\textbf{Response format.} End your response with a single line of the form:
\begin{center}
\texttt{Answer: $\langle$one of the labels above, or a numeric value$\rangle$}
\end{center}
\end{minipage}}
\caption{Prompt template used for the Gemini VLM zero-shot evaluation.}
\label{fig:vlm-prompt}
\end{figure}

\clearpage
\section{Imputation Tasks}
\label{sec:imputation}

Wearable sensor data is inherently incomplete: devices are removed, batteries die, sensors malfunction, and physiological signals drop out during specific activities. Any model operating on such data must either tolerate or recover from missingness. We define a structured imputation benchmark that evaluates methods across six masking approaches grounded in real-world failure modes, using both continuous and binary reconstruction metrics. We provide two primary tasks for imputation. First, a \textit{single-day imputation} task. Here a model has to impute based on a single day of passive data. Second, a \textit{long-context imputation} task, where the goal is to impute using longer context windows. While we do not limit the context window for the public benchmark, for practical purposes in this evaluation we limited neural models to  7-day windows. Since imputation is particularly useful if it can be applied for higher-frequency data such that the resulting imputed data is broadly usable, we set this task to take place on minute-resolution passive data.

We denote an imputation sample as $\mathbf{X} \in \mathbb{R}^{C \times T}$ with $C$ channels and $T$ time steps (minutes). For single-day methods, $T=1{,}440$; for 7-day methods, $T=10{,}080$. The original validity mask $\mathbf{M} \in \{0,1\}^{C \times T}$ indicates positions with observed data ($M_{c,t}=1$). Each masking approach produces an artificial mask $\mathbf{A} \in \{0,1\}^{C \times T}$ where $A_{c,t}=1$ marks positions to be imputed, with $\mathbf{A} \leq \mathbf{M}$ element-wise (only observed positions can be masked). Data preprocessing for imputation follows the general benchmark pre-processing pipeline outlined in Appendix \ref{app:model_input_representations}.

\subsection{Masking Approaches}
\label{sec:masking_scenarios}

\begin{figure}[h]
  \centering
  \includegraphics[width=\linewidth]{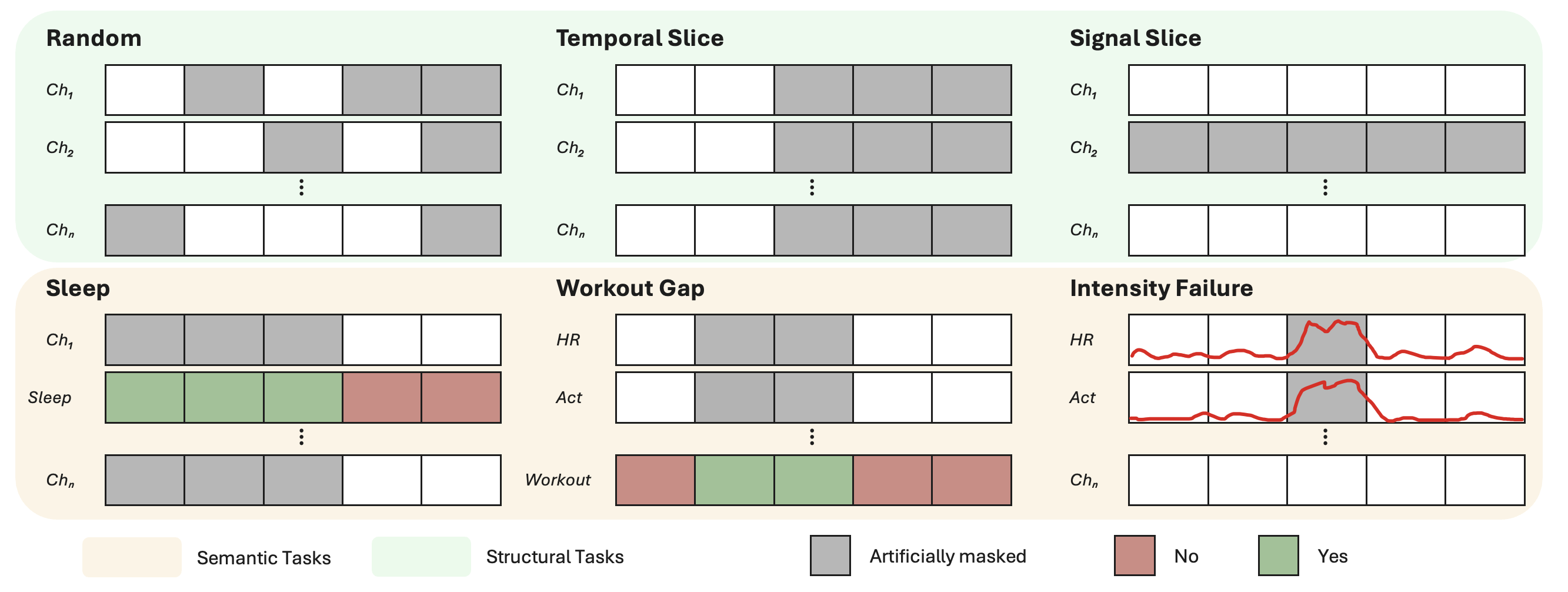}
  \caption{%
    \textbf{Imputation task structure.}
    Six masking approaches on daily minute-level multi-channel wearable
    tensors, organized into \emph{structural} (top, signal-agnostic)
    and \emph{semantic} (bottom, event-driven) tiers. Gray cells mark
    artificially masked positions on which methods are scored.
  }
  \label{fig:imputation_task_structure}
\end{figure}

We organize masking approaches into two tiers. \emph{Structural} masks simulate generic data-collection failures, while \emph{semantic} masks target physiologically meaningful gaps tied to specific activities or sensor limitations (see Figure \ref{fig:imputation_task_structure}).

The imputation masks for all imputation tasks are pre-defined. We generate the masks across the provided user split and cap each user to 91 days in the test set for imputation evaluation. The rationale for 91 days is two-fold: first, we want to avoid a few superusers with 10 years of data dominating the results; and second, for performance, this keeps the evaluation pipeline runnable in a reasonable amount of time. When comparing performance to the full run we found only minor differences and most of those were due to long-term users being over-represented.

\paragraph{Tier~1: Structural masks.}

\begin{itemize}[leftmargin=*]
  \item \textbf{Random noise.} Short channel-local bursts are masked to simulate sporadic transmission errors or transient sensor glitches. We partition the tensor into non-overlapping patches of $p=30$ consecutive minutes per channel, yielding $C \cdot \lfloor T/p \rfloor$ candidate patches. Candidates are randomly permuted and greedily selected until the number of newly masked valid positions reaches $r \cdot \|\mathbf{M}\|_1$ with mask ratio $r=0.5$.

  \item \textbf{Temporal slice.} Contiguous blocks of time are masked across all channels simultaneously, simulating periods where the device was removed or powered off. Block sizes are drawn uniformly from $[b_{\min}, b_{\max}] = [30, 60]$ minutes. The number of blocks is estimated as
  \begin{equation}
    n_b = \left\lceil \frac{\log(1-r)}{\log\,\bigl(1 - \bar{b}/T\bigr)} \right\rceil, \quad \bar{b} = \frac{b_{\min}+b_{\max}}{2},
    \label{eq:temporal_nblocks}
  \end{equation}
  with target mask ratio $r=0.25$. Each block start is sampled uniformly from valid time steps, and the resulting timestep mask is broadcast to all channels: $A_{c,t} = M_{c,t} \cdot \mathbf{1}[t \in \text{masked blocks}]$.

  \item \textbf{Signal slice.} Entire sensor channels are masked for the full observation window, simulating complete sensor failure or a missing modality. With equal probability, one of two modes is selected:
  \begin{itemize}
    \item \emph{Mode~A (individual channels):} $\lceil r \cdot |\mathcal{C}_{\text{valid}}| \rceil$ channels are sampled uniformly without replacement from the set of channels with any observed data, with $r=0.5$.
    \item \emph{Mode~B (device group):} One device group (e.g., all watch channels or all phone channels) is selected uniformly at random, and all its channels are masked.
  \end{itemize}
  For each selected channel $c$, we set $A_{c,:} = M_{c,:}$.
\end{itemize}

\paragraph{Tier~2: Semantic masks.}

\begin{itemize}[leftmargin=*]
  \item \textbf{Sleep gap.} All channels are masked during detected sleep periods, simulating the common behavior of removing a wearable device before bed. Sleep is detected at each time step as
  \begin{equation}
    s_t = \mathbf{1}\bigl[X_{\text{asleep},t} > 0\bigr] \lor \mathbf{1}\bigl[X_{\text{inbed},t} > 0\bigr],
    \label{eq:sleep_detection}
  \end{equation}
  where $X_{\text{asleep}}$ and $X_{\text{inbed}}$ denote the corresponding activity indicator channels. For all time steps where $s_t = 1$, every channel except the two sleep indicators is masked: $A_{c,t} = M_{c,t} \cdot s_t$ for $c \notin \{\text{asleep}, \text{inbed}\}$.

  \item \textbf{Workout gap.} Continuous sensor channels are masked during detected workout periods, simulating motion-artifact--induced signal dropout during vigorous exercise. A workout is detected at time $t$ when any of the binary workout-type channels is active:
  \begin{equation}
    w_t = \bigvee_{c \in \mathcal{C}_{\text{workout}}} \mathbf{1}\bigl[X_{c,t} > 0\bigr],
    \label{eq:workout_detection}
  \end{equation}
  where $\mathcal{C}_{\text{workout}}$ spans 10 activity-type channels (walking, running, cycling, etc.). Only heart rate and active energy burned channels are masked during detected workouts; binary channels are not affected.

  \item \textbf{Intensity failure.} Continuous sensor channels are masked during high-intensity activity intervals, simulating sensor saturation or clipping at elevated physiological loads. High intensity is detected where heart rate exceeds a threshold $\tau = 160$~BPM. Only contiguous runs of $\geq 5$ consecutive high-intensity minutes qualify:
  \begin{equation}
    h_t = \mathbf{1}\bigl[X_{\text{HR},t} > \tau\bigr], \quad
    \mathcal{R} = \bigl\{(t_s, t_e) : h_t = 1\;\forall\, t \in [t_s, t_e],\; t_e - t_s \geq 4\bigr\}.
    \label{eq:intensity_detection}
  \end{equation}
  Heart rate and active energy burned channels are masked at all time steps belonging to qualifying runs $\mathcal{R}$. Binary channels are not affected.
\end{itemize}

Table~\ref{tab:masking_scenarios} summarizes the approaches, their real-world motivation, and key parameters.

\begin{table}[t]
  \caption{Overview of masking approaches. ``Channels masked'' indicates whether continuous (C), binary (B), or both channel types are affected.}
  \label{tab:masking_scenarios}
  \centering
  \small
  \renewcommand{\arraystretch}{1.15}
  \begin{tabular}{llccc}
    \toprule
    Approach & Real-world failure mode & Tier & Channels masked & Key parameters \\
    \midrule
    Random noise       & Sporadic transmission errors        & Structural & C + B & $r{=}0.5$, $p{=}30$ \\
    Temporal slice     & Device removal or power-off         & Structural & C + B & $r{=}0.25$, $b \in [30,60]$ \\
    Signal slice       & Complete sensor/modality failure     & Structural & C + B & $r{=}0.5$ \\
    Sleep gap          & Device removed during sleep         & Semantic   & C + B & data-driven \\
    Workout gap        & Motion-artifact dropout in exercise & Semantic   & C only & data-driven \\
    Intensity failure  & Sensor saturation at high intensity & Semantic   & C only & $\tau{=}160$~BPM, $\geq 5$~min \\
    \bottomrule
  \end{tabular}
\end{table}

\subsection{Imputation Methods Overview}
\label{sec:baseline_methods}

We evaluate seventeen imputation methods in total. Among the deep learning architectures, BRITS, DLinear, FEDformer, and TimesNet are implemented via the PyPOTS library~\citep{du2023pypots}. For BRITS, we implement gradient clipping (maximum norm of 1.0) to prevent training instability and numerical overflow caused by large loss spikes. While we explored tuning the MIT/ORT weights for DLinear, this did not yield significant improvements; consequently, due to computational resource constraints, we opted not to extend this tuning to the other PyPots neural imputers. Finally, we excluded MICE \citep{azur2011multiple} based on computational considerations at our benchmark's scale and concerns about its missing-at-random assumption. This choice is informed by LSM-2's decision to omit MICE based on prior performance and missingness assumptions in its own data \citep{xu2025lsm}.

\subsubsection{Single-day methods}
\label{sec:single_day_methods}
The following methods operate solely on a single-day sample without access to further user history.

\bo{Statistical baselines.}
\begin{itemize}[leftmargin=*]
  \item \textbf{Mean}: replaces each masked entry with the per-channel training-set mean.
  \item \textbf{Mode}: replaces each masked entry with the per-channel training-set mode (primarily relevant for binary channels).
  \item \textbf{Linear interpolation}: interpolates linearly between the nearest observed values in each channel along the time axis.
  \item \textbf{Last observation carried forward (LOCF)}: fills each masked entry with the most recent observed value in the same channel.
  \item \textbf{Temporal mean}: computes per-channel, per-minute-of-day means from the training set by folding multi-day windows into a standard 24-hour (1440-minute) diurnal profile via modular indexing ($t \bmod 1440$), capturing population-level circadian patterns such as resting heart rate at night or step-count peaks during the day. Falls back to the global channel mean for (channel, minute) pairs with no observations.
  \item \textbf{Temporal mode}: analogous to temporal mean but uses the per-channel, per-minute-of-day mode (most frequent rounded value), primarily relevant for binary channels whose activity patterns follow regular diurnal schedules.
\end{itemize}

\bo{Learned models.}
\begin{itemize}[leftmargin=*]
  \item \textbf{BRITS}~\citep{cao2018brits}: bidirectional RNN imputation model (Appendix~\ref{sec:imputation_baselines}).
  \item \textbf{DLinear}~\citep{zeng2023dlinear}: decomposition-linear model with trend/seasonality separation (Appendix~\ref{sec:imputation_baselines}).
  \item \textbf{FEDformer}~\citep{zhou2022fedformer}: Fourier-enhanced Transformer (Appendix~\ref{sec:imputation_baselines}).
  \item \textbf{TimesNet}~\citep{wu2023timesnet}: temporal 2D convolution model over reshaped time series (Appendix~\ref{sec:imputation_baselines}).
  \item \textbf{LSM2}~\citep{xu2025lsm}: masked autoencoder with adaptive and inherited masking (Appendix~\ref{sec:LSM-2_architecture}).
\end{itemize}

\subsubsection{Long-context methods}
\label{sec:extended_history_methods}

When a user's context data across multiple weeks is available, imputation can exploit individual-level patterns. The following methods operate on 7-day windows built within each evaluation split by sorting each user's daily samples chronologically and chunking them into non-overlapping windows. Calendar gaps are preserved rather than interpolated, incomplete tail windows are left-padded with sentinel days, and each day slot carries its true calendar-day offset relative to the first non-padded day. Methods impute the full 7-day tensor, but evaluation slices predictions back to real daily segments and scores only non-padded days with nonzero artificial masks.

Personalized statistical methods compute per-user fill values from that user's samples within the same evaluation split, falling back to population-level statistics when user-specific data is insufficient.

\begin{itemize}[leftmargin=*]
  \item \textbf{Personalized mean}: replaces each masked entry with the per-channel mean computed from the user's own context samples. Falls back to the population mean for channels where the user has no observations.
  \item \textbf{Personalized mode}: replaces each masked entry with the per-channel mode from the user's context samples. Falls back to the population mode.
  \item \textbf{Personalized temporal mean}: computes per-user, per-channel, per-minute-of-day means by folding the user's multi-week context into a personalized 24-hour diurnal profile. Applies a three-level fallback chain: user-specific minute mean $\to$ user-specific channel mean $\to$ population-level temporal mean.
  \item \textbf{DLinear (7-day)}: the single-day DLinear trained on concatenated 7-day inputs ($C \times 10{,}080$ time steps; Appendix~\ref{sec:imputation_baselines}).
  \item \textbf{LSM2 (7-day)}: the daily \textsc{LSM2} encoder--decoder applied to 7-day concatenated inputs with a coarser 60-minute patch size. This reduces the weekly token count to $19 \times 24 \times 7 = 3{,}192$, making dense self-attention tractable and providing a dense weekly baseline for \textsc{LSM2-Sparse}. See Appendix~\ref{sec:LSM-2_architecture}.
  \item \textbf{LSM2-Sparse (7-day)}: a two-stage masked autoencoder that extends the daily \textsc{LSM2} to exploit multi-day context via a sparse cross-day decoder while retaining 10-minute per-day patches. Its cross-day attention layers consume the true calendar \texttt{day\_offsets} for each window and use RoPE to distinguish consecutive from gappy histories. See Appendix~\ref{sec:LSM-2_architecture}.
\end{itemize}

\subsection{Raw Metrics}
\label{sec:imputation_metrics}

We calculate two complementary metrics evaluated exclusively on artificially masked positions. For each channel $c$, let $\mathcal{A}_c = \{t : A_{c,t} = 1\}$ denote the set of artificially masked time indices.

\paragraph{MAE.} For continuous channels $\mathcal{C}_{\text{cont}}$, we compute the per-channel mean absolute error over artificially masked positions:
\begin{equation}
  \text{MAE}_c = \frac{1}{|\mathcal{A}_c|}\sum_{t \in \mathcal{A}_c}\bigl|\hat{x}_{c,t} - x_{c,t}\bigr|,
  \label{eq:mae}
\end{equation}
where $\hat{x}_{c,t}$ is the imputed value and $x_{c,t}$ the ground truth. We use MAE as the continuous error fed to the aggregate Skill Score and Average Rank; for binary channels the corresponding error is $1 - \text{ROC\,AUC}_c$ (defined next). Appendix~\ref{sec:imputation_aggregation} details how these per-channel errors are aggregated into the reported scores.

\paragraph{Macro ROC AUC.} For binary channels $\mathcal{C}_{\text{bin}}$, we compute the area under the receiver operating characteristic curve per channel using continuous-valued predictions against binarized ground truth ($y_{c,t} = \mathbf{1}[x_{c,t} > 0.5]$). The macro average is:
\begin{equation}
  \text{macro ROC AUC} = \frac{1}{|\mathcal{C}_{\text{bin}}^{*}|}\sum_{c \in \mathcal{C}_{\text{bin}}^{*}} \text{ROC\,AUC}_c,
  \label{eq:rocauc}
\end{equation}
where $\mathcal{C}_{\text{bin}}^{*} \subseteq \mathcal{C}_{\text{bin}}$ excludes channels with only a single class present in the masked ground truth. Workout gap and intensity failure mask only continuous channels and therefore have no ROC AUC scores. For aggregate scoring, each binary channel contributes the error $1 - \text{ROC\,AUC}_c$ (Appendix~\ref{sec:imputation_aggregation}).

\paragraph{Handling non-finite imputations.} Any artificially masked cell that an imputer leaves non-finite (\texttt{NaN} or \texttt{Inf}), or fails to fill, is substituted before scoring with a per-channel fallback computed on the training split: the channel mean for continuous channels and the majority class for binary channels (the \textbf{Mean}/\textbf{Mode} baseline of Appendix~\ref{sec:single_day_methods}). This mirrors the forecasting track's NaN handling and surfaces a model's inability to impute in its score rather than silently dropping those cells; we record the resulting fallback substitution rate.

\subsection{Aggregation and Scoring}
\label{sec:imputation_aggregation}

\paragraph{Tasks, scopes, and channel categories.} The headline imputation columns in Table~\ref{tab:imputation_main_results} (the overall Skill Score $S$, the Average Rank, and the per-category Activity / Physiology / Sleep / Workout / Semantic columns) are produced by aggregating the per-channel errors of Section~\ref{sec:imputation_metrics} through the unified Skill Score calculation of Appendix~\ref{sec:scoring_methodology}. We reuse the base clipping, geometric-mean, and bootstrap definitions given there (clip bounds $\ell = 0.01$, $u = 100$) and specialize them to the imputation track below, where channels split very unevenly across categories and the aggregation is therefore \emph{category-balanced}: the four reporting categories act as equal-weight buckets in a hierarchical mean rather than being pooled as a flat set of per-channel tasks. Throughout, $p \in \mathcal{P}$ indexes participants, a task $r = (s, c)$ is a (approach, channel) pair, $m$ is the evaluated method, and $b = \textsc{LOCF}$ is the reference baseline.

The unit of evaluation is a \emph{task} $r = (s, c)$: a single channel $c$ scored under a single masking approach $s$. A \emph{scope} is a collection of tasks aggregated into one reported number; each column of Table~\ref{tab:imputation_main_results} corresponds to a scope, defined by a set of masking approaches $\mathcal{S}_{\text{scope}}$ together with the channel buckets scored within them. The $C = 19$ channels partition into the continuous channels $\mathcal{C}_{\text{cont}}$ (seven channels) and the binary channels $\mathcal{C}_{\text{bin}}$ (twelve channels), and are further grouped into four reporting categories that serve as aggregation \emph{buckets}: \emph{Activity} (five continuous channels---phone and watch step counts and distances, and flights climbed), \emph{Physiology} (two continuous channels---heart rate and active energy), \emph{Sleep} (two binary channels---the asleep and in-bed indicators), and \emph{Workout} (the ten binary workout-type indicators). The \emph{structural} approaches (random noise, temporal slice, signal slice) and \emph{semantic} approaches (sleep gap, workout gap, intensity failure) are those defined in Section~\ref{sec:masking_scenarios}.

\paragraph{Per-participant error.} For each task we first reduce all participants artificially masked cells to a single error. For a continuous task ($c \in \mathcal{C}_{\text{cont}}$) we pool the absolute errors into $E^{\text{cont}}$ (superscript $\text{cont}$: \emph{continuous}); for a binary task ($c \in \mathcal{C}_{\text{bin}}$) we use that participant's pooled ROC AUC to form $\tilde{E}^{\text{bin}}$ ($\text{bin}$: \emph{binary}):
\begin{equation}
  E^{\text{cont}}_{m,p,r} = \frac{1}{N_{p,r}}\sum_{\text{masked cells}} |\hat{x} - x|, \qquad
  \tilde{E}^{\text{bin}}_{m,p,r} = \max\,\bigl(1 - \text{ROC\,AUC}_{m,p,r},\ \varepsilon\bigr),
  \label{eq:imp_peruser_error}
\end{equation}
where the sum runs over participant $p$'s artificially masked cells ($A_{c,t}=1$) for task $r$, pooled over that participant's daily samples; $N_{p,r}$ is their count and $\varepsilon = 0.005$. Participants whose binary task is single-class (an undefined AUC) are dropped. The tilde in $\tilde{E}$ marks the $\varepsilon$-floored binary error, which enters only the paired-ratio path~\eqref{eq:imp_paired_ratio}; the unfloored $1 - \text{ROC\,AUC}$ (written without a tilde) is used for the Average Rank~\eqref{eq:imp_avg_rank}.

\paragraph{Collapsed binary categories.} The reported Sleep and Workout columns, and the binary side of the overall column, collapse each binary category $\kappa \in \{\text{Sleep}, \text{Workout}\}$ into a single per-participant error per approach (superscript $\text{coll}$: \emph{collapsed}), averaging the per-channel binary errors over the category's channels that have a defined AUC for that participant and then applying the $\varepsilon$ floor once to the resulting mean:
\begin{equation}
  \tilde{E}^{\text{coll}}_{m,p,(s,\kappa)}
    = \max \, \biggl(\frac{1}{|\kappa_p|}\sum_{c \in \kappa_p}
      \bigl(1 - \text{ROC\,AUC}_{m,p,(s,c)}\bigr),\ \varepsilon\biggr),
  \label{eq:imp_collapsed_error}
\end{equation}
where $\kappa_p \subseteq \kappa$ is the set of channels in category $\kappa$ with a defined AUC for participant $p$. Each structural approach therefore contributes one Sleep task and one Workout task rather than two and ten per-channel tasks; this equal weighting prevents the ten workout channels from dominating the two sleep channels and the seven continuous channels. Continuous categories are not collapsed.

\paragraph{Per-task paired ratio.} Let $E^{\star}_{m,p,r}$ denote the per-participant error used for task $r$: $E^{\text{cont}}$ for a continuous task, $\tilde{E}^{\text{bin}}$ for a per-channel binary task, and $\tilde{E}^{\text{coll}}$ for a collapsed-category task from \eqref{eq:imp_peruser_error} and \eqref{eq:imp_collapsed_error}. For each task the ratio against the baseline is formed \emph{within} each participant and then geometrically averaged over participants (it is not a ratio of pooled errors):
\begin{equation}
  R_{m,r} = \exp\,\Bigl(\tfrac{1}{|\mathcal{P}_r|}\sum_{p\in\mathcal{P}_r}
            \log \text{clip}\bigl(E^{\star}_{m,p,r}/E^{\star}_{b,p,r},\,\ell,u\bigr)\Bigr),
  \label{eq:imp_paired_ratio}
\end{equation}
where $\mathcal{P}_r$ is the set of participants for whom both the method and baseline errors are defined and finite with $E^{\star}_{b,p,r} > 0$.

\paragraph{Skill Score per scope.} The Skill Score is a category-balanced hierarchical geometric mean over the buckets introduced above. Within a masking approach $s$, let $\mathcal{K}_s$ be the categories with at least one scored task, and let $\mathcal{T}_{s,k}$ be the tasks of bucket $k$ in approach $s$---the continuous per-channel tasks for Activity and Physiology, and the single collapsed task $(s,\kappa)$ of \eqref{eq:imp_collapsed_error} for Sleep ($\kappa = \text{Sleep}$) and Workout ($\kappa = \text{Workout}$). Per-channel binary tasks never enter directly; a binary category reaches a scope only through its collapsed task. For a scope spanning the approach set $\mathcal{S}_{\text{scope}}$,
\begin{equation}
  S_{m,\text{scope}} = 1 - \exp\,\Biggl(
      \frac{1}{|\mathcal{S}_{\text{scope}}|}\sum_{s\in\mathcal{S}_{\text{scope}}}
      \frac{1}{|\mathcal{K}_s|}\sum_{k\in\mathcal{K}_s}
      \frac{1}{|\mathcal{T}_{s,k}|}\sum_{r\in\mathcal{T}_{s,k}}
      \log \text{clip}(R_{m,r},\ell,u)\Biggr).
  \label{eq:imp_skill_scope}
\end{equation}
Reading the three means from the inside out: within each (approach, bucket) pair we average the clipped log-ratios over the bucket's tasks, then average over the buckets present in the approach, then over the approaches in the scope. Each bucket therefore has equal voice within an approach and each approach equal voice in the scope, so neither Activity's five channels nor Workout's ten can dominate the aggregate.

\paragraph{Scopes reported in the main table.} Each column fixes the approach set $\mathcal{S}_{\text{scope}}$ of \eqref{eq:imp_skill_scope}:
\begin{itemize}[leftmargin=*]
  \item \textbf{Overall ($S$):} all six approaches. Within each approach the present buckets are averaged as in \eqref{eq:imp_skill_scope}; the Activity and Physiology buckets contribute wherever their channels are masked (all six approaches), while the Sleep and Workout buckets contribute only in the three structural approaches, since binary channels are excluded from the semantic approaches.
  \item \textbf{Activity / Physiology:} a single bucket over the three structural approaches---equivalently a per-channel geometric mean over the structural approaches crossed with the five Activity channels ($15$ tasks) or the two Physiology channels ($6$ tasks).
  \item \textbf{Sleep / Workout:} the single collapsed bucket over the three structural approaches, i.e.\ a geometric mean over the collapsed tasks $\{(s, \kappa) : s \text{ structural}\}$ for $\kappa \in \{\text{Sleep}, \text{Workout}\}$ ($3$ tasks each; \eqref{eq:imp_collapsed_error}).
  \item \textbf{Semantic:} the three semantic approaches. Only continuous buckets are present (binary channels are excluded from these approaches), so each approach reduces to its Activity and/or Physiology buckets.
\end{itemize}
The per-category Activity, Physiology, Sleep, and Workout columns are restricted to the structural approaches, whereas the Overall column spans all six; the binary categories are scored only on the structural approaches, so they reach the Overall column through exactly the structural collapsed tasks that define the Sleep and Workout columns.

\paragraph{Channel restriction.} Only artificially masked channels are scored: the pairing step emits records only at masked positions, so unmasked channels contribute no tasks. In addition, binary channels are excluded from the three semantic approaches (sleep gap, workout gap, intensity failure). The net scored channels are the two Physiology channels (heart rate and active energy) for workout gap and intensity failure, and all seven continuous channels for sleep gap.

\paragraph{Average Rank.} For each task, methods are ranked by their per-participant error---$E^{\text{cont}}$ for a continuous task, and the \emph{unfloored} $1 - \text{ROC\,AUC}$ (per channel, or its per-category mean for a collapsed task) for a binary task---in ascending order with ties averaged. The per-participant ranks of method $m$ are averaged into a task rank $\bar{\rho}_{m,r}$, which is then reduced to the scope through the \emph{same} category-balanced hierarchy as the Skill Score, but with arithmetic means at every level:
\begin{equation}
  \bar{\rho}_{m,r} = \frac{1}{|\mathcal{P}_r|}\sum_{p} \text{rank}_p\bigl(E_{\cdot,p,r}\bigr), \qquad
  \rho_{m,\text{scope}} = \frac{1}{|\mathcal{S}_{\text{scope}}|}\sum_{s\in\mathcal{S}_{\text{scope}}}
      \frac{1}{|\mathcal{K}_s|}\sum_{k\in\mathcal{K}_s}
      \frac{1}{|\mathcal{T}_{s,k}|}\sum_{r\in\mathcal{T}_{s,k}}\bar{\rho}_{m,r},
  \label{eq:imp_avg_rank}
\end{equation}
where $E_{\cdot,p,r}$ is the vector of these per-participant errors across the competing methods and $\text{rank}_p$ returns method $m$'s position within it. Ranks are computed on errors, not on Skill Scores; the reported Average Rank uses the overall scope (all six approaches), as $S$ does.

\paragraph{Bootstrap confidence intervals.} For the main benchmark results, we compute participant-level bootstrap confidence intervals (CIs). We resample participants with replacement using a single shared, seeded draw matrix per split ($B = 1000$ replicates). Each participant's contribution is precomputed once---the pooled absolute errors for a continuous channel, and a single ROC AUC for a binary channel---and every draw re-aggregates these per-participant statistics through the full skill / rank pipeline. We report each score as its point estimate (computed on the held-out test split) together with a $95\%$ bootstrap percentile CI.

\subsection{Imputation Results}
\label{sec:imputation_results}
Here we present additional results of the imputation evaluation. Results for \textit{Single-day imputation} (Section~\ref{sec:single_day}) uses only the current daily sample, while \textit{long-context imputation} (Section~\ref{sec:extended_history}) additionally leverages a user's historical data as well as any additional information that could be leveraged.

Table~\ref{tab:imputation_appendix_skill_by_scenario} complements the main-paper summary in Table~\ref{tab:imputation_main_results} by listing all six masking approaches explicitly. We additionally report raw approach-level metrics for interpretability: MAE for continuous channels and macro ROC AUC for binary channels---the same per-channel errors that feed the aggregate Skill Score and Average Rank (Appendix~\ref{sec:imputation_aggregation}).

\begin{table}[t!]
    \renewcommand{\arraystretch}{1.05}
    \centering
    \captionsetup{width=\textwidth}
    \caption{\textbf{Imputation Results by Masking Scenario.} Aggregate Skill Score $S$ (in \%; $0=$LOCF reference), Average Rank $R$, and per-scenario Skill Scores across all six masking scenarios (lower is better for $R$; higher otherwise). Single-day methods above; long-context methods ($\geq 7\times 1440$ time steps) below. Gradients computed within each track. The Average Rank $R$ is a competition rank over the methods in the table; this table additionally includes LSM-2 (7-day) in the long-context pool (absent from Table~\ref{tab:imputation_main_results}), so every $R$ here is slightly higher (worse) than its counterpart there. The baseline-paired $S$ (computed against LOCF) is identical across the two tables. Values are point estimates on the held-out test split; sub/superscripts give the $95\%$ bootstrap percentile confidence interval ($B{=}1000$).}
    \label{tab:imputation_appendix_skill_by_scenario}
    \small
    \setlength{\tabcolsep}{1.5pt}
    \resizebox{\linewidth}{!}{%
    \begin{tabular}{l cccccccc}
    \toprule[1.5pt]
    \textbf{Method} & $S\uparrow$ & $R\downarrow$ & Random noise\,$\uparrow$ & Temporal slice\,$\uparrow$ & Signal slice\,$\uparrow$ & Sleep gap\,$\uparrow$ & Workout gap\,$\uparrow$ & Intensity failure\,$\uparrow$ \\
    \midrule
    \multicolumn{9}{l}{\textbf{\emph{Single-day imputation}}} \\
    \hline
    \multicolumn{9}{l}{\cellcolor[HTML]{EFEFEF}\textit{Statistical Models}} \\
    Linear & \cellcolor{customblue!78}$+21.5^{+0.7}_{-1.2}$ & \cellcolor{customblue!67}$7.0^{+0.1}_{-0.1}$ & \cellcolor{customblue!85}$+47.1^{+1.3}_{-2.0}$ & \cellcolor{customblue!98}$+56.8^{+0.9}_{-2.2}$ & $+0.0^{+0.0}_{-0.0}$ & \cellcolor{customblue!72}$-4.1^{+2.4}_{-2.9}$ & \cellcolor{customblue!51}$+15.1^{+1.3}_{-1.1}$ & \cellcolor{customblue!100}$\mathbf{-15.9}^{+5.2}_{-5.7}$ \\
    LOCF \textit{(reference)} & $0.0$ & \cellcolor{customblue!52}$8.4^{+0.1}_{-0.1}$ & $0.0$ & $0.0$ & $0.0$ & $0.0$ & $0.0$ & $0.0$ \\
    Temporal mode & \cellcolor{customblue!63}$-6.2^{+2.4}_{-2.3}$ & \cellcolor{customblue!35}$10.0^{+0.1}_{-0.1}$ & \cellcolor{customblue!49}$-34.1^{+3.4}_{-3.0}$ & \cellcolor{customblue!55}$-53.2^{+4.7}_{-4.7}$ & \cellcolor{customblue!58}$+49.9^{+0.8}_{-0.8}$ & \cellcolor{customblue!100}$\mathbf{+75.1}^{+1.6}_{-2.0}$ & $-25.7^{+3.0}_{-2.9}$ & $-346.7^{+38.9}_{-40.6}$ \\
    Mode & \cellcolor{customblue!51}$-27.3^{+2.7}_{-2.5}$ & \cellcolor{customblue!29}$10.6^{+0.1}_{-0.1}$ & \cellcolor{customblue!25}$-90.6^{+4.8}_{-4.3}$ & \cellcolor{customblue!27}$-124.5^{+7.3}_{-6.8}$ & \cellcolor{customblue!34}$+29.2^{+0.6}_{-0.6}$ & \cellcolor{customblue!100}$+74.4^{+1.6}_{-1.9}$ & \cellcolor{customblue!3}$-23.5^{+3.0}_{-2.9}$ & \cellcolor{customblue!1}$-343.9^{+38.9}_{-40.2}$ \\
    Temporal mean & \cellcolor{customblue!49}$-31.2^{+3.1}_{-3.2}$ & \cellcolor{customblue!32}$10.4^{+0.1}_{-0.1}$ & \cellcolor{customblue!57}$-16.2^{+3.6}_{-3.5}$ & \cellcolor{customblue!63}$-31.9^{+4.6}_{-5.4}$ & \cellcolor{customblue!62}$+53.7^{+1.3}_{-1.3}$ & \cellcolor{customblue!56}$-51.6^{+10.6}_{-13.3}$ & \cellcolor{customblue!19}$-10.3^{+2.3}_{-2.1}$ & \cellcolor{customblue!5}$-329.8^{+31.3}_{-34.7}$ \\
    Mean & $-119.7^{+4.7}_{-4.4}$ & $13.4^{+0.0}_{-0.0}$ & $-147.1^{+6.5}_{-5.9}$ & $-191.8^{+9.9}_{-9.5}$ & $+0.0^{+0.0}_{-0.0}$ & $-212.9^{+21.5}_{-25.7}$ & \cellcolor{customblue!15}$-14.0^{+2.3}_{-2.2}$ & \cellcolor{customblue!3}$-336.8^{+32.5}_{-35.5}$ \\
    \hline
    \multicolumn{9}{l}{\cellcolor[HTML]{EFEFEF}\textit{Neural Models}} \\
    LSM-2~\cite{xu2025lsm} & \cellcolor{customblue!100}$\mathbf{+61.4}^{+0.5}_{-1.2}$ & \cellcolor{customblue!100}$\mathbf{3.8}^{+0.1}_{-0.1}$ & \cellcolor{customblue!100}$\mathbf{+81.1}^{+0.1}_{-1.0}$ & \cellcolor{customblue!100}$\mathbf{+61.1}^{+0.6}_{-2.0}$ & \cellcolor{customblue!100}$\mathbf{+86.7}^{+0.4}_{-0.6}$ & \cellcolor{customblue!89}$+44.7^{+3.6}_{-4.1}$ & \cellcolor{customblue!100}$\mathbf{+53.6}^{+1.4}_{-1.6}$ & \cellcolor{customblue!95}$-32.7^{+8.1}_{-9.6}$ \\
    BRITS~\cite{cao2018brits} & \cellcolor{customblue!70}$+6.8^{+1.8}_{-1.9}$ & \cellcolor{customblue!59}$7.8^{+0.1}_{-0.1}$ & \cellcolor{customblue!68}$+8.4^{+2.6}_{-3.1}$ & \cellcolor{customblue!72}$-10.7^{+3.6}_{-4.1}$ & \cellcolor{customblue!52}$+45.1^{+1.8}_{-1.7}$ & \cellcolor{customblue!63}$-31.5^{+8.0}_{-9.0}$ & \cellcolor{customblue!80}$+38.1^{+2.0}_{-2.1}$ & \cellcolor{customblue!91}$-45.1^{+8.8}_{-10.2}$ \\
    DLinear~\cite{zeng2023dlinear} & \cellcolor{customblue!63}$-5.7^{+2.1}_{-2.1}$ & \cellcolor{customblue!55}$8.2^{+0.1}_{-0.1}$ & \cellcolor{customblue!69}$+10.7^{+2.8}_{-3.0}$ & \cellcolor{customblue!83}$+18.0^{+2.8}_{-3.0}$ & \cellcolor{customblue!44}$+38.5^{+1.9}_{-1.8}$ & \cellcolor{customblue!67}$-20.4^{+8.0}_{-10.0}$ & \cellcolor{customblue!21}$-9.4^{+2.4}_{-2.3}$ & \cellcolor{customblue!64}$-135.7^{+15.2}_{-17.7}$ \\
    FEDformer~\cite{zhou2022fedformer} & \cellcolor{customblue!36}$-53.7^{+3.3}_{-3.0}$ & \cellcolor{customblue!23}$11.3^{+0.1}_{-0.1}$ & \cellcolor{customblue!28}$-84.2^{+4.8}_{-4.5}$ & \cellcolor{customblue!37}$-97.1^{+7.0}_{-6.8}$ & \cellcolor{customblue!26}$+23.0^{+0.5}_{-0.5}$ & \cellcolor{customblue!73}$-1.9^{+6.7}_{-8.3}$ & \cellcolor{customblue!4}$-22.5^{+2.6}_{-2.5}$ & \cellcolor{customblue!21}$-277.5^{+29.2}_{-30.2}$ \\
    TimesNet~\cite{wu2023timesnet} & \cellcolor{customblue!30}$-66.0^{+3.5}_{-3.5}$ & \cellcolor{customblue!26}$10.9^{+0.1}_{-0.1}$ & \cellcolor{customblue!40}$-56.2^{+5.3}_{-5.0}$ & \cellcolor{customblue!24}$-131.7^{+8.8}_{-8.6}$ & \cellcolor{customblue!36}$+31.2^{+2.1}_{-2.1}$ & \cellcolor{customblue!26}$-139.2^{+16.5}_{-19.5}$ & \cellcolor{customblue!28}$-3.6^{+2.3}_{-2.2}$ & \cellcolor{customblue!33}$-238.7^{+24.1}_{-26.4}$ \\
    \midrule
    \multicolumn{9}{l}{\textbf{\emph{Long-context imputation ($\geq 7 \times 1440$ time steps)}}} \\
    \hline
    \multicolumn{9}{l}{\cellcolor[HTML]{EFEFEF}\textit{Statistical Models}} \\
    Pers.\ temp.\ mean & \cellcolor{customblue!59}$-7.7^{+2.8}_{-2.8}$ & \cellcolor{customblue!42}$9.1^{+0.1}_{-0.1}$ & \cellcolor{customblue!63}$-7.3^{+4.0}_{-4.0}$ & \cellcolor{customblue!69}$-20.4^{+5.1}_{-5.5}$ & \cellcolor{customblue!70}$+63.8^{+1.1}_{-1.2}$ & \cellcolor{customblue!75}$+16.1^{+5.8}_{-6.6}$ & \cellcolor{customblue!30}$-1.0^{+2.0}_{-2.0}$ & \cellcolor{customblue!17}$-294.3^{+28.7}_{-32.4}$ \\
    Pers.\ mode & \cellcolor{customblue!49}$-26.1^{+2.6}_{-2.4}$ & \cellcolor{customblue!28}$10.5^{+0.1}_{-0.1}$ & \cellcolor{customblue!29}$-89.8^{+4.8}_{-4.4}$ & \cellcolor{customblue!31}$-123.5^{+7.2}_{-6.8}$ & \cellcolor{customblue!30}$+29.7^{+0.6}_{-0.6}$ & \cellcolor{customblue!100}$\mathbf{+76.0}^{+1.6}_{-1.9}$ & $-25.2^{+2.8}_{-2.8}$ & $-348.4^{+38.0}_{-40.7}$ \\
    Pers.\ mean & $-114.1^{+4.4}_{-4.3}$ & $13.3^{+0.1}_{-0.1}$ & $-160.3^{+6.8}_{-6.0}$ & $-207.7^{+10.0}_{-9.7}$ & $+4.3^{+1.0}_{-1.0}$ & $-166.2^{+18.4}_{-21.2}$ & \cellcolor{customblue!18}$-11.0^{+2.1}_{-2.1}$ & \cellcolor{customblue!7}$-325.2^{+31.4}_{-35.0}$ \\
    \hline
    \multicolumn{9}{l}{\cellcolor[HTML]{EFEFEF}\textit{Neural Models}} \\
    LSM-2-Sparse (7-day) & \cellcolor{customblue!100}$\mathbf{+64.7}^{+0.4}_{-1.2}$ & \cellcolor{customblue!100}$\mathbf{3.3}^{+0.1}_{-0.1}$ & \cellcolor{customblue!100}$\mathbf{+82.4}^{+0.1}_{-1.0}$ & \cellcolor{customblue!100}$\mathbf{+64.5}^{+0.5}_{-2.0}$ & \cellcolor{customblue!100}$\mathbf{+89.0}^{+0.2}_{-0.6}$ & \cellcolor{customblue!89}$+48.9^{+3.4}_{-3.9}$ & \cellcolor{customblue!100}$\mathbf{+55.6}^{+1.3}_{-1.4}$ & \cellcolor{customblue!100}$\mathbf{-23.4}^{+8.1}_{-9.7}$ \\
    LSM-2 (7-day) & \cellcolor{customblue!90}$+46.9^{+1.0}_{-1.5}$ & \cellcolor{customblue!78}$5.5^{+0.1}_{-0.1}$ & \cellcolor{customblue!93}$+64.9^{+0.8}_{-1.6}$ & \cellcolor{customblue!91}$+39.6^{+2.1}_{-2.7}$ & \cellcolor{customblue!97}$+86.2^{+0.4}_{-0.7}$ & \cellcolor{customblue!83}$+34.0^{+4.3}_{-5.3}$ & \cellcolor{customblue!86}$+44.1^{+1.5}_{-1.7}$ & \cellcolor{customblue!75}$-106.3^{+13.9}_{-17.1}$ \\
    DLinear (7-day)~\cite{zeng2023dlinear} & \cellcolor{customblue!48}$-28.3^{+2.5}_{-2.6}$ & \cellcolor{customblue!39}$9.5^{+0.1}_{-0.1}$ & \cellcolor{customblue!58}$-20.3^{+3.4}_{-3.5}$ & \cellcolor{customblue!75}$-4.5^{+3.8}_{-4.2}$ & \cellcolor{customblue!27}$+27.3^{+1.7}_{-1.8}$ & \cellcolor{customblue!42}$-64.9^{+11.3}_{-13.6}$ & \cellcolor{customblue!9}$-18.1^{+2.5}_{-2.4}$ & \cellcolor{customblue!61}$-150.1^{+16.0}_{-18.8}$ \\
    \bottomrule[1.5pt]
    \end{tabular}%
    }
\end{table}

\subsubsection{Single-day imputation}
\label{sec:single_day}

Among single-day methods, \textsc{LSM-2} retains the strongest overall per-approach profile, especially on the structural approaches, while the constant mode-based baselines remain hardest to beat on the sleep gap. Table~\ref{tab:imputation_appendix_raw_single_day} lists the raw MAE and ROC AUC values for every approach.

\begin{table}[H]
    \renewcommand{\arraystretch}{1.05}
    \centering
    \captionsetup{width=\textwidth}
    \caption{\textbf{Single-Day Imputation Raw Metrics.} Scenario-level raw metrics on the test split. Each MAE entry is the mean per-channel MAE across applicable continuous channels; each ROC AUC entry is the macro-average across applicable binary channels. Sleep gap, workout gap, and intensity failure are scored on continuous channels only (binary channels are excluded from these semantic scenarios), so ROC AUC is not reported there. Values are point estimates over the test cohort.}
    \label{tab:imputation_appendix_raw_single_day}
    \small
    \setlength{\tabcolsep}{1.5pt}
    \resizebox{\linewidth}{!}{%
    \begin{tabular}{l ccccccccc}
    \toprule[1.5pt]
    \textbf{Method} & Random MAE$\downarrow$ & Random AUC$\uparrow$ & Temporal MAE$\downarrow$ & Temporal AUC$\uparrow$ & Signal MAE$\downarrow$ & Signal AUC$\uparrow$ & Sleep MAE$\downarrow$ & Workout MAE$\downarrow$ & Intensity MAE$\downarrow$ \\
    \hline
    \multicolumn{10}{l}{\cellcolor[HTML]{EFEFEF}\textit{Statistical Models}} \\
    Linear & \cellcolor{customblue!66}$44.3$ & \cellcolor{customblue!68}$0.829$ & \cellcolor{customblue!98}$42.7$ & \cellcolor{customblue!81}$0.867$ & $65.9$ & \cellcolor{customblue!1}$0.500$ & \cellcolor{customblue!55}$28.9$ & \cellcolor{customblue!35}$1512.3$ & \cellcolor{customblue!95}$1103.3$ \\
    Temporal mean & \cellcolor{customblue!35}$61.5$ & \cellcolor{customblue!39}$0.688$ & \cellcolor{customblue!80}$62.7$ & \cellcolor{customblue!40}$0.680$ & \cellcolor{customblue!17}$60.6$ & \cellcolor{customblue!42}$0.690$ & \cellcolor{customblue!64}$24.8$ & \cellcolor{customblue!14}$1753.9$ & \cellcolor{customblue!8}$3299.8$ \\
    LOCF \textit{(reference)} & \cellcolor{customblue!53}$51.2$ & \cellcolor{customblue!45}$0.715$ & \cellcolor{customblue!91}$50.5$ & \cellcolor{customblue!54}$0.742$ & $65.9$ & \cellcolor{customblue!1}$0.500$ & \cellcolor{customblue!54}$29.7$ & \cellcolor{customblue!33}$1537.4$ & \cellcolor{customblue!100}$\mathbf{979.0}$ \\
    Temporal mode & \cellcolor{customblue!59}$47.8$ & \cellcolor{customblue!12}$0.557$ & \cellcolor{customblue!92}$48.8$ & \cellcolor{customblue!13}$0.558$ & \cellcolor{customblue!61}$47.2$ & \cellcolor{customblue!13}$0.557$ & \cellcolor{customblue!100}$\mathbf{6.7}$ & $1922.3$ & $3492.7$ \\
    Mode & \cellcolor{customblue!59}$47.8$ & $0.500$ & \cellcolor{customblue!92}$48.8$ & $0.500$ & \cellcolor{customblue!61}$47.2$ & \cellcolor{customblue!1}$0.500$ & \cellcolor{customblue!100}$\mathbf{6.7}$ & $1922.3$ & $3492.7$ \\
    Mean & \cellcolor{customblue!25}$66.9$ & $0.500$ & \cellcolor{customblue!76}$67.1$ & $0.500$ & $65.9$ & \cellcolor{customblue!1}$0.500$ & $56.2$ & \cellcolor{customblue!11}$1796.2$ & \cellcolor{customblue!6}$3348.5$ \\
    \hline
    \multicolumn{10}{l}{\cellcolor[HTML]{EFEFEF}\textit{Neural Models}} \\
    LSM-2~\cite{xu2025lsm} & \cellcolor{customblue!100}$\mathbf{25.1}$ & \cellcolor{customblue!100}$\mathbf{0.983}$ & \cellcolor{customblue!100}$\mathbf{40.1}$ & \cellcolor{customblue!100}$\mathbf{0.951}$ & \cellcolor{customblue!100}$\mathbf{35.0}$ & \cellcolor{customblue!100}$\mathbf{0.954}$ & \cellcolor{customblue!98}$8.0$ & \cellcolor{customblue!100}$\mathbf{749.0}$ & \cellcolor{customblue!96}$1083.5$ \\
    DLinear~\cite{zeng2023dlinear} & \cellcolor{customblue!61}$47.1$ & \cellcolor{customblue!60}$0.787$ & \cellcolor{customblue!93}$48.3$ & \cellcolor{customblue!72}$0.826$ & \cellcolor{customblue!26}$58.0$ & \cellcolor{customblue!20}$0.588$ & \cellcolor{customblue!62}$25.4$ & \cellcolor{customblue!30}$1571.0$ & \cellcolor{customblue!47}$2301.5$ \\
    BRITS~\cite{cao2018brits} & $81.0$ & \cellcolor{customblue!40}$0.691$ & $150.4$ & \cellcolor{customblue!39}$0.674$ & \cellcolor{customblue!70}$44.2$ & \cellcolor{customblue!19}$0.583$ & \cellcolor{customblue!4}$54.2$ & \cellcolor{customblue!75}$1044.6$ & \cellcolor{customblue!95}$1115.4$ \\
    TimesNet~\cite{wu2023timesnet} & \cellcolor{customblue!37}$60.1$ & \cellcolor{customblue!19}$0.589$ & \cellcolor{customblue!80}$62.1$ & \cellcolor{customblue!6}$0.528$ & \cellcolor{customblue!16}$60.8$ & \cellcolor{customblue!12}$0.554$ & \cellcolor{customblue!7}$53.0$ & \cellcolor{customblue!13}$1767.7$ & \cellcolor{customblue!20}$2996.4$ \\
    FEDformer~\cite{zhou2022fedformer} & \cellcolor{customblue!45}$55.8$ & \cellcolor{customblue!7}$0.533$ & \cellcolor{customblue!88}$53.9$ & \cellcolor{customblue!11}$0.549$ & \cellcolor{customblue!38}$54.2$ & $0.498$ & \cellcolor{customblue!76}$18.7$ & \cellcolor{customblue!12}$1784.8$ & \cellcolor{customblue!39}$2514.0$ \\
    \bottomrule[1.5pt]
    \end{tabular}%
    }
\end{table}

\subsubsection{Long-context imputation}
\label{sec:extended_history}

Table~\ref{tab:imputation_appendix_raw_long_context} lists the raw metrics for every long-context method.

\begin{table}[H]
    \renewcommand{\arraystretch}{1.05}
    \centering
    \captionsetup{width=\textwidth}
    \caption{\textbf{Long-Context Imputation Raw Metrics.} Scenario-level raw metrics on the test split. Each MAE entry is the mean per-channel MAE across applicable continuous channels; each ROC AUC entry is the macro-average across applicable binary channels. Sleep gap, workout gap, and intensity failure are scored on continuous channels only (binary channels are excluded from these semantic scenarios), so ROC AUC is not reported there. Values are point estimates over the test cohort.}
    \label{tab:imputation_appendix_raw_long_context}
    \small
    \setlength{\tabcolsep}{1.5pt}
    \resizebox{\linewidth}{!}{%
    \begin{tabular}{l ccccccccc}
    \toprule[1.5pt]
    \textbf{Method} & Random MAE$\downarrow$ & Random AUC$\uparrow$ & Temporal MAE$\downarrow$ & Temporal AUC$\uparrow$ & Signal MAE$\downarrow$ & Signal AUC$\uparrow$ & Sleep MAE$\downarrow$ & Workout MAE$\downarrow$ & Intensity MAE$\downarrow$ \\
    \hline
    \multicolumn{10}{l}{\cellcolor[HTML]{EFEFEF}\textit{Statistical Models}} \\
    Pers.\ temp.\ mean & \cellcolor{customblue!13}$57.6$ & \cellcolor{customblue!66}$0.715$ & \cellcolor{customblue!20}$58.5$ & \cellcolor{customblue!69}$0.724$ & \cellcolor{customblue!20}$55.5$ & \cellcolor{customblue!72}$0.757$ & \cellcolor{customblue!78}$15.9$ & \cellcolor{customblue!19}$1694.8$ & \cellcolor{customblue!12}$3194.9$ \\
    Pers.\ mean & $62.7$ & $0.196$ & $62.9$ & $0.201$ & $60.9$ & $0.244$ & $49.6$ & \cellcolor{customblue!10}$1803.1$ & \cellcolor{customblue!6}$3343.4$ \\
    Pers.\ mode & \cellcolor{customblue!40}$47.6$ & \cellcolor{customblue!38}$0.499$ & \cellcolor{customblue!64}$48.7$ & \cellcolor{customblue!39}$0.499$ & \cellcolor{customblue!50}$47.1$ & \cellcolor{customblue!36}$0.500$ & \cellcolor{customblue!100}$\mathbf{6.6}$ & $1922.2$ & $3492.7$ \\
    \hline
    \multicolumn{10}{l}{\cellcolor[HTML]{EFEFEF}\textit{Neural Models}} \\
    LSM-2 (7-day) & \cellcolor{customblue!68}$37.0$ & \cellcolor{customblue!96}$0.954$ & \cellcolor{customblue!63}$48.8$ & \cellcolor{customblue!94}$0.917$ & \cellcolor{customblue!85}$37.4$ & \cellcolor{customblue!98}$0.946$ & \cellcolor{customblue!92}$9.8$ & \cellcolor{customblue!87}$856.4$ & \cellcolor{customblue!78}$1565.1$ \\
    LSM-2-Sparse (7-day) & \cellcolor{customblue!100}$\mathbf{25.1}$ & \cellcolor{customblue!100}$\mathbf{0.985}$ & \cellcolor{customblue!100}$\mathbf{40.7}$ & \cellcolor{customblue!100}$\mathbf{0.960}$ & \cellcolor{customblue!100}$\mathbf{33.4}$ & \cellcolor{customblue!100}$\mathbf{0.960}$ & \cellcolor{customblue!96}$8.1$ & \cellcolor{customblue!100}$\mathbf{694.0}$ & \cellcolor{customblue!100}$\mathbf{1024.2}$ \\
    DLinear (7-day)~\cite{zeng2023dlinear} & \cellcolor{customblue!42}$46.8$ & \cellcolor{customblue!63}$0.693$ & \cellcolor{customblue!62}$49.0$ & \cellcolor{customblue!74}$0.764$ & \cellcolor{customblue!24}$54.2$ & \cellcolor{customblue!41}$0.534$ & \cellcolor{customblue!54}$26.3$ & \cellcolor{customblue!25}$1612.4$ & \cellcolor{customblue!45}$2370.0$ \\
    \bottomrule[1.5pt]
    \end{tabular}%
    }
\end{table}

\subsection{Train/Evaluation Mask Alignment Ablation}
\label{app:mask_alignment}

To assess whether similarity between training and evaluation masks explains \textsc{LSM-2}'s imputation advantage, we retrained \textsc{DLinear} and \textsc{BRITS} with two artificial-masking regimes.
The default \textsc{DLinear} objective combines observed reconstruction (ORT) and masked imputation (MIT), using point-wise random masking at ratio $0.2$ (Appendix~\ref{sec:imputation_baselines}).
The default \textsc{BRITS} objective uses observed-value reconstruction and forward/backward consistency losses, without a masked-imputation term.
For \textsc{DLinear}, we retain the objective and vary the masks; for \textsc{BRITS}, both retrained variants add a masked-imputation loss on the artificially hidden values.

\paragraph{Training and evaluation.}
The first regime uses random 10-minute patches with mask ratio $0.5$.
The second draws with equal probability from the evaluation's three structural mask families (Appendix~\ref{sec:masking_scenarios}): random 10-minute patches, 30--60-minute temporal slices across channels, and signal slices over individual channels or device groups, each with target mask ratio $0.5$.
This mixture follows \textsc{LSM-2}'s artificial-masking strategy (Appendix~\ref{sec:LSM-2_architecture}), but shares only the evaluation's mask families, not its exact parameters: evaluation uses 30-minute random patches and a temporal-slice ratio of $0.25$.
The retrained variants use the same training data, splits, normalization, and each model's published architecture and optimizer settings.
\textsc{DLinear}'s ORT/MIT weights, tuned under point-wise random masking at ratio $0.2$, remain fixed.
Checkpoints are selected by validation MAE under random 10-minute patch masking at ratio $0.2$, using the same validation masks across variants.
All six baseline variants are evaluated on the same 1,777 test participants across all six masking scenarios, using the released evaluation pipeline and the \textsc{LOCF} reference.

\begin{table}[htbp]
  \centering
  \footnotesize
  \setlength{\tabcolsep}{6pt}
  \caption{\textbf{Mask-alignment ablation.} Aggregate imputation Skill Score $S$ (in \%; $0=\TN{LOCF}$ reference) on the full test split. \textsc{LSM-2} is shown for comparison. The \textsc{DLinear} reference is an independent retraining of the released recipe with a fresh seed; the released checkpoint in Table~\ref{tab:imputation_main_results} scores $-5.7$. The \textsc{BRITS} reference is the released checkpoint re-evaluated in the same pipeline, matching Table~\ref{tab:imputation_main_results}. For compute reasons, the retrained \textsc{BRITS} variants ran for four epochs; \textsc{DLinear} used up to 25 epochs with early stopping.}
  \label{tab:mask_alignment}
  \begin{tabular}{lrrr}
    \toprule
    \textbf{Method} & \makecell{Default /\\reference} & \makecell{Random\\10-min patches} & \makecell{Structural\\masks} \\
    \midrule
    \textsc{DLinear}~\citep{zeng2023dlinear} & $-9.8$ & $+35.0$ & $+31.1$ \\
    \textsc{BRITS}~\citep{cao2018brits} & $+6.8$ & $+49.0$ & $+48.8$ \\
    \textsc{LSM-2}~\citep{xu2025lsm} & $+61.4$ & -- & -- \\
    \bottomrule
  \end{tabular}
\end{table}

\paragraph{Results and limits.}
Random-patch training substantially improves aggregate Skill Score relative to the reference variants (\textsc{DLinear}: $-9.8 \to +35.0$; \textsc{BRITS}: $+6.8 \to +49.0$; Table~\ref{tab:mask_alignment}).
Replacing random patches with the structural mixture yields similar or slightly worse scores ($+31.1$ and $+48.8$, respectively); both remain below daily \textsc{LSM-2} ($+61.4$).
These results suggest that \textsc{LSM-2}'s advantage is unlikely to stem solely from a benchmark design similarity between training and evaluation masks and suggests instead that masking generally is critical for imputation performance, which current timeseries optimized approaches like \textsc{DLinear} and \textsc{BRITS} do not exploit by default but could easily be enhanced with.

\subsection{Imputation Models}
\label{app:imputation_models}

\subsubsection{\textsc{LSM-2} Reimplementation and Adaptations}
\label{sec:LSM-2_architecture}

We provide a detailed description of our reimplementation of Google's \textsc{LSM-2} masked autoencoder~\cite{xu2025lsm}. We implement three variants that differ in temporal scope and attention strategy: a single-day model (\textsc{LSM-2}) that mimics the original as faithfully as possible, a 7-day variant (\textsc{LSM-2-Weekly}), and a sparse 7-day model (\textsc{LSM-2-Sparse}). All share the same basic patch embedding and masking approaches; they differ in how attention is organized across days and in patch size.

Throughout, we denote the number of sensor channels as $C$, the per-day sequence length as $L$ (minutes), the patch size as $p$ (minutes), and the number of patches per channel per day as $T = L/p$. The total token count per day is $N_{\mathrm{day}} = C \cdot T$.

\bo{\textsc{LSM-2} (Daily).}
\begin{itemize}[leftmargin=*]
    \item \bo{Input and tokenization.}
The daily model receives a single-day input $\mathbf{X} \in \mathbb{R}^{C \times L}$ with $C{=}19$ channels and $L{=}1{,}440$ minutes. Non-overlapping patches of $p{=}10$ minutes are extracted per channel and projected to $d_{\mathrm{enc}}$-dimensional token embeddings via a shared 1D convolution:
\begin{equation}
  \mathbf{h}_i = \mathrm{Conv1D}\!\bigl(\mathbf{X}[\text{patch } i]\bigr) + \mathbf{e}^{\mathrm{pos}}_i, \quad \mathbf{h}_i \in \mathbb{R}^{d_{\mathrm{enc}}},
  \label{eq:LSM-2_patch_embed}
\end{equation}
where $\mathbf{e}^{\mathrm{pos}}_i$ is a fixed 2D sinusoidal positional embedding that encodes both the channel index and the intra-day time index of patch $i$, each axis contributing $d_{\mathrm{enc}}/2$ dimensions. This produces $N_{\mathrm{day}} = C \cdot T = 19 \times 144 = 2{,}736$ tokens per day.
    \item \bo{Encoder and decoder.}
The encoder is a 12-layer 1D (Vision) Transformer with $d_{\mathrm{enc}}{=}384$ and 6 attention heads. The decoder uses 4 Transformer layers with $d_{\mathrm{dec}}{=}256$ and 4 attention heads. Self-attention operates over all $N_{\mathrm{day}}$ tokens, giving a per-layer complexity of $O(N_{\mathrm{day}}^2) = O\!\bigl((CT)^2\bigr)$.
    \item \bo{Adaptive and Inherited Masking (AIM).}
AIM jointly handles real-world missingness and self-supervised masking by combining two binary masks per token: the \emph{inherited mask} $M^I_i$ where $M^I_i = \mathbf{1}[\text{patch } i \text{ contains NaN}]$ from actual data gaps, and an \emph{artificial mask} $M^A_i$ generated by one of three strategies applied with equal probability:
\begin{itemize}[nosep]
  \item \emph{Random}: each patch masked independently with probability $r{=}0.5$.
  \item \emph{Temporal slice}: 50\% of time indices masked across all channels.
  \item \emph{Sensor slice}: 50\% of channels masked across all time steps.
\end{itemize}
A priority score determines which tokens are physically removed from the sequence:
\begin{equation}
  s_i = 100 \cdot (M^I_i \lor M^A_i) + \epsilon_i, \quad \epsilon_i \sim \mathrm{Uniform}(0,1).
  \label{eq:aim_priority}
\end{equation}
Tokens are sorted by $s_i$ in descending order; the $(1{-}\rho) \cdot N_{\mathrm{day}}$ tokens with lowest priority (most likely observed) are retained, with $\rho{=}0.5$. Any retained token that is nonetheless masked ($M^I_i \lor M^A_i = 1$) receives an attention logit of $-\infty$, preventing it from being attended to while still receiving information for reconstruction.
    \item \bo{Reconstruction loss.}
The decoder predicts the raw patch values $\hat{\mathbf{x}}_i \in \mathbb{R}^p$ for all positions. The loss is computed only over artificially masked patches with valid ground truth:
\begin{equation}
  \mathcal{L} = \frac{1}{|\mathcal{S}|}\sum_{i \in \mathcal{S}} w_{c(i)} \cdot \ell\!\bigl(\hat{\mathbf{x}}_i,\, \mathbf{x}_i\bigr), \quad
  \ell = \begin{cases}
    \mathrm{MSE} & \text{if } c(i) \in \mathcal{C}_{\mathrm{cont}},\\[2pt]
    \mathrm{BCE} & \text{if } c(i) \in \mathcal{C}_{\mathrm{bin}},
  \end{cases}
  \label{eq:LSM-2_loss}
\end{equation}
where $\mathcal{S} = \{i : M^A_i = 1 \;\land\; M^I_i = 0\}$, $c(i)$ denotes the channel of patch $i$, $w_{c(i)}$ is an optional per-channel weight, $\mathcal{C}_{\mathrm{cont}} = \{0,\ldots,6\}$ are continuous channels (steps, distance, flights, exercise time, stand time, heart rate, active energy), and $\mathcal{C}_{\mathrm{bin}} = \{7,\ldots,18\}$ are binary channels (sleep stages, workout types). Binary channels use BCE with logit predictions. Because the Apple Watch does not record heart rate every minute, zero-valued minutes within a heart rate patch represent absent measurements rather than true observations; the MSE for heart rate patches is therefore computed only over non-zero minutes within each patch. All reported results were derived with equal channel loss weighting.
\end{itemize}

\bo{\textsc{LSM-2-Weekly} (Dense).}
\textsc{LSM-2-Weekly} applies the same encoder--decoder architecture to a 7-day concatenated input $\mathbf{X} \in \mathbb{R}^{C \times DL}$ with $D{=}7$ days. To keep the total token count tractable for dense self-attention, the patch size is increased to $p'{=}60$ minutes (6$\times$ coarser than the daily model). This yields $T' = L/p' = 24$ patches per channel per day and a total of
\[
  N_{\mathrm{week}} = C \cdot D \cdot T' = 19 \times 7 \times 24 = 3{,}192 \text{ tokens}.
\]
Full self-attention over the entire week gives a per-layer complexity of $O(N_{\mathrm{week}}^2) = O\!\bigl((CDT')^2\bigr)$. This scales quadratically with the number of days $D$, meaning that extending the context window further would become prohibitively expensive. The coarser temporal resolution ($60$ vs.\ $10$ minutes) also limits the model's ability to capture fine-grained temporal patterns.

\bo{\textsc{LSM-2-Sparse}.}
\textsc{LSM-2-Sparse} achieves multi-day context while preserving the fine-grained $p{=}10$-minute patch resolution of the daily model and allows for irregularly sampled days (thus missing days). It uses a two-stage architecture: a frozen per-day encoder (based on the regular daily model) followed by a sparse cross-day decoder.
\begin{itemize}[leftmargin=*]
    \item \bo{Per-day Encoder.}
Each of the $D{=}7$ daily slices $\mathbf{X}_d \in \mathbb{R}^{C \times L}$ is encoded independently by the daily \textsc{LSM-2} encoder, producing $N_{\mathrm{day}} = CT = 2{,}736$ latent tokens per day. The encoder weights are loaded from the pre-trained daily checkpoint and frozen during weekly training; only the decoder parameters below are learned. 
    \item \bo{Sparse Cross-Day Decoder.}
The decoder consists of 4 Transformer layers ($d_{\mathrm{dec}}{=}256$, 4 heads) that alternate between two attention patterns:
    \begin{itemize}
        \item \bo{Day-local layers (layers 0, 2).}
Standard self-attention restricted to tokens within a single day. Each day's $N_{\mathrm{day}} = CT$ tokens are processed independently, giving per-layer complexity $O\!\bigl((CT)^2\bigr)$, identical to the daily model and independent of $D$.
        \item \bo{Cross-day window layers (layers 1, 3).}
Tokens are regrouped into temporal windows that span all days to keep the memory layout contiguous and thus allow to keep the benefits of modern optimizations like FlashAttention out of the box. Per-day tokens are reshaped from their channel-major layout into windows:
\begin{equation}
  \underbrace{(B,\; D,\; C \cdot T,\; d)}_{\text{per-day tokens}}
  \;\longrightarrow\;
  \underbrace{(B \cdot W,\; D \cdot C \cdot P_w,\; d)}_{\text{per-window tokens}},
  \label{eq:cross_day_reshape}
\end{equation}
where $W = T / P_w$ is the number of windows per day and $P_w = w/p$ is the number of patches per window ($w$ is the window width in minutes). Each window thus contains tokens from all $D$ days, all $C$ channels, within one contiguous time-of-day slot of $w$ minutes. Self-attention is applied independently within each window, with per-layer complexity $O\!\bigl((D \cdot C \cdot P_w)^2\bigr)$.

In our configuration, $w{=}120$ minutes, $P_w = 120/10 = 12$, and $W = 144/12 = 12$ windows per day. Each window contains $D \cdot C \cdot P_w = 7 \times 19 \times 12 = 1{,}596$ tokens.
    \end{itemize}
    \item 
\bo{Rotary position embeddings (RoPE) for day offsets.}
In cross-day layers, all tokens from the same day share a calendar-day offset $\delta_d \in \{0, \ldots, D{-}1\}$. RoPE encodes this offset into queries $\mathbf{q}$ and keys $\mathbf{k}$ as:
\begin{equation}
  \mathbf{q}'_i = \mathbf{q}_i \odot \cos(\delta_d \cdot \boldsymbol{\theta}) \;-\; \bar{\mathbf{q}}_i \odot \sin(\delta_d \cdot \boldsymbol{\theta}),
  \label{eq:rope_day}
\end{equation}
where $\boldsymbol{\theta}_j = 10000^{-2j/d_h}$ for head dimension $d_h$, and $\bar{\mathbf{q}}_i$ denotes the rotation partner (first and second halves of the head dimension swapped). Keys are rotated analogously. This enables the model to handle non-contiguous calendar days when some days are absent from a user's history, as the relative offset between any two days is encoded implicitly through the rotation angles. While we limit our experiments to 7-days, it is straightforward to extend to more days due to how RoPE encodes relative day offset differences.
    \item \bo{Attention Complexity Comparison.}
Table~\ref{tab:LSM-2_complexity} summarizes the per-layer attention complexity of each variant. We define $T = L/p$ (patches per channel per day for the given patch size), $P_w$ (patches per window), and $T' = L/p'$ (patches per channel per day in the weekly-dense model).
\end{itemize}

\begin{table}[h!]
  \centering
  \caption{\bo{Per-layer self-attention complexity for \textsc{LSM-2} variants.} The ``Tokens / layer'' column shows the concrete sequence length for our configuration ($C{=}19$, $D{=}7$, $T{=}144$, $T'{=}24$, $P_w{=}12$).}
  \label{tab:LSM-2_complexity}
  \small
  \renewcommand{\arraystretch}{1.15}
  \begin{tabular}{lllcc}
    \toprule
    Variant & Layer type & Complexity & Tokens / layer & Multi-day \\
    \midrule
    \textsc{LSM-2} (Daily)        & Full           & $O\!\bigl((CT)^2\bigr)$     & $2{,}736$  & No  \\
    \textsc{LSM-2-Weekly}         & Full           & $O\!\bigl((CDT')^2\bigr)$   & $3{,}192$  & Yes \\
    \textsc{LSM-2-Sparse}         & Day-local      & $O\!\bigl((CT)^2\bigr)$     & $2{,}736$  & --- \\
    \textsc{LSM-2-Sparse}         & Cross-day      & $O\!\bigl((DCP_w)^2\bigr)$  & $1{,}596$  & Yes \\
    \bottomrule
  \end{tabular}
\end{table}

The sparse decoder decouples the number of days $D$ from the temporal resolution $T$. Day-local layers have complexity $O\!\bigl((CT)^2\bigr)$, identical to the daily model---multi-day context adds zero overhead. Cross-day layers scale as $O\!\bigl((DCP_w)^2\bigr)$ with $P_w \ll T$ ($P_w/T = 12/144 \approx 0.08$ in our setting), making cross-day attention strictly cheaper than day-local attention.

In contrast, the dense weekly model couples all dimensions into a single sequence with cost $O\!\bigl((CDT')^2\bigr)$, even with $6\times$ coarser patches, its per-layer cost exceeds the sparse model's. In summary, \textsc{LSM-2-Sparse} achieves 7-day context at the same maximum per-layer cost as the single-day model while preserving 10-minute temporal resolution and thus allows us to handle longer context for imputation and forecasting tasks.

\bo{Hyperparameter Selection.}
Both stages of the \textsc{LSM-2} pretraining pipeline, the daily MAE encoder and the sparse cross-day decoder, were tuned via Bayesian optimization~\citep{snoek2012practical} sweeps conducted with Weights \& Biases. Each sweep uses Hyperband early termination~\citep{li2018hyperband} with minimum iterations $= 5$ and reduction factor $\eta{=}3$, optimizing the validation reconstruction loss on the validation split. Hyperparameters not listed in the tables below are held fixed at their default values.

Table~\ref{tab:LSM-2_daily_hpo} reports the search space and selected configuration for the daily \textsc{LSM-2} encoder (15 Bayesian trials, 25 training epochs, cosine LR schedule).

\begin{table}[ht]
  \caption{Daily \textsc{LSM-2} encoder HPO search space and selected configuration (Bayesian optimization, 15 trials, 25 epochs, cosine schedule).}
  \label{tab:LSM-2_daily_hpo}
  \centering
  \small
  \renewcommand{\arraystretch}{1.15}
  \begin{tabularx}{0.98\linewidth}{>{\raggedright\arraybackslash}p{0.42\linewidth} >{\raggedright\arraybackslash}X >{\centering\arraybackslash}p{0.2\linewidth}}
    \toprule
    \textbf{Hyperparameter} & \textbf{Search space} & \textbf{Selected value} \\
    \midrule
    Learning rate (\texttt{lr}) & log-U$[10^{-6}, 10^{-3}]$ & $2.447 \times 10^{-4}$ \\
    Weight decay (\texttt{wd}) & log-U$[10^{-6}, 10^{-2}]$ & $1.500 \times 10^{-3}$ \\
    Batch size & $\{16, 32, 64, 80\}$ & $16$ \\
    \bottomrule
  \end{tabularx}
\end{table}

Table~\ref{tab:LSM-2_sparse_hpo} reports the search space and selected configuration for the \textsc{LSM-2-Sparse} weekly decoder, which is trained with the daily encoder frozen. The sweep runs 30 Bayesian trials over optimizer, masking, and architecture hyperparameters (50 training epochs, cosine LR schedule).

\begin{table}[ht]
  \caption{\textsc{LSM-2-Sparse} weekly decoder HPO search space and selected configuration (Bayesian optimization, 30 trials, 50 epochs, cosine schedule).}
  \label{tab:LSM-2_sparse_hpo}
  \centering
  \small
  \renewcommand{\arraystretch}{1.15}
  \begin{tabularx}{0.98\linewidth}{>{\raggedright\arraybackslash}p{0.42\linewidth} >{\raggedright\arraybackslash}X >{\centering\arraybackslash}p{0.2\linewidth}}
    \toprule
    \textbf{Hyperparameter} & \textbf{Search space} & \textbf{Selected value} \\
    \midrule
    Learning rate (\texttt{lr}) & log-U$[10^{-5}, 5 \times 10^{-3}]$ & $2.12 \times 10^{-4}$ \\
    Weight decay (\texttt{wd}) & log-U$[10^{-6}, 10^{-2}]$ & $3.19 \times 10^{-5}$ \\
    Decoder depth (\texttt{decoder\_depth}) & $\{2, 4\}$ & $4$ \\
    Mask ratio (\texttt{mask\_ratio}) & $\{0.3, 0.5, 0.75\}$ & $0.5$ \\
    \bottomrule
  \end{tabularx}
\end{table}

\subsubsection{Imputation Deep Learning Baselines}
\label{sec:imputation_baselines}
Besides LSM2, all deep learning imputation baselines are trained via the PyPOTS library~\citep{du2023pypots} on minute-resolution daily tensors ($C{=}19$ channels, $T{=}1{,}440$ time steps). For BRITS, gradient clipping (max norm $1.0$) is applied to prevent loss spikes from sporadic large gradients; the other PyPOTS baselines (DLinear, FEDformer, TimesNet) are trained without gradient clipping. During training, each channel is standardized by subtracting the training-set mean and dividing by the training-set standard deviation; the same scaler is applied to validation and test splits.

For each learned imputation model (BRITS, DLinear, FEDformer, TimesNet), we perform hyperparameter optimization on \mbox{\openmhc{}-XS}, using Bayesian optimization with Hyperband early stopping ($\eta{=}3$), running 15 trials per model and selecting the configuration that minimizes validation MAE. Per-model search spaces and selected configurations are reported in each model's section below.

\bo{DLinear.}
DLinear~\citep{zeng2023dlinear} is a time series baseline that decomposes the input sequence into trend and seasonal components and applies separate linear projections to each part.
DLinear is used in two roles in this benchmark: as a forecasting baseline (see Appendix~\ref{sec:deeplearningforecast}) and as an imputation baseline. The architecture is identical across both roles; the two roles differ only in training objective and hyperparameter optimization, detailed below. The selected hyperparameters reflect this: the forecasting variant favors a coarser moving-average window (301 minutes) and a larger batch size, while the imputation variant prefers a much finer window (51 minutes), a smaller batch size, and an order-of-magnitude larger learning rate, with additional ORT/MIT loss weights tuned in a separate sweep. We use the PyPOTS implementation for both~\citep{du2023pypots}.

For imputation, the Observed Reconstruction Term (ORT) penalizes reconstruction error on observed positions, while the Masked Imputation Term (MIT) penalizes error on randomly masked positions. For DLinear, a two-stage search is used as it would not converge under a single-stage joint sweep (although this only marginally helped after all): the first stage selects architecture and optimizer hyperparameters, after which the loss weights (ORT and MIT) are tuned in a second stage with the remaining parameters fixed. Table~\ref{tab:dlinear_imp_config} reports the search space and selected configuration.

\begin{table}[h!]
  \caption{DLinear selected configuration for imputation (Bayesian optimization, 15 trials, 25 epochs, patience 10). Loss weights tuned in a separate second-stage sweep.}
  \label{tab:dlinear_imp_config}
  \centering
  \small
  \renewcommand{\arraystretch}{1.15}
  \begin{tabularx}{0.98\linewidth}{>{\raggedright\arraybackslash}p{0.42\linewidth} >{\raggedright\arraybackslash}X >{\centering\arraybackslash}p{0.2\linewidth}}
    \toprule
    \textbf{Hyperparameter} & \textbf{Search space} & \textbf{Selected value} \\
    \midrule
    Learning rate (\texttt{lr}) & log-U$[10^{-5}, 10^{-1}]$ & $6.469 \times 10^{-3}$ \\
    Batch size & $\{128, 256, 512, 1024\}$ & $128$ \\
    Moving avg.\ window size & $\{25, 51, 101, 201, 301\}$ & $51$ \\
    \texttt{d\_model} & fixed & $256$ \\
    ORT weight & log-U$[0.01, 10]$ & $0.010$ \\
    MIT weight & log-U$[0.01, 10]$ & $9.546$ \\
    \bottomrule
  \end{tabularx}
\end{table}

\textit{Long-context (7-day) imputation.}
DLinear is also extended to the long-context imputation track. We train the same DLinear architecture on concatenated 7-day inputs ($C \times 10{,}080$ time steps), allowing the model to exploit cross-day temporal patterns; the training objective (PyPOTS ORT/MIT) is identical to the single-day imputation variant. All architecture hyperparameters are shared with the single-day model; only the sequence length and batch size differ.

\bo{BRITS.}
BRITS (Bidirectional Recurrent Imputation for Time Series)~\citep{cao2018brits} is an RNN-based model that learns temporal dynamics in both forward and backward directions, jointly estimating missing values and the underlying data-generating process. At each time step, the model combines its recurrent hidden state with feature-level regression to produce imputation estimates, and a consistency loss encourages agreement between the forward and backward passes. Table~\ref{tab:brits_imp_config} reports the search space and selected configuration.

\begin{table}[h!]
    \vspace{-3mm}
  \caption{BRITS selected configuration for imputation (Bayesian optimization, 15 trials, 25 epochs, patience 24).}
  \label{tab:brits_imp_config}
  \centering
  \small
  \renewcommand{\arraystretch}{1.15}
  \begin{tabularx}{0.98\linewidth}{>{\raggedright\arraybackslash}p{0.42\linewidth} >{\raggedright\arraybackslash}X >{\centering\arraybackslash}p{0.2\linewidth}}
    \toprule
    \textbf{Hyperparameter} & \textbf{Search space} & \textbf{Selected value} \\
    \midrule
    Learning rate (\texttt{lr}) & log-U$[10^{-5}, 10^{-2}]$ & $2.780 \times 10^{-3}$ \\
    Batch size & $\{128, 256, 512\}$ & $256$ \\
    RNN hidden size & $\{64, 128\}$ & $128$ \\
    \bottomrule
  \end{tabularx}
\end{table}

\bo{FEDformer.}
FEDformer (Frequency Enhanced Decomposed Transformer)~\citep{zhou2022fedformer} replaces standard self-attention with Fourier-enhanced blocks that perform attention in the frequency domain, capturing global temporal dependencies with linear complexity. Like DLinear, it uses a seasonal-trend decomposition, but applies Transformer layers to each component. We use the Fourier version with random mode selection.
For FEDformer, we tune the learning rate, batch size, and moving-average window size for the imputation task, while keeping the remaining architecture parameters fixed. Table~\ref{tab:fedformer_imp_config} reports the search space and selected configuration.

\begin{table}[H]
    \vspace{-3mm}
  \caption{FEDformer selected configuration for imputation (Bayesian optimization, 15 trials, 50 epochs, patience 10).}
  \label{tab:fedformer_imp_config}
  \centering
  \small
  \renewcommand{\arraystretch}{1.15}
  \begin{tabularx}{0.98\linewidth}{>{\raggedright\arraybackslash}p{0.42\linewidth} >{\raggedright\arraybackslash}X >{\centering\arraybackslash}p{0.2\linewidth}}
    \toprule
    \textbf{Hyperparameter} & \textbf{Search space} & \textbf{Selected value} \\
    \midrule
    Learning rate (\texttt{lr}) & log-U$[10^{-5}, 10^{-1}]$ & $1.000 \times 10^{-3}$ \\
    Batch size & $\{128, 256, 512\}$ & $512$ \\
    Moving avg.\ window size & $\{25, 51, 101, 201, 301\}$ & $25$ \\
    \texttt{n\_layers} & fixed & $2$ \\
    \texttt{d\_model} & fixed & $512$ \\
    \texttt{d\_ffn} & fixed & $512$ \\
    \texttt{n\_heads} & fixed & $8$ \\
    Fourier modes & fixed & $64$ \\
    Dropout & fixed & $0.1$ \\
    \bottomrule
  \end{tabularx}
\end{table}

\bo{TimesNet.}
TimesNet~\citep{wu2023timesnet} reshapes 1D time series into 2D tensors based on learned period lengths, then applies Inception-style 2D convolutions (``TimesBlocks'') to jointly capture intra-period and inter-period variations. The top-$k$ most salient periods are identified via FFT and processed in parallel. 
For TimesNet, we also tune the learning rate, batch size, and dropout for the imputation task, while keeping the remaining architecture parameters fixed. Table~\ref{tab:timesnet_imp_config} reports the search space and selected configuration.

\begin{table}[H]
    \vspace{-3mm}
  \caption{TimesNet selected configuration for imputation (Bayesian optimization, 15 trials, 50 epochs, patience 5).}
  \label{tab:timesnet_imp_config}
  \centering
  \small
  \renewcommand{\arraystretch}{1.15}
  \begin{tabularx}{0.98\linewidth}{>{\raggedright\arraybackslash}p{0.42\linewidth} >{\raggedright\arraybackslash}X >{\centering\arraybackslash}p{0.2\linewidth}}
    \toprule
    \textbf{Hyperparameter} & \textbf{Search space} & \textbf{Selected value} \\
    \midrule
    Learning rate (\texttt{lr}) & log-U$[10^{-5}, 10^{-1}]$ & $5.000 \times 10^{-3}$ \\
    Batch size & $\{64, 128, 256\}$ & $256$ \\
    Dropout & U$[0, 0.5]$ & $0.4$ \\
    \texttt{n\_layers} & fixed & $2$ \\
    \texttt{d\_model} & fixed & $128$ \\
    \texttt{d\_ffn} & fixed & $512$ \\
    \texttt{top\_k} & fixed & $5$ \\
    \texttt{n\_kernels} & fixed & $6$ \\
    Weight decay & fixed & $10^{-4}$ \\
    \bottomrule
  \end{tabularx}
\end{table}

\clearpage
\section{Forecasting}
\label{sec:forecasting}

In Appendix \ref{sec:forecasting_task}, we formalize the forecasting task, specifically the data preprocessing and evaluation setup.
In Appendix \ref{sec:forecastingModels}, we discuss the forecasting models we train and evaluate.
Finally, in Appendix \ref{sec:forecastingResults}, we report additional forecasting model results.

\subsection{Forecasting Task}
\label{sec:forecasting_task}

\subsubsection{Data Preprocessing}
\label{sec:forecasting_data_preprocessing}

We formulate a wearable data forecasting task by constructing hourly user-level trajectories. 
We first aggregate the raw minute-level data to hourly resolution; see Section~\ref{app:model_input_representations} for details. For each individual $i \in [1 \colon n]$, the hourly observations are ordered chronologically and concatenated across days to form a multivariate trajectory
\[
Y_{i,1:T_i} = [Y_{i,1}, Y_{i,2}, \ldots, Y_{i,T_i}], \qquad Y_{i,t} = \bigl(Y_{i,t}^{(c)}\bigr)_{c=1}^{C},
\]
where $T_i$ is the trajectory length for individual $i$ and $C=19$ is the number of channels. These $19$ channels consist of activity-related measurements from iPhone and Apple Watch, sleep indicators, and workout indicators; see Table~\ref{tab:healthkit_summary}. Each vector $Y_{i,t}$ contains the hourly values of all wearable channels at time step $t$. 
The goal of the forecasting model is to predict future vectors $Y_{i,t}$ given past observations.

\bo{Missingness Mask.}
Apple HealthKit does not provide information to differentiate whether a particular measurement is zero or missing. In order to prevent the model from predicting all zeros (e.g., due to days in which the individual did not wear their watch), we use the following crude procedure to define what we consider as ``missing'' versus a ``true zero''.

Specifically, we construct the following binary mask:
\[
M_{i,1:T_i} = [M_{i, 1}, M_{i, 2}, \ldots, M_{i, T_i}], \qquad M_{i,t} = \bigl(M_{i, t}^{(c)}\bigr)_{c=1}^{C} \in \{0,1\}^{C}
\]

where $M_{i,t}^{(c)} = 1$ indicates that channel $c$ at time step $t$ for individual $i$ is treated as ``observed'', while $M_{i,t}^{(c)}=0$ indicates that it is treated as ``missing''. We apply the zero-to-NaN transform defined in Section~\ref{app:model_input_representations} to the raw time-series data $Y_{i,t}^{(c)}$. We then set the corresponding entries of $M_{i,t}^{(c)}$ to 0 wherever the transformed values are NaN, and to 1 otherwise.

\subsubsection{Evaluation Procedure}

We consider forecast horizons $H = 24$, corresponding to predicting the next 24 hours. Using the individual-level trajectory data $(Y_{i,1:T_i}, M_{i, 1:T_i})$, we form a sequence of forecasting examples for each $t \in \MC{T}_i$ (we discuss how we choose $\MC{T}_i$ below):
\begin{align}
    (Y_{i,1:t}, M_{i, 1:t}) \rightarrow Y_{i, t+1:t+H}. %
    \label{eqn:forecastingEx}
\end{align}
That is, the model takes as input the historical target sequence available up to the forecasting horizon, and the historical validity mask, and predicts the future target sequence over the next $H$ steps.

To form our training, validation, and test sets, we follow the same official  splits used in the health outcome prediction and imputation tasks. 

\bo{Specifying Evaluation Timepoints $\MC{T}_i$.}
We only evaluate on forecasting samples, akin to \eqref{eqn:forecastingEx}, starting at values of $t \in \MC{T}_i$. Specifically,
\begin{itemize}[leftmargin=*]
    \item $\MC{T}_i$ only includes timepoints $t$ that correspond to midnight
    according to the individual's local time. For each participant, forecast
    targets therefore cover distinct, non-overlapping 24-hour periods.

    \item We require that all days used to form $Y_{i,t+1:t+H}$ are valid days. In addition, we require at least three of the most recent seven days in the historical sequence $Y_{i,1:t}$ to be valid. A day is considered valid if its minute-level data pass wear-time filter and low-variance filter described in Sections~\ref{app:model_input_representations}.
    
    \item For evaluation, we restrict $|\MC{T}_i| \leq 100$ by subsampling at most 100 timesteps $t$ that satisfy all the above criteria.
\end{itemize}

\begin{figure}
    \centering
    \vspace{-5mm}
    \includegraphics[width=0.7\linewidth]{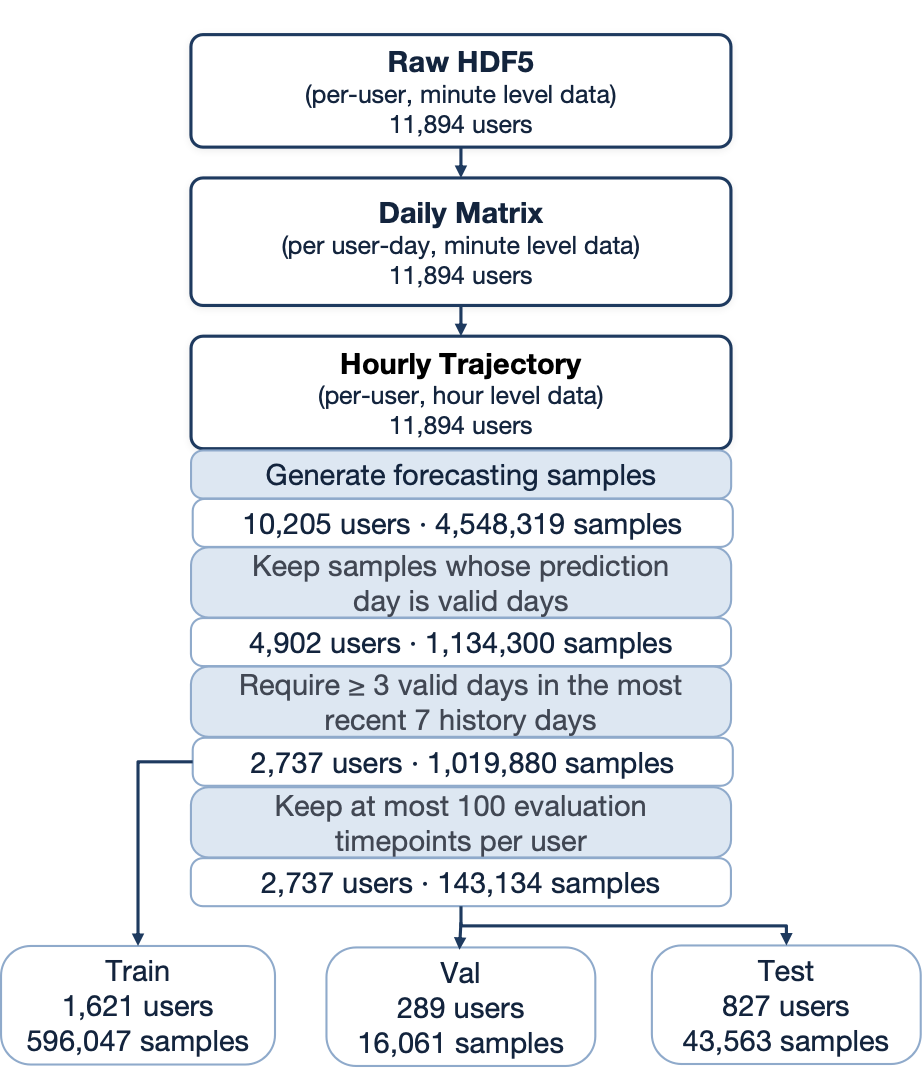}
    \caption{Sample generation pipeline for the forecasting task. The initial participant filtering process is shown in Figure~\ref{fig:consort}, and the construction details of the Daily Matrix are described in Section~\ref{app:daily_matrix_construction}.}
    \label{fig:sample-generation-workflow}
\end{figure}

The complete sample generation workflow, as well as the number of users and samples retained at each stage, is shown in Figure~\ref{fig:sample-generation-workflow}. After applying these filters, the test cohort comprises 827 users contributing 43{,}563 forecasting sub-trajectories, with a mean of 52.7 (median 48) evaluation days per user; 352 of the 827 users hit the per-user cap of 100 days.

\bo{Forecasting Model Predictions.}
The model outputs
\[
\hat{Y}_{i,t+1:t+H} = [\hat{Y}_{i,t+1}, \ldots, \hat{Y}_{i,t+H}],
\qquad
\hat{Y}_{i,t+h} = (\hat{Y}_{i,t+h}^{(c)})_{c=1}^{C},
\]
where $\hat{Y}_{i,t+h}^{(c)}$ denotes the predicted value of channel $c$ at forecast step $h$.


\bo{Handling NaN Point Forecasts.}
If a model returns \texttt{NaN} for any point forecast, we substitute the corresponding Seasonal Naive prediction at that channel and forecast step before scoring.

\subsubsection{Evaluation Metrics}

Among the 19 channels shown in Table~\ref{tab:healthkit_summary}, there are two types of data: continuous and binary.

\begin{itemize}[leftmargin=*]
    \item For continuous channels, we use MAE and QL to evaluate the quality of point forecasts and probabilistic forecasts, respectively. For models that do not natively support probabilistic forecasting, we report MAE only.
    \item For binary channels, we use AUROC to evaluate the prediction results.
\end{itemize}

\bo{Mean Absolute Error (MAE).}For continuous channels, the primitive evaluation quantity is the masked absolute error
\[
AE_{i,t,h}^{(c)}
=
M_{i,t+h}^{(c)}
\left|Y_{i,t+h}^{(c)}-\hat{Y}_{i,t+h}^{(c)}\right|,
\]
where $i$ indexes participants, $t \in \mathcal{T}_i$ indexes valid forecasting
windows, $h \in \{0,\ldots,H-1\}$ indexes the forecast horizon, and
$M_{i,t+h}^{(c)}$ indicates whether channel $c$ is observed at that target time.

For participant $i$ and continuous channel $c$, we pool the absolute error over all observed forecast hours across that participant's windows---equivalently, weighting each window by its number of observed forecast hours:
\[
MAE_{i}^{(c)}
=
\frac{
    \sum_{t \in \mathcal{T}_{i}}
    \sum_{h=0}^{H-1}
    AE_{i,t,h}^{(c)}
}{
    \sum_{t \in \mathcal{T}_{i}}
    \sum_{h=0}^{H-1}
    M_{i,t+h}^{(c)}
}.
\]





\bo{Area Under the Receiver Operating Characteristic Curve (AUROC).}

For binary channels, we evaluate predicted scores using AUROC. For each participant and binary channel, we pool the valid binary target hours across all of that participant's forecast windows and compute a single AUROC over the pooled (target, score) pairs. A participant--channel whose pooled targets contain only a single class does not define an AUROC value and is therefore excluded.

For model-level comparison, each participant thus contributes one AUROC value per binary channel. The main results group binary channels into sleep and workout categories by averaging the available participant-level channel AUROC values within each group (larger AUROC is better). For Skill Scores, AUROC is converted to an error scale as $e = 1 - \text{AUROC}$, floored at $\varepsilon = 0.005$, and then compared with the Seasonal Naive baseline using the same paired participant-level procedure as for the continuous metrics. The category-balanced aggregation of these per-channel skills and ranks into the headline $S$ and $R$ is detailed in Section~\ref{sec:forecasting_aggregation}.

\subsubsection{Aggregation and Scoring}
\label{sec:forecasting_aggregation}

The headline forecasting columns, the aggregate Skill Score $S$, the Average Rank $R$, and the
per-category Skill Scores, are produced from the per-participant errors above through the unified
Skill Score calculation of Appendix~\ref{sec:scoring_methodology} (clip bounds $\ell = 0.01$,
$u = 100$; baseline $b = \textsc{Seasonal Naive}$), specialized to the forecasting track. Because the
$19$ channels split very unevenly across sensor types, the aggregation is \emph{category-balanced}:
the channels partition into four reporting categories that act as equal-weight buckets---\emph{Activity}
(channels $0$--$4$: phone and watch step counts and distances, and flights climbed; continuous, scored
by MAE), \emph{Physiology} (channels $5$--$6$: heart rate and active energy; continuous, MAE),
\emph{Sleep} (channels $7$--$8$: the asleep and in-bed indicators; binary, AUROC), and \emph{Workout}
(channels $9$--$18$: the ten workout-type indicators; binary, AUROC).

\bo{Per-channel paired ratio.} For each channel $c$ we form the ratio against the baseline
\emph{within} each participant and geometrically average it over participants $\MC{P}^{(c)}$:
\begin{equation*}
R_{m}^{(c)}
= \exp\,\Bigl(\tfrac{1}{|\mathcal{P}^{(c)}|}\sum_{p\in\mathcal{P}^{(c)}}
  \log \mathrm{clip}\bigl(\ E_{m,p}^{(c)} / E_{b,p}^{(c)},\,\ell,u\bigr)\Bigr),
\label{eq:fc_paired_ratio}
\end{equation*}
where $E_{m,p}^{(c)}$ is participant $p$'s error for model $m$ on channel $c$---the pooled MAE
$MAE_i^{(c)}$ for a continuous channel and $\max(1-\mathrm{AUROC}_{m,p}^{(c)},\,\varepsilon)$ with
$\varepsilon = 0.005$ for a binary channel---and $\mathcal{P}^{(c)}$ is the set of participants for whom
both errors are defined and finite with $E_{b,p}^{(c)} > 0$.

\bo{Category-balanced Skill Score.} Let a scope be a set of categories $\mathcal{K}$ and let
$\mathcal{C}_k$ denote the channels of category $k$. The Skill Score averages the clipped log-ratios
over the channels within each category, then over the categories with equal weight:
\begin{equation}
S_{m,\mathcal{K}}
= 1 - \exp\,\Bigl(
  \frac{1}{|\mathcal{K}|}\sum_{k\in\mathcal{K}}
  \frac{1}{|\mathcal{C}_k|}\sum_{c\in\mathcal{C}_k}
  \log \mathrm{clip}(R_{m}^{(c)},\,\ell,u)\Bigr).
\label{eq:fc_skill_scope}
\end{equation}
The reported per-category columns take $\mathcal{K} = \{k\}$ (a single inner mean over that category's
channels), while the aggregate $S$ takes $\mathcal{K}$ to be all four categories. Each category
therefore has equal voice, so the ten Workout channels cannot dominate the two Sleep or the seven
continuous channels.

\bo{Category-balanced Average Rank.} For each channel $c$, models are ranked \emph{within} each
participant by their per-participant error---$E_i^{(c)}$ for a continuous channel and the
\emph{unfloored} $1-\mathrm{AUROC}$ for a binary channel (ascending, ties averaged)---and these
per-participant ranks are averaged into a task rank $\bar{\rho}_{m,c}$. The scope rank then follows
the same equal-weight category hierarchy, with arithmetic means at every level (rank is already
scale-free, so there is no $\log/\exp$):
\begin{equation}
\rho_{m,\mathcal{K}}
= \frac{1}{|\mathcal{K}|}\sum_{k\in\mathcal{K}}
  \frac{1}{|\mathcal{C}_k|}\sum_{c\in\mathcal{C}_k}\bar{\rho}_{m}^{(c)}.
\label{eq:fc_avg_rank}
\end{equation}
The reported Average Rank $R$ uses the overall (four-category) scope, as $S$ does.

Unlike the imputation track (Appendix~\ref{sec:imputation_aggregation}), the forecasting aggregation
has no scenario level and keeps per-channel binary tasks---a binary category is balanced through the
equal-weight category means of \eqref{eq:fc_skill_scope} and \eqref{eq:fc_avg_rank} rather than by
collapsing its channels into a single per-participant task.

\bo{Bootstrap confidence intervals.}
Although forecast targets do not overlap, forecast days from the same
participant may remain correlated. We therefore resample test participants
with replacement, keeping each participant's evaluated days together,
using the participant-level bootstrap described in
Appendix~\ref{sec:scoring_methodology}. We use the same participant draws
across models, recompute the aggregate scores and ranks for each of
$1{,}000$ replicates, and report $95\%$ percentile confidence intervals.

\subsection{Forecasting Models}
\label{sec:forecastingModels}

Since our data are multivariate time series, we aimed to include models that natively support multivariate forecasting, in order to better exploit cross-channel information. This primarily drove the decision around which time series foundation models to evaluate. For models that do not support this setting (e.g., statistical models), we perform univariate forecasting for each channel separately and then aggregate the results into multivariate forecasts.

\subsubsection{Statistical models}
We fit statistical baselines separately for each participant and channel using only that participant's observed history at test time. This per-series fitting follows standard practice for local forecasting models in benchmarks such as GIFT-Eval~\citep{aksu2024gift} and fev-bench~\citep{shchur2025fevbench}.

Statistical models' inference is performed on a CPU server equipped with 8 cores. Because historical data in some forecasting examples in our benchmark contain particularly long time series, we impose a maximum input length on each forecasting examples due to time constraints, truncating each historical data to 336 time steps (14 days).

We consider three statistical baselines:
\begin{itemize}[leftmargin=*]
    \item \textbf{Seasonal Naive} \citep{hyndman2018forecasting}: a simple forecasting baseline that repeats the value observed at the corresponding seasonal lag in the past. In our hourly setting, it predicts each future time step using the value from the same hour of the previous day.
    
    \item \textbf{Auto\_ARIMA} (implemented with SkTime library \citep{LoningSktime2019}): an automatically configured AutoRegressive Integrated Moving Average model that selects the ARIMA orders from data. It captures linear temporal dependencies, trends, and autocorrelation in the historical series.
    
    \item \textbf{Auto\_ETS} (implemented with SkTime library \citep{LoningSktime2019}): an automatically configured exponential smoothing model with error, trend, and seasonality components. It is designed to model level, trend, and seasonal patterns in time series.
\end{itemize}

\subsubsection{Deep Learning Models (Trained from Scratch): DLinear, MixLinear, SegRNN}
\label{sec:deeplearningforecast}
We evaluate three deep learning forecasting models trained from scratch on the benchmark’s training split: DLinear~\citep{zeng2023dlinear}, MixLinear~\citep{ma2024mixlinear}, and SegRNN~\citep{lin2025segrnn}. 
To train these models, we run targeted Bayesian optimization~\citep{snoek2012practical} using Weights \& Biases sweeps. To limit compute cost, the sweep is capped at 15 trials and uses Hyperband early termination \citep{li2018hyperband} with minimum iterations $=5$ and reduction factor $\eta=3$. Table~\ref{tab:hpo-search-space} lists the range of parameters we search over for each model.

During training, the input trajectories are standardized using a StandardScaler fitted only on the training split, where each channel is transformed by subtracting the training-set mean and dividing by the training-set standard deviation. The same scaler is then applied to the validation and test splits to avoid data leakage. For all deep learning models, we use implementations readily available in the PyPOTS library \citep{du2023pypots}.

 \begin{table}[ht]
    \centering
    \caption{\bo{Hyperparameter optimization search space for model training.}}
    \label{tab:hpo-search-space}
    \small
    \renewcommand{\arraystretch}{1.15}
    \begin{tabular}{p{2.5cm} p{4.0cm} p{3.5cm} p{5.0cm}}
    \toprule
    \textbf{Model} & \multicolumn{3}{c}{\textbf{Hyperparameter}} \\
    \midrule

    \textbf{DLinear}
    & Batch size
    & Learning rate
    & Window size of moving average \\
    \textbf{Search Range}
    & $[128, 256, 512, 1024]$
    & $[10^{-5} : 10^{-1}]$
    & $[25, 51, 101, 201, 301]$ \\

    \midrule

    \textbf{MixLinear}
    & Batch size
    & Learning rate
    & Segment length \\
    \textbf{Search Range}
    & $[256, 512, 1024]$
    & $[10^{-5} : 10^{-1}]$
    & $[2, 4, 6, 8]$ \\

    \midrule

    \textbf{SegRNN}
    & Batch size
    & Learning rate
    & Segment length \\
    \textbf{Search Range}
    & $[64, 128, 256, 512]$
    & $[10^{-5} : 10^{-1}]$
    & $[6, 12, 24]$ \\

    \bottomrule
    \end{tabular}
\end{table}

\bo{DLinear.}
DLinear~\citep{zeng2023dlinear} is a time series baseline that decomposes the input sequence into trend and seasonal components and applies separate linear projections to each part.
DLinear was also used for imputation (identical architecture, different loss function/hyperparameter optimization, and time-scale). The forecasting variant favors a coarser moving-average window (301 hours) and a larger batch size. As with the imputation variant, we use the PyPOTS implementation ~\citep{du2023pypots}.

Trained with standard next-horizon MSE on the benchmark training split (StandardScaler fit on train, applied to validation and test); 15 Bayesian trials with Hyperband early termination ($\eta{=}3$), 30 epochs, patience 10. The full forecasting HPO search space across deep learning baselines is summarized in Table~\ref{tab:hpo-search-space}; the selected configuration is in Table~\ref{tab:dlinear_config}.

\begin{table}[h!]
  \caption{DLinear selected configuration (Bayesian optimization, 15 trials, 30 epochs, patience 10).}
  \label{tab:dlinear_config}
  \centering
  \small
  \renewcommand{\arraystretch}{1.15}
  \begin{tabularx}{0.98\linewidth}{>{\raggedright\arraybackslash}p{0.42\linewidth} >{\raggedright\arraybackslash}X >{\centering\arraybackslash}p{0.2\linewidth}}
    \toprule
    \textbf{Hyperparameter} & \textbf{Search space} & \textbf{Selected value} \\
    \midrule
    Learning rate (\texttt{lr}) & log-U$[10^{-5}, 10^{-1}]$ & $3.686 \times 10^{-4}$ \\
    Batch size & $\{128, 256, 512, 1024\}$ & $512$ \\
    Moving avg.\ window size & $\{25, 51, 101, 201, 301\}$ & $301$ \\
    \bottomrule
  \end{tabularx}
\end{table}

\bo{MixLinear.}
MixLinear~\citep{ma2024mixlinear} is an extremely lightweight forecasting model that combines segment-based linear modeling in the time domain with adaptive low-rank filtering in the frequency domain.

\begin{table}[h!]
  \caption{\bo{MixLinear selected configuration (Bayesian optimization, 15 trials, 30 epochs, patience 10).}}
  \label{tab:mixlinear_config}
  \centering
  \small
  \renewcommand{\arraystretch}{1.15}
  \begin{tabularx}{0.98\linewidth}{>{\raggedright\arraybackslash}p{0.42\linewidth} >{\raggedright\arraybackslash}X >{\centering\arraybackslash}p{0.2\linewidth}}
    \toprule
    \textbf{Hyperparameter} & \textbf{Search space} & \textbf{Selected value} \\
    \midrule
    Learning rate (\texttt{lr}) & log-U$[10^{-5}, 10^{-1}]$ & $2.214 \times 10^{-4}$ \\
    Batch size & $\{256, 512, 1024\}$ & $1024$ \\
    Segment length (\texttt{period\_len}) & $\{2, 4, 6, 8\}$ & $6$ \\
    \bottomrule
  \end{tabularx}
\end{table}

\bo{SegRNN.}
SegRNN~\citep{lin2025segrnn} is an RNN-based model designed for long-horizon forecasting through segment-wise iterations and parallel multi-step forecasting.

\begin{table}[h!]
  \caption{SegRNN selected configuration (Bayesian optimization, 15 trials, 30 epochs, patience 10).}
  \label{tab:segrnn_config}
  \centering
  \small
  \renewcommand{\arraystretch}{1.15}
  \begin{tabularx}{0.98\linewidth}{>{\raggedright\arraybackslash}p{0.42\linewidth} >{\raggedright\arraybackslash}X >{\centering\arraybackslash}p{0.2\linewidth}}
    \toprule
    \textbf{Hyperparameter} & \textbf{Search space} & \textbf{Selected value} \\
    \midrule
    Learning rate (\texttt{lr}) & log-U$[10^{-5}, 10^{-1}]$ & $2.025 \times 10^{-3}$ \\
    Batch size & $\{64, 128, 256, 512\}$ & $256$ \\
    Segment length (\texttt{seg\_len}) & $\{6, 12, 24\}$ & $12$ \\
    \bottomrule
  \end{tabularx}
\end{table}

\subsubsection{ToTo}
\label{sec:toto_architecture}
We additionally evaluate \textsc{Toto}~\citep{cohen2025time} both in zero-shot and fine-tuned settings. The fine-tuned model is also used as a feature extractor for health outcome prediction.

\bo{Architecture.}
\textsc{Toto}~\citep{cohen2025time} is a Transformer decoder-only foundation model for multivariate time series forecasting. It models temporal and cross-variate dependencies through proportional factorized attention and uses causal patch-wise normalization with a Student-$t$ mixture output head. We use the publicly available \texttt{Datadog/Toto-Open-Base-1.0} checkpoint from Hugging Face (PyPI version 0.1.4), which was trained for a maximum context length of 4{,}096 tokens.

\bo{Fine-tuning on MHC data.}
\textsc{Toto} is fine-tuned on the MHC forecasting training split following the official implementation.\footnote{\url{https://github.com/datadog/toto}} The fine-tuned checkpoint corresponds to epoch~24 (step~116{,}225, validation loss~$-1.3597$). The forecasting task uses a 336-step (14-day) historical context to predict the next 24 hours across all 19 wearable channels.

\begin{table}[h!]
  \caption{\bo{\textsc{Toto} fine-tuning configuration.}}
  \label{tab:toto_finetuning}
  \centering
  \small
  \renewcommand{\arraystretch}{1.15}
  \begin{tabularx}{1.0\linewidth}{>{\raggedright\arraybackslash}p{0.25\linewidth} >{\raggedright\arraybackslash}X}
    \toprule
    \textbf{Parameter} & \textbf{Value} \\
    \midrule
    Base model & Datadog/Toto-Open-Base-1.0 \\
    Max context length & 2{,}048 \\
    Number of channels ($C$) & 19 \\
    Forecast horizon & 24 hours \\
    Best checkpoint & epoch 24, val\_loss $= -1.3597$ \\
    \midrule
    Fine-tuning method & LoRA ($r{=}32$, $\alpha{=}32$, dropout $= 0.097$) \\
    Optimizer & AdamW ($\beta_1{=}0.9$, $\beta_2{=}0.999$, weight decay $= 0.01$) \\
    Learning rate & $8.56 \times 10^{-4}$ peak; warmup/stable/decay $= 200/1000/1000$ steps; min $= 10^{-5}$ \\
    Batch size & 128 \\
    Epochs & 25 \\
    \bottomrule
  \end{tabularx}
\end{table}

\subsubsection{Chronos-2}
\label{sec:chronos2_architecture}

We additionally evaluate \textsc{Chronos-2}~\citep{ansari2025chronos} both in zero-shot and fine-tuned settings. The fine-tuned model is also used as a feature extractor for health outcome prediction.

\bo{Architecture.}
Chronos-2~\citep{ansari2025chronos} is a Transformer encoder-only time-series foundation model that introduces a group attention mechanism to enable information sharing across related series. It natively handles univariate, multivariate, and covariate-informed forecasting in a unified framework. We use the publicly available \texttt{amazon/chronos-2} checkpoint from Hugging Face (PyPI version 2.2.2), which supports a maximum context length of 8{,}192 tokens.

\bo{Fine-tuning on MHC data.}
Chronos-2 is fine-tuned on the MHC forecasting training split using LoRA (Low-Rank Adaptation~\cite{hu2022lora}), following the official implementation.\footnote{\url{https://github.com/amazon-science/chronos-forecasting}} The fine-tuning uses a 168-step (7-day) minimum past context with 336-step (14-day) input sequences to predict the next 24 hours across 19 channels.

\begin{table}[h!]
  \caption{\bo{Chronos-2 LoRA fine-tuning configuration.}}
  \label{tab:chronos2_finetuning}
  \centering
  \small
  \renewcommand{\arraystretch}{1.15}
  \begin{tabularx}{0.80\linewidth}{>{\raggedright\arraybackslash}p{0.45\linewidth} >{\raggedright\arraybackslash}X}
    \toprule
    \textbf{Parameter} & \textbf{Value} \\
    \midrule
    Base model & amazon/chronos-2 \\
    Max context length & 8{,}192 \\
    Number of channels ($C$) & 19 \\
    Input steps (\texttt{n\_steps}) & 336 \\
    Prediction steps (\texttt{n\_pred\_steps}) & 24 \\
    Context length (minimum past) & 168 \\
    \midrule
    Fine-tuning method & LoRA \\
    LoRA rank ($r$) & 16 \\
    LoRA alpha ($\alpha$) & 16 \\
    LoRA dropout & 0.0 \\
    Optimizer learning rate & $4.65 \times 10^{-5}$ \\
    Batch size & 760 \\
    Epochs & 25 \\
    \bottomrule
  \end{tabularx}
\end{table}

\bo{Hyperparameter selection.}
The LoRA fine-tuning hyperparameters are selected via Bayesian optimization~\citep{snoek2012practical} using Weights \& Biases sweeps, capped at 15 trials with Hyperband early termination (minimum iterations~$=3$, reduction factor $\eta{=}3$). The optimization metric is \texttt{eval/loss} (minimize). Table~\ref{tab:chronos2_hpo} reports the search space and selected values.

\begin{table}[h!]
  \caption{\bo{Chronos-2 LoRA HPO search space and selected values.}}
  \label{tab:chronos2_hpo}
  \centering
  \small
  \renewcommand{\arraystretch}{1.15}
  \begin{tabularx}{0.98\linewidth}{>{\raggedright\arraybackslash}p{0.35\linewidth} >{\raggedright\arraybackslash}X >{\centering\arraybackslash}p{0.18\linewidth}}
    \toprule
    \textbf{Hyperparameter} & \textbf{Search space} & \textbf{Selected} \\
    \midrule
    Learning rate & Log-uniform $[10^{-7}, 10^{-4}]$ & $4.65 \times 10^{-5}$ \\
    LoRA rank ($r$) & $\{4, 8, 16\}$ & 16 \\
    LoRA alpha ($\alpha$) & $\{8, 16, 32\}$ & 16 \\
    LoRA dropout & Uniform $[0.0, 0.2]$ & 0.0 \\
    \bottomrule
  \end{tabularx}
\end{table}

\subsection{Additional Results}
\label{sec:forecastingResults}

This section presents detailed channel-level Skill Scores for each model, reported separately for continuous in Table~\ref{tab:continuous_channel_skill_scores} and binary channels in Table~\ref{tab:binary_channel_skill_scores}. The detailed channel information can be found in Table~\ref{tab:healthkit_summary}.

Overall, these findings are consistent with the results reported in Table~\ref{tab:forecasting_grouped_model_summary}. Notably, Toto performs very poorly on the HeartRate channel, with a Skill Score of $-8.7$, but its performance improves substantially after fine-tuning. In addition, although DLinear achieves a Workout-category Skill Score of $5.5$, this improvement appears to be driven primarily by strong performance on only a few specific workout channels.

\begin{figure}[h!]
    \centering
    \caption{\bo{Example forecasting results across three channels: StepCount(iPhone), HeartRate, and ActiveEnergyBurned.} Each panel shows the 48-hour historical context, followed by the 24-hour forecasting horizon with both ground-truth observations and model predictions. FT denotes fine-tuned.}
    \includegraphics[width=1\linewidth]{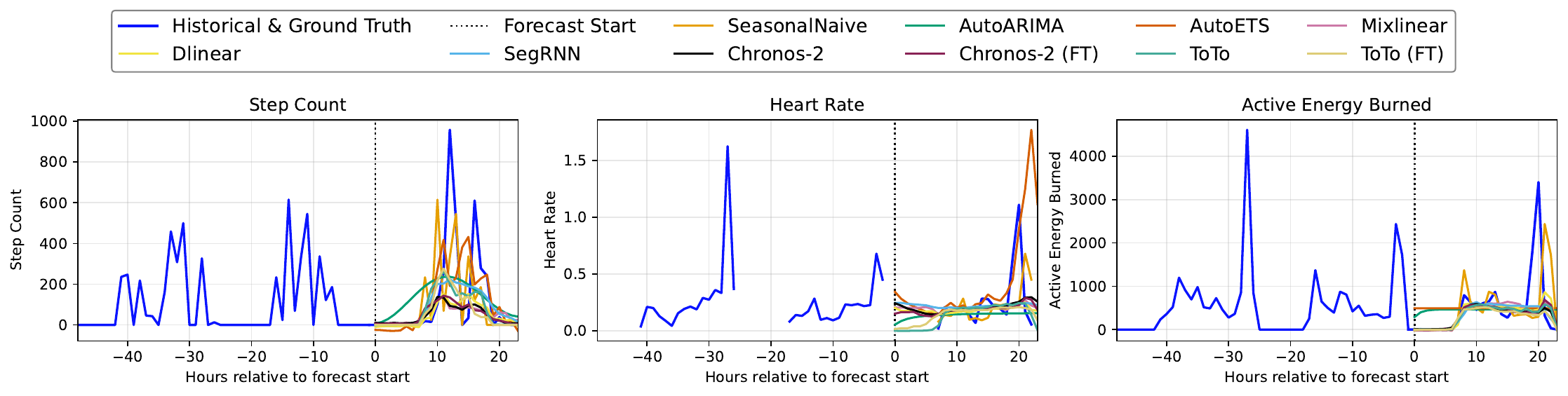}
    \label{fig:forecasting_example}
\end{figure}

In addition, Figure~\ref{fig:forecasting_example} provides an example of model forecasts on our dataset.

\clearpage
\onecolumn

\begingroup

\captionsetup{
    width=0.98\textwidth,
    font=normalsize
}

\small
\renewcommand{\arraystretch}{1.16}
\setlength{\tabcolsep}{2.2pt}


\begin{xltabular}{\textwidth}{
    >{\raggedright\arraybackslash}p{2.5cm}
    *{7}{>{\centering\arraybackslash}X}
}

\caption{
\bo{Channel-level forecasting performance on continuous channels.}
The table reports the Skill Score (in \%) of each model relative to the Seasonal Naive baseline for seven continuous channels.
Subscripts and superscripts indicate the $95\%$ participant-level bootstrap percentile confidence interval based on $1{,}000$ resamples. FT denotes fine-tuned.
}
\label{tab:continuous_channel_skill_scores}
\\

\toprule[1.4pt]

\textbf{Method}
& \shortstack{\textbf{StepCount}\\\textbf{(iPhone)}}
& \shortstack{\textbf{Distance}\\\textbf{(iPhone)}}
& \shortstack{\textbf{Flights}\\\textbf{Climbed}}
& \shortstack{\textbf{StepCount}\\\textbf{(Watch)}}
& \shortstack{\textbf{Distance}\\\textbf{(Watch)}}
& \textbf{HeartRate}
& \shortstack{\textbf{Energy}\\\textbf{Burned}}
\\

\midrule
\endfirsthead

\multicolumn{8}{l}{
    \small\itshape
    Table~\thetable\ continued from the previous page
}
\\

\toprule[1.4pt]

\textbf{Method}
& \shortstack{\textbf{StepCount}\\\textbf{(iPhone)}}
& \shortstack{\textbf{Distance}\\\textbf{(iPhone)}}
& \shortstack{\textbf{Flights}\\\textbf{Climbed}}
& \shortstack{\textbf{StepCount}\\\textbf{(Watch)}}
& \shortstack{\textbf{Distance}\\\textbf{(Watch)}}
& \textbf{HeartRate}
& \shortstack{\textbf{Energy}\\\textbf{Burned}}
\\

\midrule
\endhead

\midrule
\multicolumn{8}{r}{
    \small\itshape Continued on the next page
}
\\
\endfoot

\bottomrule[1.4pt]
\endlastfoot

\rowcolor[HTML]{EFEFEF}
\multicolumn{8}{l}{\textit{Statistical Models}} \\

Seasonal Naive
& \est{+0.0}{0.0}{0.0}
& \est{+0.0}{0.0}{0.0}
& \est{+0.0}{0.0}{0.0}
& \est{+0.0}{0.0}{0.0}
& \est{+0.0}{0.0}{0.0}
& \est{+0.0}{0.0}{0.0}
& \est{+0.0}{0.0}{0.0} \\

AutoARIMA
& \est{-1.5}{1.6}{2.0}
& \est{-1.5}{1.5}{1.7}
& \est{-5.7}{1.7}{1.7}
& \est{-0.2}{1.2}{1.4}
& \est{-0.3}{1.3}{1.4}
& \est{-11.2}{2.9}{2.7}
& \est{-6.9}{1.4}{1.5} \\

AutoETS
& \cellcolor{customblue!5}\est{+1.1}{1.4}{1.8}
& \cellcolor{customblue!5}\est{+1.8}{1.5}{1.7}
& \est{-4.0}{1.5}{1.6}
& \cellcolor{customblue!5}\est{+1.8}{1.8}{1.5}
& \cellcolor{customblue!5}\est{+2.3}{1.1}{1.2}
& \est{-49.0}{3.9}{4.2}
& \est{-8.0}{2.7}{3.5} \\

\specialrule{\lightrulewidth}{0pt}{0pt}
\rowcolor[HTML]{EFEFEF}
\multicolumn{8}{l}{\textit{Neural Models}} \\

MixLinear
& \cellcolor{customblue!45}\est{+19.3}{1.4}{1.5}
& \cellcolor{customblue!50}\est{+20.9}{1.3}{1.4}
& \cellcolor{customblue!85}\est{+36.0}{2.2}{2.1}
& \cellcolor{customblue!45}\est{+19.6}{1.3}{1.5}
& \cellcolor{customblue!50}\est{+19.8}{1.3}{1.6}
& \cellcolor{customblue!25}\est{+10.9}{2.8}{3.3}
& \cellcolor{customblue!40}\est{+15.7}{1.7}{1.9} \\

DLinear
& \cellcolor{customblue!55}\est{+22.4}{1.2}{1.4}
& \cellcolor{customblue!55}\est{+22.9}{1.2}{1.4}
& \cellcolor{customblue!80}\est{+33.9}{1.9}{1.9}
& \cellcolor{customblue!55}\est{+22.7}{1.2}{1.4}
& \cellcolor{customblue!55}\est{+22.7}{1.2}{1.4}
& \cellcolor{customblue!30}\est{+12.3}{2.9}{3.2}
& \cellcolor{customblue!50}\est{+20.5}{1.5}{1.8} \\

SegRNN
& \cellcolor{customblue!60}\est{+25.4}{1.2}{1.2}
& \cellcolor{customblue!60}\est{+24.8}{1.4}{1.5}
& \cellcolor{customblue!65}\est{+27.6}{2.4}{2.5}
& \cellcolor{customblue!60}\est{+24.6}{1.2}{1.4}
& \cellcolor{customblue!60}\est{+24.7}{1.2}{1.4}
& \cellcolor{customblue!40}\est{+17.4}{2.4}{2.6}
& \cellcolor{customblue!60}\est{+24.1}{1.2}{1.5} \\

\specialrule{\lightrulewidth}{0pt}{0pt}
\rowcolor[HTML]{EFEFEF}
\multicolumn{8}{l}{\textit{Foundation Models}} \\

Chronos-2
& \cellcolor{customblue!70}\est{+29.3}{1.2}{1.3}
& \cellcolor{customblue!70}\est{+29.6}{1.3}{1.4}
& \cellcolor{customblue!95}\est{+41.1}{2.8}{2.4}
& \cellcolor{customblue!60}\est{+25.4}{0.9}{1.0}
& \cellcolor{customblue!60}\est{+25.7}{0.9}{0.9}
& \cellcolor{customblue!70}\est{+28.9}{1.0}{1.0}
& \cellcolor{customblue!60}\est{+24.0}{0.9}{0.9} \\

Chronos-2 (FT)
& \cellcolor{customblue!70}\est{\mathbf{+29.4}}{1.2}{1.3}
& \cellcolor{customblue!70}\est{\mathbf{+29.8}}{1.3}{1.4}
& \cellcolor{customblue!100}\est{\mathbf{+41.3}}{2.8}{2.5}
& \cellcolor{customblue!60}\est{\mathbf{+25.7}}{0.9}{1.0}
& \cellcolor{customblue!65}\est{\mathbf{+26.0}}{0.9}{0.9}
& \cellcolor{customblue!70}\est{\mathbf{+29.2}}{1.0}{1.0}
& \cellcolor{customblue!60}\est{+24.5}{0.9}{0.9} \\

Toto
& \cellcolor{customblue!70}\est{+27.9}{1.3}{1.3}
& \cellcolor{customblue!70}\est{+28.3}{1.4}{1.4}
& \cellcolor{customblue!100}\est{\mathbf{+41.3}}{2.6}{2.4}
& \cellcolor{customblue!55}\est{+23.1}{1.1}{1.1}
& \cellcolor{customblue!55}\est{+23.6}{1.2}{1.1}
& \est{-8.7}{3.2}{3.1}
& \cellcolor{customblue!50}\est{+20.0}{1.2}{1.2} \\

Toto (FT)
& \cellcolor{customblue!65}\est{+27.2}{1.1}{1.1}
& \cellcolor{customblue!65}\est{+27.5}{1.2}{1.3}
& \cellcolor{customblue!95}\est{+41.2}{2.7}{2.5}
& \cellcolor{customblue!60}\est{+24.9}{1.0}{1.0}
& \cellcolor{customblue!60}\est{+25.1}{0.9}{0.9}
& \cellcolor{customblue!65}\est{+27.5}{1.2}{1.2}
& \cellcolor{customblue!60}\est{\mathbf{+24.6}}{0.8}{0.8} \\

\end{xltabular}

\vspace{0.8em}

\begin{xltabular}{\textwidth}{
    >{\raggedright\arraybackslash}X
    *{6}{>{\centering\arraybackslash}m{1.78cm}}
}

\caption{
\bo{Channel-level forecasting performance on binary channels.}
The table reports the Skill Score (in \%) of each model relative to the Seasonal Naive baseline for twelve binary channels, displayed in two connected panels. Subscripts and superscripts indicate the $95\%$ participant-level bootstrap percentile confidence interval based on $1{,}000$ resamples. FT denotes fine-tuned.
}
\label{tab:binary_channel_skill_scores}
\\

\toprule[1.4pt]

\textbf{Method}
& \textbf{Asleep}
& \textbf{In Bed}
& \textbf{Walking}
& \textbf{Cycling}
& \textbf{Running}
& \textbf{Other}
\\

\midrule
\endfirsthead

\multicolumn{7}{l}{
    \small\itshape
    Table~\thetable\ continued from the previous page
}
\\

\toprule[1.4pt]

\textbf{Method}
& \textbf{Asleep}
& \textbf{In Bed}
& \textbf{Walking}
& \textbf{Cycling}
& \textbf{Running}
& \textbf{Other}
\\

\midrule
\endhead

\midrule
\multicolumn{7}{r}{
    \small\itshape Continued on the next page
}
\\
\endfoot

\bottomrule[1.4pt]
\endlastfoot

\rowcolor[HTML]{EFEFEF}
\multicolumn{7}{l}{\textit{Statistical Models}} \\

Seasonal Naive
& \est{+0.0}{0.0}{0.0}
& \est{+0.0}{0.0}{0.0}
& \est{+0.0}{0.0}{0.0}
& \est{+0.0}{0.0}{0.0}
& \est{+0.0}{0.0}{0.0}
& \est{+0.0}{0.0}{0.0} \\

AutoARIMA
& \cellcolor{customblue!20}\est{+13.7}{6.2}{6.3}
& \est{-0.2}{6.9}{6.8}
& \est{-9.5}{8.5}{10.2}
& \cellcolor{customblue!5}\est{+3.9}{9.2}{11.2}
& \cellcolor{customblue!20}\est{+14.0}{9.7}{10.6}
& \cellcolor{customblue!40}\est{+26.8}{11.2}{10.8} \\

AutoETS
& \cellcolor{customblue!55}\est{+38.1}{4.0}{4.0}
& \cellcolor{customblue!50}\est{+37.0}{3.9}{3.8}
& \cellcolor{customblue!15}\est{\mathbf{+8.5}}{7.0}{7.9}
& \cellcolor{customblue!20}\est{\mathbf{+15.0}}{8.0}{8.6}
& \cellcolor{customblue!45}\est{\mathbf{+31.5}}{8.1}{8.9}
& \cellcolor{customblue!45}\est{+31.2}{11.4}{11.7} \\

\specialrule{\lightrulewidth}{0pt}{0pt}
\rowcolor[HTML]{EFEFEF}
\multicolumn{7}{l}{\textit{Neural Models}} \\

MixLinear
& \cellcolor{customblue!90}\est{+64.4}{2.5}{2.3}
& \cellcolor{customblue!90}\est{+64.8}{1.8}{1.8}
& \est{-58.5}{13.2}{15.2}
& \est{-0.9}{11.6}{14.0}
& \est{-1.3}{11.6}{12.9}
& \cellcolor{customblue!45}\est{\mathbf{+32.6}}{9.1}{9.6} \\

DLinear
& \cellcolor{customblue!95}\est{\mathbf{+69.5}}{1.8}{1.9}
& \cellcolor{customblue!100}\est{\mathbf{+73.4}}{1.7}{1.6}
& \est{-2.8}{6.9}{8.2}
& \est{-31.4}{14.2}{18.1}
& \cellcolor{customblue!5}\est{+0.3}{13.7}{15.9}
& \cellcolor{customblue!5}\est{+3.6}{12.4}{13.5} \\

SegRNN
& \cellcolor{customblue!90}\est{+66.1}{2.5}{2.6}
& \cellcolor{customblue!95}\est{+70.1}{1.8}{2.1}
& \est{-0.5}{7.2}{8.8}
& \est{-1.5}{11.6}{14.8}
& \est{-3.1}{10.4}{12.0}
& \cellcolor{customblue!5}\est{+3.5}{12.4}{13.5} \\

\specialrule{\lightrulewidth}{0pt}{0pt}
\rowcolor[HTML]{EFEFEF}
\multicolumn{7}{l}{\textit{Foundation Models}} \\

Chronos-2
& \cellcolor{customblue!80}\est{+59.0}{3.0}{2.9}
& \cellcolor{customblue!90}\est{+65.4}{2.3}{2.3}
& \cellcolor{customblue!5}\est{+3.3}{7.9}{9.8}
& \cellcolor{customblue!15}\est{+9.3}{10.1}{11.8}
& \cellcolor{customblue!20}\est{+12.6}{10.3}{12.2}
& \cellcolor{customblue!30}\est{+19.1}{9.8}{10.4} \\

Chronos-2 (FT)
& \cellcolor{customblue!85}\est{+60.9}{2.9}{2.8}
& \cellcolor{customblue!90}\est{+66.6}{2.1}{2.2}
& \cellcolor{customblue!10}\est{+5.6}{8.0}{9.2}
& \cellcolor{customblue!15}\est{+9.1}{10.1}{11.7}
& \cellcolor{customblue!20}\est{+14.2}{10.3}{11.5}
& \cellcolor{customblue!25}\est{+16.4}{9.6}{10.2} \\

Toto
& \cellcolor{customblue!65}\est{+47.9}{3.3}{3.2}
& \cellcolor{customblue!75}\est{+53.1}{2.7}{2.8}
& \est{-9.1}{7.6}{9.0}
& \cellcolor{customblue!5}\est{+0.5}{8.0}{10.8}
& \cellcolor{customblue!10}\est{+7.2}{8.4}{9.5}
& \cellcolor{customblue!20}\est{+11.6}{9.8}{12.1} \\

Toto (FT)
& \cellcolor{customblue!60}\est{+45.1}{3.3}{3.2}
& \cellcolor{customblue!65}\est{+47.1}{2.6}{2.7}
& \est{-2.4}{7.1}{8.2}
& \cellcolor{customblue!10}\est{+6.5}{8.5}{9.5}
& \cellcolor{customblue!15}\est{+11.5}{8.7}{9.5}
& \cellcolor{customblue!20}\est{+12.9}{9.6}{12.4} \\

\end{xltabular}

\vspace{0.4em}

\begin{xltabular}{\textwidth}{
    >{\raggedright\arraybackslash}X
    *{6}{>{\centering\arraybackslash}m{1.78cm}}
}

\multicolumn{7}{l}{
    \small\itshape
    Table~\ref{tab:binary_channel_skill_scores} continued:
    remaining binary channels
}
\\[0.15em]

\toprule[1.4pt]

\textbf{Method}
& \textbf{Cardio}
& \textbf{Strength}
& \textbf{Elliptical}
& \textbf{HIIT}
& \textbf{Functional}
& \textbf{Yoga}
\\

\midrule
\endfirsthead

\multicolumn{7}{l}{
    \small\itshape
    Table~\ref{tab:binary_channel_skill_scores} continued from the previous page
}
\\

\toprule[1.4pt]

\textbf{Method}
& \textbf{Cardio}
& \textbf{Strength}
& \textbf{Elliptical}
& \textbf{HIIT}
& \textbf{Functional}
& \textbf{Yoga}
\\

\midrule
\endhead

\midrule
\multicolumn{7}{r}{
    \small\itshape Continued on the next page
}
\\
\endfoot

\bottomrule[1.4pt]
\endlastfoot

\rowcolor[HTML]{EFEFEF}
\multicolumn{7}{l}{\textit{Statistical Models}} \\

Seasonal Naive
& \est{+0.0}{0.0}{0.0}
& \est{+0.0}{0.0}{0.0}
& \est{+0.0}{0.0}{0.0}
& \est{+0.0}{0.0}{0.0}
& \est{+0.0}{0.0}{0.0}
& \est{+0.0}{0.0}{0.0} \\

AutoARIMA
& \cellcolor{customblue!40}\est{+28.0}{34.1}{44.5}
& \cellcolor{customblue!20}\est{+14.1}{18.5}{24.3}
& \cellcolor{customblue!35}\est{+25.9}{11.0}{12.2}
& \cellcolor{customblue!55}\est{+39.6}{16.9}{21.3}
& \cellcolor{customblue!60}\est{\mathbf{+41.8}}{14.4}{16.4}
& \cellcolor{customblue!55}\est{+39.5}{11.2}{12.0} \\

AutoETS
& \cellcolor{customblue!25}\est{+16.8}{34.3}{42.8}
& \cellcolor{customblue!40}\est{\mathbf{+27.8}}{15.3}{18.6}
& \cellcolor{customblue!55}\est{\mathbf{+38.8}}{11.1}{12.9}
& \cellcolor{customblue!70}\est{\mathbf{+48.8}}{13.7}{14.6}
& \cellcolor{customblue!55}\est{+40.3}{13.4}{15.8}
& \cellcolor{customblue!65}\est{\mathbf{+44.4}}{10.3}{11.3} \\

\specialrule{\lightrulewidth}{0pt}{0pt}
\rowcolor[HTML]{EFEFEF}
\multicolumn{7}{l}{\textit{Neural Models}} \\

MixLinear
& \cellcolor{customblue!50}\est{+36.7}{28.9}{49.0}
& \est{-41.3}{22.5}{30.2}
& \est{-7.7}{16.5}{19.2}
& \est{-43.9}{33.8}{52.0}
& \est{-5.8}{21.7}{30.8}
& \est{-24.6}{18.8}{24.6} \\

DLinear
& \cellcolor{customblue!40}\est{+26.7}{43.7}{94.7}
& \cellcolor{customblue!40}\est{+27.4}{14.9}{18.8}
& \cellcolor{customblue!40}\est{+29.2}{13.5}{16.8}
& \est{-38.8}{34.4}{49.8}
& \cellcolor{customblue!15}\est{+8.2}{21.5}{26.9}
& \cellcolor{customblue!15}\est{+8.4}{16.9}{18.9} \\

SegRNN
& \est{-30.2}{30.7}{50.1}
& \est{-2.6}{23.8}{32.1}
& \cellcolor{customblue!35}\est{+25.9}{14.2}{14.9}
& \cellcolor{customblue!20}\est{+12.1}{20.8}{32.3}
& \cellcolor{customblue!10}\est{+6.5}{18.6}{25.9}
& \cellcolor{customblue!10}\est{+6.4}{20.5}{25.0} \\

\specialrule{\lightrulewidth}{0pt}{0pt}
\rowcolor[HTML]{EFEFEF}
\multicolumn{7}{l}{\textit{Foundation Models}} \\

Chronos-2
& \cellcolor{customblue!55}\est{+39.5}{30.1}{53.3}
& \cellcolor{customblue!25}\est{+16.3}{16.6}{22.3}
& \cellcolor{customblue!30}\est{+19.4}{17.0}{19.5}
& \cellcolor{customblue!15}\est{+11.2}{23.9}{31.3}
& \est{-7.0}{26.8}{34.2}
& \cellcolor{customblue!25}\est{+16.6}{16.5}{17.2} \\

Chronos-2 (FT)
& \cellcolor{customblue!45}\est{+32.6}{35.5}{68.0}
& \cellcolor{customblue!30}\est{+21.7}{16.1}{20.1}
& \cellcolor{customblue!30}\est{+21.5}{17.1}{20.4}
& \cellcolor{customblue!20}\est{+13.3}{20.0}{23.6}
& \cellcolor{customblue!10}\est{+6.9}{24.5}{33.3}
& \cellcolor{customblue!35}\est{+25.0}{16.6}{17.8} \\

Toto
& \cellcolor{customblue!15}\est{+9.6}{34.3}{51.1}
& \est{-4.8}{15.7}{22.9}
& \cellcolor{customblue!35}\est{+22.9}{13.8}{15.3}
& \cellcolor{customblue!25}\est{+16.9}{24.3}{34.5}
& \cellcolor{customblue!40}\est{+26.3}{13.3}{16.3}
& \cellcolor{customblue!40}\est{+29.7}{13.5}{14.5} \\

Toto (FT)
& \cellcolor{customblue!85}\est{\mathbf{+61.0}}{29.1}{71.6}
& \cellcolor{customblue!5}\est{+1.5}{16.9}{23.8}
& \cellcolor{customblue!20}\est{+13.6}{12.9}{12.9}
& \cellcolor{customblue!20}\est{+11.8}{18.8}{27.3}
& \cellcolor{customblue!35}\est{+24.0}{14.1}{17.3}
& \cellcolor{customblue!35}\est{+24.3}{14.7}{16.5} \\

\end{xltabular}

\endgroup

\begin{figure}[h!]
    \centering
    \caption{
    \textbf{Ablation results for different forecasting start times.}
    We report Aggregate Skill Score. FT denotes fine-tuned. Values are point estimates on the
    held-out test split; subscripts and superscripts indicate the $95\%$ participant-level bootstrap percentile confidence interval ($1{,}000$ resamples).
    }
    \includegraphics[width=0.8\linewidth]{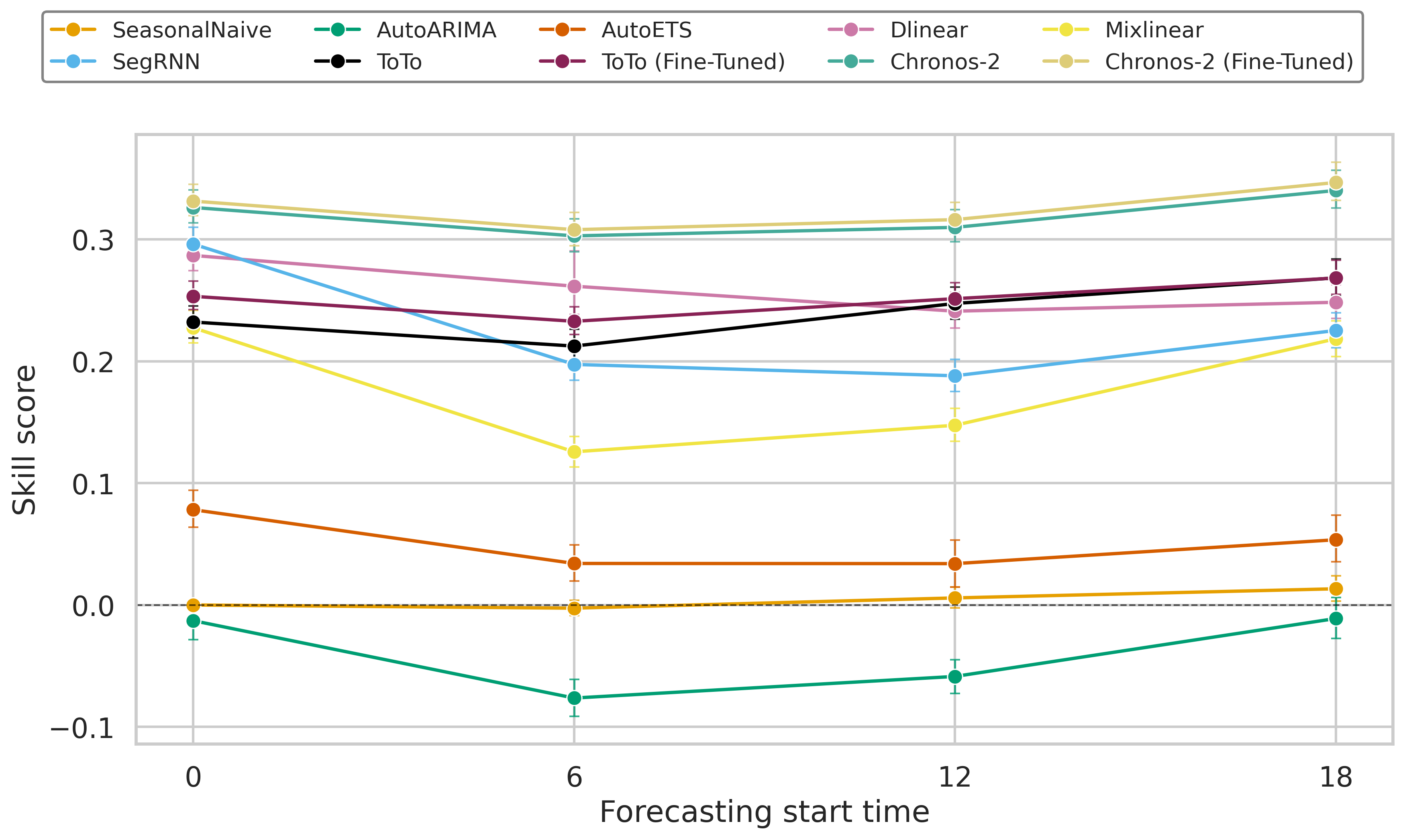}
    \label{fig:forecasting_start_time_ablation_skill_score_ci}
\end{figure}

The results reported in Table~\ref{tab:forecasting_grouped_model_summary} are all based on forecasts generated from a forecasting start time of 00:00. That is, each prediction covers a full calendar day from 00:00 to 24:00. To examine whether different forecasting start times affect model performance, we additionally evaluated each model with forecasting start times of 06:00, 12:00, and 18:00. Specifically, the forecasting window was shifted to [00{:}00 + \text{forecasting start time}, \text{forecasting start time on the next day}]. To ensure comparability across different settings, we used the seasonal naive model with a forecasting start time of 00:00 as the baseline for computing the Skill Score. The results are shown in Figure~\ref{fig:forecasting_start_time_ablation_skill_score_ci}. We observe that different forecasting start times have a noticeable impact on the three deep learning models (DLinear, MixLinear, SegRNN), which may be because these models were trained using datasets with a forecasting start time of 00:00.

\clearpage
\twocolumn

\onecolumn
\section{Full List of Linked Variables}
\label{app:label_registry}
This section lists every variable exposed through the labels API in our accompanying codebase (\texttt{src/labels/}). 
The list contains $170$ variables in total: $7$ longitudinal HealthKit-derived metrics extracted from raw Apple Watch, and $163$ variables obtained from or derived from the MHC app's questionnaires and app profile. Variables are grouped into $16$ semantic categories. The \textbf{Source} column distinguishes HealthKit (\texttt{HK}) from survey-derived (\texttt{Survey}) labels. The \textbf{Role} column marks each label as a prediction \emph{target} (\texttt{T}, exposed through \texttt{TARGET\_NAMES}) or an auxiliary \emph{context} covariate (\texttt{C}). 
The \textbf{Type} column reports the value type enforced by \texttt{src/labels/api.py}.

{\footnotesize
\renewcommand{\arraystretch}{1.10}
\setlength{\tabcolsep}{4pt}
\begin{longtable}{@{}>{\ttfamily\raggedright\arraybackslash}p{0.42\linewidth}>{\raggedright\arraybackslash}p{0.24\linewidth}llp{0.13\linewidth}@{}}
\caption{Complete labels registry for the OpenMHC release ($N=170$).
HealthKit (HK) labels are derived from raw Apple Watch
\texttt{HKQuantityTypeIdentifier} records; survey labels include
self-reported survey and app profile responses (such as height, weight,
bedtime and wake-up time) and derived variables. Roles are T
(predictive target) or C (auxiliary context covariate).}
\label{tab:full_label_registry} \\
\toprule
\textnormal{\textbf{Label}} & \textbf{Category} & \textbf{Source} & \textbf{Role} & \textbf{Type} \\
\midrule
\endfirsthead
\multicolumn{5}{@{}l}{\emph{Table~\ref{tab:full_label_registry} continued from previous page.}} \\
\toprule
\textnormal{\textbf{Label}} & \textbf{Category} & \textbf{Source} & \textbf{Role} & \textbf{Type} \\
\midrule
\endhead
\midrule
\multicolumn{5}{r@{}}{\emph{Continued on next page.}} \\
\endfoot
\bottomrule
\endlastfoot
\texttt{BMI\_\allowbreak categories} & anthropometrics & Survey & T & ordinal \\
\texttt{BMI\_\allowbreak values} & anthropometrics & Survey & T & continuous \\
\texttt{WeightKilograms} & anthropometrics & Survey & T & continuous \\
\texttt{field\_\allowbreak HeightCentimeters} & anthropometrics & Survey & C & continuous \\
\texttt{Diabetes} & cardiometabolic\_labs & Survey & T & binary \\
\texttt{Hdl} & cardiometabolic\_labs & Survey & T & continuous \\
\texttt{Hypertension} & cardiometabolic\_labs & Survey & T & binary \\
\texttt{Ldl} & cardiometabolic\_labs & Survey & T & continuous \\
\texttt{SystolicBloodPressure} & cardiometabolic\_labs & Survey & T & continuous \\
\texttt{TotalCholesterol} & cardiometabolic\_labs & Survey & T & continuous \\
\texttt{blood\_\allowbreak pressure\_\allowbreak categories} & cardiometabolic\_labs & Survey & T & ordinal \\
\texttt{field\_\allowbreak BloodGlucose} & cardiometabolic\_labs & Survey & C & continuous \\
\texttt{framingham\_\allowbreak risk} & cardiometabolic\_labs & Survey & T & continuous \\
\texttt{Atrial fibrillation (Afib)} & cardiovascular\_disease\_history & Survey & T & binary \\
\texttt{CAD} & cardiovascular\_disease\_history & Survey & T & binary \\
\texttt{Cerebrovascular Disease} & cardiovascular\_disease\_history & Survey & T & binary \\
\texttt{Congenital Heart} & cardiovascular\_disease\_history & Survey & T & binary \\
\texttt{Heart Failure or CHF} & cardiovascular\_disease\_history & Survey & T & binary \\
\texttt{PH} & cardiovascular\_disease\_history & Survey & T & binary \\
\texttt{Peripheral/Systemic Vascular Disease} & cardiovascular\_disease\_history & Survey & T & binary \\
\texttt{cardiovascular\_\allowbreak disease} & cardiovascular\_disease\_history & Survey & T & binary \\
\texttt{field\_\allowbreak family\_\allowbreak history} & cardiovascular\_disease\_history & Survey & C & multi\_categorical \\
\texttt{field\_\allowbreak medications\_\allowbreak to\_\allowbreak treat} & cardiovascular\_disease\_history & Survey & C & multi\_categorical \\
\texttt{field\_\allowbreak antibiotics} & covid\_19 & Survey & C & multi\_categorical \\
\texttt{field\_\allowbreak building} & covid\_19 & Survey & C & ordinal \\
\texttt{field\_\allowbreak conditions} & covid\_19 & Survey & C & multi\_categorical \\
\texttt{field\_\allowbreak covid} & covid\_19 & Survey & C & ordinal \\
\texttt{field\_\allowbreak covid\_\allowbreak serologic} & covid\_19 & Survey & C & ordinal \\
\texttt{field\_\allowbreak daily\_\allowbreak activities} & covid\_19 & Survey & C & ordinal \\
\texttt{field\_\allowbreak days\_\allowbreak admitted} & covid\_19 & Survey & C & continuous \\
\texttt{field\_\allowbreak exposure} & covid\_19 & Survey & C & ordinal \\
\texttt{field\_\allowbreak face\_\allowbreak covering} & covid\_19 & Survey & C & ordinal \\
\texttt{field\_\allowbreak healthcare\_\allowbreak worker} & covid\_19 & Survey & C & categorical \\
\texttt{field\_\allowbreak most\_\allowbreak intense\_\allowbreak care} & covid\_19 & Survey & C & ordinal \\
\texttt{field\_\allowbreak self\_\allowbreak isolating} & covid\_19 & Survey & C & ordinal \\
\texttt{field\_\allowbreak severity} & covid\_19 & Survey & C & ordinal \\
\texttt{field\_\allowbreak severity\_\allowbreak covid} & covid\_19 & Survey & C & ordinal \\
\texttt{field\_\allowbreak symptoms\_\allowbreak past\_\allowbreak week} & covid\_19 & Survey & C & multi\_categorical \\
\texttt{field\_\allowbreak symptoms\_\allowbreak week\_\allowbreak preceding} & covid\_19 & Survey & C & multi\_categorical \\
\texttt{BiologicalSex} & demographics & Survey & T & binary \\
\texttt{age} & demographics & Survey & T & continuous \\
\texttt{field\_\allowbreak Ethnicity\_\allowbreak heartage} & demographics & Survey & C & categorical \\
\texttt{field\_\allowbreak FitzpatrickSkinType} & demographics & Survey & C & ordinal \\
\texttt{field\_\allowbreak education} & demographics & Survey & C & ordinal \\
\texttt{field\_\allowbreak ethnicity} & demographics & Survey & C & ordinal \\
\texttt{field\_\allowbreak race} & demographics & Survey & C & multi\_categorical \\
\texttt{field\_\allowbreak alcohol} & diet & Survey & C & ordinal \\
\texttt{field\_\allowbreak fish} & diet & Survey & C & continuous \\
\texttt{field\_\allowbreak fruit} & diet & Survey & C & continuous \\
\texttt{field\_\allowbreak grains} & diet & Survey & C & continuous \\
\texttt{field\_\allowbreak sodium} & diet & Survey & C & multi\_categorical \\
\texttt{field\_\allowbreak sugar\_\allowbreak drinks} & diet & Survey & C & continuous \\
\texttt{field\_\allowbreak vegetable} & diet & Survey & C & continuous \\
\texttt{field\_\allowbreak country} & geography & Survey & C & categorical \\
\texttt{field\_\allowbreak ruca\_\allowbreak primary\_\allowbreak mean} & geography & Survey & C & continuous \\
\texttt{field\_\allowbreak state} & geography & Survey & C & categorical \\
\texttt{Watch\_\allowbreak BasalEnergyBurned} & healthkit\_watch\_metrics & HK & T & continuous \\
\texttt{Watch\_\allowbreak HeartRateVariabilitySDNN} & healthkit\_watch\_metrics & HK & T & continuous \\
\texttt{Watch\_\allowbreak RespiratoryRate} & healthkit\_watch\_metrics & HK & T & continuous \\
\texttt{Watch\_\allowbreak RestingHeartRate} & healthkit\_watch\_metrics & HK & T & continuous \\
\texttt{Watch\_\allowbreak StandTime} & healthkit\_watch\_metrics & HK & T & continuous \\
\texttt{Watch\_\allowbreak VO2Max} & healthkit\_watch\_metrics & HK & T & continuous \\
\texttt{Watch\_\allowbreak WalkingHeartRateAverage} & healthkit\_watch\_metrics & HK & T & continuous \\
\texttt{field\_\allowbreak beneficial} & mindset\_measures & Survey & C & ordinal \\
\texttt{field\_\allowbreak body\_\allowbreak remarkable\_\allowbreak self\_\allowbreak healing} & mindset\_measures & Survey & C & ordinal \\
\texttt{field\_\allowbreak body\_\allowbreak self\_\allowbreak healing\_\allowbreak from\_\allowbreak most\_\allowbreak conditions\_\allowbreak and\_\allowbreak diseases} & mindset\_measures & Survey & C & ordinal \\
\texttt{field\_\allowbreak body\_\allowbreak self\_\allowbreak healing\_\allowbreak in\_\allowbreak many\_\allowbreak different\_\allowbreak circumstances} & mindset\_measures & Survey & C & ordinal \\
\texttt{field\_\allowbreak chronic\_\allowbreak illness\_\allowbreak body\_\allowbreak betrayal} & mindset\_measures & Survey & C & ordinal \\
\texttt{field\_\allowbreak chronic\_\allowbreak illness\_\allowbreak body\_\allowbreak blame} & mindset\_measures & Survey & C & ordinal \\
\texttt{field\_\allowbreak chronic\_\allowbreak illness\_\allowbreak body\_\allowbreak coping} & mindset\_measures & Survey & C & ordinal \\
\texttt{field\_\allowbreak chronic\_\allowbreak illness\_\allowbreak body\_\allowbreak failure} & mindset\_measures & Survey & C & ordinal \\
\texttt{field\_\allowbreak chronic\_\allowbreak illness\_\allowbreak body\_\allowbreak handling} & mindset\_measures & Survey & C & ordinal \\
\texttt{field\_\allowbreak chronic\_\allowbreak illness\_\allowbreak body\_\allowbreak management} & mindset\_measures & Survey & C & ordinal \\
\texttt{field\_\allowbreak chronic\_\allowbreak illness\_\allowbreak body\_\allowbreak meaning} & mindset\_measures & Survey & C & ordinal \\
\texttt{field\_\allowbreak chronic\_\allowbreak illness\_\allowbreak challenge} & mindset\_measures & Survey & C & ordinal \\
\texttt{field\_\allowbreak chronic\_\allowbreak illness\_\allowbreak empowering} & mindset\_measures & Survey & C & ordinal \\
\texttt{field\_\allowbreak chronic\_\allowbreak illness\_\allowbreak handling} & mindset\_measures & Survey & C & ordinal \\
\texttt{field\_\allowbreak chronic\_\allowbreak illness\_\allowbreak impact} & mindset\_measures & Survey & C & ordinal \\
\texttt{field\_\allowbreak chronic\_\allowbreak illness\_\allowbreak management} & mindset\_measures & Survey & C & ordinal \\
\texttt{field\_\allowbreak chronic\_\allowbreak illness\_\allowbreak more\_\allowbreak meaning\_\allowbreak in\_\allowbreak life} & mindset\_measures & Survey & C & ordinal \\
\texttt{field\_\allowbreak chronic\_\allowbreak illness\_\allowbreak positive\_\allowbreak opportunity} & mindset\_measures & Survey & C & ordinal \\
\texttt{field\_\allowbreak chronic\_\allowbreak illness\_\allowbreak relatively\_\allowbreak normal\_\allowbreak life} & mindset\_measures & Survey & C & ordinal \\
\texttt{field\_\allowbreak chronic\_\allowbreak illness\_\allowbreak runing\_\allowbreak life} & mindset\_measures & Survey & C & ordinal \\
\texttt{field\_\allowbreak chronic\_\allowbreak illness\_\allowbreak spoil} & mindset\_measures & Survey & C & ordinal \\
\texttt{field\_\allowbreak convenient} & mindset\_measures & Survey & C & ordinal \\
\texttt{field\_\allowbreak disease} & mindset\_measures & Survey & C & ordinal \\
\texttt{field\_\allowbreak easy} & mindset\_measures & Survey & C & ordinal \\
\texttt{field\_\allowbreak fun} & mindset\_measures & Survey & C & ordinal \\
\texttt{field\_\allowbreak indulgent} & mindset\_measures & Survey & C & ordinal \\
\texttt{field\_\allowbreak muscles} & mindset\_measures & Survey & C & ordinal \\
\texttt{field\_\allowbreak pleasurable} & mindset\_measures & Survey & C & ordinal \\
\texttt{field\_\allowbreak relaxing} & mindset\_measures & Survey & C & ordinal \\
\texttt{field\_\allowbreak social} & mindset\_measures & Survey & C & ordinal \\
\texttt{field\_\allowbreak unhealthy} & mindset\_measures & Survey & C & ordinal \\
\texttt{field\_\allowbreak weight} & mindset\_measures & Survey & C & ordinal \\
\texttt{field\_\allowbreak chestPain} & parq\_readiness & Survey & C & binary \\
\texttt{field\_\allowbreak chestPainInLastMonth} & parq\_readiness & Survey & C & binary \\
\texttt{field\_\allowbreak dizziness} & parq\_readiness & Survey & C & binary \\
\texttt{field\_\allowbreak heartCondition} & parq\_readiness & Survey & C & binary \\
\texttt{field\_\allowbreak jointProblem} & parq\_readiness & Survey & C & binary \\
\texttt{field\_\allowbreak physicallyCapable} & parq\_readiness & Survey & C & binary \\
\texttt{field\_\allowbreak prescriptionDrugs} & parq\_readiness & Survey & C & binary \\
\texttt{field\_\allowbreak activity1\_\allowbreak intensity} & physical\_activity & Survey & C & ordinal \\
\texttt{field\_\allowbreak activity1\_\allowbreak option} & physical\_activity & Survey & C & binary \\
\texttt{field\_\allowbreak activity1\_\allowbreak time} & physical\_activity & Survey & C & continuous \\
\texttt{field\_\allowbreak activity1\_\allowbreak type} & physical\_activity & Survey & C & categorical \\
\texttt{field\_\allowbreak activity2\_\allowbreak intensity} & physical\_activity & Survey & C & ordinal \\
\texttt{field\_\allowbreak activity2\_\allowbreak option} & physical\_activity & Survey & C & binary \\
\texttt{field\_\allowbreak activity2\_\allowbreak time} & physical\_activity & Survey & C & continuous \\
\texttt{field\_\allowbreak activity2\_\allowbreak type} & physical\_activity & Survey & C & categorical \\
\texttt{field\_\allowbreak atwork} & physical\_activity & Survey & C & ordinal \\
\texttt{field\_\allowbreak moderate\_\allowbreak act} & physical\_activity & Survey & C & continuous \\
\texttt{field\_\allowbreak phys\_\allowbreak activity} & physical\_activity & Survey & C & ordinal \\
\texttt{vigorous\_\allowbreak act} & physical\_activity & Survey & T & continuous \\
\texttt{work} & physical\_activity & Survey & T & binary \\
\texttt{field\_\allowbreak riskfactors1} & risk\_perception & Survey & C & ordinal \\
\texttt{field\_\allowbreak riskfactors2} & risk\_perception & Survey & C & ordinal \\
\texttt{field\_\allowbreak riskfactors3} & risk\_perception & Survey & C & ordinal \\
\texttt{field\_\allowbreak riskfactors4} & risk\_perception & Survey & C & ordinal \\
\texttt{GoSleepTime\_\allowbreak categories} & sleep & Survey & T & ordinal \\
\texttt{WakeUpTime\_\allowbreak categories} & sleep & Survey & T & ordinal \\
\texttt{field\_\allowbreak GoSleepTime} & sleep & Survey & C & continuous \\
\texttt{field\_\allowbreak WakeUpTime} & sleep & Survey & C & continuous \\
\texttt{field\_\allowbreak sleep\_\allowbreak diagnosis2} & sleep & Survey & C & multi\_categorical \\
\texttt{field\_\allowbreak sleep\_\allowbreak time} & sleep & Survey & C & continuous \\
\texttt{field\_\allowbreak sleep\_\allowbreak time1} & sleep & Survey & C & continuous \\
\texttt{field\_\allowbreak sleep\_\allowbreak time\_\allowbreak daily} & sleep & Survey & C & continuous \\
\texttt{sleep\_\allowbreak diagnosis1} & sleep & Survey & T & binary \\
\texttt{sleep\_\allowbreak time\_\allowbreak categories} & sleep & Survey & T & ordinal \\
\texttt{field\_\allowbreak device\_\allowbreak activity\_\allowbreak band} & study\_metadata & Survey & C & binary \\
\texttt{field\_\allowbreak device\_\allowbreak iphone} & study\_metadata & Survey & C & binary \\
\texttt{field\_\allowbreak device\_\allowbreak other} & study\_metadata & Survey & C & binary \\
\texttt{field\_\allowbreak device\_\allowbreak smartwatch} & study\_metadata & Survey & C & binary \\
\texttt{field\_\allowbreak labwork} & study\_metadata & Survey & C & categorical \\
\texttt{field\_\allowbreak phone\_\allowbreak on\_\allowbreak user} & study\_metadata & Survey & C & ordinal \\
\texttt{field\_\allowbreak cannabisSmoking} & tobacco\_vaping\_cannabis & Survey & C & ordinal \\
\texttt{field\_\allowbreak cannabisVaping} & tobacco\_vaping\_cannabis & Survey & C & ordinal \\
\texttt{field\_\allowbreak currentCannabisSmoking} & tobacco\_vaping\_cannabis & Survey & C & ordinal \\
\texttt{field\_\allowbreak currentCannabisVaping} & tobacco\_vaping\_cannabis & Survey & C & ordinal \\
\texttt{field\_\allowbreak currentSmokeless} & tobacco\_vaping\_cannabis & Survey & C & ordinal \\
\texttt{field\_\allowbreak currentSmoking} & tobacco\_vaping\_cannabis & Survey & C & ordinal \\
\texttt{field\_\allowbreak currentVaping} & tobacco\_vaping\_cannabis & Survey & C & ordinal \\
\texttt{field\_\allowbreak durationQuitSmokeless} & tobacco\_vaping\_cannabis & Survey & C & categorical \\
\texttt{field\_\allowbreak durationQuitSmoking} & tobacco\_vaping\_cannabis & Survey & C & categorical \\
\texttt{field\_\allowbreak durationQuitVaping} & tobacco\_vaping\_cannabis & Survey & C & categorical \\
\texttt{field\_\allowbreak everQuitSmokeless} & tobacco\_vaping\_cannabis & Survey & C & binary \\
\texttt{field\_\allowbreak everQuitSmoking} & tobacco\_vaping\_cannabis & Survey & C & binary \\
\texttt{field\_\allowbreak everQuitVaping} & tobacco\_vaping\_cannabis & Survey & C & binary \\
\texttt{field\_\allowbreak lastCannabisSmoking} & tobacco\_vaping\_cannabis & Survey & C & ordinal \\
\texttt{field\_\allowbreak lastCannabisVaping} & tobacco\_vaping\_cannabis & Survey & C & ordinal \\
\texttt{field\_\allowbreak onsetSmokeless} & tobacco\_vaping\_cannabis & Survey & C & continuous \\
\texttt{field\_\allowbreak onsetSmoking} & tobacco\_vaping\_cannabis & Survey & C & continuous \\
\texttt{field\_\allowbreak onsetVaping} & tobacco\_vaping\_cannabis & Survey & C & continuous \\
\texttt{field\_\allowbreak pastCannabisSmoking} & tobacco\_vaping\_cannabis & Survey & C & ordinal \\
\texttt{field\_\allowbreak pastCannabisVaping} & tobacco\_vaping\_cannabis & Survey & C & ordinal \\
\texttt{field\_\allowbreak pastSmokeless} & tobacco\_vaping\_cannabis & Survey & C & ordinal \\
\texttt{field\_\allowbreak pastVaping} & tobacco\_vaping\_cannabis & Survey & C & ordinal \\
\texttt{field\_\allowbreak readinessQuitSmokeless} & tobacco\_vaping\_cannabis & Survey & C & ordinal \\
\texttt{field\_\allowbreak readinessQuitSmoking} & tobacco\_vaping\_cannabis & Survey & C & ordinal \\
\texttt{field\_\allowbreak readinessQuitVaping} & tobacco\_vaping\_cannabis & Survey & C & ordinal \\
\texttt{field\_\allowbreak smokingHistory} & tobacco\_vaping\_cannabis & Survey & C & binary \\
\texttt{field\_\allowbreak tobaccoProducts} & tobacco\_vaping\_cannabis & Survey & C & multi\_categorical \\
\texttt{field\_\allowbreak tobaccoProductsEver} & tobacco\_vaping\_cannabis & Survey & C & multi\_categorical \\
\texttt{feel\_\allowbreak worthwhile1} & wellbeing & Survey & T & ordinal \\
\texttt{feel\_\allowbreak worthwhile2} & wellbeing & Survey & T & ordinal \\
\texttt{feel\_\allowbreak worthwhile3} & wellbeing & Survey & T & ordinal \\
\texttt{feel\_\allowbreak worthwhile4} & wellbeing & Survey & T & ordinal \\
\texttt{happiness} & wellbeing & Survey & T & continuous \\
\texttt{happiness\_\allowbreak categories} & wellbeing & Survey & T & ordinal \\
\texttt{satisfiedwith\_\allowbreak life} & wellbeing & Survey & T & ordinal 

\end{longtable}
}

\clearpage
\section*{NeurIPS Paper Checklist}

\begin{enumerate}

\item {\bf Claims}
    \item[] Question: Do the main claims made in the abstract and introduction accurately reflect the paper's contributions and scope?
    \item[] Answer: \answerYes{} 
    \item[] Justification: Both the introduction and abstract accurately outline the scope and contribution of this paper.

\item {\bf Limitations}
    \item[] Question: Does the paper discuss the limitations of the work performed by the authors?
    \item[] Answer: \answerYes{} 
    \item[] Justification: 
    Limitations are discussed in the Discussion's Limitations paragraph and Appendix~\ref{sec:limitations}. 

\item {\bf Theory assumptions and proofs}
    \item[] Question: For each theoretical result, does the paper provide the full set of assumptions and a complete (and correct) proof?
    \item[] Answer: \answerNA{} 
    \item[] Justification: We propose no new theoretical results.

    \item {\bf Experimental result reproducibility}
    \item[] Question: Does the paper fully disclose all the information needed to reproduce the main experimental results of the paper to the extent that it affects the main claims and/or conclusions of the paper (regardless of whether the code and data are provided or not)?
    \item[] Answer: \answerYes{} 
    \item[] Justification: Setup and approaches are introduced in the main paper and elaborated in detail across the appendix: research implementation and ethical oversight (Appendix~\ref{sec:implementation}), evaluation methodology including the paired user-level bootstrap (Appendix~\ref{sec:scoring_methodology}), per-track preprocessing, hyperparameters, and baselines (Appendices~\ref{app:preprocessing}, \ref{app:health_outcome_prediction}, \ref{sec:imputation_baselines}, \ref{sec:deeplearningforecast}), and compute resources (Appendix~\ref{sec:computing_resources}). The full code base for replicating all reported results is available at the GitHub repository linked in the Introduction (the public research code is linked within that repository as well), and the dataset is distributed via Harvard Dataverse under a Data Use Agreement (described in the Dataset section). For the Gemini baselines, Appendix~\ref{app:gemini_baselines} reports the model identifiers and inference settings. Model checkpoints for trained models are available on Hugging Face (Appendix~\ref{app:release_notes}). For full transparency we also release the data processing code, allowing interested readers to inspect every assumption and design decision in the pipeline.

\item {\bf Open access to data and code}
    \item[] Question: Does the paper provide open access to the data and code, with sufficient instructions to faithfully reproduce the main experimental results, as described in supplemental material?
    \item[] Answer: \answerYes{} 
    \item[] Justification: All code and data are released. All code is published on GitHub; this includes all training, evaluation, and data-processing pipelines together with configuration files, reproducibility instructions, and a labels API for downstream extension. The data are distributed through Harvard Dataverse with a click-through DUA, with a separate 5\% development subsample for fast iteration. Beyond the raw inputs we also ship intermediate processed artifacts (daily and hourly Hugging Face Arrow datasets, label-validity masks, window indices, and forecasting sample indices) so that downstream users can reproduce results without rebuilding the pipeline from scratch.

\item {\bf Experimental setting/details}
    \item[] Question: Does the paper specify all the training and test details (e.g., data splits, hyperparameters, how they were chosen, type of optimizer) necessary to understand the results?
    \item[] Answer: \answerYes{} 
    \item[] Justification: Exact data splits are released alongside the dataset as JSON files keyed by participant ID (60/10/30 train/val/test at the participant level, plus a 5\% development subsample). Per-track preprocessing and inclusion criteria are documented in Appendix~\ref{app:preprocessing} (with sensitivity analyses in Appendix~\ref{app:inclusion_ablations}). Hyperparameter search spaces, selected configurations, and optimizer settings are provided in the per-track baseline appendices (Appendices~\ref{app:health_outcome_prediction}, \ref{sec:imputation_baselines}, \ref{sec:deeplearningforecast}). All configurations are also versioned in the code repository as YAML config files for reproducibility.

\item {\bf Experiment statistical significance}
    \item[] Question: Does the paper report error bars suitably and correctly defined or other appropriate information about the statistical significance of the experiments?
    \item[] Answer: \answerYes{} 
    \item[] Justification: We report $95\%$ bootstrap confidence intervals using $B{=}1{,}000$ paired participant-level resamples (Appendix~\ref{sec:scoring_methodology}); the construction for worst-subgroup skill is detailed in Appendix~\ref{app:subgroup_reporting}. Forecast targets are non-overlapping 24-hour windows, and resampling keeps each participant's evaluated days together (Appendix~\ref{sec:forecasting_aggregation}).
    The unified evaluation methodology, the skill score, and the average rank are described in Appendix~\ref{sec:scoring_methodology}; performance stratified by demographic subgroups is reported as described in Section~\ref{sec:evaluation}. Statistical tests and sensitivity analyses for data filtering choices are reported in Appendix~\ref{app:inclusion_ablations}. Sensitivity of model rankings to evaluation choices is reported in Appendix~\ref{app:aggregation_sensitivity}.

\item {\bf Experiments compute resources}
    \item[] Question: For each experiment, does the paper provide sufficient information on the computer resources (type of compute workers, memory, time of execution) needed to reproduce the experiments?
    \item[] Answer: \answerYes{} 
    \item[] Justification: \justificationYes{Compute resources are documented in Appendix~\ref{sec:computing_resources}, including the academic HPC clusters and heterogeneous NVIDIA GPU mix used, single-run wall-clock ranges, hardware and timing tables for selected pretraining and inference/evaluation components (Tables~\ref{tab:compute_pretraining} and~\ref{tab:compute_inference}), cumulative GPU-hours for reported experiments, additional GPU-hours for hyperparameter sweeps and exploratory runs, and the at-list-price equivalent on Vertex AI for the frontier-LLM and agentic API-based baselines (which consumed no local GPU compute by the authors and incurred zero actual cost under a courtesy research allocation).}
    
\item {\bf Code of ethics}
    \item[] Question: Does the research conducted in the paper conform, in every respect, with the NeurIPS Code of Ethics \url{https://neurips.cc/public/EthicsGuidelines}?
    \item[] Answer: \answerYes{} 
    \item[] Justification: \justificationYes{Yes, we read the NeurIPS Code of Ethics and confirm that the conducted research conforms to it in every aspect.}

\item {\bf Broader impacts}
    \item[] Question: Does the paper discuss both potential positive societal impacts and negative societal impacts of the work performed?
    \item[] Answer: \answerYes{} 
    \item[] Justification: Both positive and negative societal impacts are discussed in the Discussion. 
    On the positive side, the dataset and benchmark are intended to democratize access to large-scale wearable and mobile health data and to accelerate methods development on real-world data characteristics (missingness, irregular sampling, demographic heterogeneity) that have previously been studied predominantly behind industry walls, with potential downstream benefits for cardiovascular and digital-health research at population scale. On the negative side, demographic skew (predominantly white, male, late-thirties, US metropolitan, Apple-ecosystem participants) limits generalizability and may reinforce bias if downstream users do not adjust for it. We report rank and skill by age and sex, together with worst-subgroup skill, to make differences across demographic groups visible (Section~\ref{sec:evaluation}, Appendix~\ref{app:subgroup_reporting}). 
    
\item {\bf Safeguards}
    \item[] Question: Does the paper describe safeguards that have been put in place for responsible release of data or models that have a high risk for misuse (e.g., pre-trained language models, image generators, or scraped datasets)?
    \item[] Answer: \answerYes{} 
    \item[] Justification: Distribution is access-gated via Harvard Dataverse behind a click-through Data Use Agreement, approved by Stanford's Privacy Office, that restricts use to personal, non-commercial research and prohibits commercial use, any attempt at re-identification, and distribution of any portion of the data without written permission from the Stanford University School of Medicine; users must report potentially identifying information and any re-identification to Stanford's Privacy Office, and may train generative AI models only if no data can be obtained or recreated from them. 
    Stanford's Privacy Office and multiple researchers independently and manually reviewed all variables. The ZIP code/postcode field was removed. Three-digit US ZIP prefixes were used to derive state or territory and a rounded mean primary Rural-Urban Commuting Area (RUCA) code. 
    Fitzpatrick skin type was grouped into three levels, 
    race and ethnicity categories were coarsened into larger groups, and four rare outcomes were removed. 
    Dates were randomly shifted by a per-participant offset (Appendix~\ref{app:data_access}). Appendix~\ref{app:data_access} summarizes the DUA terms, including the restriction to personal, non-commercial research and the prohibition on diagnosis or patient care.

\item {\bf Licenses for existing assets}
    \item[] Question: Are the creators or original owners of assets (e.g., code, data, models), used in the paper, properly credited and are the license and terms of use explicitly mentioned and properly respected?
    \item[] Answer: \answerYes{} 
    \item[] Justification: The OpenMHC dataset is derived from the My Heart Counts Cardiovascular Health Study; the upstream data descriptor and study-design publications are cited in the Dataset section~\citep{hershman2019physical, mcconnell2017feasibility}. The dataset itself is owned by Stanford University and distributed on Harvard Dataverse under a custom Data Use Agreement (no SPDX equivalent; the full agreement text is shown to all downloaders at download time and is reachable through the Dataverse landing page). Appendix~\ref{app:asset_licenses} identifies the reused code, pretrained checkpoints, scientific libraries and collection software, names their licenses, and links to the upstream terms. It also identifies the Gemini API service terms and the RUCA reference's catalogue license. Original methods are credited in the corresponding model subsections; our LSM-2 and WBM models are reimplementations trained on OpenMHC, while Chronos-2 and Toto use upstream checkpoints and adapted code.

\item {\bf New assets}
    \item[] Question: Are new assets introduced in the paper well documented and is the documentation provided alongside the assets?
    \item[] Answer: \answerYes{} 
    \item[] Justification: The OpenMHC dataset (full and 5\% development subsample) is documented through a layered set of artifacts: (i) two Croissant 1.1 metadata files (one per Dataverse deposit), each containing the full Croissant-RAI block required by the NeurIPS 2026 Evaluations \& Datasets Track (Data Limitations, Data Biases, Personal/Sensitive Information, Data Use Cases, Social Impact, Synthetic-Data flag, Provenance) and structured \texttt{prov:wasGeneratedBy} entries describing the Data Collection and Data Preprocessing activities and their attributed agents; (ii) a Data Use Agreement shown at download time; (iii) per-deposit READMEs and a labels README describing every label and its release-time disposition; and (iv) a deposit-level RELEASE\_NOTES file documenting the privacy / re-identification audit. The MHC-Benchmark code base is released on GitHub with detailed READMEs, configuration files, reproduction scripts, and a labels API for downstream extension.

\item {\bf Crowdsourcing and research with human subjects}
    \item[] Question: For crowdsourcing experiments and research with human subjects, does the paper include the full text of instructions given to participants and screenshots, if applicable, as well as details about compensation (if any)? 
    \item[] Answer: \answerYes{} 
    \item[] Justification: Screenshots of the consent flow are provided 
    alongside a description of the data collection application in Appendix~\ref{subsec:app_details}; the original \textit{Scientific Data} descriptor of the My Heart Counts study \citep{hershman2019physical} provides the original collection protocol. Sensor data were collected passively via Apple HealthKit, and participants were prompted by the application to complete questionnaires. As a long-term, fully-remote mobile-health study, compliance was not penalised in any shape or form nor were any monetary incentives provided.

\item {\bf Institutional review board (IRB) approvals or equivalent for research with human subjects}
    \item[] Question: Does the paper describe potential risks incurred by study participants, whether such risks were disclosed to the subjects, and whether Institutional Review Board (IRB) approvals (or an equivalent approval/review based on the requirements of your country or institution) were obtained?
    \item[] Answer: \answerYes{} 
    \item[] Justification: Participants in the My Heart Counts study were informed of all potential risks and provided digital informed consent (e-consent) at enrollment, with separate broad-vs-narrow consent tiers governing whether their de-identified data could be redistributed to qualified researchers worldwide; only the broad-consent subset is included in OpenMHC. The study was reviewed and approved by Stanford University's Research Compliance Office under Protocol \#IRB-31409, mentioned in the Dataset section of the paper and described in detail in Appendix~\ref{sec:implementation}. The release of OpenMHC was additionally reviewed and approved by Stanford's Privacy Office.

\item {\bf Declaration of LLM usage}
    \item[] Question: Does the paper describe the usage of LLMs if it is an important, original, or non-standard component of the core methods in this research? Note that if the LLM is used only for writing, editing, or formatting purposes and does \emph{not} impact the core methodology, scientific rigor, or originality of the research, declaration is not required.
    \item[] Answer: \answerYes{} 
    \item[] Justification: We declare the following LLM use cases for transparency. (i) \textbf{As evaluated baselines.} \textsc{Gemini-2.5-Pro} and \textsc{Gemini-3.1-Pro-Preview} are evaluated as zero-shot and few-shot baselines on the downstream-prediction tasks under three prompting strategies (statistics, vision, and agentic), with a Gemini-2.5-Pro--based variant of the Personal Health Agent Data Science Agent~\citep{merrill2026transforming, heydari2025anatomy} as the agentic configuration; the methodology, prompting templates, cohort construction, and per-task results are detailed in Appendix~\ref{app:gemini_baselines} and Appendix~\ref{app:gemini_vision_details}. These models are evaluated as baselines for comparison rather than proposed as part of our core contribution. (ii) \textbf{As a coding assistant.} Claude Code has been used during development and code review; all AI-suggested code has been modified and verified by at least two authors of the paper. (iii) \textbf{For proofreading and editorial feedback.} LLMs may also have been used by individual authors for proofreading, stylistic feedback, and minor editorial suggestions on draft sections; the substantive content, scientific claims, and structural decisions of the manuscript are authored by the team.

\end{enumerate}

\end{document}